%% file: neurips_2026.tex
\pdfoutput=1
\documentclass{article}

\PassOptionsToPackage{numbers, sort&compress}{natbib}

\usepackage[final,eandd]{neurips_2026}

\input{preamble}

\newcommand{\availability}{\space\textbf{Code:} \url{https://github.com/LOGO-CUHKSZ/TRL-Bench};
  data is released on Hugging Face.\footnote{%
    \begin{tabular}[t]{@{}l@{\ }l@{}}
      \textsc{TRL-CTbench}: & \url{https://huggingface.co/datasets/logo-lab/trl-ctbench}\\
      \textsc{TRL-Rbench}:  & \url{https://huggingface.co/datasets/logo-lab/trl-rbench}\\
      \textsc{TRL-DLTE}:    & \url{https://huggingface.co/datasets/logo-lab/trl-dlte}%
    \end{tabular}}}

\author{%
  \normalfont
  \textbf{Wei Pang}$^{1,*}$, \textbf{Xiangru Jian}$^{2,*}$, \textbf{Hehan Li}$^{1,*}$, \textbf{Zhixuan Yu}$^{1,*}$, \textbf{Alex Xue}$^{2,*}$ \\[3pt]
  \textbf{Jinyang Li}$^{3}$, \textbf{Zhengyuan Dong}$^{2}$, \textbf{Xinjian Zhao}$^{1}$, \textbf{Hao Xu}$^{4}$ \\[3pt]
  \textbf{Chao Zhang}$^{5}$, \textbf{Reynold Cheng}$^{3}$, \textbf{M.~Tamer Özsu}$^{2}$, \textbf{Tianshu Yu}$^{1,\dagger}$ \\[8pt]
  \small $^{1}$The Chinese University of Hong Kong, Shenzhen \quad $^{2}$University of Waterloo \\[2pt]
  \small $^{3}$The University of Hong Kong \quad $^{4}$The University of Sydney \quad $^{5}$Université Lyon 1 \\[6pt]
  \small $^{*}$Core contributors \qquad $^{\dagger}$Corresponding author: \texttt{yutianshu@cuhk.edu.cn}
}

\begin{document}

\input{body}

\newpage
\input{checklist.tex}

\end{document}

%% file: preamble.tex
\usepackage[utf8]{inputenc}
\usepackage[T1]{fontenc}
\usepackage{microtype}
\usepackage[hidelinks]{hyperref}
\usepackage{url}
\usepackage{booktabs}
\usepackage{tabularx}
\usepackage{array}
\usepackage{multirow}
\usepackage{makecell}
\usepackage{adjustbox}
\usepackage{amsfonts}
\usepackage{amsmath,amssymb}
\usepackage{nicefrac}
\usepackage{graphicx}
\usepackage{wrapfig}
\usepackage{needspace}
\usepackage{longtable}
\usepackage{svg}
\usepackage{subcaption}
\usepackage{float}
\usepackage{algorithm}
\usepackage{algpseudocode}
\usepackage{enumitem}
\usepackage{xcolor}
\usepackage{pifont}
\newcommand{\cmark}{\ding{51}}
\newcommand{\xmark}{\ding{55}}
\usepackage[normalem]{ulem}
\usepackage{colortbl}
\definecolor{rankfirst}{RGB}{253,213,155}
\definecolor{ranksecond}{RGB}{173,206,240}
\definecolor{rankthird}{RGB}{216,193,230}
\newcommand{\cellf}{\cellcolor{rankfirst}}
\newcommand{\cells}{\cellcolor{ranksecond}}
\newcommand{\cellt}{\cellcolor{rankthird}}
\newlength{\ulthick}
\renewcommand{\uline}[1]{%
  \leavevmode
  \setbox0=\hbox{#1}%
  \copy0\kern-\wd0
  {\color{black}\vrule height0pt depth\ulthick width\wd0}%
}
\usepackage{titletoc}

\titlecontents{lsection}[0em]{\smallskip\bfseries}{\thecontentslabel\quad}{}{~\titlerule*[6pt]{.}~\thecontentspage}
\titlecontents{lsubsection}[1.5em]{\smallskip}{\thecontentslabel\quad}{}{~\titlerule*[6pt]{.}~\thecontentspage}

\newcolumntype{Y}{>{\raggedright\arraybackslash}X}
\providecommand{\met}[1]{\scriptsize #1}
\providecommand{\nrhdr}{\makecell{\textsc{NR}\\\met{$\downarrow$}}}

\newcommand{\score}[4]{\makecell[c]{$#1 \pm #2$\\$#3 \pm #4$}}
\newcommand{\NA}{\textsc{na}}
\newcommand{\std}[1]{{\scriptsize\textcolor{gray}{$\pm$#1}}}
\newcommand{\benchmark}{\textsc{TRL-Bench}}
\newcommand{\na}{\textcolor{gray}{--}}
\newcommand{\eg}{\emph{e.g.}}
\newcommand{\ie}{\emph{i.e.}}

\setlist[itemize]{leftmargin=1.4em, topsep=2pt, itemsep=2pt}
\setlist[enumerate]{leftmargin=1.6em, topsep=2pt, itemsep=2pt}
\graphicspath{{figures/}}

\title{\benchmark: Standardizing Cross-Paradigm Representation-Level Evaluation of Tabular Encoders}

%% file: body.tex
\maketitle

\begin{abstract}
Tabular encoders are usually evaluated inside task-specific end-to-end pipelines, so models from different training paradigms are difficult to compare directly even when they operate on similar tabular signals. We introduce \benchmark{}, a multi-granular \emph{tabular representation learning} (TRL) benchmark that standardizes cross-paradigm representation-level evaluation: each encoder exports row-, column-, or table embeddings through its supported wrapper, and shared lightweight heads probe them across three suites: \textsc{TRL-CTbench} (column/table), \textsc{TRL-Rbench} (row), and \textsc{TRL-DLTE} (compositional Data-Lake Table Enrichment spanning all three granularities). To support this standardized setting, we release curated benchmark assets and task reformulations, including 50 OpenML tables with 123 verified targets, 16 row-pair linkage rewrites, and a 47,772-table DLTE lake derived from 1,379 parent tables. Across 20 models and 16 tasks, \benchmark\ shows that once downstream conditions are standardized, encoder quality is capability-specific rather than captured by a single leaderboard. In \textsc{TRL-CTbench}, generic text encoders often lead on tasks with strong surface-text signal, while tabular specialists win where their pretraining objective aligns with the task. In \textsc{TRL-Rbench}, within-table prediction and cross-table linkage favor different training regimes, with atomic linkage performance correlating strongly with the row-matching stage of DLTE pipelines. In \textsc{TRL-DLTE}, the strongest pipelines combine capability-matched specialists rather than reuse a single encoder, and top end-to-end quality depends on non-additive compositional fit rather than per-stage marginal rank alone. \benchmark\ provides a common protocol for measuring reusable signal in exported tabular representations under shared downstream conditions.\availability
\end{abstract}

\input{sections/introduction}
\input{sections/related_work}
\input{sections/benchmark_design}
\input{sections/experiments}
\input{sections/conclusion}

\begin{ack}
This project was supported by Shanghai Municipal Science and Technology Major Project under grant 2025SHZDZX025D05. The authors declare no competing interests.
\end{ack}

\clearpage
\bibliographystyle{plainnat}
\bibliography{references}

\input{sections/appendix}

%% file: sections/introduction.tex
\vspace{-4pt}
\section{Introduction}

Tables have long been recognized as the fundamental data structures for storing structured data, and there has been considerable work on using them across a wide range of analytical workloads. Recent work has produced strong row-, column-, and table-level encoders for reasoning over tabular data. Many of these are useful as reusable components: tables can be encoded once and their embeddings indexed and reused across tasks and large multi-table corpora such as data lakes, where per-task fine-tuning is often impractical~\citep{badaro2023transformers,starmie,fan2023tablediscovery}. In such encode-once, reuse-many settings, the representation itself, not the task-specific wrapper, is the object of evaluation. Yet these encoders are still mostly evaluated inside task-specific end-to-end pipelines, so models from different training paradigms are difficult to compare directly: a strong result may come from the wrapped predictor, training budget, and task-specific adaptation as much as from the encoder itself. This motivates a comparability question: \emph{under one shared evaluation protocol over the exported representations, how do heterogeneous tabular encoders actually differ?}

\begin{figure*}[t]
\centering
\makebox[\textwidth][c]{%
\begin{minipage}{\textwidth}
\centering
\captionsetup{skip=3pt}
\includegraphics[width=\linewidth,trim=4pt 4pt 4pt 4pt,clip]{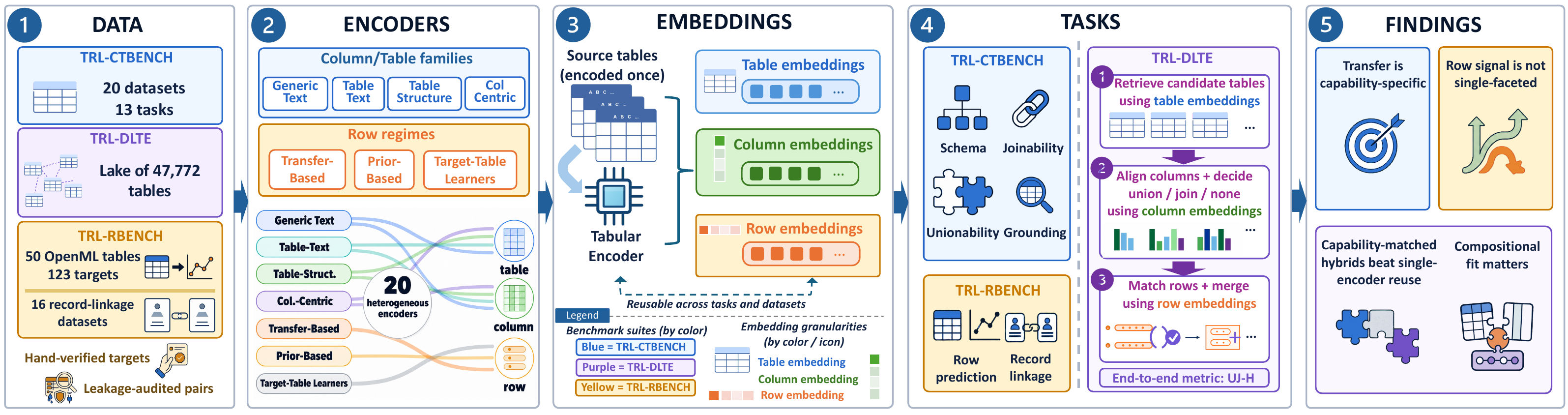}
\caption{\textbf{TRL-Bench at a glance.} Each model is processed once through its supported wrapper to export row-, column-, or table embeddings, and shared lightweight modules then evaluate those embeddings across \textsc{TRL-CTbench} (schema, joinability, unionability, grounding), \textsc{TRL-Rbench} (row prediction, record linkage), and \textsc{TRL-DLTE} (multi-stage data-lake enrichment).}
\label{fig:pipeline_overview}
\end{minipage}%
}
\vspace{-8pt}
\end{figure*}

\benchmark\ is designed around that question and complements end-to-end task benchmarks by isolating reusable representation quality under shared downstream conditions (Figure~\ref{fig:pipeline_overview}). Each model is run once through its supported wrapper to export the row-, column-, or table embeddings it exposes, and shared lightweight downstream modules evaluate those embeddings across tasks rather than re-optimizing the encoder end to end. Throughout, we use ``encoder'' operationally to denote any tabular model that exposes reusable row-, column-, or table-level embeddings.

To make this evaluation comprehensive rather than task-specific, \benchmark\ treats retrieval, schema alignment, linkage, prediction, and grounding as atomic capabilities that serve as reusable building blocks for downstream tabular systems in the encode-once, reuse-many setting. The three suites measure these capabilities at the granularities where embeddings are reused: \textsc{TRL-CTbench} for column/table transfer, \textsc{TRL-Rbench} for row transfer, and \textsc{TRL-DLTE} for compositional data-lake table enrichment.

When the benchmark is used to test 20 models and 16 tasks, three empirical findings emerge. First, once downstream conditions are standardized, transfer is capability-specific: in \textsc{TRL-CTbench}, generic text encoders often lead on tasks with strong surface-text signal, while the remaining wins are better explained by pretraining--task alignment than by any single dominant encoder class. Second, row signal is not single-faceted: within-table prediction and noisy cross-table linkage separate model families by training scope. Third, compositional fit shapes pipeline quality: in \textsc{TRL-DLTE}, the best pipelines are capability-matched hybrids that consistently outperform single-encoder reuse. Per-stage marginals are informative but do not determine the top pipelines. End-to-end quality depends on how well retrieval, column alignment, and row matching compose, not on per-stage rank in isolation.

\noindent\textbf{Contributions.}
\begin{enumerate}[topsep=2pt,itemsep=2pt,parsep=0pt]
 \item \textbf{Standardized cross-paradigm protocol:} heterogeneous encoder packages export row-, column-, or table-level embeddings through their supported interfaces, and shared lightweight readouts evaluate them under common task definitions, enabling direct comparison without end-to-end fine-tuning.
 \item \textbf{Comprehensive benchmark for reusable tabular signal:} \textsc{TRL-CTbench}, \textsc{TRL-Rbench}, and \textsc{TRL-DLTE} cover column/table transfer, row transfer, and compositional enrichment over 16 tasks and 87 datasets from SATO~\citep{sato}, SOTAB~\citep{sotab}, WikiCT~\citep{turl}, Spider~\citep{spider}, Valentine~\citep{valentine}, OpenML~\citep{openml}, and DeepMatcher/WDC~\citep{deepmatcher,wdcproducts} (Appendix~\ref{app:dataset-inventory}).
 \item \textbf{Curated assets and task reformulations:} we contribute (a)~20 column/table datasets standardized for representation-level evaluation, (b)~50 OpenML-derived row-prediction tables with 123 hand-verified targets, (c)~16 record-linkage datasets rewritten as explicit row-pair matching tasks, and (d)~a 47,772-table enrichment lake built from 1,379 parent tables, together with representation-centric rewrites of heterogeneous source tasks such as WTQ~\citep{wikitablequestions} and DeepMatcher.
 \item \textbf{A cross-paradigm empirical study:} across 20 models and 16 tasks, we show that no single pretraining recipe behaves as a universal tabular representation, and we identify structural gaps in model choice, transfer scope, and pipeline composition that single-paradigm or single-granularity evaluations cannot isolate.
\end{enumerate}

%% file: sections/related_work.tex
\vspace{-6pt}
\section{Related Work and Positioning}

We situate \benchmark\ relative to both prior tabular model families and existing benchmark resources. Appendix~\ref{app:extended-related} gives additional citations and comparisons.

\noindent\textbf{Related work: model families and evaluation traditions.}
Prior tabular work is fragmented by granularity. Row-level models focus on supervised prediction and transfer \cite{tabtransformer,saint,subtab,tabbinning,vime,scarf,dae,transtab,tabpfn,tabpfnv2,tabicl,fttransformer,excelformer,tabr,carte,xtab}, while column/table models target schema semantics, grounding, retrieval, and discovery \cite{bert,gte,tapas,tabert,turl,tuta,tabbie,starmie,tabsketchfm}. These families are usually evaluated in task-specific settings rather than under a shared multi-granular representation-level protocol. Observatory~\citep{observatory} is the closest prior resource that compares frozen tabular embeddings across model families, but it measures perturbation- and invariance-style intrinsic properties such as sample fidelity and order insignificance rather than downstream task performance.

\begin{table}[t]
\centering
\footnotesize
\setlength{\tabcolsep}{3pt}
\renewcommand{\arraystretch}{0.85}
\caption{Comparison with prior tabular evaluation resources (\cmark\,=\,supported). \emph{Comp.}: tasks composed across multiple representation levels. \emph{Cross-paradigm}: multiple training paradigms under one protocol. \emph{Repr.-level eval.}: representation-level evaluation as primary intended use. \emph{Downstream transfer}: downstream task performance vs.\ intrinsic properties. \emph{Task family}: broad problem class (\eg, row prediction, semantic typing). Observatory reports intrinsic properties only, hence ``--''.}
\label{tab:related_comparison}
\begin{tabular}{lcccccccc}
\toprule
\textbf{Benchmark} & \makecell{\textbf{Col.}} & \makecell{\textbf{Table}} & \makecell{\textbf{Row}} & \makecell{\textbf{Comp.}} & \makecell{\textbf{Cross-}\\\textbf{paradigm}} & \makecell{\textbf{Repr.-level}\\\textbf{eval.}} & \makecell{\textbf{Downstream}\\\textbf{transfer}} & \makecell{\textbf{\# Task}\\\textbf{fam.}} \\
\midrule
OpenML suites~\cite{openml,openmlcc18,openmlctr23}      & \xmark & \xmark & \cmark & \xmark & \xmark & \xmark & \cmark & 1 \\
TabArena~\cite{tabarena}           & \xmark & \xmark & \cmark & \xmark & \xmark & \xmark & \cmark & 1 \\
DeepMatcher~\cite{deepmatcher}        & \xmark & \xmark & \cmark & \xmark & \xmark & \xmark & \cmark & 1 \\
LakeBench~\cite{lakebench_srinivas,lakebench_deng}          & \cmark & \cmark & \xmark & \xmark & \xmark & \xmark & \cmark & 2 \\
SANTOS / TUS~\cite{tus_original,santos}       & \cmark & \cmark & \xmark & \xmark & \xmark & \xmark & \cmark & 1 \\
Valentine~\cite{valentine}          & \cmark & \xmark & \xmark & \xmark & \xmark & \xmark & \cmark & 1 \\
SemTab / SOTAB~\cite{semtab,sotab}     & \cmark & \xmark & \xmark & \xmark & \xmark & \xmark & \cmark & 1 \\
Observatory~\cite{observatory}        & \cmark & \cmark & \cmark & \xmark & \cmark & \cmark & \xmark & -- \\
\midrule
\benchmark\ (ours) & \cmark & \cmark & \cmark & \cmark & \cmark & \cmark & \cmark & 7 \\
\bottomrule
\end{tabular}
\end{table}

\noindent\textbf{Positioning of \benchmark.}
Prior benchmarks each target narrow task scopes (Table~\ref{tab:related_comparison}) \cite{openml,openmlcc18,openmlctr23,tabarena,deepmatcher,wdcproducts,semtab,sotab,valentine,tus_original,santos,lakebench_srinivas,lakebench_deng} and usually compare models within a single task family or end-to-end pipeline rather than under a shared representation-level protocol. \benchmark\ complements these resources by standardizing heterogeneous tabular encoders into a shared representation-level evaluation protocol. Its main distinctions are: (i) multi-granular evaluation across columns, rows, and tables, (ii) direct cross-paradigm comparison under common task definitions and lightweight downstream heads, and (iii) a compositional DLTE benchmark testing whether strong atomic capabilities compose into an end-to-end pipeline. Because this standardized comparison operates on exported representations, it applies to models that expose reusable row-, column-, or table-level embeddings, either natively or via a natural extraction point in the architecture. Generative table LLMs~\cite{tablellama,tablegpt2} generally do not provide such an interface, while heavily task-specific fine-tuned systems~\cite{ditto,doduo,omnitab} are formulated as end-to-end predictors rather than reusable representations.

%% file: sections/benchmark_design.tex
\vspace{-4pt}
\section{Benchmark Design}
\label{sec:benchmark}

\benchmark\ asks a comparability question: once heterogeneous tabular encoders are evaluated under one shared representation-level protocol, how do they differ across rows, columns, and tables, and how do they compose end-to-end? The benchmark has three suites, \textsc{TRL-CTbench}, \textsc{TRL-Rbench}, and \textsc{TRL-DLTE}. Figure~\ref{fig:pipeline_overview} gives the high-level view.

\suppressfloats[t]
\begin{figure*}[t]
\centering
\includegraphics[width=\textwidth]{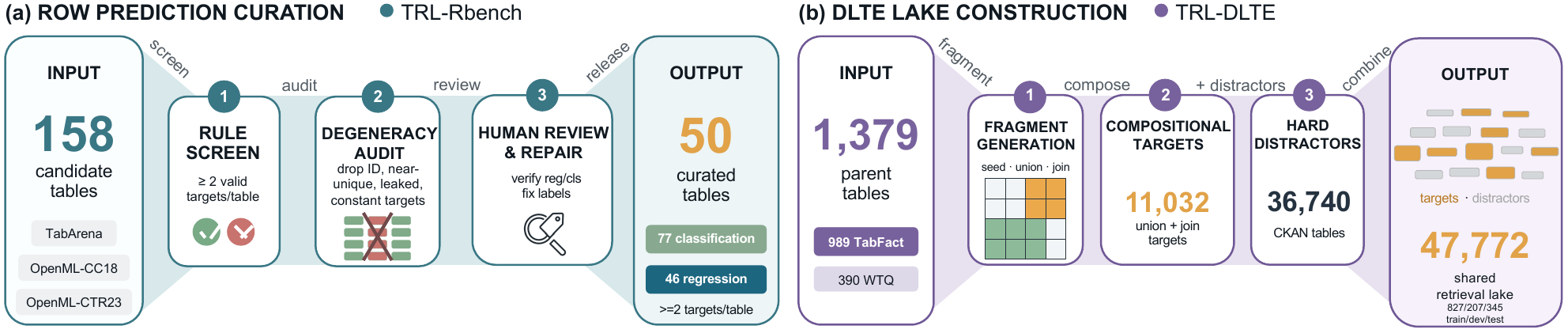}
\caption{Curation of \textsc{TRL-Rbench} row-prediction tables and assembly of the \textsc{TRL-DLTE} lake. (a)~Row prediction curation: 158 candidate tables filtered through rule screening, degeneracy audit, and human review with label repair into 50 tables with 123 targets. (b)~DLTE lake assembly: 1,379 TabFact/WTQ parents fragmented into seed queries and union/join targets at four noise tiers. 11,032 targets are embedded alongside 36,740 CKAN distractors in a 47,772-table lake.}
\label{fig:construction}
\end{figure*}

\subsection{Problem Setting and Standardized Representation-Level Protocol}
\label{sec:benchmark-protocol}

We call a tabular representation good when one exported embedding meets two established criteria: simple readouts can recover task information from it (\emph{recoverability}, the probing tradition~\citep{alain2016probes}), and it serves many downstream tasks (\emph{transferability}~\citep{bengio2013representation}). \benchmark\ therefore standardizes evaluation at the level of exported representations and does not force a single input serialization. For a table $T$ with columns $C(T)=(c_1,\ldots,c_M)$ and rows $R(T)=(r_1,\ldots,r_N)$, a tabular encoder $f_\theta$ may expose column, row, or table representations,
\[
E^{\mathrm{col}}(T)=(e^{\mathrm{col}}_1,\ldots,e^{\mathrm{col}}_M),\qquad
E^{\mathrm{row}}(T)=(e^{\mathrm{row}}_1,\ldots,e^{\mathrm{row}}_N),\qquad
e^{\mathrm{tbl}}(T).
\]
For each task, write $\mathbf{e}$ for the relevant exported encoder output(s) (a column, row, table, or pair thereof). A \emph{downstream module} $r$ then maps these embeddings to the task output, and \benchmark\ uses three downstream-module types. \emph{Training-free modules} $r(\mathbf{e})$ have no task-specific learned parameters and operate directly on embedding geometry, e.g., cosine ranking for schema matching or union search, or $k$-means for column clustering. \emph{Learned modules} $r_\psi(\mathbf{e})$ are lightweight supervised probes trained on exported encoder embeddings, e.g., column type prediction, join/union classification, join search, row prediction, and record linkage. \emph{Query-conditioned modules} $r_\psi(q,\mathbf{e})$ additionally consume a frozen text-query embedding $q=f_{\mathrm{text}}(\text{query})$, e.g., a dual-projection head for table retrieval or a decoder for table QA. Operationalizing \emph{recoverability}, all downstream heads are intentionally lightweight and held fixed across encoder families, so comparisons reflect the choice of exported embedding under common readouts rather than the choice of downstream predictor. \emph{Transferability} is then tested by reusing each exported embedding across the multiple tasks within its suite, and, for models that expose multiple granularities, across suites as well (Sec.~\ref{sec:ctbench}--\ref{sec:dlte}).

The learned-module category follows a unified supervised-probe protocol. For every supervised probe task, we train both a linear head and a one-hidden-layer MLP (hidden size 256, robust to head size and depth, Appendix Table~\ref{tab:row-head-sweep}) on exported encoder embeddings with Adam~\cite{adam} under standardized settings (Appendix~\ref{app:reproducibility}). We use the arithmetic average of the two as the canonical score: the linear head tests linearly accessible signal, and the MLP tests whether a small nonlinear readout can recover it. Per-head diagnostics, including a cosine reference for record linkage, are reported in Appendices~\ref{app:ablation_head_col} and~\ref{app:ablation_rl_head}. For table-level tasks, if a model supports multiple table aggregations (e.g., \texttt{cls}, \texttt{col-mean}, \texttt{tok-mean}), we apply this protocol to each aggregation and report the one selected on development data (per-model, full ablations in Appendix~\ref{app:aggregation_ablation}). Training-free tasks (Column Clustering, Union Search, Schema Matching), query-conditioned tasks (Table QA, Table Retrieval), and the DLTE pipeline (Sec.~\ref{sec:dlte}) follow their corresponding module types defined above.

Pairwise tasks use fixed combination operators, typically concatenation. Appendix Tables~\ref{tab:appendix-col-table-tasks}--\ref{tab:appendix-row-dlte-tasks} give per-task details. A model is evaluated only when the required granularity is natively exposed or obtainable by supported pooling. Appendix~\ref{app:model-inventory} summarizes supported granularities, and Appendix~\ref{app:input-pipelines} states the benchmark's wrapper policy. Each model runs under the interface its authors designate, with the exceptions listed in Appendix~\ref{app:input-pipelines}, so \benchmark\ is a comparison of encoder packages under their supported interfaces.

\subsection{\textsc{TRL-CTbench}: Column- and Table-Level Transfer}
\label{sec:ctbench}

\textsc{TRL-CTbench} contains 13 tasks: 8 column-level and 5 table-level, grouped into schema understanding, joinability, unionability, and grounding.

\noindent\textbf{Schema understanding.}
These tasks consume a single column embedding or an ordered intra-table column pair and test whether exported column representations expose semantic type and intra-table structure. They include column type prediction, column clustering, and column relation prediction.

\noindent\textbf{Joinability.}
Joinability asks whether two tables are \emph{complementary}, \ie, one can add attributes to the other through overlapping columns. Tasks are join search, column overlap, and table-level join classification. Since raw cosine between column embeddings does not directly model directional value containment (\ie, whether one column's values are contained in another's), the main join-search setting uses a minimal learned projection head on top of the exported embeddings.

\noindent\textbf{Unionability.}
Unionability asks whether two tables are \emph{stackable} after schema alignment. Tasks are union search, schema matching, union classification, union regression, and table subset. Union search uses SANTOS~\cite{santos}, UGEN~\cite{ugen}, and TUS~\cite{tus_original}, with an added low-overlap TUS-hard variant (Appendix~\ref{app:tus_hard}) that removes positive pairs whose directed column containment is $\ge 0.70$, creating an explicit contrast between the original TUS setting and one without high value-containment positives.

\noindent\textbf{Grounding.}
Grounding tests whether a representation can ground a natural-language query in structured table content. These tasks are \emph{query-conditioned}: a frozen text encoder embeds the question, and a lightweight head combines that query embedding with table-side representations exported by the evaluated model. Table QA uses question and column embeddings with a lightweight decoder, while table retrieval trains a dual projection head over query and table embeddings.

\noindent\textbf{Task adaptation and curation.}
Most source tasks were not originally formulated for standardized
representation-level evaluation. We therefore standardize table
identifiers, align label schemas, rewrite end-to-end datasets into
representation-centric variants, and clean dataset layouts where
needed. Across the benchmark, this rewriting is concrete: for
WikiTableQuestions~\cite{wikitablequestions}, we replace joint encoder
fine-tuning with exported column/question embeddings plus a
lightweight decoder. For the four
pairwise tasks whose original splits exhibit table-level overlap (join
classification, column overlap, union classification, and union
regression), we enforce table-disjoint train/dev/test splits
that prevent any test table from appearing during training. The remaining supervised
table-pair task (table subset) already has table-disjoint splits in
the source data. Appendix Tables~\ref{tab:appendix-col-table-tasks}--\ref{tab:appendix-row-dlte-tasks}
list per-task evaluation modes, split types, and metrics. Appendix~\ref{app:hygiene}
summarizes the benchmark's shortcut and leakage mitigations. Appendices~\ref{app:curation-rationale} and~\ref{app:licenses} give per-dataset curation rationales and licenses.

\subsection{\textsc{TRL-Rbench}: Row-Level Transfer}
\label{sec:rbench}

\textsc{TRL-Rbench} asks whether exported row embeddings transfer both within a table and across tables. It contains row prediction and record linkage.

\noindent\textbf{Row prediction.}
For row prediction, the encoder sees only the observed columns $X$, produces one target-agnostic embedding per row, and that same embedding is reused to predict each curated target column $y_k \in Y = \{y_1, \ldots, y_K\}$ ($K \geq 2$) with a lightweight probe under the protocol of Sec.~\ref{sec:benchmark-protocol}. This asks whether a single row embedding can be reused across multiple targets from the same table, including tables with mixed classification and regression targets. The suite contains 50 OpenML-derived tables with 123 curated targets (77 classification, 46 regression), filtered from 158 candidates from TabArena~\cite{tabarena}, OpenML-CC18~\cite{openmlcc18}, and OpenML-CTR23~\cite{openmlctr23}. Every released table has 2--3 targets (mean 2.46). Human curators selected target columns, repaired label issues where needed, verified classification-versus-regression typing, and removed degenerate or leaked targets such as constant columns, near-duplicate targets, and label columns recoverable from the input (Figure~\ref{fig:construction}). All target columns are excluded from encoder input.

\noindent\textbf{Record linkage.}
Record linkage complements intra-table prediction with inter-table matching: given a pair of rows from two tables, predict whether they refer to the same entity. We adopt 16 datasets from two entity-matching benchmark families~\cite{deepmatcher,wdcproducts,wdcproducts_lspm}: 8 clean DeepMatcher~\cite{deepmatcher} benchmarks, 4 dirty DeepMatcher variants with synthetic schema noise, and 4 size variants of the WDC Products Large-Scale Product Matching (LSPM) benchmark. For analysis, we split these into \emph{Clean Linkage} (the 8 clean DeepMatcher datasets) and \emph{Robust Linkage} (the 4 dirty DeepMatcher + 4 WDC datasets). We rewrite these sources as explicit row tables with labeled row pairs, retain the original source pair-disjoint splits, which are the canonical evaluation protocol in the entity-matching literature, and feed paired exported row embeddings via concatenation to a lightweight supervised probe under the protocol of Sec.~\ref{sec:benchmark-protocol}. Appendix~\ref{app:linkage_split_audit} reports per-source train/test pair and row-overlap statistics, and documents the removal of label-equivalent columns before any encoder consumes a row.
\vspace{-4pt}
\subsection{\textsc{TRL-DLTE}: Multi-Stage Data Lake Table Enrichment}
\label{sec:dlte}

Atomic tasks test local transfer, but not whether row-, column-, and table-level representations compose into a full multi-stage pipeline. \textsc{TRL-DLTE} addresses this gap. We start from a complete \emph{parent table}, the ground-truth table to be reconstructed. From it, we remove a block of rows and a block of columns. The remaining subtable is the \emph{seed query}. The removed rows form the \emph{union target} (same schema as the seed, additional rows), and the removed columns form the \emph{join target} (same rows as the seed, additional attributes). Given only the seed and a data lake, the system must recover both targets by retrieving relevant tables, deciding whether each candidate contributes by union, join, or neither, aligning columns, matching rows, and merging the result.

We build \textsc{TRL-DLTE} from filtered parent tables drawn from TabFact~\cite{tabfact} and WTQ~\cite{wikitablequestions}. Figure~\ref{fig:construction}(b) summarizes the construction and counts: each parent is fragmented at four cumulative noise tiers (clean, schema, cell, hard) into a seed query, a union target, and a join target, and the targets are inserted into a shared retrieval lake together with CKAN distractors. Seeds serve only as queries and are not lake members. Parent tables are split before fragmentation so that train/dev/test remain parent-disjoint.

Evaluation has three stages. Stage~1 retrieves candidate tables using \emph{table} embeddings. Stage~2 aligns columns and predicts union/join/none using \emph{column} embeddings. Stage~3 matches rows and merges content using \emph{row} embeddings. Pipelines can use a single multi-granular model or combine different specialists across stages, making DLTE a composition test for the benchmark as a whole. Full stage-wise operator specifications, including the Stage-2 threshold calibration procedure and the dev-based selection of the headline pipeline reported in Sec.~\ref{sec:exp-dlte}, are in Appendix~\ref{app:dlte-operators}.

We introduce $\mathrm{UJ\text{-}H}$ as the primary end-to-end score, which summarizes recovery of both the union and join targets. Let $R_{\text{union}}$ be the fraction of removed-row-block cells recovered in the seed columns, and $R_{\text{join}}$ the fraction of removed-column-block cells recovered for the seed rows. $\mathrm{UJ\text{-}H}$ is the per-query harmonic mean of these two recalls, averaged over queries (zero when both recalls vanish):
\begingroup
\setlength{\abovedisplayskip}{4pt}
\setlength{\belowdisplayskip}{4pt}
\setlength{\abovedisplayshortskip}{2pt}
\setlength{\belowdisplayshortskip}{4pt}
\[
  \mathrm{UJ\text{-}H} =
  \frac{2\,R_{\text{union}}\,R_{\text{join}}}
       {R_{\text{union}} + R_{\text{join}}}
\]
\endgroup
penalizing pipelines that succeed on only one enrichment path. We additionally report Cell~$F_1$ in Appendix~\ref{app:cellf1}. It is a multiset $F_1$ over recovered cells pooled across the removed-row and removed-column blocks, used as a complementary diagnostic of pooled cell-recovery yield.

%% file: sections/experiments.tex
\vspace{-8pt}
\section{Experiments}
\label{sec:experiments}
\vspace{-4pt}
We now use the standardized representation-level protocol of Sec.~\ref{sec:benchmark-protocol} to compare heterogeneous tabular encoders across the three benchmark suites introduced in Sec.~\ref{sec:benchmark}: \textsc{TRL-CTbench}, \textsc{TRL-Rbench}, and \textsc{TRL-DLTE}.
\vspace{-6pt}
\subsection{Experimental Setup}
\label{sec:exp-setup}

We follow the standardized representation-level protocol in Sec.~\ref{sec:benchmark-protocol}. Here we summarize only the choices needed to read the result tables.

\noindent\textbf{Compared models.}
We evaluate 20 models spanning generic text encoders, table-aware and structure-aware encoders, column-specialized models, target-table self-supervised learners, and meta-pretrained priors. Appendix Table~\ref{tab:full-models} lists the full inventory, and Appendix~\ref{app:openai_ablation} reports an embedding ablation.

\noindent\textbf{Baselines and controls.}
Each task reports the applicable simple non-neural or analytical baselines alongside learned encoders (e.g., TF-IDF for column-level tasks, value overlap for search). ``Best$^{\ast}$'' in Table~\ref{tab:main_results} denotes the strongest applicable non-neural baseline, with markers $a$--$d$ identifying which one. Full specifications and 5-seed per-dataset results are in Appendix~\ref{app:baselines}.

\noindent\textbf{Reporting and Metrics.}
CTBench reports raw metrics plus per-family normalized-rank (NR, lower is better) aggregates. For table-level tasks, the main comparison uses the supported aggregation selected on development data (Appendix~\ref{app:aggregation_ablation}). Row prediction averages over 123 targets (77 classification and 46 regression, with \textsc{TabTransformer} covering 63 due to its categorical-feature requirement), with linkage split into \emph{Clean Linkage} (DM-C) and \emph{Robust Linkage} (DM-D~+~WDC NR aggregate). DLTE uses $\mathrm{UJ\text{-}H}$ as the primary end-to-end score. To avoid selection bias from reporting the maximum over 1{,}120 test evaluations, headline DLTE pipelines are selected on the development split by $\mathrm{UJ\text{-}H}$ within the relevant candidate set and evaluated once on test. Top-50 frequencies, per-stage marginals, and Oracle-RA (Stages~1--2 replaced by ground truth) are test-set descriptive analyses over the full pipeline space. Stage-2 thresholds are calibrated on development with macro-$F_1$, independently of headline pipeline selection (Appendix~\ref{app:dlte-operators}). NR averages a model's normalized rank over units of an aggregate (tasks within a CTBench family, target columns, or linkage datasets), excluding missing units. Formulas and standard metric definitions are in Appendix~\ref{app:metrics}. Appendices~\ref{app:table-footprint}, \ref{app:family-summary}, \ref{app:robustness}, and~\ref{app:counting} cover table footprints, a family-level summary figure, Observatory-style robustness checks, and dataset counting. Significance tests (Friedman/Wilcoxon) are in Appendix~\ref{app:statistical} and embedding-generation costs in Appendix~\ref{app:efficiency}. Summed over all jobs, running the full benchmark (all suites, five seeds) takes an estimated 1{,}200 L40S GPU-hours, mostly for embedding generation, plus about 700 hours of CPU-only jobs for downstream evaluation and DLTE matching. In all tables, $\downarrow$ marks lower-is-better. The colors \colorbox{rankfirst}{\strut\textbf{orange}}/\colorbox{ranksecond}{\strut\uline{blue}}/\colorbox{rankthird}{light purple} highlight ranks 1/2/3 per column, and $\dagger$ marks table-disjoint splits.

\subsection{Column- and Table-Level Results}
\label{sec:exp-col-table}

Table~\ref{tab:main_results} reports the main column/table results across 13 \textsc{TRL-CTbench} tasks, evaluated on the 10 of 20 models that natively expose column or table embeddings, spanning schema understanding, joinability, unionability, and grounding. Two empirical patterns stand out.

\input{tables/column_table_level.tex}

\noindent\textbf{Generic-text rankings track surface-text signal.}
Family-level NR (lower is better) for \textsc{BERT} and \textsc{GTE} is lowest on Schema or Join and highest on Grounding (\textsc{BERT} $0.000/0.344/0.260/0.397$ and \textsc{GTE} $0.190/0.095/0.343/0.429$ for Schema/Join/Union/Grounding). A task-level surface-text audit (Appendix~\ref{app:ctbench_lexical}) is consistent with this interpretation: generic text encoders are strongest on tasks where headers and short cell strings carry most of the signal, and the CTBench tasks where a tabular specialist beats them (Union Search, Schema Matching, Table Subset, and Table QA) all sit in the Union and Grounding families, which instead reward cross-table alignment geometry and grounded table understanding.

\noindent\textbf{Pretraining alignment matters beyond surface cues.}
The four CTBench tasks won by tabular specialists are each consistent with their winner's pretraining design. Column-centric \textsc{Starmie}'s contrastive objective matches the cosine-scoring setup for Union Search (0.662 MAP) and Schema Matching (0.764 R@GT). On Table Subset, the top three are all tabular specialists (\textsc{TAPAS} 0.567 $F_1$, \textsc{TAPEX} 0.557, \textsc{TABBIE} 0.546), placing structural pretraining above text serialization on this task. In Grounding, the Table-Text family leads at the family level: \textsc{TaBERT}'s joint text-table pretraining delivers the best Grounding NR (0.198), while task-level wins split between structure-aware \textsc{TURL} (Table QA, 0.277 Acc) and generic-text \textsc{GTE} (Table Retrieval, 0.476 MRR). Although \textsc{GTE} is classified as a generic text model, it is pretrained with a retrieval-contrastive objective~\cite{gte}. Its Table Retrieval win is itself an instance of pretraining-task alignment rather than an exception.
\vspace{-4pt}
\subsection{Row-Level Results}
\label{sec:exp-row}

Table~\ref{tab:r_main_results} reports row-level transfer under the shared probe protocol of Sec.~\ref{sec:benchmark-protocol}. Three empirical patterns stand out.

\input{tables/row_level.tex}

\noindent\textbf{Prediction and linkage decouple by model family.}
Prior-based \textsc{TabICL} leads prediction (AUROC 0.816, Macro-$F_1$ 0.671, SGM 0.505), while linkage leaders are dominated by transfer-oriented encoders: \textsc{BERT} on Clean Linkage ($F_1$ 0.418, NR 0.096) and \textsc{GTE} on Robust Linkage (NR 0.048), with \textsc{TransTab} second on Robust Linkage (NR 0.096) via its cross-table contrastive objective. Target-table SSL methods are generally competitive on prediction but weak on linkage.

\noindent\textbf{Intra- and inter-table transfer follows training scope.}
Row prediction operates within a table (intra-table transfer), while record linkage operates across tables (inter-table transfer). Target-table SSL methods, trained from scratch on each target, fit locally: they are competitive on prediction (mean NR 0.48/0.47 on classification/regression) but trail on linkage (0.65/0.65 on Clean/Robust). Transfer-based encoders, applying one shared model to every table, produce comparable row spaces: they lead linkage (mean NR 0.19/0.19) but sit mid-pack on prediction (0.50/0.64). Two designs combine both axes. \textsc{TransTab} layers a cross-table contrastive objective onto per-table SSL, taking second on Robust Linkage (NR 0.096) while staying competitive on prediction. \textsc{TabICL} layers target-table adaptation onto a shared meta-pretrained prior, leading prediction and ranking 5th of 14 on Robust Linkage (NR 0.394). Combining intra-table and inter-table transfer is thus hard but achievable through different design paths, an open direction for further study.

\noindent\textbf{Geometric diagnostics cross-validate the task-based rankings.}
Alongside the task-based protocol, we also evaluate exported row embeddings through task-free geometric diagnostics (Appendix~\ref{app:taskfree}). The two axes agree: row-linkage utility correlates with embedding anisotropy at $|\bar\rho|\approx0.80$ (Linear head, $\alpha_{\mathrm{req}}$), and regression utility correlates with effective rank at $|\bar\rho|\approx0.36$ (MLP head, RankMe$^{\star}$). See Figure~\ref{fig:rankrank_all}. Intrinsic embedding geometry thus offers a task-free lens on row-level transfer, with broadly consistent agreement across the full diagnostic family.
\vspace{-8pt}
\Needspace{12\baselineskip}
\subsection{Compositional Results on \textsc{TRL-DLTE}}
\label{sec:exp-dlte}
\input{tables/dlte_table.tex}

We analyze the full $10 \times 8 \times 14 = 1120$ pipeline space (Stage-1 table encoders~$\times$~Stage-2 column models~$\times$~Stage-3 row models, with pool composition given in Appendix~\ref{app:dlte-rankings}). Table~\ref{tab:dlte_main} and Figure~\ref{fig:dlte_scatter} summarize the resulting top-50 memberships, per-stage marginals, and full pipeline landscape. Three findings matter most.

\noindent\textbf{Capability-matched hybrids beat single-encoder reuse.}
Under this dev-selection protocol, the best hybrid \textsc{TUTA}/\textsc{GTE}/\textsc{GTE} reaches 0.229 $\mathrm{UJ\text{-}H}$, 0.090 above the best dev-selected monolithic \textsc{BERT}/\textsc{BERT}/\textsc{BERT} (0.139). Development and test rankings are similar over the 1{,}120 pipelines (Spearman $\rho = 0.96$, top-50 overlap 42/50, see Appendix~\ref{app:dlte-rankings}), so the dev-selected hybrid result is consistent with the broader test landscape. Frontier presence tracks atomic-task leadership at Stages~2--3 (Table~\ref{tab:dlte_main}): the Stage-2 frontier picks all lead at least one CTBench task (e.g., \textsc{BERT} on Schema NR, \textsc{GTE} on Join NR, Join Search, and Table Retrieval, \textsc{TURL} on Table QA), and the Stage-3 top-three frontier picks (\textsc{GTE}, \textsc{TransTab}, \textsc{TabICL}) are exactly the top three row models on Oracle-RA (Appendix~\ref{app:oracle_ra}). \textsc{GTE} and \textsc{TransTab} additionally take the top two slots on Robust Linkage NR, with \textsc{TUTA} third.

\Needspace{18\baselineskip}
\input{tables/dlte_figure.tex}
\vspace{-4pt}
\noindent\textbf{Compositional fit shapes pipeline quality.}
Atomic strength is thus necessary for frontier presence but not sufficient for the best assembly. Per-stage marginals summarize average main effects, not optimal compositions. The test Stage-1/2/3 marginal leaders assemble to \textsc{Starmie}/\textsc{TABBIE}/\textsc{TransTab} at 0.134 $\mathrm{UJ\text{-}H}$ (Fig.~\ref{fig:dlte_scatter}), well below both the test rank-1 pipeline \textsc{Starmie}/\textsc{GTE}/\textsc{GTE} (0.253) and the dev-selected hybrid \textsc{TUTA}/\textsc{GTE}/\textsc{GTE} (0.229). Development marginals assemble to a different pipeline, \textsc{Starmie}/\textsc{BERT}/\textsc{TransTab}, scoring 0.231 on test (Fig.~\ref{fig:dlte_scatter}). Marginals therefore carry signal, but the 0.097 swing between the two marginal assemblies, despite this split stability, shows that top marginal ranks are a lossy selection rule. Two decouplings in Table~\ref{tab:dlte_main} explain why. At Stage~1, atomic retrieval strength decouples from compositional utility: \textsc{GTE} leads CTBench Table Retrieval (0.476 MRR, Table~\ref{tab:main_results}) and DLTE target recall@100 (0.801), yet ranks only third on the Stage-1 $\mathrm{UJ\text{-}H}$ marginal, behind \textsc{Starmie} and \textsc{TUTA}. At Stage~2, average marginal strength decouples from top-pipeline membership: \textsc{TABBIE} leads the Stage-2 marginal (unrounded) but never enters the top-50, where \textsc{GTE} and \textsc{TURL} dominate. DLTE therefore rewards compositional fit, defined here as non-additive compatibility among retrieval, column alignment, and row matching, not independent per-stage rank alone. These patterns hold separately on TabFact-only and WTQ-only parents (Spearman $\rho=0.871$ cross-source rank agreement across all 1{,}120 canonical pipelines, see Appendix~\ref{app:dlte-source-split}).

\noindent\textbf{A shared identity-resolution capability across RBench and DLTE.}
DLTE Stage~3 (row matching) is the compositional counterpart of RBench's Record Linkage task: both test whether exported row embeddings can resolve cross-table row identity under noise. Oracle-RA (Appendix~\ref{app:oracle_ra}) shows that the two settings share a strong row-model rank signal: Stage-3 row-model rankings agree with the RBench Robust Linkage NR ranking at Spearman $|\rho|=0.80$ ($p=6.3{\times}10^{-4}$). The agreement is strongest at the top two models, \textsc{GTE} and \textsc{TransTab}, with only a small middle-pack shift at \textsc{TabICL} (third on Oracle-RA versus fifth on Robust Linkage NR). This identifies a shared \emph{identity-resolution} capability of frozen row embeddings, surfaced consistently by both atomic record linkage (RBench) and compositional row matching (DLTE Stage~3). This Stage-3 row-model ranking persists across all four DLTE noise tiers (Appendix~\ref{app:dlte-rankings}).

All DLTE conclusions are scoped to controlled reconstruction with complete gold: seeds and targets are fragments of one parent table. Appendix~\ref{app:dlte_robustness} checks the per-stage rankings on independently constructed data.

\vspace{-8pt}
\subsection{Three Open Gaps Across the Suites}
\label{sec:reading-across}

Read jointly, the three suites expose three structural gaps that single-paradigm or single-granularity evaluations cannot isolate. At the level of model choice, \textsc{TRL-CTbench} (Sec.~\ref{sec:exp-col-table}) reveals a \emph{specialization gap}, where no pretraining recipe behaves like a universal representation. At the level of transfer, \textsc{TRL-Rbench} (Sec.~\ref{sec:exp-row}) reveals a \emph{transfer-scope gap}, where intra-table adaptation and cross-table comparability pull learning in different directions. At the level of deployment, \textsc{TRL-DLTE} (Sec.~\ref{sec:exp-dlte}) reveals a \emph{composition gap}, where granularity choices interact rather than stack independently. Together, these mark where tabular representation research has yet to converge on a unified account.

%% file: tables/column_table_level.tex
\begin{table*}[t]
\centering
\caption{%
 Column- and table-level results on 13 \textsc{TRL-CTbench} tasks spanning four capability families.
 Join, Union, and Grounding are evaluated at both column and table granularity
 (superscripts ${}^{\text{c}}$/${}^{\text{t}}$).
 \textsc{NR} reports the mean normalized rank within each family.
 Dashes indicate unsupported granularities. Superscripts give each model's adaptation track: F frozen transfer, S target-table SSL. Formula of metrics in Appendix~\ref{app:metrics}.
}
\label{tab:main_results}
\setlength{\ulthick}{3.5pt}%
\resizebox{\textwidth}{!}{%
\begin{tabular}{ll ccc|c | ccc|c | ccccc|c | cc|c}
\toprule

& &
\multicolumn{4}{c|}{\makecell{\textbf{Schema}}} &
\multicolumn{4}{c|}{\makecell{\textbf{Join}}} &
\multicolumn{6}{c|}{\makecell{\textbf{Union}}} &
\multicolumn{3}{c}{\makecell{\textbf{Grounding}}} \\

\cmidrule(lr){3-6}
\cmidrule(lr){7-10}
\cmidrule(lr){11-16}
\cmidrule(lr){17-19}

\textbf{Family} & \textbf{Model} &
 \makecell{Col\\[-2pt]Type$^{\text{c}}$\\\met{$F_1\uparrow$}} &
 \makecell{Col\\[-2pt]Clust$^{\text{c}}$\\\met{NMI$\uparrow$}} &
 \makecell{Col\\[-2pt]Rel$^{\text{c}}$\\\met{$F_1\uparrow$}} &
 \nrhdr &
 \makecell{Join\\[-2pt]Search$^{\text{c}}$\\\met{MAP$\uparrow$}} &
 \makecell{Col\\[-2pt]Overlap$^{\text{c}\dagger}$\\\met{nRMSE$\downarrow$}} &
 \makecell{Join\\[-2pt]Class.$^{\text{t}\dagger}$\\\met{$F_1\uparrow$}} &
 \nrhdr &
 \makecell{Union\\[-2pt]Search$^{\text{c}}$\\\met{MAP$\uparrow$}} &
 \makecell{Schema\\[-2pt]Match$^{\text{c}}$\\\met{R@GT$\uparrow$}} &
 \makecell{Union\\[-2pt]Class.$^{\text{t}\dagger}$\\\met{$F_1\uparrow$}} &
 \makecell{Union\\[-2pt]Reg.$^{\text{t}\dagger}$\\\met{nRMSE$\downarrow$}} &
 \makecell{Tbl\\[-2pt]Subset$^{\text{t}}$\\\met{$F_1\uparrow$}} &
 \nrhdr &
 \makecell{Tbl\\[-2pt]QA$^{\text{c}}$\\\met{Acc$\uparrow$}} &
 \makecell{Tbl\\[-2pt]Ret.$^{\text{t}}$\\\met{MRR$\uparrow$}} &
 \nrhdr \\

\midrule

Baseline
 & Best$^{\ast}$
 & 0.813$^{a}$ & 0.400$^{a}$ & 0.015 & ---
 & 0.155$^{b}$ & 1.014 & 0.516 & ---
 & 0.574$^{c}$ & 0.473$^{d}$ & 0.500 & 1.138 & 0.458 & ---
 & 0.204 & 0.131 & --- \\[-3pt]
 &
 & \std{.007} & \std{.001} & \std{.001} &
 & \std{.000} & \std{.003} & \std{.014} &
 & \std{.000} & \std{.000} & \std{.004} & \std{.024} & \std{.027} &
 & \std{.004} & \std{.120} & \\

\midrule

\multirow{4}{*}{Generic Text}
 & BERT$^{\mathrm{F}}$
 & \cellf$\textbf{0.926}$ & \cellf$\textbf{0.516}$ & \cellf$\textbf{0.826}$ & \cellf\textbf{0.000}
 & \cells\uline{$0.434$} & \cellf$\textbf{0.786}$ & $0.496$ & 0.344
 & $0.564$ & \cellt$0.423$ & \cellf$\textbf{0.857}$ & \cellf$\textbf{0.595}$ & $0.545$ & \cells\uline{0.260}
 & $0.255$ & \cellt$0.367$ & 0.397 \\[-3pt]
 &
 & \cellf\std{.001} & \cellf\std{.000} & \cellf\std{.002} & \cellf
 & \cells\std{.000} & \cellf\std{.001} & \std{.020} &
 & \std{.000} & \cellt\std{.000} & \cellf\std{.002} & \cellf\std{.005} & \std{.005} & \cells
 & \std{.004} & \cellt\std{.008} & \\
 & GTE$^{\mathrm{F}}$
 & \cells\uline{$0.922$} & \cellt$0.466$ & \cells\uline{$0.811$} & \cells\uline{0.190}
 & \cellf$\textbf{0.469}$ & \cellt$0.817$ & \cellf$\textbf{0.535}$ & \cellf\textbf{0.095}
 & \cellt$0.608$ & $0.417$ & $0.839$ & \cells\uline{$0.600$} & $0.544$ & \cellt0.343
 & $0.245$ & \cellf$\textbf{0.476}$ & 0.429 \\[-3pt]
 &
 & \cells\std{.002} & \cellt\std{.001} & \cells\std{.002} & \cells
 & \cellf\std{.002} & \cellt\std{.002} & \cellf\std{.019} & \cellf
 & \cellt\std{.000} & \std{.000} & \std{.001} & \cells\std{.009} & \std{.002} & \cellt
 & \std{.002} & \cellf\std{.003} & \\
\midrule

\multicolumn{19}{l}{\textit{Tabular-Pretrained}} \\

\multirow{6}{*}{\quad Table-Text}
 & TaBERT$^{\mathrm{F}}$
 & $0.874$ & \cells\uline{$0.514$} & $0.760$ & \cellt0.381
 & \cellt$0.406$ & $0.855$ & $0.498$ & 0.593
 & \cells\uline{$0.619$} & \cells\uline{$0.430$} & $0.760$ & $0.615$ & $0.540$ & 0.435
 & \cellt$0.267$ & \cells\uline{$0.372$} & \cellf\textbf{0.198} \\[-3pt]
 &
 & \std{.003} & \cells\std{.003} & \std{.002} & \cellt
 & \cellt\std{.002} & \std{.002} & \std{.034} &
 & \cells\std{.000} & \cells\std{.000} & \std{.004} & \std{.009} & \std{.004} &
 & \cellt\std{.005} & \cells\std{.013} & \cellf \\
 & TAPAS$^{\mathrm{F}}$
 & $0.868$ & $0.448$ & $0.769$ & 0.476
 & $0.320$ & $0.823$ & $0.509$ & 0.471
 & $0.529$ & $0.415$ & $0.837$ & \cellt$0.612$ & \cellf$\textbf{0.567}$ & 0.419
 & $0.254$ & $0.295$ & 0.579 \\[-3pt]
 &
 & \std{.001} & \std{.001} & \std{.003} &
 & \std{.001} & \std{.002} & \std{.015} &
 & \std{.000} & \std{.000} & \std{.004} & \cellt\std{.004} & \cellf\std{.004} &
 & \std{.003} & \std{.006} & \\
 & TAPEX$^{\mathrm{F}}$
 & --- & --- & --- & ---
 & --- & --- & $0.514$ & \cellt0.333
 & --- & --- & \cells\uline{$0.854$} & $0.622$ & \cells\uline{$0.557$} & \cellf\textbf{0.222}
 & --- & $0.332$ & \cells\uline{0.333} \\[-3pt]
 &
 &  &  &  &
 &  &  & \std{.023} & \cellt
 &  &  & \cells\std{.002} & \std{.005} & \cells\std{.004} & \cellf
 &  & \std{.005} & \cells \\
\cmidrule(lr){2-19}

\multirow{6}{*}{\quad Table-Struct.}
 & TABBIE$^{\mathrm{F}}$
 & \cellt$0.892$ & $0.262$ & \cellt$0.785$ & 0.476
 & $0.208$ & $0.862$ & \cellt$0.527$ & 0.693
 & $0.410$ & $0.205$ & $0.833$ & $0.663$ & \cellt$0.546$ & 0.705
 & \cells\uline{$0.276$} & $0.170$ & 0.516 \\[-3pt]
 &
 & \cellt\std{.004} & \std{.007} & \cellt\std{.002} &
 & \std{.001} & \std{.002} & \cellt\std{.017} &
 & \std{.000} & \std{.000} & \std{.002} & \std{.001} & \cellt\std{.004} &
 & \cells\std{.004} & \std{.004} & \\
 & TURL$^{\mathrm{F}}$
 & $0.814$ & $0.406$ & $0.758$ & 0.667
 & $0.299$ & \cells\uline{$0.809$} & \cells\uline{$0.532$} & \cells\uline{0.323}
 & $0.575$ & $0.340$ & $0.814$ & $0.657$ & $0.507$ & 0.673
 & \cellf$\textbf{0.277}$ & $0.199$ & \cellt0.389 \\[-3pt]
 &
 & \std{.001} & \std{.004} & \std{.002} &
 & \std{.000} & \cells\std{.001} & \cells\std{.014} & \cells
 & \std{.000} & \std{.001} & \std{.001} & \std{.009} & \std{.004} &
 & \cellf\std{.005} & \std{.010} & \cellt \\
 & TUTA$^{\mathrm{F}}$
 & --- & --- & --- & ---
 & --- & --- & $0.468$ & 1.000
 & --- & --- & $0.810$ & $0.652$ & $0.447$ & 0.778
 & --- & $0.235$ & 0.556 \\[-3pt]
 &
 &  &  &  &
 &  &  & \std{.010} &
 &  &  & \std{.003} & \std{.007} & \std{.008} &
 &  & \std{.013} & \\
\cmidrule(lr){2-19}

\multirow{4}{*}{\quad Col.-Centric}
 & Starmie$^{\mathrm{S}}$
 & $0.789$ & $0.404$ & $0.698$ & 0.810
 & $0.316$ & $0.847$ & $0.510$ & 0.529
 & \cellf$\textbf{0.662}$ & \cellf$\textbf{0.764}$ & \cellt$0.853$ & $0.662$ & $0.539$ & 0.356
 & $0.266$ & $0.018$ & 0.714 \\[-3pt]
 &
 & \std{.004} & \std{.000} & \std{.001} &
 & \std{.001} & \std{.001} & \std{.019} &
 & \cellf\std{.000} & \cellf\std{.000} & \cellt\std{.002} & \std{.003} & \std{.005} &
 & \std{.005} & \std{.002} & \\
 & TabSketchFM$^{\mathrm{F}}$
 & $0.566$ & $0.252$ & $0.373$ & 1.000
 & $0.265$ & $0.946$ & $0.505$ & 0.841
 & $0.531$ & $0.155$ & $0.737$ & $0.671$ & $0.545$ & 0.810
 & $0.235$ & $0.218$ & 0.833 \\[-3pt]
 &
 & \std{.001} & \std{.007} & \std{.003} &
 & \std{.001} & \std{.001} & \std{.011} &
 & \std{.000} & \std{.000} & \std{.002} & \std{.006} & \std{.007} &
 & \std{.005} & \std{.011} & \\

\midrule
\multicolumn{2}{l}{\textit{All-model mean}} & 0.831 & 0.409 & 0.723 & --- & 0.340 & 0.843 & 0.509 & --- & 0.562 & 0.394 & 0.819 & 0.635 & 0.534 & --- & 0.259 & 0.268 & --- \\
\bottomrule
\end{tabular}%
}
\vspace{2pt}
{\raggedright\footnotesize
${}^{\ast}$Best baseline per task: unmarked\,=\,Random;
${}^{a}$TF-IDF,
${}^{b,c}$value overlap,
${}^{d}$Jaccard. The \emph{All-model mean} row averages each column over the models with a value, and \textsc{TAPEX} and \textsc{TUTA} are evaluated on the 5 table-level tasks only.
Table-level results report, for each model, the supported table aggregation among \textsc{cls}, \textsc{col-mean}, and \textsc{tok-mean} selected on development data under the unified supervised-probe protocol of Sec.~\ref{sec:benchmark-protocol}; full aggregation ablations in Tables~\ref{tab:aggregation_ablation} and~\ref{tab:aggregation_ablation_linear}.\par}
\end{table*}

%% file: tables/row_level.tex
\begin{table*}[t]
\centering
\caption{%
 Row-level results across four evaluation categories.
 Row prediction averages over 77 classification and 46 regression targets (\textsc{TabTransformer}: 63 targets due to categorical-feature requirement).
 SGM is $\mathrm{SGM}_{0.01}(\mathrm{nRMSE})$.
 Linkage columns report binary $F_1$ (match class) on DM-C, DM-D, and WDC. See Appendix~\ref{app:metrics} for the convention. The \emph{All-model mean} row averages each column over the 14 models.
 Rank columns aggregate ranks over individual targets (classification, regression) or datasets (linkage). Row groups are adaptation tracks: F frozen transfer, P prior-based, S target-table SSL. Formula of metrics in Appendix~\ref{app:metrics}.
}
\label{tab:r_main_results}
\resizebox{\textwidth}{!}{%
\begin{tabular}{ll cc|c | c|c | c|c | cc|c}
\toprule

& &
\multicolumn{3}{c|}{\textbf{\large Classification}} &
\multicolumn{2}{c|}{\textbf{\large Regression}} &
\multicolumn{2}{c|}{\textbf{\large Clean Linkage}} &
\multicolumn{3}{c}{\textbf{\large Robust Linkage}} \\

\cmidrule(lr){3-5}
\cmidrule(lr){6-7}
\cmidrule(lr){8-9}
\cmidrule(lr){10-12}

\textbf{Family} & \textbf{Model} &
 \makecell{AUROC\,\met{$\uparrow$}} &
 \makecell{Macro\\[-2pt]$F_1$\,\met{$\uparrow$}} &
 \makecell{\textsc{NR}\,\met{$\downarrow$}} &
 \makecell{SGM\,\met{$\downarrow$}} &
 \makecell{\textsc{NR}\,\met{$\downarrow$}} &
 \makecell{DM-C\\[-2pt]$F_1$\,\met{$\uparrow$}} &
 \makecell{\textsc{NR}\,\met{$\downarrow$}} &
 \makecell{DM-D\\[-2pt]$F_1$\,\met{$\uparrow$}} &
 \makecell{WDC\\[-2pt]$F_1$\,\met{$\uparrow$}} &
 \makecell{\textsc{NR}\,\met{$\downarrow$}} \\

\midrule

\multirow{2}{*}{Baseline}
 & Dummy
 & $0.500$\,\std{.000} & $0.304$\,\std{.000} & {---}
 & $1.004$\,\std{.000} & {---}
 & $0.000$\,\std{.000} & {---} & $0.000$\,\std{.000} & $0.000$\,\std{.000} & {---} \\
 & Random
 & $0.506$\,\std{.001} & $0.348$\,\std{.001} & {---}
 & $1.103$\,\std{.000} & {---}
 & $0.179$\,\std{.007} & {---} & $0.223$\,\std{.003} & $0.128$\,\std{.003} & {---} \\

\midrule

\multirow{4}{*}{\makecell[l]{Transfer-\\Based (F)}}
 & BERT
 & $0.791$\,\std{.000} & $0.635$\,\std{.000} & \cellt0.378
 & $0.704$\,\std{.002} & 0.559
 & \cellf\textbf{0.418}\,\std{.004} & \cellf\textbf{0.096} & \cells\uline{0.464}\,\std{.005} & \cellt$0.236$\,\std{.003} & 0.163 \\
 & GTE
 & $0.770$\,\std{.000} & $0.610$\,\std{.001} & 0.544
 & $0.765$\,\std{.001} & 0.714
 & \cells\uline{0.392}\,\std{.006} & \cells\uline{0.173} & \cellf\textbf{0.516}\,\std{.005} & \cells\uline{0.311}\,\std{.001} & \cellf\textbf{0.048} \\
 & TABBIE
 & $0.770$\,\std{.001} & $0.599$\,\std{.001} & 0.541
 & $0.766$\,\std{.003} & 0.643
 & $0.365$\,\std{.005} & 0.250 & $0.330$\,\std{.011} & $0.140$\,\std{.009} & 0.404 \\
 & TUTA
 & $0.720$\,\std{.000} & $0.553$\,\std{.002} & 0.551
 & $0.725$\,\std{.003} & 0.632
 & \cellt$0.377$\,\std{.011} & \cellt0.231 & \cellt$0.451$\,\std{.006} & $0.227$\,\std{.005} & \cellt0.154 \\

\midrule

\multirow{2}{*}{\makecell[l]{Prior-\\Based (P)}}
 & TabICL
 & \cellf\textbf{0.816}\,\std{.001} & \cellf\textbf{0.671}\,\std{.002} & \cellf\textbf{0.164}
 & \cellf\textbf{0.505}\,\std{.001} & \cellf\textbf{0.139}
 & $0.316$\,\std{.011} & 0.423 & $0.318$\,\std{.006} & $0.147$\,\std{.006} & 0.394 \\
 & TabPFN
 & $0.793$\,\std{.001} & $0.621$\,\std{.002} & 0.492
 & $0.607$\,\std{.002} & 0.499
 & $0.254$\,\std{.008} & 0.596 & $0.251$\,\std{.023} & $0.087$\,\std{.010} & 0.663 \\

\midrule

\multirow{8}{*}{\makecell[l]{Target-Table\\Learners (S)}}
 & VIME
 & \cells\uline{0.794}\,\std{.000} & $0.640$\,\std{.001} & 0.385
 & \cells\uline{0.556}\,\std{.003} & \cells\uline{0.367}
 & $0.257$\,\std{.026} & 0.529 & $0.259$\,\std{.027} & $0.099$\,\std{.006} & 0.596 \\
 & SCARF
 & \cellt$0.794$\,\std{.001} & \cellt$0.642$\,\std{.001} & \cells\uline{0.371}
 & \cellt$0.571$\,\std{.002} & 0.399
 & $0.266$\,\std{.010} & 0.510 & $0.258$\,\std{.021} & $0.069$\,\std{.008} & 0.683 \\
 & DAE
 & $0.793$\,\std{.001} & \cells\uline{0.643}\,\std{.001} & 0.379
 & $0.575$\,\std{.002} & \cellt0.392
 & $0.241$\,\std{.013} & 0.644 & $0.252$\,\std{.022} & $0.098$\,\std{.004} & 0.615 \\
 & TabBinning
 & $0.792$\,\std{.001} & $0.640$\,\std{.001} & 0.396
 & $0.573$\,\std{.002} & 0.397
 & $0.256$\,\std{.011} & 0.615 & $0.279$\,\std{.025} & $0.068$\,\std{.004} & 0.673 \\
 & SAINT
 & $0.768$\,\std{.002} & $0.595$\,\std{.004} & 0.543
 & $0.617$\,\std{.009} & 0.561
 & $0.167$\,\std{.047} & 0.712 & $0.176$\,\std{.048} & $0.133$\,\std{.004} & 0.606 \\
 & SubTab
 & $0.731$\,\std{.001} & $0.550$\,\std{.002} & 0.798
 & $0.782$\,\std{.003} & 0.779
 & $0.094$\,\std{.005} & 0.933 & $0.121$\,\std{.005} & $0.009$\,\std{.005} & 0.962 \\
 & TabTransformer
 & $0.768$\,\std{.003} & $0.594$\,\std{.004} & 0.497
 & $0.666$\,\std{.003} & 0.447
 & $0.083$\,\std{.017} & 0.942 & $0.089$\,\std{.041} & $0.020$\,\std{.004} & 0.942 \\
 & TransTab
 & $0.778$\,\std{.001} & $0.608$\,\std{.001} & 0.477
 & $0.611$\,\std{.020} & 0.441
 & $0.339$\,\std{.009} & 0.346 & $0.423$\,\std{.032} & \cellf\textbf{0.400}\,\std{.025} & \cells\uline{0.096} \\

\midrule
\multicolumn{2}{l}{\textit{All-model mean}} & 0.777 & 0.614 & {---} & 0.645 & {---} & 0.273 & {---} & 0.299 & 0.146 & {---} \\
\bottomrule
\end{tabular}%
}
\end{table*}

%% file: tables/dlte_table.tex
\begin{wraptable}[26]{r}{0.48\textwidth}
\vspace{-22pt}
\centering
\scriptsize
\setlength{\tabcolsep}{3pt}
\renewcommand{\arraystretch}{0.9}
\caption{\textbf{Per-stage view of the DLTE pipeline space} (5-round avg, test set). Top 5 models per stage by \emph{top-50 frequency} (\#, the number of the 50 best pipelines that use the model). $UJ$-$H$ and Cell~$F_1$ are \emph{full-grid marginal means}: each model's average over every pipeline in the $1{,}120$-pipeline grid that uses it at that stage. Superscripts: adaptation track, coded as in Table~\ref{tab:r_main_results}. Bottom row lists marginal leaders.}
\label{tab:dlte_main}
\begin{tabular}{l l r r r}
\toprule
\textbf{Stage} & \textbf{Model} & \textbf{\#} & \textbf{$UJ$-$H$} & \textbf{Cell~$F_1$} \\
\midrule
\multirow{5}{*}{Stage 1 (Tbl)} & \textbf{Starmie}$^{\mathrm{S}}$ & \textbf{19} & \textbf{0.144} & \textbf{0.601} \\
 & TUTA$^{\mathrm{F}}$ & 18 & 0.138 & 0.593 \\
 & TAPEX$^{\mathrm{F}}$ & 4 & 0.127 & 0.558 \\
 & GTE$^{\mathrm{F}}$ & 3 & 0.129 & 0.591 \\
 & BERT$^{\mathrm{F}}$ & 3 & 0.128 & 0.589 \\
\midrule
\multirow{5}{*}{Stage 2 (Col)} & \textbf{TURL}$^{\mathrm{F}}$ & \textbf{16} & \textbf{0.143} & 0.597 \\
 & \textbf{GTE}$^{\mathrm{F}}$ & \textbf{16} & 0.141 & 0.601 \\
 & BERT$^{\mathrm{F}}$ & 10 & 0.135 & 0.595 \\
 & TAPAS$^{\mathrm{F}}$ & 7 & 0.132 & 0.600 \\
 & TaBERT$^{\mathrm{F}}$ & 1 & 0.128 & \textbf{0.628} \\
\midrule
\multirow{5}{*}{Stage 3 (Row)} & \textbf{TransTab}$^{\mathrm{S}}$ & \textbf{12} & \textbf{0.132} & 0.574 \\
 & \textbf{GTE}$^{\mathrm{F}}$ & \textbf{12} & 0.131 & 0.574 \\
 & \textbf{TabICL}$^{\mathrm{P}}$ & \textbf{12} & 0.130 & 0.566 \\
 & TUTA$^{\mathrm{F}}$ & 7 & 0.130 & \textbf{0.584} \\
 & BERT$^{\mathrm{F}}$ & 6 & 0.128 & 0.581 \\
\midrule
\multicolumn{5}{@{}l@{}}{\textit{Full-grid marginal-mean leaders ($UJ$-$H$, unrounded means):}} \\
\multicolumn{5}{@{}l@{}}{\textit{\textbf{Starmie} 0.144 / \textbf{TABBIE} 0.143 / \textbf{TransTab} 0.132.}} \\
\bottomrule
\end{tabular}
\vspace{-10pt}
\end{wraptable}

%% file: tables/dlte_figure.tex
\begin{wrapfigure}[20]{l}{0.49\textwidth}
\vspace{-8pt}
\centering
\includegraphics[width=\linewidth]{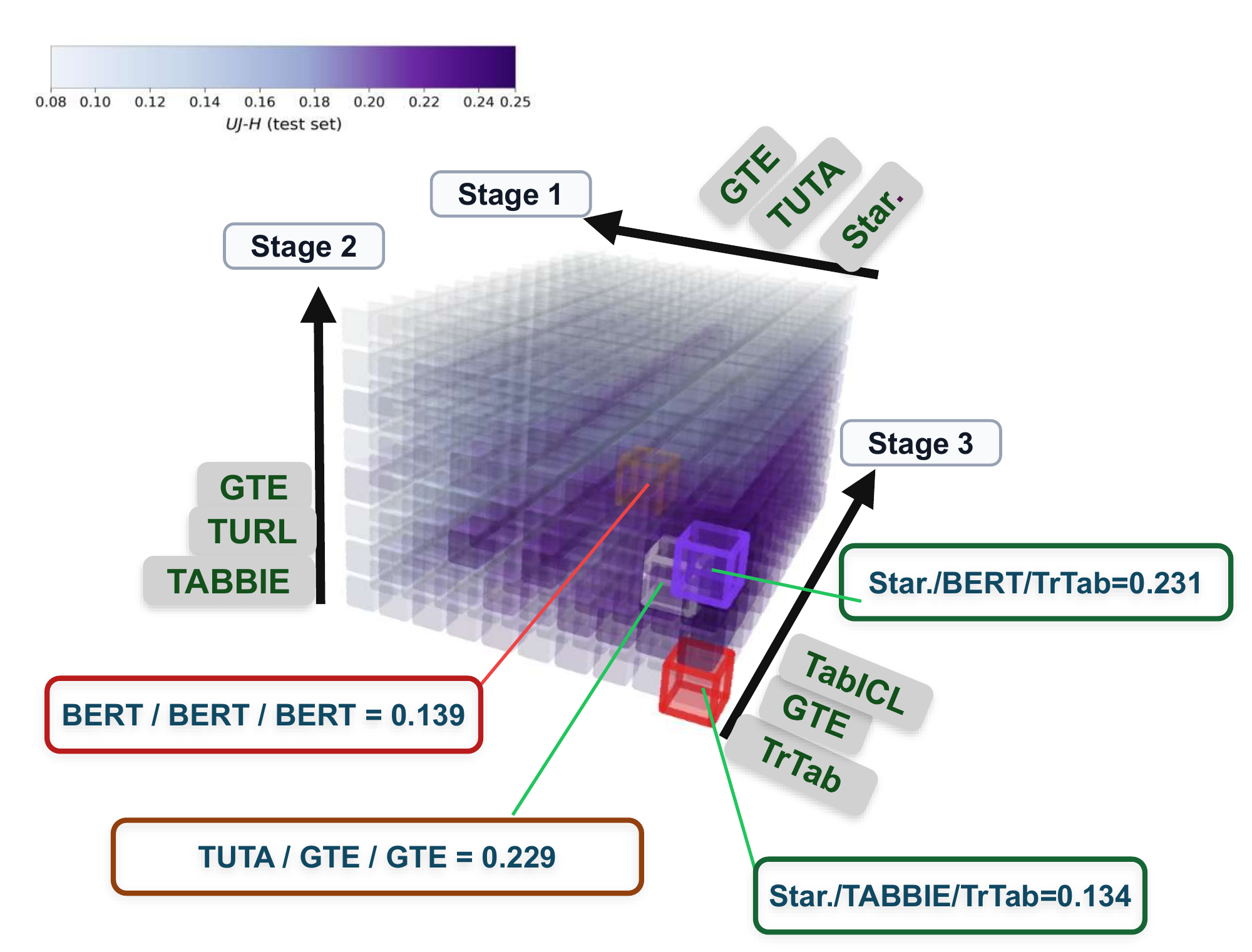}
\caption{\textbf{DLTE pipeline landscape} (test split). Axes sort by per-stage marginal $UJ\text{-}H$. Color encodes end-to-end $UJ\text{-}H$. Halo boxes mark marginal and dev-selected compositions.}
\label{fig:dlte_scatter}
\vspace{-10pt}
\end{wrapfigure}

%% file: sections/conclusion.tex
\section{Conclusion}

\benchmark\ reframes tabular encoder evaluation around the artifact many
downstream systems actually reuse: exported embeddings. Under a single
representation-level protocol, heterogeneous encoder packages become comparable
through shared readouts of their exported embeddings, with each model's
adaptation track labeled beside its scores. By making this setting measurable across
diverse encoders, datasets, and downstream uses, \benchmark\ provides a
common reference point for building tabular models as portable
representation learners rather than one-off task solvers. Appendices~\ref{sec:limitations} and~\ref{app:broader-impact} discuss limitations and broader impact.

%% file: sections/appendix.tex
\appendix

\let\TRLoldsection\section
\renewcommand{\section}{\clearpage\TRLoldsection}

\newpage
\begin{center}
{\Large\bfseries Appendix}
\end{center}
\vspace{0.5em}
\startcontents[appendices]
\printcontents[appendices]{l}{1}{\setcounter{tocdepth}{2}}
\newpage

\providecommand{\std}{}
\renewcommand{\std}[1]{{\tiny $\pm$ #1}}
\providecommand{\score}{}
\renewcommand{\score}[4]{$#1$\,\std{#2} / $#3$\,\std{#4}}

\section{Limitations}
\label{sec:limitations}

\benchmark\ is designed to standardize cross-paradigm comparison at the representation level, not to report the best end-to-end system for each task. Its scores therefore answer a narrower question, namely what common lightweight readouts can extract from exported embeddings, rather than replacing fully adapted-system benchmarks. The protocol standardizes task definitions and downstream evaluation, not raw model preprocessing: each encoder is run in its documented operating regime through its standard supported wrapper rather than a single forced serialization, which improves fidelity to the original models but leaves some wrapper-induced variation. Appendix~\ref{app:input-pipelines} summarizes this policy, and the exact wrapper settings are documented in the released code. Beyond the table-aggregation and query-encoder ablations (Appendices~\ref{app:aggregation_ablation} and~\ref{app:ablation_qe}), we do not measure how scores change under alternative wrappers. By design, every encoder is read out by the same linear and MLP heads with fixed settings, untuned per model or per task. This keeps the comparison fair, and scores characterize representation quality under one common readout rather than each encoder's best achievable downstream number. Finally, dataset counts and normalized-rank summaries are convenience aggregates and should be read alongside the per-task results.

\section{Extended Related Work and Scope}
\label{app:extended-related}

This section provides the fuller narrative context omitted from the compressed main-text related-work discussion. Table~\ref{tab:related_comparison} in the main paper remains the compact structural summary.

\paragraph{Row-level tabular learning and evaluation.}
A large body of work in tabular deep learning focuses on row-level predictive modeling, using feature-token architectures, denoising losses, contrastive self-supervision, cross-table transfer, or prior-based/meta-pretrained predictors \cite{tabtransformer,saint,subtab,tabbinning,vime,scarf,dae,transtab,tabpfn,tabpfnv2,tabicl}. See \citet{borisov2022} and \citet{shwartzziv2022} for surveys. Recent strong backbones such as FT-Transformer, ExcelFormer, TabR, CARTE, and XTab further strengthen this row-level tradition \cite{fttransformer,excelformer,tabr,carte,xtab}. Because these methods are usually presented as end-to-end row predictors rather than as encoders exposing reusable row embeddings under a shared multi-granular representation-level protocol, we treat them as important row-level references rather than as direct baselines for \benchmark{}'s encode-once comparison. These methods have substantially improved supervised tabular prediction, but their evaluation is still centered on row-level predictive benchmarks, which do not test whether one exported row representation can be reused across multiple targets from the same table or transferred to inter-table tasks such as record linkage.

\paragraph{Column- and table-centric representation learning.}
A separate line of work studies schema semantics, table-language grounding, retrieval, and discovery. Generic text encoders can be applied to serialized tables \cite{bert,gte}, while table-aware and column-aware models \cite{tapas,tabert,tapex,turl,tuta,tabbie,starmie,tabsketchfm} inject structural or contrastive inductive bias through pretraining. Their evaluations are typically task-specific (semantic typing, relation prediction, question answering, table retrieval, or data discovery) rather than unified transfer across columns, rows, and tables. Across both row-level and column/table-level work, evaluation remains largely single-granularity and task-specific, which makes cross-granularity transfer hard to assess. A notable exception is Observatory~\citep{observatory}, which characterizes learned representations along perturbation- and invariance-style properties such as sample fidelity, perturbation robustness, and order insignificance, but does not relate those measurements to downstream transfer on the same encoder outputs. Our robustness appendix directly applies Observatory's three diagnostics to the models in our benchmark pool, and Appendix~\ref{app:taskfree} additionally pairs downstream row-level evaluation with intrinsic embedding-geometry diagnostics (spectral spread, spectral shape, spatial structure) from the broader representation-learning literature on the same exported row embeddings used by the benchmark protocol, cross-validating the two views. The main benchmark suites themselves are organized around downstream task transfer under a shared representation-level protocol.

\paragraph{Task-specific benchmarks and broader evaluation suites.}
The tabular ecosystem already contains strong benchmarks for semantic typing and schema understanding \cite{semtab,sato,sotab}, schema matching \cite{valentine}, join and union discovery \cite{santos,lakebench_srinivas,lakebench_deng,starmie}, entity matching \cite{deepmatcher,wdcproducts}, table question answering \cite{wikitablequestions}, question-to-table retrieval \cite{nqtables}, and row prediction \cite{openml,openmlcc18,openmlctr23,tabarena}. LakeBench~\citep{lakebench_srinivas} is the closest prior resource, covering both join and union discovery with systematic model comparisons, but it does not evaluate row-level transfer, record linkage, or multi-stage enrichment composition, and it does not separate training regimes. TabArena~\citep{tabarena} provides carefully curated row-level benchmarking with living maintenance, but evaluates supervised end-to-end prediction rather than frozen-representation transfer. These resources are indispensable ingredients for \benchmark, but they differ substantially in split design, task formulation, supervision, and permitted adaptation, and they generally do not combine representation-level downstream evaluation with explicit separation of training regimes under a common protocol. Prior work on data lakes and table augmentation, including discovery systems such as D3L~\citep{d3l} and Juneau~\citep{juneau}, typically evaluates retrieval, join recommendation, or schema alignment stages in isolation \cite{santos,lakebench_srinivas,lakebench_deng,starmie}. \textsc{TRL-DLTE} instead benchmarks their composition under a common representation-centric protocol.

\paragraph{Scope and model selection.}
\benchmark\ complements these model papers and task-specific benchmarks rather than replacing them. Rather than reproducing each task family's strongest end-to-end stack, it standardizes comparison around transferable signal already present in a representation. Accordingly, the main leaderboard follows one eligibility rule: a model is eligible when it exposes reusable embeddings at one or more granularities (row, column, or table) through a documented interface, at the $\sim$0.5M--500M parameter scale studied in this paper. Within this band, the pool takes representatives of each model family (Table~\ref{tab:full-models}). End-to-end predictors without a stable embedding interface appear only as unranked reference points, never as leaderboard rows (e.g., the raw-feature gradient-boosted trees of Appendix~\ref{app:baselines}). End-to-end generative table LLMs and heavily task-specific fine-tuned systems (e.g., TableLlama, TableGPT2, Ditto, DODUO, OmniTab) are complementary but out of scope for this comparison~\cite{tablellama,tablegpt2,ditto,doduo,omnitab}: the former typically operate at $7$B+ scale and do not expose compatible stable multi-granular embeddings, while the latter belong to a separate task-specific end-to-end paradigm. The benchmark differs from prior work in three ways. First, it is explicitly \emph{multi-granular} across columns, rows, and tables. Second, it is explicitly \emph{cross-paradigm}, comparing externally pretrained transfer, target-table self-supervision, and prior-based predictors under one interface. Third, it includes a \emph{compositional data-lake enrichment benchmark} that tests whether strong atomic representations actually compose into a useful end-to-end system.

\paragraph{Other tabular ML threads.}
Tabular data is studied across many research threads in addition to representation learning. Tabular synthesis spans GANs, diffusion models, language models, and relational generators~\cite{ctgan,tabddpm,tabsyn,great_tab,sdv,clavaddpm}, with privacy-preserving variants targeting differential-privacy guarantees~\cite{pategan,privbayes}. Other threads include AutoML and hyperparameter-optimization stacks for tabular prediction~\cite{autogluon}, causal inference and treatment-effect estimation on tabular covariates~\cite{dml}, anomaly and outlier detection~\cite{iforest}, and missing-value imputation~\cite{gain}. Although diverse in goal, several of these threads increasingly intersect with representation learning, whether through learned latent spaces in generative models or embedding-based scoring in anomaly and outlier detection, making the quality of reusable encoders a cross-cutting concern that \benchmark{} is designed to characterize.

\paragraph{Recent tabular-embedding benchmarks.}
Three recent resources benchmark exported tabular embeddings. TEmBed~\citep{tembed} evaluates eight frozen or in-context embedders on six single-stage tasks at multiple levels, \citet{hoppe2025} compare task-agnostic row embeddings on outlier detection and supervised prediction, and \citet{younes2026} evaluate table-level embeddings on intrinsic consistency, discriminability, and robustness. Their readouts differ from ours. TEmBed scores prediction with an XGBoost~\citep{xgboost} head, and Hoppe et al.\ use KNN and MLP heads with tuned hyperparameters, whereas \benchmark\ averages a linear and a one-hidden-layer MLP probe, selects table aggregations and headline pipelines on development data, and reports five seeds. Hoppe et al.\ draw their supervised tasks from TabArena-Lite~\citep{tabarena}, which contains 35 of the 50 \textsc{TRL-Rbench} tables, so the two benchmarks share data but differ in readout. TEmBed also scores cell-level embeddings. A cell embedding can be computed from the cell text alone and largely measures value semantics rather than table structure, whereas the structure-aware encoders we study (e.g., \textsc{TURL}, \textsc{TABBIE}, \textsc{TUTA}) contextualize cell-level representations with table structure to form the row, column, and table representations evaluated here, which is why cell-level evaluation lies outside the benchmark's scope.

\section{Model Inventory}
\label{app:model-inventory}

\begin{table}[H]
 \centering
 \footnotesize
 \setlength{\tabcolsep}{5pt}
 \renewcommand{\arraystretch}{1.15}
 \caption{Model inventory in \benchmark. Parameter counts come from the loaded checkpoint (frozen encoders), the official Python package (TabICL, TabPFN), or the trained model under the default training configuration (target-table SSL, per-model architectures in Appendix~\ref{app:input-pipelines}). $^\ddagger$Retrained per dataset, reported at the median column count ($31$, targets included) across the $50$ \textsc{TRL-Rbench} OpenML tables. \emph{Checkpoint}: Hugging Face model id or upstream release file of the weights behind the reported results (Appendix~\ref{app:reproducibility}). $^\S$Package versions: \texttt{tabpfn} 6.4.1 with \texttt{tabpfn-extensions} 0.2.2, and \texttt{tabicl} 0.1.4. The \textsc{TabPFN} checkpoint is TabPFN-2.5~\cite{tabpfn25}. $^\P$No upstream checkpoint is released. The weights come from our run of the upstream pretraining code. \emph{Per dataset}: trained on each target dataset, so no shared checkpoint exists (\textsc{Starmie} starts from \texttt{roberta-base}).}
 \label{tab:full-models}
 \resizebox{\textwidth}{!}{%
 \begin{tabular}{l l l l l c c l}
 \toprule
 \textbf{Model} & \textbf{Granularity} & \textbf{Family} & \textbf{Adaptation} & \textbf{Pretraining source} & \textbf{Dim} & \textbf{Params} & \textbf{Checkpoint} \\
 \midrule
 BERT & Column, Row, Table & Generic text encoder & Frozen & General text & 768 & 110M & {\scriptsize\texttt{bert-base-uncased}} \\
 GTE & Column, Row, Table & Generic text encoder & Frozen & General text & 768 & 110M & {\scriptsize\texttt{thenlper/gte-base}} \\
 TabSketchFM & Column, Table & Column-specialized & Frozen & Tables only & 768 & 135M & {\scriptsize\makecell[l]{\texttt{epoch=10-step=}\\\texttt{27786.ckpt}$^\P$}} \\
 TAPAS & Column, Table & Table-Text & Frozen & Tables + NL & 768 & 111M & {\scriptsize\texttt{google/tapas-base}} \\
 TaBERT & Column, Table & Table-Text & Frozen & Tables + NL & 768 & 181M & {\scriptsize\texttt{tabert\_base\_k3}} \\
 TABBIE & Column, Row, Table & Structure-aware & Frozen & Tables only & 768 & 170M & {\scriptsize\texttt{mix.tar.gz}} \\
 Starmie & Column, Table & Column-specialized & Target-table SSL & Target-table unlabeled data (contrastive) & 768 & 125M & {\scriptsize\emph{per dataset}} \\
 TURL & Column, Table & Structure-aware & Frozen & Entity-linked tables & 312 & 314M & {\scriptsize\makecell[l]{\texttt{pretrained/}\\\texttt{pytorch\_model.bin}}} \\
 TUTA & Table, Row & Structure-aware & Frozen & Tables only & 768 & 134M & {\scriptsize\texttt{tuta.bin}} \\
 TAPEX & Table & Table-Text & Frozen & Tables + SQL execution & 768 & 139M & {\scriptsize\texttt{microsoft/tapex-base}} \\
 TabICL$^\S$ & Row & Prior-based & Meta-pretrained & Tabular tasks/examples & 512 & 27M & {\scriptsize\makecell[l]{\texttt{tabicl-classifier-}\\\texttt{v1.1-0506.ckpt}}} \\
 TabPFN$^\S$ & Row & Prior-based & Meta-pretrained & Synthetic priors & 192 & 11M & {\scriptsize\makecell[l]{\texttt{tabpfn-v2.5-classifier-}\\\texttt{v2.5\_default.ckpt}}} \\
 SCARF & Row & Target-table SSL & Target-table SSL & Target-table unlabeled data & 512 & 0.7M$^\ddagger$ & {\scriptsize\emph{per dataset}} \\
 SAINT & Row & Target-table SSL & Target-table SSL & Target-table unlabeled data & 512 & 38M & {\scriptsize\emph{per dataset}} \\
 SubTab & Row & Target-table SSL & Target-table SSL & Target-table unlabeled data & 512 & 0.8M$^\ddagger$ & {\scriptsize\emph{per dataset}} \\
 TabTransformer & Row & Target-table SSL & Target-table SSL & Target-table unlabeled data & 512 & 0.7M$^\ddagger$ & {\scriptsize\emph{per dataset}} \\
 TabBinning & Row & Target-table SSL & Target-table SSL & Target-table unlabeled data & 512 & 0.7M$^\ddagger$ & {\scriptsize\emph{per dataset}} \\
 TransTab & Row & Target-table SSL & Target-table SSL & Target-table unlabeled data & 512 & 19M & {\scriptsize\emph{per dataset}} \\
 VIME & Row & Target-table SSL & Target-table SSL & Target-table unlabeled data & 512 & 0.7M$^\ddagger$ & {\scriptsize\emph{per dataset}} \\
 DAE & Row & Target-table SSL & Target-table SSL & Target-table unlabeled data & 512 & 0.6M$^\ddagger$ & {\scriptsize\emph{per dataset}} \\
 \bottomrule
 \end{tabular}%
 }
\end{table}

\section{Model Input Policy}
\label{app:input-pipelines}

Tabular encoders expose different native interfaces, for example serialized text, structured cells, or direct feature tensors, and therefore come with model-specific preprocessing and context regimes. In \benchmark, each encoder is evaluated under its standard public operating regime: the preprocessing and context configuration described in its source paper or released with its official code, using the paper-default or released-default settings whenever available. When several supported settings are possible, we choose the most permissive deterministic setting that stays within the model's documented operating regime and preserves benchmark coverage. We deliberately do not force a single common input serialization or context length, because doing so would push most models outside the regime in which they were designed and validated, and would distort rather than equalize the comparison. The benchmark should therefore be read as a comparison of encoder packages under their supported interfaces, with the per-model launch scripts (exact tokenization, truncation, and configuration) provided in the codebase.

Of the 20 models, 7 target-table SSL methods (\textsc{VIME}, \textsc{SCARF}, \textsc{SubTab}, \textsc{TabBinning}, \textsc{TabTransformer}, \textsc{SAINT}, \textsc{DAE}) are implemented through the TabularS3L framework \cite{tabulars3l}, and 2 generic text encoders (\textsc{BERT}, \textsc{GTE}) are applied to serialized tabular input. These 7 methods are included to cover the target-table SSL regime (\emph{Adaptation} column of Table~\ref{tab:full-models}), not as state-of-the-art supervised tabular predictors. Wrappers also introduce model-specific choices that we note here. \textsc{TABBIE} and \textsc{TUTA} row embeddings are synthesized via per-row mini-tables (no published row-extraction head exists in either model). \textsc{TabPFN} is run with \texttt{ignore\_pretraining\_limits=True} and \textsc{TURL} with \texttt{max\_entities=12000}, both library-supported settings that admit inputs beyond the models' pretrained sizes. \textsc{DAE}'s wrapper uses TabularS3L's tabular Swap corruption \cite{vime}; see also \cite{dae} for the original image-domain DAE. Per-model citations are in Table~\ref{tab:wrapper_sources}. Per-model launch scripts (exact tokenization, truncation, output extraction) live in the released codebase.

For target-table SSL methods, the Table~\ref{tab:full-models} parameter counts are measured from the trained model under the default training configuration: hidden size $512$ with $3$ hidden layers for the MLP-based learners; \textsc{SAINT} uses a $6$-layer $d{=}512$ feature-token transformer; \textsc{TransTab} uses a $2$-layer $d{=}512$ encoder whose size is dominated by a $\sim$$16$M shared token-embedding table. Because these models are retrained per dataset, the backbone grows approximately linearly with input dimension (column counts, targets included, range from $7$ to $1{,}777$ across the $50$ \textsc{TRL-Rbench} OpenML tables), yielding an empirical upper bound of roughly $1.6$--$2.0$M parameters for the largest-feature dataset.

\begin{table*}[t]
\centering
\small
\setlength{\tabcolsep}{6pt}
\caption{Per-model source provenance. The \emph{Source} column lists the model's source paper and (where relevant) the implementation library. Wrapper scripts (exact tokenization, truncation, output extraction) live in \texttt{models/<model>/} in the released codebase. Wrapper-introduced choices are documented in the policy paragraph above.}
\label{tab:wrapper_sources}
\begin{tabularx}{\textwidth}{l l X}
\toprule
\textbf{Model} & \textbf{Family} & \textbf{Source} \\
\midrule
BERT           & Generic text     & \cite{bert} \\
GTE            & Generic text     & \cite{gte} \\
\midrule
TAPAS          & Table-Text       & \cite{tapas} \\
TaBERT         & Table-Text       & \cite{tabert} \\
TAPEX          & Table-Text       & \cite{tapex} \\
\midrule
Starmie        & Column-spec.     & \cite{starmie} \\
TabSketchFM    & Column-spec.     & \cite{tabsketchfm} \\
\midrule
TABBIE         & Struct.-aware    & \cite{tabbie} \\
TURL           & Struct.-aware    & \cite{turl} \\
TUTA           & Struct.-aware    & \cite{tuta} \\
\midrule
TabICL         & Prior-based      & \cite{tabicl} \\
TabPFN         & Prior-based      & \cite{tabpfn,tabpfnv2,tabpfn25} \\
\midrule
TransTab       & Target-table SSL & \cite{transtab} \\
VIME           & Target-table SSL & \cite{vime}; \cite{tabulars3l} \\
SCARF          & Target-table SSL & \cite{scarf}; \cite{tabulars3l} \\
SubTab         & Target-table SSL & \cite{subtab}; \cite{tabulars3l} \\
TabBinning     & Target-table SSL & \cite{tabbinning}; \cite{tabulars3l} \\
TabTransformer & Target-table SSL & \cite{tabtransformer}; \cite{tabulars3l} \\
SAINT          & Target-table SSL & \cite{saint}; \cite{tabulars3l} \\
DAE            & Target-table SSL & \cite{dae}; \cite{vime}; \cite{tabulars3l} \\
\bottomrule
\end{tabularx}
\end{table*}

\section{Appendix Task Summary}
\label{app:task-summary}

\begin{table*}[t]
 \centering
 \scriptsize
 \setlength{\tabcolsep}{3pt}
 \caption{Summary of column- and table-level tasks in \benchmark. \emph{Module form} is the fine-grained downstream-module instantiation. These map to the three downstream-module types of Sec.~\ref{sec:benchmark-protocol}: Geometry $\rightarrow$ training-free, Probe and Learned proj.\ $\rightarrow$ learned, and Dual proj.\ and Decoder $\rightarrow$ query-conditioned. DLTE's Pipeline module (Table~\ref{tab:appendix-row-dlte-tasks}) is a stage-wise composition specified in Sec.~\ref{sec:dlte}. \emph{Split}: $\dagger$ = table-disjoint (re-split by this project), orig.\ = table-disjoint in source data, -- = no split (training-free task).}
 \label{tab:appendix-col-table-tasks}
 \begin{tabularx}{\textwidth}{l l l l X l l}
 \toprule
 \textbf{Task} & \textbf{Level} & \textbf{Module form} & \textbf{Split} & \textbf{Protocol summary} & \textbf{Metric} & \textbf{Sources} \\
 \midrule
 Column Type Pred. & Column & Probe & Original & Frozen column embeddings with linear / MLP probe & $F_1$ & SATO, SOTAB \\
 Column Clustering & Column & Geometry & -- & Training-free clustering on frozen column embeddings & NMI & SATO, SOTAB \\
 Column Relation Pred. & Column & Probe & Original & Ordered column-pair probe over concatenated frozen embeddings & $F_1$ & WikiCT (relation) \\
 Join Search & Column & Learned proj. & Query-disjoint & Retrieval over frozen column embeddings with a small learned projection & MAP & OpenData variants \\
 Column Overlap & Column & Probe & Table-disjoint$^\dagger$ & Frozen column-pair regression probe & nRMSE & Wiki Containment \\
 Union Search & Column & Geometry & -- & Retrieval under one-to-one column alignment over frozen embeddings & MAP & SANTOS, UGEN, TUS \\
 Schema Matching & Column & Geometry & -- & Training-free ranking of cross-table column pairs by cosine similarity & R@GT & Valentine \\
 Table QA & Column & Decoder & Original & Table QA with frozen column representations on the table side & Accuracy & WTQ \\
 Join Classification & Table & Probe & Table-disjoint$^\dagger$ & Frozen table-pair probe with lightweight head & $F_1$ & Spider Join \\
 Union Classification & Table & Probe & Table-disjoint$^\dagger$ & Frozen table-pair probe with lightweight head & $F_1$ & Wiki Union \\
 Union Regression & Table & Probe & Table-disjoint$^\dagger$ & Frozen table-pair regression probe & nRMSE & ECB Union \\
 Table Subset & Table & Probe & Tbl-disjoint (orig.) & Frozen table-pair probe with lightweight head & $F_1$ & CKAN Subset \\
 Table Retrieval & Table & Dual proj. & Original & Dual-projection question-to-table retrieval over frozen table embeddings & MRR & NQ-Tables \\
 \bottomrule
 \end{tabularx}
\end{table*}

\begin{table*}[t]
 \centering
 \scriptsize
 \setlength{\tabcolsep}{3pt}
 \caption{Summary of row-level and compositional tasks in \benchmark. See Table~\ref{tab:appendix-col-table-tasks} caption for column definitions.}
 \label{tab:appendix-row-dlte-tasks}
 \begin{tabularx}{\textwidth}{l l l l X l l}
 \toprule
 \textbf{Task} & \textbf{Level} & \textbf{Module form} & \textbf{Split} & \textbf{Protocol summary} & \textbf{Metric} & \textbf{Sources} \\
 \midrule
 Row Prediction & Row & Probe & Fixed 80/10/10 & One frozen row embedding per record, reused across multiple targets from the same table & \makecell[l]{Macro-$F_1$,\\AUROC, SGM} & 50 OpenML tables \\
 Record Linkage & Row & Probe & Original (source) & Pair classification over concatenated frozen row embeddings; headline averages linear and MLP probe heads & $F_1$ & DeepMatcher, WDC \\
 DLTE & \makecell[l]{Tbl + Col\\+ Row} & Pipeline & Parent-disjoint & Three-stage retrieval $\rightarrow$ alignment $\rightarrow$ merge over frozen encoder outputs; operators in Appendix~\ref{app:dlte-operators} & $\mathrm{UJ\text{-}H}$, Cell~$F_1$ & TabFact, WTQ \\
 \bottomrule
 \end{tabularx}
\end{table*}

\begin{table*}[t]
 \centering
 \scriptsize
 \setlength{\tabcolsep}{4pt}
 \caption{Grouping of the 16 benchmark tasks by downstream-module type (Sec.~\ref{sec:benchmark-protocol}). The three primary module types are training-free, learned, and query-conditioned. DLTE is handled separately as a multi-stage pipeline (Sec.~\ref{sec:dlte}). The fine-grained \emph{Module form} entries in Tables~\ref{tab:appendix-col-table-tasks}--\ref{tab:appendix-row-dlte-tasks} refine this taxonomy.}
 \label{tab:readout_taxonomy}
 \begin{tabularx}{\textwidth}{l>{\raggedright\arraybackslash}X}
 \toprule
 \textbf{Downstream-module type} & \textbf{Tasks} \\
 \midrule
 Training-free & Column Clustering; Union Search; Schema Matching \\
 Learned & Column Type Pred.; Column Relation Pred.; Join Search; Column Overlap; Join Classification; Union Classification; Union Regression; Table Subset; Row Prediction; Record Linkage \\
 Query-conditioned & Table QA; Table Retrieval \\
 Pipeline & DLTE \\
 \bottomrule
 \end{tabularx}
\end{table*}

\subsection{Benchmark Protocol Adaptations under Frozen Multi-Granular Transfer}
\label{app:hygiene}

The protocol of Section~\ref{sec:benchmark-protocol} fixes the common evaluation infrastructure (frozen embeddings, shared lightweight readouts), but several reused source tasks need targeted protocol adaptations to remain meaningful tests of multi-granular representation transfer. Each paragraph below states one consideration, the choice we adopt in \benchmark, and the empirical evidence motivating that choice within this setting.

\paragraph{Cross-table generalization in reused pair tasks.}
\emph{Affected tasks:} Join Classification, Column Overlap, Union Classification, Union Regression.
\emph{Choice:} table-disjoint train/dev/test splits.
\emph{Why:} These source tasks were created for different objectives; for our transfer-oriented use, table-disjoint splits ensure test tables are unseen during training and lower Join Classification $F_1$ by $0.212$ on average relative to pair-random splits (Table~\ref{tab:strict_ablation}).

\paragraph{High-overlap positives in union search.}
\emph{Affected task:} Union Search (TUS).
\emph{Choice:} TUS-hard variant (containment $\geq 0.70$ removed).
\emph{Why:} In our frozen retrieval setting, removing the highest-overlap positives helps distinguish lexical overlap from broader union signal; the value-overlap baseline drops from $1.000$ to $0.008$ and rankings shift substantially (Table~\ref{tab:tus_hard}).

\paragraph{Degenerate or mislabeled targets.}
\emph{Affected task:} Row Prediction.
\emph{Choice:} rule-based data gates and stratified selection from $158$ candidate tables to the $50$ released ones, then human review, label repair, and degeneracy audits of their targets ($131 \to 123$; Appendix~\ref{app:dataset-inventory}).
\emph{Why:} Removes constant-column targets, near-duplicate targets, and labeling issues from the OpenML candidate pool before reusable row-transfer evaluation.

\paragraph{Label-equivalent columns leaking match identity.}
\emph{Affected datasets:} Record Linkage on WDC and Fodors--Zagats.
\emph{Choice:} remove \texttt{cluster\_id} and \texttt{identifiers} (WDC) and \texttt{class} (Fodors--Zagats) before any encoder consumes a row.
\emph{Why:} These columns are deterministic functions of the match label (Appendix~\ref{app:linkage_split_audit}).

\paragraph{Train/test row overlap in reused linkage sources.}
\emph{Affected task:} Record Linkage.
\emph{Choice:} retain source pair-disjoint splits, audit per-source row overlap, and report a strict row-disjoint ablation on the $10$ viable sources.
\emph{Why:} Pair-disjoint splits are the entity-matching canon; the strict ablation confirms rankings are stable across all $14$ row models (Spearman $\rho=0.94$, $p=5.6\times 10^{-7}$; Appendix~\ref{app:linkage_split_audit}).

\paragraph{End-to-end scoring across removed blocks.}
\emph{Affected task:} DLTE.
\emph{Choice:} $\mathrm{UJ\text{-}H}$ (harmonic mean of union and join recall) as the primary metric.
\emph{Why:} $\mathrm{UJ\text{-}H}$ directly tracks recovery of both removed blocks; Cell~$F_1$ (pooled cell-recovery yield) is reported as a complementary diagnostic in Appendix~\ref{app:cellf1}.

\paragraph{Retrieval difficulty in DLTE.}
\emph{Affected task:} DLTE (Stage~1).
\emph{Choice:} include $36{,}740$ CKAN distractor tables in the lake.
\emph{Why:} A large distractor pool makes retrieval a meaningful table-representation test; the lake contains $11{,}032$ targets among $47{,}772$ tables total.

\paragraph{Shared query-side signal in hybrid retrieval.}
\emph{Affected task:} Table Retrieval.
\emph{Choice:} model-only mode (no query-encoder table embedding concatenated on the table side).
\emph{Why:} Hybrid mode adds a strong common signal from the query encoder and compresses model differences into a narrow MRR band (Table~\ref{tab:hybrid_ablation}), whereas model-only better isolates table-side transfer.

\section{Full Dataset Inventory}
\label{app:dataset-inventory}

\paragraph{CTBench datasets (20).}
Schema Understanding: SATO, SOTAB, WikiCT (relation).
Joinability: OpenData (main), OpenData CAN, OpenData USA, OpenData UK/SG, Wiki Containment (\texttt{wiki\_containment}), Spider Join (\texttt{spider\_join}).
Unionability: SANTOS, UGEN-v1, UGEN-v2, TUS, TUS-hard, Valentine, Wiki Union (\texttt{wiki\_union}), ECB Union (\texttt{ecb\_union}), CKAN Subset (\texttt{ckan\_subset}).
Grounding: WTQ (WikiTableQuestions), NQ-Tables.

\begin{figure}[!h]
\centering
\includegraphics[width=0.95\linewidth]{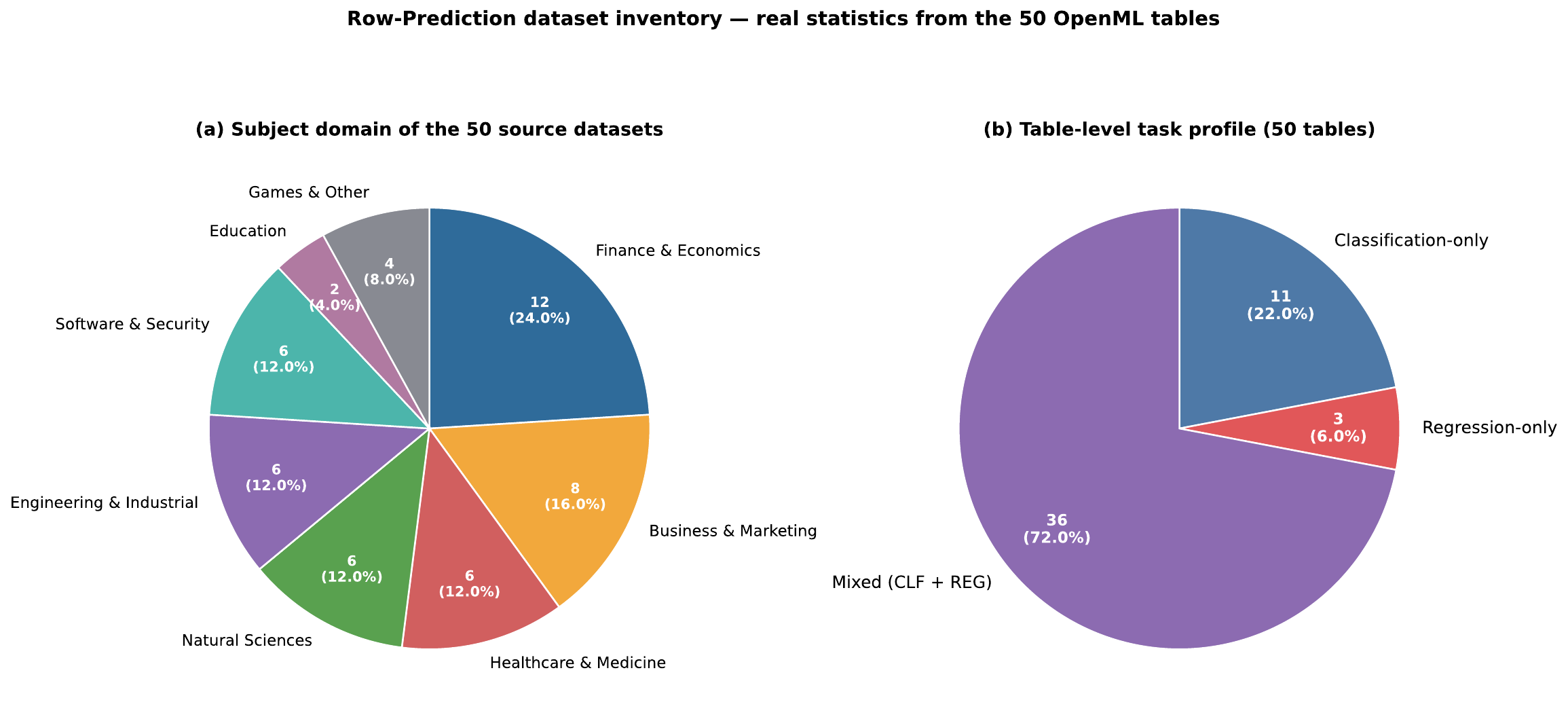}
\caption{\textbf{Row-Prediction dataset inventory.} Real statistics computed from the 50 source datasets and their per-target metadata. \textbf{(a)}~Subject-domain distribution of the 50 OpenML tables, hand-curated from each dataset's public OpenML description: 12 Finance \& Economics, 8 Business \& Marketing, 6 Healthcare \& Medicine, 6 Natural Sciences, 6 Engineering \& Industrial, 6 Software \& Security, 2 Education, and 4 Games \& Other. \textbf{(b)}~Table-level task profile of the 50 tables: 11 host only classification targets, 3 host only regression targets, and 36 host \emph{both} classification and regression targets on the same table, enabling the ``one frozen row embedding reused across multiple targets'' protocol of \benchmark.}
\label{fig:row_prediction_pie}
\end{figure}

\paragraph{Row prediction (50 OpenML tables, 123 targets).}
OpenML dataset IDs: 3, 38, 458, 1063, 1486, 4534, 6332, 40668, 40966, 40978, 44958, 44967, 44975, 44984, 44992, 46906, 46907, 46908, 46910, 46911, 46912, 46915, 46916, 46918, 46919, 46920, 46922, 46923, 46927, 46929, 46930, 46932, 46933, 46934, 46935, 46937, 46939, 46940, 46950, 46952, 46955, 46956, 46958, 46960, 46961, 46963, 46964, 46969, 46979, 46980.
Per-table target counts range from 2 to 3 (77 classification + 46 regression).
Sourced from TabArena v0.1, OpenML-CC18, and OpenML-CTR23, deduplicated by OpenML ID: 158 candidate tables. Data gates (at least 500 rows, target missingness at most 30\%, 2 to 50 classes or non-degenerate variance, and at least five input columns after target removal) leave 146. Stratified selection over size, width, and categorical-ratio bins yields the 50 released tables. Of the 81 candidate auxiliary targets entering verification, 34 came from columns outside the automatic suggestions, and the final second target overrides the first AI-assisted suggestion on 38 of the 50 tables. Staged verification rule checks, with a logged reason for every removal or repair, took 131 candidate targets to the released 123.

\paragraph{Row-prediction annotation protocol.}
Five of the authors reviewed all 50 tables together, in one review interface and against one written guideline, and saved one consensus record per table. For each table they inspected the dataset description, column names and types, missingness, cardinality, a data preview, every column that passes the hard target conditions, and the automatic suggestions, namely the rule-selected first pick for the second target and AI-assisted drafts of candidate targets $y_2$--$y_5$, which served only as editable suggestions. A column became a target only if it is a meaningful prediction object, remains predictable from the other inputs once all targets are removed, is not a duplicate or deterministic proxy of another target, and carries no post-outcome information. Each record stores the chosen targets with their task types, a rationale, the inputs expected to predict each target, and leakage and temporal flags. Programmatic checks after saving compared the records with the frozen target list. The records hold the team's consensus only, so no inter-annotator agreement is reported.

\paragraph{Target repair log.}
Table~\ref{tab:repair_log} lists every removal, repair, and addition between the 131 targets saved in the annotation step and the 123 released targets. The rules were fixed in advance. A classification target is removed when its majority class reaches 98\% or it lacks two valid classes with at least 5 rows each. A regression target needs at least 20 valid rows and non-degenerate variance. A split is repaired when a class is missing from training or appears only at test. A target is near-trivial when a linear baseline reaches accuracy $\geq 99.5\%$ or ROC AUC $\geq 99.9\%$, stably over 10 label shuffles, and is removed unless removal would leave its table with one target, with at most one such exception per table and five overall. All reported results use the final targets and splits.

\begin{table}[h]
\centering
\scriptsize
\setlength{\tabcolsep}{3pt}
\caption{Row-prediction target repair log, from 131 to 123 targets (OpenML table ID / target column).}
\label{tab:repair_log}
\begin{tabularx}{\textwidth}{l >{\raggedright\arraybackslash}p{0.36\textwidth} l X}
\toprule
\textbf{Step} & \textbf{Target} & \textbf{Action} & \textbf{Reason} \\
\midrule
131$\to$130 & 46908 / \texttt{as\_000} & Removed & Absent from the packaged data and results, near-constant \\
130$\to$126 & 46906 / \texttt{packing}, 3 / \texttt{hdchk}, 46939 / \texttt{appetency} & Removed & Majority class 98.89\%, 99.57\%, 98.21\% \\
 & 46908 / \texttt{AirPressureSystemFailure} & Removed & Majority class 98.17\% \\
126$\to$127 & 46908 / \texttt{AirPressureSystemFailure} & Restored & Original failure target (76{,}000 rows, over 1{,}000 minority rows), single-table cap of 98.5\% keeps the table at two targets \\
127$\to$123 & Tables 46939, 46952, 46958 & Re-split & Unusable class structure, new fixed seeds 51, 129, 166 \\
 & 1063 / \texttt{v(g)}, \texttt{branchCount} & Removed & Split not repairable by re-seeding; the re-seeded split (seed 2) is kept \\
 & 1063 / \texttt{loc} & Added (regression) & Keeps the table at two targets \\
 & 44992 / \texttt{GpuName}, \texttt{CpuName}, 40966 / \texttt{class} & Removed & Near-trivial \\
 & 44992 / \texttt{GpuMemoryType}, 46933 / \texttt{molecule\_structure\_property\_1564}, 458 / \texttt{Author}, 40668 / \texttt{d1}, 46964 / \texttt{wine\_color} & Kept as exceptions & Near-trivial, but removal would leave one target \\
\bottomrule
\end{tabularx}
\end{table}

\paragraph{Per-table curation.}
Table~\ref{tab:row_per_table} lists each of the 50 tables with its domain, size, released targets, and the repair-log actions that affected it.

\input{tables/row_curation_per_table.tex}

\paragraph{Record linkage (16 datasets).}
DeepMatcher clean (8): amazon-google, beer, dblp-acm, dblp-scholar, fodors-zagats, itunes-amazon, walmart-amazon, abt-buy.
DeepMatcher dirty (4): dblp-acm, dblp-scholar, itunes-amazon, walmart-amazon.
WDC Products (4 sizes): small ($\sim$2.5K pairs), medium ($\sim$8K pairs), large ($\sim$18K pairs), xlarge ($\sim$30K pairs).

\paragraph{DLTE.}
Parent tables: 1,379 (989 from TabFact, 390 from WTQ). Split: 827/207/345 train/dev/test. Fragments: 5,516 seeds + 5,516 union targets + 5,516 join targets = 16,548 total. Distractors: 36,740 CKAN tables. Total lake: 47,772 tables. Noise tiers: clean, schema, cell, hard.

\paragraph{DLTE parent-table and fragmentation parameters.}
TabFact and WTQ were chosen as two large public pools of entity-anchored web tables with the size profile the fragmentation procedure below requires; the rest of the construction is deterministic, seeded code. Parent filtering keeps tables with 30--200 rows and 5--20 columns, removes artifact columns, and requires a viable entity-key column, preferring text columns with maximal uniqueness, which yields the 1,379 parents above. Fragmentation draws the seed table as 35--50\% of the non-key columns and 60--80\% of the rows; the union target keeps the same columns with complementary rows, and the join target keeps the key column plus the remaining columns. The four cumulative noise tiers are T0 (clean), T1 (schema: column-name perturbation with $p=0.70$ plus column shuffling), T2 (cell: T1 plus 5\% non-key cell corruption plus row shuffling), and T3 (hard: T2 plus 10\% key perturbation plus 2--5 spurious rows and 1--2 spurious columns). Everything is seeded, with global seed 42.

\subsection{Per-Dataset Curation Rationale}
\label{app:curation-rationale}

Selection is criterion-driven rather than ad hoc, and processing follows one pipeline applied uniformly across \textsc{TRL-CTbench} and the record-linkage datasets of \textsc{TRL-Rbench}: acquire the source asset, normalize it into the benchmark's loader schema, apply any task-specific intervention (the Choice-line paragraphs of Appendix~\ref{app:hygiene}), and fix the splits (pair-disjoint with a per-source row-overlap audit for linkage, table-disjoint where tables recur). Three criteria drove which assets enter that pipeline: \emph{capability coverage} (each task family needs a dataset that probes it at the intended granularity), \emph{comparability} (where a subfield has a canonical benchmark, we adopt it so results stay interpretable against prior work, adding a harder variant only where our own audits showed saturation), and \emph{redistribution feasibility} (permissively licensed assets are rehosted directly; the rest ship as reconstruction scripts and manifests, Appendix~\ref{app:licenses}). Table~\ref{tab:curation_rationale} instantiates this with one rationale row per dataset (or closely related dataset group); dataset-specific processing beyond the shared four-step pipeline is pointed to its own Choice-line paragraph rather than repeated here.

\begin{table*}[t]
\centering
\scriptsize
\setlength{\tabcolsep}{4pt}
\caption{Per-dataset curation rationale (Appendix~\ref{app:table-footprint} gives per-dataset size statistics, Appendix~\ref{app:hygiene} gives the Choice-line detail this table points to). \textsc{TRL-CTbench} and record-linkage rows follow the shared four-step pipeline above, and the \emph{Why this asset} column gives each dataset's comparability or coverage rationale.}
\label{tab:curation_rationale}
\begin{tabularx}{\textwidth}{l >{\raggedright\arraybackslash}p{0.17\textwidth} >{\raggedright\arraybackslash}X >{\raggedright\arraybackslash}p{0.27\textwidth}}
\toprule
\textbf{Family} & \textbf{Dataset(s)} & \textbf{Why this asset} & \textbf{Dataset-specific step} \\
\midrule
Schema Understanding & SATO & Canonical column-typing benchmark & Std.\ 4-step pipeline \\
Schema Understanding & SOTAB & Larger, harder companion to SATO & Std.\ 4-step pipeline \\
Schema Understanding & WikiCT (relation) & Only entity-linked source with ordered column-pair relation labels & Std.\ 4-step pipeline \\
Joinability & OpenData (4 regional variants) & Canonical join-search corpora across four open-data portals & Std.\ 4-step pipeline \\
Joinability & Wiki Containment & Only source with a continuous directional value-containment target & Table-disjoint split \\
Joinability & Spider Join & Canonical cross-database join benchmark & Table-disjoint, grouped by database prefix \\
Unionability & SANTOS & Canonical union-search benchmark & Std.\ 4-step pipeline \\
Unionability & UGEN-v1, UGEN-v2 & Synthetic-ground-truth union corpora complementing SANTOS/TUS & Std.\ 4-step pipeline \\
Unionability & TUS, TUS-hard & Canonical union-search benchmark; hard variant added after our audit showed high-overlap positives saturate it & TUS-hard containment filter (Appendix~\ref{app:hygiene}) \\
Unionability & Valentine & Canonical schema-matching benchmark; most pairs are built by partitioning real source tables & Std.\ 4-step pipeline \\
Unionability & Wiki Union & Table-disjoint pairwise union-classification source & Table-disjoint split \\
Unionability & ECB Union & Only source with a continuous unionability regression target & Table-disjoint split \\
Unionability & CKAN Subset & Table-disjoint-in-source subset-detection task from open-government portals & Std.\ 4-step pipeline \\
Grounding & WTQ & Canonical table-QA benchmark & Joint fine-tuning $\to$ exported embeddings + decoder \\
Grounding & NQ-Tables & Canonical question-to-table retrieval benchmark & Question-disjoint dev carve (Appendix~\ref{app:aggregation_ablation}) \\
Row Prediction & 50 OpenML tables & TabArena/OpenML-CC18/OpenML-CTR23 pool for comparability with established row-level suites (curation funnel in \S\ref{app:dataset-inventory}) & Human review, label repair, degeneracy audit \\
Record Linkage & DeepMatcher clean (8) & Canonical clean entity-matching benchmark family & Label-equivalent-column removal on Fodors--Zagats (Appendix~\ref{app:hygiene}) \\
Record Linkage & DeepMatcher dirty (4) & Synthetic-schema-noise variants stress-testing linkage robustness & Std.\ 4-step pipeline \\
Record Linkage & WDC Products (4 sizes) & Canonical large-scale product-matching benchmark at 4 scales & \texttt{cluster\_id}/\texttt{identifiers} removal (Appendix~\ref{app:hygiene}) \\
DLTE & TabFact + WTQ parent lake & Two large public pools of entity-anchored web tables with the size profile fragmentation requires & Generation parameters above \\
\bottomrule
\end{tabularx}
\end{table*}

\section{Task-Local Baselines}
\label{app:baselines}

Each downstream task in \benchmark\ includes simple task-local baselines in addition to learned encoders. These baselines serve three distinct roles. \emph{Embedding baselines} (Random, TF-IDF) produce frozen vectors that flow through the same downstream pipeline as neural encoders. They test whether the pipeline itself drives performance rather than the embedding. \emph{Embedding-free baselines} (Inverted-Index Containment, Hungarian Set Match, Jaccard and Distribution matchers from Valentine) bypass the pipeline and operate on raw table data. They represent classical, task-specific methods and provide a reference for what is achievable without learned representations. \emph{Analytical baselines} (Chance, Dummy) compute expected performance from dataset statistics alone. Table~\ref{tab:baselines_coverage} lists per-task applicability. The baselines themselves are documented below.

\input{tables/baselines_coverage}

\paragraph{Random embeddings.}
Random vectors of the same dimension as the neural encoder replace learned embeddings throughout the downstream pipeline, at every supported granularity (column, row, table). Any benchmark entry that significantly under-performs Random indicates a failure mode. Entries that match Random indicate that the probe head, not the encoder, is doing the work.

\paragraph{Chance and Dummy.}
\emph{Chance} is an analytical floor: for classification tasks, the expected score from random class assignment proportional to class frequencies; for retrieval, the expected recall from uniform random ranking. \emph{Dummy} is the strictly stronger majority-class (classification) or mean (regression) predictor trained on frozen embeddings. It exposes cases where the label distribution alone is enough. Every supervised probe task reports a dummy head alongside the linear and MLP heads.

\paragraph{TF-IDF embeddings.}
A character-$n$-gram TF-IDF vectorizer (char$_{\mathrm{wb}}$ analyzer, range $[3,5]$, $256$ dimensions) produces a per-column embedding from the column header concatenated with up to 50 sampled cell values. Fit per dataset. Restricted to the two column-level tasks with compatible serialization (column type prediction and column clustering) on the \texttt{sato} and SOTAB datasets.

\paragraph{TF-IDF row embeddings.}
A row-level analogue of the column TF-IDF baseline, used as the string-similarity reference for record linkage. Each row is serialized as \texttt{col: val | col: val | $\ldots$} (the same template used by the GTE / BERT row encoders), then character-$n$-gram TF-IDF (char$_{\mathrm{wb}}$, range $[3,5]$, $512$ dimensions) is fit per dataset on the union of tableA and tableB rows so that paired rows live in a shared vocabulary. The resulting row vectors flow through the same downstream record-linkage probe as the neural row encoders, including all four heads (cosine threshold, linear, MLP, dummy). Because record linkage is dominated by surface character overlap on many sources (Sec.~\ref{sec:rbench}), this baseline is the appropriate floor for a learned row encoder: an encoder that does not beat TF-IDF row at matching head type is not capturing match signal beyond raw character overlap.

\paragraph{Jaccard token-overlap row embeddings.}
A second row-level non-neural baseline that complements char-TF-IDF row at the token level. Each row is serialized identically and tokenized into word unigrams. Rows are then represented as L2-normalized binary token-presence vectors (\texttt{TfidfVectorizer(analyzer='word', binary=True, use\_idf=False, norm='l2')}, $512$ dimensions). Cosine of two such vectors equals the Ochiai coefficient $|A\cap B|/\sqrt{|A|\cdot|B|}$, a token-overlap similarity that is a near-monotone transform of the token Jaccard $|A\cap B|/|A\cup B|$, so under the cosine-threshold head this baseline reads as a token-overlap threshold close to a Jaccard threshold. Under the learned MLP / linear heads the per-token presence indicators remain directly accessible to the probe. Together, char-TF-IDF row (sub-word) and Jaccard row (token-level) form the two-way string-similarity floor against which a neural row encoder must compete on linkage. On the canonical avg(MLP, linear) probe protocol, TF-IDF row reaches $F_1 = 0.380 / 0.495 / 0.227$ on DM-C / DM-D / WDC, and Jaccard row reaches $F_1 = 0.353 / 0.481 / 0.255$ on the same subsets, both well above the Random / Dummy floors (Table~\ref{tab:baselines_analytical}) and below the strongest learned encoders (Table~\ref{tab:r_main_results}).

\paragraph{Inverted-Index Containment.}
An embedding-free join-search baseline. Given a query column $Q$ and a lake column $C$, the score is the containment $|Q \cap C| / |Q|$ computed via an inverted index over normalized cell values (posting-list prune at $10{,}000$ to remove ubiquitous values). This is a strong baseline in the JOSIE~\cite{josie} lineage of work, since LakeBench-style ground truth is defined by value overlap.

\paragraph{Hungarian Set Match.}
The union-search counterpart to Inverted-Index Containment. Column-to-column containment scores are assembled into a bipartite matrix and solved with the Hungarian algorithm~\cite{hungarian}. The table-level score is the mean of matched column scores. Identical to the embedding-based union-search pipeline except cosine similarity is replaced by value containment.

\paragraph{Valentine matchers (Jaccard, Distribution).}
Two embedding-free schema-matching baselines from the Valentine library. \emph{Jaccard} scores column pairs by a weighted combination of header character 3-gram Jaccard and value-set Jaccard. \emph{Distribution} scores by comparing value distributions (KS/EMD for numerical columns, frequency for categorical). Both are evaluated on the Valentine benchmark under the same Recall@GT metric as embedding-based schema matching.

\paragraph{Cosine-threshold (record linkage).}
An unsupervised head that replaces the learned linear/MLP probe with a single cosine-similarity threshold between paired row embeddings, tuned on the validation split. It is reported as a fourth head in Table~\ref{tab:ablation_rl_head} and consolidated in Table~\ref{tab:baselines_cosine} here. It is the only baseline that specifically exercises geometry rather than supervised readout.

\paragraph{Per-dataset results.}
Tables~\ref{tab:baselines_nonneural}--\ref{tab:baselines_cosine} report 5-seed means ($\pm$ std) for every baseline on every dataset where it applies. Table~\ref{tab:baselines_nonneural} gives the embedding-free matchers. Table~\ref{tab:baselines_analytical} summarizes Random, Dummy, and TF-IDF by task-level means (probe tasks) or by classification/regression/linkage block (row-level tasks). Table~\ref{tab:baselines_cosine} gives Cosine-threshold across the 14 row models used in Table~\ref{tab:r_main_results}. Overlap with the main results tables is intentional and serves as a cross-validation check: e.g., Jaccard's Valentine R@GT matches the Best$^{*}$ marker (d) of Table~\ref{tab:main_results}, and Cosine-threshold values match the \emph{Cos} column of Table~\ref{tab:ablation_rl_head}.

\input{tables/baselines_nonneural}
\input{tables/baselines_analytical}
\input{tables/baselines_cosine}

\paragraph{Frozen reuse trades off against label-abundant per-table fitting.}
With abundant labels the encode-once constraint has a measurable cost: raw-feature gradient-boosted trees, trained per target with fixed settings, exceed the best frozen probe. On a common pool of 70 classification targets, XGBoost~\citep{xgboost} reaches AUROC 0.860 and LightGBM~\citep{lightgbm} 0.857, against 0.849 for the best probe (\textsc{TabICL}, 0.816 over all 77 classification targets in Table~\ref{tab:r_main_results}). These trees are unranked reference points, not leaderboard entries, as are the trees of the label-efficiency sweep (Table~\ref{tab:label_efficiency}). The row suite therefore evaluates what a frozen row representation carries for reuse across tasks, robust record linkage and the compositional identity resolution of \textsc{TRL-DLTE}'s Stage~3 (Sec.~\ref{sec:exp-dlte}), rather than substituting for label-abundant per-table fitting, with the prior-based \textsc{TabICL} strongest on row prediction.

\paragraph{Label-efficiency sweep.}
Table~\ref{tab:label_efficiency} compares frozen probes with raw-feature trees when only $n$ training rows per target are available, for $n$ from 32 to 2{,}048. The probes are logistic-regression heads ($C{=}1$) on frozen \textsc{BERT}, \textsc{GTE}, \textsc{TabICL}, and \textsc{TabPFN} row embeddings. The best of the four probes wins the majority of targets against both tree models up to $n{=}256$. It wins fewer than half of the targets against XGBoost from $n{=}512$ and against LightGBM from $n{=}1{,}024$. Coverage narrows at the three largest budgets (72, 63, and 52 of 77 targets) because a budget is skipped when it exceeds 90\% of the target's labeled training rows.

\input{tables/label_efficiency.tex}

\section{Metric Definitions and Normalized Rank}
\label{app:metrics}

\paragraph{Normalized rank (NR).}
NR denotes the mean normalized rank of a model across the finest evaluation unit $u$ available for a given aggregate:
\[
  \mathrm{NR}(m) \;=\; \frac{1}{|\mathcal{U}(m)|} \sum_{u \in \mathcal{U}(m)} \frac{\mathrm{rank}_u(m) - 1}{N_u - 1}\,,
\]
where $N_u$ is the number of models with a score on unit $u$, $\mathcal{U}(m)$ is the subset of units on which $m$ is scored, and ties are broken by \texttt{min}. Missing units are excluded from $m$'s own average and do not penalize other models' per-unit ranks. Lower is better. The unit $u$ differs by aggregate: \textsc{CTBench} family NRs average ranks over the individual tasks in a family (\eg, Schema NR over ColType, ColClust, ColRel); row-prediction NRs average ranks over individual target columns (77 classification, 46 regression; classification ranks are averaged separately over AUROC and Macro~$F_1$); and Clean/Robust Linkage NRs average ranks over individual datasets (8 clean DM; 4 dirty DM $+$ 4 WDC). Because dynamic range widens when fewer units are averaged, absolute NR magnitudes should be read within a column and not compared across suites.

\paragraph{Task metrics.}
$F_1$ scores are macro-averaged over classes by default. This covers all multi-class tasks in the benchmark, including column type prediction, row-prediction classification, and the table-pair classification tasks (join classification, union classification, table subset). Record linkage is the sole exception: because it is a binary classification task, we follow the entity-matching convention established by DeepMatcher~\cite{deepmatcher} and WDC Products~\cite{wdcproducts} and report \emph{binary $F_1$ on the match (positive) class}, equivalent to \texttt{sklearn.metrics.f1\_score(\ldots, average='binary', pos\_label=1)}. Group-level linkage scores (DM-C, DM-D, WDC, and the ``All (16 pairs)'' columns of Tables~\ref{tab:ablation_dim} and~\ref{tab:ablation_rl_head}) are the unweighted mean of per-dataset binary $F_1$. For regression tasks, we use $\mathrm{nRMSE} = \sqrt{1-R^2}$, a monotone transform of the coefficient of determination (lower is better; values above 1 correspond to negative $R^2$). To aggregate nRMSE across the 46 regression targets, we report the shifted geometric mean $\mathrm{SGM}_\varepsilon$ with $\varepsilon=0.01$:
\[
\mathrm{SGM}_\varepsilon(x_1,\ldots,x_K)
  = \Bigl(\,\prod_{i=1}^{K}(x_i+\varepsilon)\Bigr)^{1/K} - \varepsilon\,,
\]
which reduces sensitivity to outlier targets while penalizing consistently poor performance. The shift prevents the product from collapsing when any $x_i$ is exactly zero. AUROC for row-prediction classification is the area under the ROC curve, computed per target column and then averaged across the 77 classification targets; binary AUROC is used for binary targets and weighted one-vs-rest AUROC is used for multi-class targets, following the standard \texttt{sklearn.metrics.roc\_auc\_score} convention. MAP denotes mean average precision computed over the full ranked list (not truncated to a fixed $K$), except for union search, where it is computed over each query's top-$K$ retrieved tables (for the encoders, $K{=}10$ on SANTOS and UGEN and $K{=}60$ on TUS and TUS-hard). R@GT (Recall at Ground Truth) follows the Valentine convention~\cite{valentine}: all $m\times n$ candidate column pairs are ranked by cosine similarity, the top~$k$ pairs are retained with $k=|\text{ground truth}|$, and recall is the fraction of true correspondences among them. R@$k$ (Recall at $k$) is the fraction of ground-truth targets present in the top-$k$ retrieved items; we report $k{=}100$ as the Stage-1 retrieval diagnostic in DLTE. NMI is standard normalized mutual information with arithmetic averaging. MRR is mean reciprocal rank. Acc on Table QA is exact-match denotation accuracy on WikiTableQuestions, following the source convention~\cite{wikitablequestions}. $\mathrm{UJ\text{-}H}$ is defined in Sec.~\ref{sec:dlte}, and Cell~$F_1$, a complementary DLTE diagnostic, is defined in Appendix~\ref{app:cellf1}.

\section{Table-Footprint Coverage Across Suites}
\label{app:table-footprint}

\begin{figure}[H]
\centering
\includegraphics[width=0.99\linewidth]{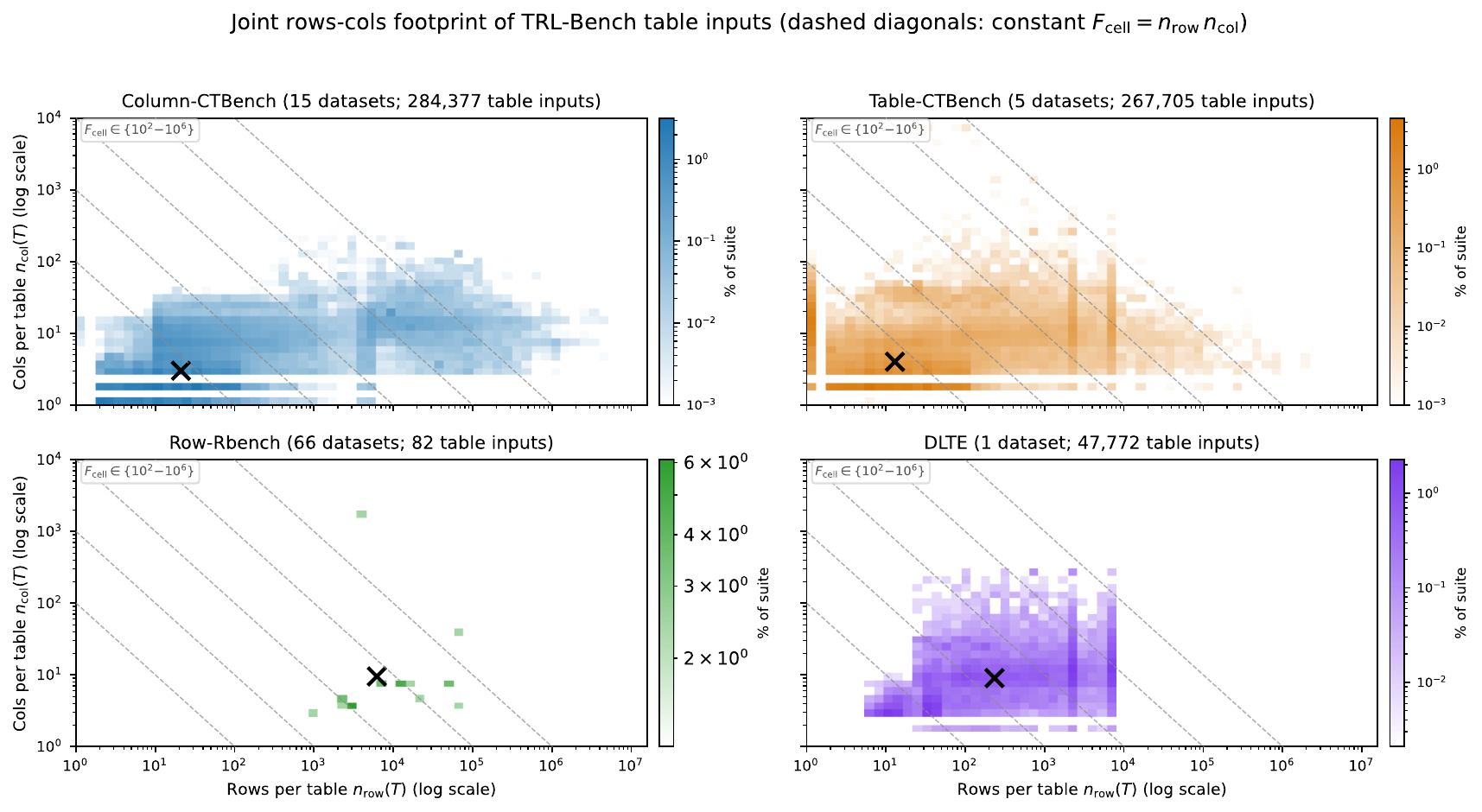}
\caption{
\textbf{Joint rows-columns footprint of TRL-Bench table inputs.}
Each panel plots a $2$D density of $(n_{\mathrm{row}}(T), n_{\mathrm{col}}(T))$ over the counted loadable table inputs of one suite, on log--log axes. Bin intensities are normalized within each suite to ``\% of suite'', so a smaller suite is not visually dominated by a larger one. Gray dashed diagonals mark constant-footprint contours $F_{\mathrm{cell}} \in \{10^{2},\,10^{3},\,10^{4},\,10^{5},\,10^{6}\}$. The black $\times$ in each panel sits at the suite-wise median of $n_{\mathrm{row}}$ and the suite-wise median of $n_{\mathrm{col}}$ (computed independently along each axis).
}
\label{fig:table-footprint}
\end{figure}

\begin{figure}[H]
\centering
\includegraphics[width=0.99\linewidth]{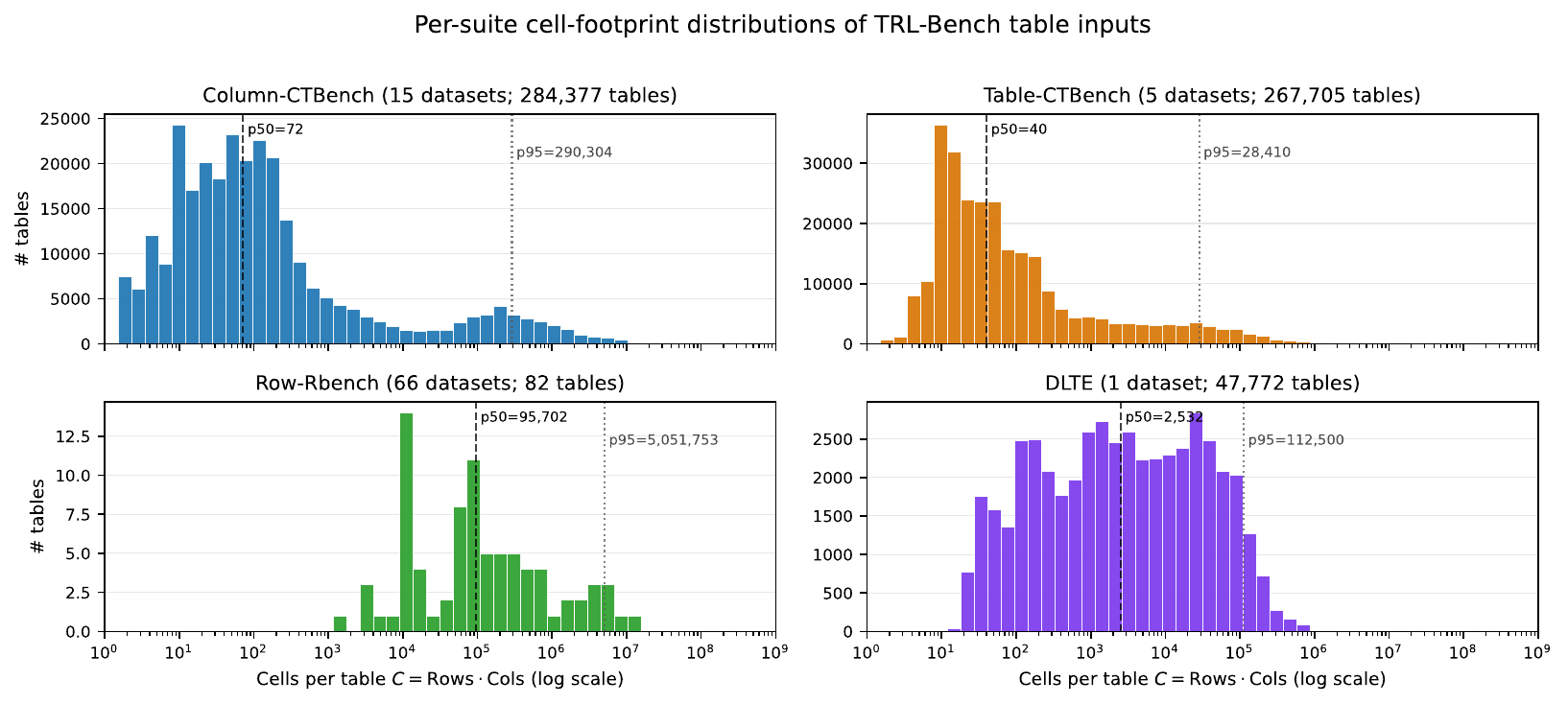}
\caption{
\textbf{Per-suite cell-footprint distributions of TRL-Bench table inputs.}
Each panel histograms the footprint values $F_{\mathrm{cell}}(T) = n_{\mathrm{row}}(T)\,n_{\mathrm{col}}(T)$ for the counted loadable table inputs in one suite, using $50$ common logarithmically spaced bins over $10^{0}$--$10^{9}$ cells.
The black dashed line marks the median footprint within that suite, and the gray dotted line marks the corresponding $95$th percentile.
Each panel's $y$-axis reports the number of counted table inputs per bin and is scaled independently because the four suites differ by orders of magnitude in the number of counted inputs.
}
\label{fig:table-footprint-hist}
\end{figure}

For each table $T$, let $n_{\mathrm{row}}(T)$ and $n_{\mathrm{col}}(T)$ denote its row and column counts (equivalently, $|R(T)|$ and $|C(T)|$ in the notation of Sec.~\ref{sec:benchmark-protocol}), and define its \emph{cell footprint} as $F_{\mathrm{cell}}(T) = n_{\mathrm{row}}(T)\,n_{\mathrm{col}}(T)$. A \emph{loadable table input} is a concrete table object returned by a benchmark evaluation loader for a table-valued role (feature table, left/right entity table, query table, or lake/corpus table), before model-specific serialization or truncation. We count each distinct table once within each role; if the same physical table appears under multiple roles, it is counted once per role. Thus row-prediction datasets contribute one feature table each, record-linkage datasets contribute their left and right entity tables, column- and table-CTBench datasets contribute their table-valued query/lake/corpus inputs as applicable, and \textsc{TRL-DLTE} is summarized by its 47{,}772 retrieval-lake candidates. Labeled table pairs, their labels, and train/valid/test split indices are metadata rather than additional table inputs: a table referenced by many pairs or splits is still counted once in its role. Figure~\ref{fig:table-footprint} shows the joint $(n_{\mathrm{row}}, n_{\mathrm{col}})$ density per suite, Figure~\ref{fig:table-footprint-hist} marginalizes to $F_{\mathrm{cell}}$ on a logarithmic axis with $50$ common bins, and Table~\ref{tab:table-size-buckets} lists per-dataset summary statistics for the $87$ dataset-source entries, grouped into seven benchmark categories: Schema, Joinability, Unionability, and Grounding from CTBench, Row Prediction and Record Linkage from RBench, and DLTE on its own. These statistics are descriptive only: benchmark scores are computed per task and are not weighted by table count or cell footprint.

\input{tables/table_size_buckets}

\section{Family-Level Performance Summary Figure}
\label{app:family-summary}

\begin{figure}[H]
\centering
\includegraphics[width=\textwidth]{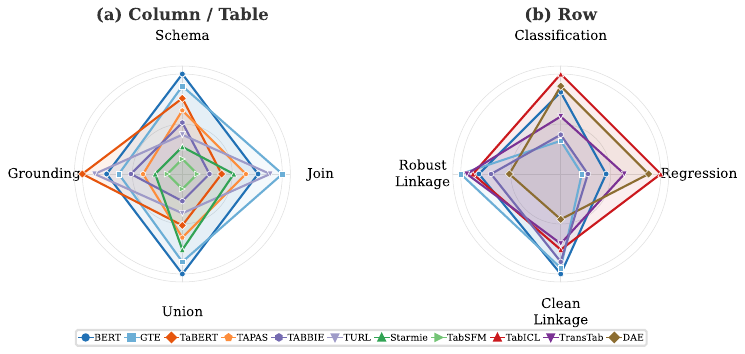}
\caption{\textbf{Granularity-dependent transfer profiles.} Radar plots summarize family-level performance for representative models on the two atomic suites. Panel~(a) compares column/table encoders across the four \textsc{TRL-CTbench} capability families: Schema, Join, Union, and Grounding. Panel~(b) compares row encoders across Classification, Regression, Clean Linkage, and Robust Linkage in \textsc{TRL-Rbench}. \textsc{BERT}, \textsc{GTE}, and \textsc{TABBIE} appear in both panels because they expose both column- and row-level embeddings. The remaining models in each panel are granularity specialists and are evaluated only within their supported suite. Values are rank-normalized within each axis (farther from center is better). No single model dominates all axes, and strengths shift substantially with representation granularity and task family. This figure is a qualitative summary of Tables~\ref{tab:main_results} and~\ref{tab:r_main_results}.}
\label{fig:triple_radar}
\end{figure}

\section{CTBench Diagnostics and Ablations}

\subsection{Observational Lexical-Accessibility Proxies for CTBench}
\label{app:ctbench_lexical}

To support the Sec.~\ref{sec:exp-col-table} reading that generic text encoders remain strong on many column/table tasks through \emph{lexical accessibility}, we report two observational proxies in Table~\ref{tab:ctbench_lexical_proxies} computed directly from the main CTBench results. The first is the gap between the strongest applicable non-neural baseline and the best neural encoder on the task: small gaps (after direction correction) indicate that surface lexical statistics, such as Jaccard, TF-IDF, value overlap, and Valentine matchers, already recover most of the task signal. The second is the direction-corrected \emph{generic-text advantage}, the mean score of \textsc{BERT}/\textsc{GTE} minus the mean of tabular specialists (\textsc{TaBERT}/\textsc{TAPAS}/\textsc{Starmie}/\textsc{TURL}) on the same task. Generic text encoders lead on 10 of 13 CTBench tasks, with the strongest advantages on tasks whose input is dominated by short natural-language text (table retrieval $+0.200$, join search $+0.116$, column type prediction $+0.088$, column relation prediction $+0.072$, and column clustering $+0.048$), consistent with these tasks being accessible from headers and short cell strings. The three exceptions where the four-specialist mean beats the two-generic-text mean are exactly the tasks whose pretraining objective is tightly aligned: \textsc{Starmie}'s column-level contrastive objective wins schema matching ($-0.067$) and union search ($-0.010$), and \textsc{TURL}'s table-language modeling wins Table QA ($-0.016$). Sec.~\ref{sec:exp-col-table} additionally counts Table Subset as a specialist-won task because the individual best model is \textsc{TAPAS} (0.567 $F_1$, beating both \textsc{BERT} and \textsc{GTE}); on this mean-vs.-mean proxy Table Subset is borderline ($+0.006$ to generic text) since the four-specialist mean dilutes a single-specialist win. This is an observational hint, not a causal test. A direct header-masking ablation would provide stronger evidence and is left as future work. Within the current data, however, the ordering aligns with the main-text reading: CTBench task families differ systematically in whether text- or structure-pretrained encoders provide the dominant signal.

\input{tables/ctbench_lexical/ctbench_lexical_proxies.tex}

\subsection{Probe Head Complexity for Column/Table-Level Tasks}
\label{app:ablation_head_col}

\input{tables/ablation_head_col}

Main column- and table-level results follow the unified supervised-probe protocol of Sec.~\ref{sec:benchmark-protocol} (averaging linear and MLP heads). This appendix reports the linear and MLP components separately to show per-task head sensitivity. Table~\ref{tab:ablation_head} compares MLP, linear, and dummy probes across five column/table-level tasks.

\paragraph{Frozen embeddings carry most of the signal.}
The $\Delta_{\text{L-D}}$ column shows that the linear probe already far outperforms the dummy baseline on every task: averages of $+0.65$ (ColType), $+0.72$ (ColRel), $+0.12$ (JoinCls), $+0.39$ (UnionCls), and $+0.05$ (TblSubset). This confirms that the frozen representations encode task-relevant structure without any task-specific training.

\paragraph{MLP adds little over linear on column tasks.}
For ColType and ColRel, the average $\Delta_{\text{M-L}}$ is $+0.00$ and $+0.02$ respectively, indicating that a linear probe is sufficient. For UnionCls and TblSubset, the MLP gains are more consistent ($+0.19$ and $+0.24$ on average), suggesting that nonlinear separation helps when table-pair geometry is more complex.

\paragraph{JoinCls favors linear probes.}
The average $\Delta_{\text{M-L}}$ for JoinCls is $-0.02$, and several models show negative gaps (e.g., BERT $-0.05$, TaBERT $-0.06$). This indicates that the MLP head overfits on the relatively small join classification training sets, and a linear probe is the more reliable choice for this task.

\subsection{Aggregation Ablation for Table-Level Embeddings}
\label{app:aggregation_ablation}

Each model's table aggregation is selected on development data rather than on the test metric; for Table Retrieval, we release a development subset of 966 questions, carved from the training questions and question-disjoint from the test set, together with the benchmark. On this subset we also select each model's query encoder (Appendix~\ref{app:ablation_qe}). Selecting on the test metric instead would give the same Table Retrieval pick for 9 of the 10 models. For models that expose multiple candidate table embeddings, we compare three aggregation strategies:
\textsc{cls} (the \texttt{[CLS]} token of the linearized table),
\textsc{col-mean} (the mean of per-column embeddings), and
\textsc{tok-mean} (the mean of all non-padding token hidden states).
Tables~\ref{tab:aggregation_ablation} and~\ref{tab:aggregation_ablation_linear} report the full MLP- and linear-probe breakdowns, respectively.

\paragraph{Development-split construction.}
Table~\ref{tab:dev_split_sizes} lists, for each table-level task, the development-partition size actually used for this selection and how that partition is constructed. The three table-disjoint pairwise tasks (join classification, union classification, union regression) and table subset select on their standard held-out validation partition (Appendix~\ref{app:reproducibility}); Table Retrieval selects on the released question-disjoint development subset described above. Selection is always performed on this development data and reported on the test partition; Appendix~\ref{app:dlte-rankings} separately cross-validates dev-vs-test rank stability for the analogous pipeline-selection setting in \textsc{TRL-DLTE}.

\input{tables/dev_split_sizes}

\paragraph{Task-level averages.}
With an MLP probe, \textsc{tok-mean} gives the best average on UnionCls, UnionReg (lower is better), and TblRet, while \textsc{col-mean} is narrowly best on JoinCls and TblSubset. With a linear probe, \textsc{col-mean} is best on JoinCls and UnionReg, while \textsc{tok-mean} is best on UnionCls and TblSubset. However, several gaps are tiny (e.g., 0.494/0.500/0.497 on MLP JoinCls for \textsc{cls}/\textsc{col-mean}/\textsc{tok-mean} and 0.724/0.725/0.734 on linear UnionCls), so these averages should be read as rough tendencies rather than definitive rankings.

\paragraph{Per-model variation.}
The per-model tables show some encoder-specific preferences: \textsc{TABBIE} often peaks with \textsc{cls} under MLP, whereas \textsc{TAPAS} and several text-pretrained models more often favor \textsc{col-mean} or \textsc{tok-mean}. Nevertheless, the model-dependent variation is modest relative to the task-dependent variation, so we do not treat aggregation choice as a major axis of analysis. The main comparison in Table~\ref{tab:main_results} reports each encoder's dev-selected aggregation averaged over MLP and linear probes.

\input{tables/aggregation_ablation}

\input{tables/aggregation_ablation_linear}

\subsection{Join Search: Direct Cosine vs.\ Learned Projection}

\input{tables/join_search_learned_table.tex}

\paragraph{Interpretation.}
Direct cosine is an informative no-training control, but in our frozen-transfer reuse of join search the relation of interest is directional containment rather than pure semantic similarity. The learned projection is therefore the canonical main-table setting: it remains a minimal probe over frozen embeddings while matching the transfer objective more closely.

\subsection{Union Search: TUS vs.\ TUS-hard}
\label{app:tus_hard}

Table~\ref{tab:tus_hard} compares model performance on TUS (original) and TUS-hard (low-overlap variant). TUS-hard filters out positive pairs whose directed column containment is at least $0.70$, removing the highest-overlap 36\% of positives so that our frozen-transfer evaluation can better separate lexical overlap from broader union signal.

\input{tables/tus_hard_comparison}

\paragraph{Interpretation.}
On original TUS, the value-overlap baseline achieves MAP 1.000 and \textsc{BERT}, \textsc{GTE}, and \textsc{TURL} all score above 0.95. On TUS-hard, the baseline drops to 0.008 and the ranking shifts substantially (Spearman $\rho = -0.67$). The models most robust under this low-overlap variant are \textsc{Starmie} and \textsc{TaBERT}, whose pretraining objectives emphasize cross-table structure. We therefore report TUS-hard alongside TUS in \benchmark\ not as a replacement for the original resource, but as a complementary variant for the frozen-transfer setting, where it is useful to distinguish overlap-driven retrieval from transfer that remains helpful when overlap is limited.

\subsection{Union Search: Pylon}
\label{app:pylon_union}

\textsc{Pylon}~\cite{pylon} is a table-union-search benchmark built from 1{,}746 real GitHub tables, 124 of which its authors hand-curated into three human ``unionable'' topic cliques (35 scholarly-article, 41 job-posting, 48 music-playlist; gold labels are full clique membership). Pylon is evaluated for the 8 column-capable models under \benchmark's standard union-search harness (per-column Munkres bipartite matching, with the $\mathrm{hnsw}/K/\text{threshold}{=}0.1$ configuration of TUS/TUS-hard, Table~\ref{tab:pylon_union}). $K{=}60$, the smallest swept $k$ at or above the largest topic clique (48), gives the R@60 headline, and MAP@10 uses the $K{=}10$ cutoff of SANTOS and UGEN. \textsc{Starmie} ranks first under both metrics.

\input{tables/pylon_union.tex}

\paragraph{Coverage.}
All 8 evaluated models load the full 1{,}746-table corpus except \textsc{TabSketchFM}, whose table-level preprocessing fails on 4 tables; all 4 are unlabeled distractors, so 0 queries or gold-clique members are affected. This is caught by a skip-and-log guard in the \textsc{TabSketchFM} wrapper that logs and excludes the failing table rather than aborting the extraction batch.

\paragraph{Hosting.}
Pylon is downloaded from its official source.

We do not compare against Pylon's own reported scores, which use a different retrieval protocol and model pool; the dataset's human-curated gold, not the original paper's numbers, is this appendix's reference throughout.

\subsection{Table Retrieval: Model-Only vs.\ Hybrid Mode}
\label{app:hybrid_ablation}

Table~\ref{tab:hybrid_ablation} compares model-only and hybrid retrieval modes. In model-only mode, the projection head operates solely on the model's own table embedding. In hybrid mode, the model's table embedding is concatenated with the query encoder's (MPNet~\cite{mpnet} or sentence-T5~\cite{sentencet5}) table embedding before projection, bridging the modality gap between the model's table space and the query space.

\input{tables/hybrid_ablation}

\paragraph{Interpretation.}
Hybrid mode uniformly improves MRR for every model, with gains ranging from $+0.057$ (\textsc{GTE}) to $+0.509$ (\textsc{Starmie}). However, all hybrid MRRs converge to a narrow band of $0.509$--$0.555$, regardless of the model's own retrieval quality ($0.018$--$0.476$ in model-only mode). This indicates that the added query-encoder table embedding contributes most of the shared signal in the augmented setting. For \benchmark's main comparison we therefore use model-only mode, because it isolates the transfer quality of the model's own table representation rather than performance in a pipeline with added query-side table evidence.

\clearpage
\subsection{Query Encoder Sensitivity}
\label{app:ablation_qe}

\input{tables/ablation_qe}

Table~\ref{tab:ablation_qe} compares MPNet~\cite{mpnet} and sentence-T5~\cite{sentencet5} (ST5) as query encoders across table retrieval (MRR) and table QA (accuracy).
MPNet consistently outperforms ST5 on table retrieval for almost every model (average $\Delta{=}{+}0.02$), with the largest gain for TaBERT ($+0.05$).
The pattern reverses for table QA: ST5 is better for all models with available results (average $\Delta{=}{-}0.03$), suggesting ST5's longer context pretraining better supports semantic parsing.
In both tasks the absolute differences are small (${\leq}0.05$), indicating that query encoder choice has limited sensitivity on these tasks.
In the main evaluation, Table QA averages the two query encoders, and Table Retrieval selects the query encoder per model on development data together with the table aggregation (Appendix~\ref{app:aggregation_ablation}).

\subsection{Pair-Level Random vs.\ Table-Disjoint Split Ablation}
\label{app:strict_ablation}

Four table-pair tasks (join classification, column overlap, union classification, and union regression) support both the original pair-level random splits and the table-disjoint splits used in our frozen-transfer evaluation, where training and test tables are separated so the task measures cross-table generalization. Table~\ref{tab:strict_ablation} compares the two protocols.

\input{tables/strict_ablation}

\paragraph{Interpretation.}
Under the table-disjoint protocol, performance is uniformly lower. The effect is largest on join classification, where the mean per-model $F_1$ drop is 0.212 (aggregate means: $0.736 \to 0.523$), and union regression shows the next-largest change (average nRMSE increases by 0.110). This is consistent with our transfer-oriented setting being harder: models must generalize to unseen tables rather than to new pairs drawn from already observed tables. Relative rankings are largely preserved, suggesting that table-disjoint evaluation mainly changes the difficulty level and the degree of cross-table separation required by our protocol. We therefore use table-disjoint splits as the default when repurposing these tasks for frozen cross-table transfer.

\section{RBench Diagnostics and Ablations}

\subsection{Row-Prediction Probe Diagnostics}

\begin{table}[t]
\centering
\small
\setlength{\tabcolsep}{5pt}
\caption{Row-prediction probe-head sweep. Here $d$ counts linear layers. The default one-hidden-layer MLP ($h{=}256,d{=}2$) is near-optimal for classification. Deeper/wider heads do not provide a consistent regression benefit.}
\label{tab:row-head-sweep}
\begin{tabular}{lcc}
\toprule
Head config & Avg Macro-$F_1$ & Avg SGM$\downarrow$ \\
\midrule
$h{=}256,d{=}2$ (default) & 0.6173 & 0.757 \\
$h{=}128,d{=}3$ & 0.6173 & 0.750 \\
$h{=}256,d{=}3$ & 0.6144 & 0.772 \\
$h{=}512,d{=}4$ & 0.6064 & 0.755 \\
$h{=}512,d{=}5$ & 0.5983 & 0.750 \\
\bottomrule
\end{tabular}
\end{table}

\paragraph{Representative linear-vs.-MLP cases.}
For classification, several models are already strongest with a linear probe: the text-transfer encoders \textsc{BERT} (0.6322 MLP $\to$ 0.6360 linear) and \textsc{GTE} (0.6004 $\to$ 0.6200), as well as the target-table contrastive learner \textsc{TransTab} (0.6016 $\to$ 0.6151). By contrast, the feature-corruption target-table SSL models benefit primarily on regression, e.g., \textsc{DAE} improves from 0.5984 to 0.5357 SGM, \textsc{SCARF} from 0.5917 to 0.5462, and \textsc{TabBinning} from 0.5929 to 0.5448 when moving from linear to MLP.

\paragraph{Dimensionality check.}
Dimensionality alone does not explain the row ranking. Even after upgrading SSL row encoders to 768 dimensions, \textsc{TabICL} (512-d) remains strongest on both classification and regression in this dim-controlled comparison (0.6744 Macro-$F_1$ and 0.4873 SGM, computed on the dim-ablation subset; the main-table values are 0.671 and 0.505 under the standard probe protocol).
Embedding dimensions range from 192 (\textsc{TabPFN}) to 768 (generic text and table-aware models), with most SSL row models at 512. Because the MLP probe uses a fixed hidden size of~256, the first-layer parameter count scales linearly with input dimension (e.g., $768\times256$ vs.\ $192\times256$). The linear-probe comparison above serves as a dimension-proportional control: linear probes have exactly $d\times C$ parameters (where $d$ is the embedding dimension and $C$ the number of classes), so they do not introduce a fixed-width bottleneck. The main row ranking is consistent across linear and MLP probes, indicating that the ranking reflects embedding quality rather than probe-capacity artifacts.
For pairwise tasks (record linkage), concatenation doubles the input dimension, producing 1536-d inputs for 768-d models and 1024-d for 512-d models with the same hidden-size-256 MLP. The same linear-probe consistency check applies.

\begin{figure}[t]
\centering
\includegraphics[width=\linewidth]{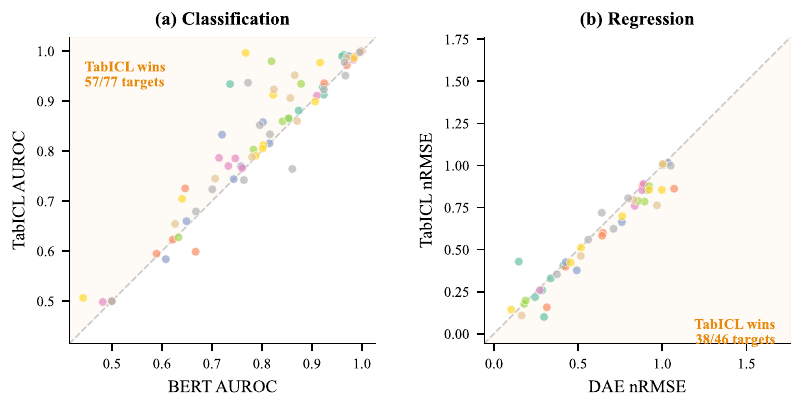}
\caption{Per-target pairwise comparison of \textsc{TabICL} against the strongest comparator. (a)~Classification AUROC: \textsc{TabICL} vs.\ \textsc{BERT} across 77~targets. (b)~Regression nRMSE: \textsc{TabICL} vs.\ \textsc{DAE} across 46~targets. Points above the diagonal in~(a) and below it in~(b) indicate \textsc{TabICL} wins. Each color represents a different source dataset.}
\label{fig:row_scatter}
\end{figure}

\paragraph{Per-target comparison.}
Figure~\ref{fig:row_scatter} plots per-target scores for \textsc{TabICL} against the strongest comparator on each regime: \textsc{BERT} for classification (AUROC) and \textsc{DAE} for regression (nRMSE). \textsc{TabICL} wins on 57/77 classification targets and 38/46 regression targets, confirming that its advantage is broad rather than driven by a few outlier tasks.

\paragraph{Adaptation regime explains a substantial fraction of row-level variance.}
Table~\ref{tab:row_regime_variance} quantifies the Sec.~\ref{sec:exp-row} claim that training regime (externally pretrained transfer vs.\ target-table self-supervision vs.\ prior-based meta-pretraining) materially shapes row-level rankings, using the per-sub-task normalized-rank aggregates from Table~\ref{tab:r_main_results}. The decomposition uses the classical 1-way ANOVA sums-of-squares with regime identity as the factor, so between-regime variance and within-regime variance sum to the total sub-task variance (an $\eta^2$-style fraction). The regime effect is largest on the two linkage sub-tasks: between-regime variance accounts for 64\% of cross-model NR variance on clean linkage and 48\% on robust linkage. The Kruskal--Wallis test rejects the null of identical regime distributions at $\alpha=0.05$ on clean linkage ($p=0.014$) and is borderline on robust linkage ($p=0.064$). Regression shows a smaller regime effect (41\% between/total) with a marginally significant Kruskal--Wallis statistic ($p=0.090$), while classification has the weakest effect (17\% between/total, $p=0.373$), consistent with the observation that frozen text encoders and target-table learners both reach competitive classification AUROC for different reasons. The direction of the regime effect also depends on the sub-task: Transfer and Prior-based encoders dominate linkage, whereas Target-Table learners are more competitive on prediction regression. This supports keeping the three regimes separately rather than flattening them into one leaderboard. (We note that the Prior-based regime contains only two models, so the Kruskal--Wallis $\chi^2$ approximation is borderline at the $\alpha=0.05$ level for robust linkage. The between/total $\eta^2$ summary does not share this small-sample limitation.)

\input{tables/row_regime/row_regime_table.tex}

\clearpage
\subsection{Embedding Dimension for Record Linkage}
\label{app:ablation_dim}

Table~\ref{tab:ablation_dim} reports binary $F_1$ (match class) for eight target-table learners across five embedding dimensions (64, 128, 256, 512, 768) on the record linkage task.
Performance increases monotonically with dimension for most models: the average All-pairs $F_1$ rises from 0.058 at $d{=}64$ to 0.139 at $d{=}768$.
The WDC group is consistently the hardest across all dimensions, reflecting the greater heterogeneity of product-matching pairs.
TransTab improves substantially with larger embeddings (0.139 $\to$ 0.254 overall), while SubTab remains near zero regardless of dimension, suggesting its representations lack pairwise match signal at any scale.
The main evaluation keeps the 512-d training default for all target-table SSL models (Appendix~\ref{app:input-pipelines}), although on record linkage larger widths score higher for most of them.

\input{tables/ablation_dim}

\noindent\textbf{A common-dimension check (row level, 5 seeds).} The native-width sweep above is complemented by a criterion comparison at a shared budget: for the same record-linkage pairs, each model's native embedding is additionally projected to 192 dimensions by (i)~PCA (fit on the training split) and (ii)~a criterion-free Gaussian random projection, at 5 seeds (42/52/62/72/82); \textsc{TabPFN}'s native embedding is already 192-d and serves as the fixed reference point with no projected arm. Table~\ref{tab:pca_linkage} reports the resulting binary $F_1$ per model and arm, aggregated over all 16 record-linkage pairs. Spearman rank correlation between the native ranking and each projected arm (Table~\ref{tab:pca_linkage}, bottom rows) shows the random-projection control tracks the native ranking at least as well as PCA at the same budget, indicating this preservation is substantially a width effect rather than an artifact specific to PCA's variance criterion.

\input{tables/pca_linkage.tex}

\clearpage
\subsection{Probe Head for Record Linkage}
\label{app:ablation_rl_head}

\input{tables/ablation_rl_head}

Table~\ref{tab:ablation_rl_head} compares cosine similarity, linear probing, an MLP probe, and a dummy baseline across 14 models on the record linkage task.

\paragraph{Cosine outperforms linear for most models.}
The average $\Delta_{\text{L-C}}$ is $-0.20$ overall, meaning supervised linear probing \emph{hurts} relative to unsupervised cosine matching. This holds across all three dataset families and is most severe on WDC ($-0.37$), where linear probes average only $F_1{=}0.049$ versus cosine's $0.415$.

\paragraph{A nonlinear head recovers most of what the linear probe leaves on the table.}
Averaged across all 16 pairs, the MLP head lifts $F_1$ by $+0.164$ over the linear probe ($0.330$ vs.\ $0.166$). The gap is largest on WDC ($+0.194$) where linear probes nearly collapse, and still substantial on DM-D ($+0.170$) and DM-C ($+0.145$). At the model level, WDC produces the most dramatic reversals for \textsc{TransTab} ($0.041$ Lin $\to$ $0.760$ MLP), \textsc{GTE} ($0.072 \to 0.550$), and \textsc{BERT} ($0.093 \to 0.379$). These models encode usable entity-matching structure that a linear probe cannot access in 5-seed training. Overall, the MLP score $0.330$ (All) sits below cosine's $0.365$, with MLP beating cosine on DM-D ($+0.031$) and matching it on DM-C ($+0.001$). Cosine retains its lead on WDC ($-0.172$).

\paragraph{WDC is the hardest for linear probes, and still hard for some MLP probes.}
The linear probe nearly collapses on WDC for target-table SSL models (e.g., SubTab: 0.000, TabTransf.: 0.022), whereas cosine similarity remains non-trivial (${\sim}0.38$--$0.43$). The MLP head partially rescues this for \textsc{TransTab} ($0.760$, target-table SSL with cross-table contrastive objective), \textsc{GTE} ($0.550$), \textsc{BERT} ($0.379$), and \textsc{TabICL} ($0.245$), but it stays near-zero for the target-table SSL encoders on WDC (SubTab: 0.017, TabTransf.: 0.018, TabBinning: 0.095, SCARF: 0.102). These embeddings appear to lack WDC-relevant entity-matching signal rather than merely hiding it in non-linear form.

\paragraph{Implication for the main evaluation.}
Main record linkage results follow the unified supervised-probe protocol of Sec.~\ref{sec:benchmark-protocol}, which averages the linear and MLP probe heads. We headline avg(linear, MLP) rather than cosine because it is the only head that applies uniformly to every row sub-task (cosine is undefined for prediction) and because the linear and MLP heads span the linear-vs-nonlinear capacity axis at a fixed supervision level. Averaging does not privilege either regime, which matters because some encoder/source combinations carry linkage signal that is linearly accessible (text encoders on DeepMatcher) while others need a nonlinear readout (\textsc{TransTab}, \textsc{GTE}, \textsc{BERT} on WDC). This appendix disentangles the two: the MLP head recovers a meaningful slice of the transferable matching signal that cosine can extract, while the linear probe under-reads it, especially on WDC. Averaging the two therefore pulls headline numbers downward for models with strong nonlinear structure (\textsc{TransTab}, \textsc{GTE}, \textsc{BERT} on WDC), and the cosine column here is best read as a training-free reference rather than an upper bound. The relative strength of learned versus training-free matching varies substantially across dataset families, with cosine stronger on WDC and the MLP head competitive on the DeepMatcher benchmarks.

\clearpage
\subsection{Record Linkage Split and Leakage Audit}
\label{app:linkage_split_audit}

The 16 record-linkage sources keep their original DeepMatcher~\cite{deepmatcher} and WDC LSPM~\cite{wdcproducts_lspm} pair-disjoint splits, so essentially no pair appears in both train and test (pair overlap $\le 0.02\%$ across all 16 datasets in Table~\ref{tab:linkage_split_audit}). Because each tableA / tableB row participates in many candidate pairs, however, the same row can appear on both sides of the split: in 11 of 16 sources, more than half of the test-side rows on both tableA and tableB already appear in the train+valid pair lists (e.g., Abt-Buy: $94.2\%$ / $95.0\%$; DBLP-Scholar: $95.8\%$ / $69.5\%$). Beer, iTunes-Amazon, iTunes-Amazon-D, WDC-medium, and WDC-small are the five sources with cross-split row overlap below $50\%$ on at least one side. We keep all sources to remain comparable with the entity-matching literature.

\paragraph{Row-pair reformulation.}
Each linkage source is stored as two row tables (\texttt{tableA} and \texttt{tableB}) with stable row identifiers and three pair lists (train, validation, test) of (left row, right row, label) triples taken from the source splits. The WDC LSPM v2 offer pairs are converted to the same format, one pair of row tables per size. Encoders embed each row once from its retained columns, and a pair is scored by a probe over the concatenation of its two row embeddings (Sec.~\ref{sec:rbench}).

\input{tables/linkage_split_audit/linkage_split_audit.tex}

\paragraph{Removal of label-equivalent columns.}
WDC LSPM v2 raw records ship with three columns that do not belong in feature input: \texttt{cluster\_id} is the gold product cluster identifier (column equality reproduces the test label at $99.5\%$ precision and $99.6\%$ recall on every WDC size), and \texttt{identifiers} contains GTIN/MPN unique product IDs ($99.8\%$ precision and $37.5\%$ recall) that are excluded from features in the standard WDC LSPM evaluation protocol. Fodors-Zagats analogously exposes a \texttt{class} column whose value is the entity cluster ID ($100\%$ precision, $100\%$ recall on the test split). \benchmark\ removes \texttt{cluster\_id} and \texttt{identifiers} from the four WDC tables and \texttt{class} from Fodors-Zagats before any encoder serializes a row, so the gold label cannot enter the row representation as a feature. The retained columns are \texttt{brand}, \texttt{category}, \texttt{description}, \texttt{keyValuePairs}, \texttt{price}, \texttt{specTableContent}, \texttt{title} for WDC, and \texttt{name}, \texttt{addr}, \texttt{city}, \texttt{phone}, \texttt{type} for Fodors-Zagats. After removal, no retained column is label-equivalent under the same equality rule: the retained column with the highest $F_1$ reaches $28.3\%$ precision at $100\%$ recall on every WDC size (\texttt{category}) and $100\%$ precision at $72.7\%$ recall on Fodors-Zagats (\texttt{name}).

\paragraph{Row-disjoint strict-test ablation.}
To check that the row-overlap reported above does not distort cross-model rankings, we build a row-disjoint variant of each source: train+valid stay as-is, and the test pair list is filtered to pairs whose tableA \emph{and} tableB rows do not appear in train+valid. The filter strips most pairs from the high-overlap sources, so we report the strict ablation only on the 10 sources whose strict-test stays $\geq 30$ pairs and keeps a $\geq 10\%$ minority class: 6 DeepMatcher viable sources (Amazon-Google, Beer, iTunes-Amazon~/~iTunes-Amazon-Dirty, Walmart-Amazon~/~Walmart-Amazon-Dirty) and 4 WDC sizes; the 6 skipped sources (Abt-Buy, Fodors-Zagats, DBLP-ACM~/~-Dirty, DBLP-Scholar~/~-Dirty) lose either viable pair count or label balance. Re-running the unified probe protocol of Sec.~\ref{sec:benchmark-protocol} on the strict-test subset at seed~42 and comparing per-model strict-test NR rankings with canonical 5-seed-mean original-protocol rankings on the same 10 viable strict-test sources, the strict-vs-original linkage rankings are highly correlated across all 14 row models: Spearman $\rho = 0.94$ ($p = 5.6 \times 10^{-7}$) over the full 10-source set, $\rho = 0.97$ on DM-viable (6 sources), and $\rho = 0.95$ on WDC (4 sources). The model-family conclusions of Sec.~\ref{sec:exp-row} (Transfer-Based encoders dominate Robust Linkage; Target-Table SSL trail) hold under the strict-test ablation. Row overlap does not change the qualitative reading. Table~\ref{tab:strict_absolute} reports per-source strict-test pair counts, positive rates, and absolute $F_1$ for the two top-ranked Robust Linkage models (\textsc{GTE} and \textsc{TransTab}).

\input{tables/linkage_split_audit/strict_absolute.tex}

\clearpage
\subsection{Intrinsic Embedding-Geometry Diagnostics for Row Encoders}
\label{app:taskfree}

We complement \textsc{TRL-Rbench}'s downstream scores with an intrinsic embedding-geometry analysis of the same exported row embeddings used by the standardized protocol. Using eight established diagnostics from the broader representation-learning literature, covering \emph{spectral spread}, \emph{spectral shape}, and \emph{spatial structure} (``task-free'' metrics in the prior literature), we ask which geometric properties of row-embedding spaces co-rank with downstream utility on row prediction and record linkage. To our knowledge, prior tabular benchmark resources do not pair representation-level downstream evaluation with intrinsic embedding-geometry diagnostics on the same encoder outputs under a common protocol.

\paragraph{Diagnostic families and formulas.}
We group the eight task-free diagnostics used in this analysis into three complementary families that we introduce here to organise the discussion: \textbf{Spectral Spread}, \textbf{Spectral Shape}, and \textbf{Spatial Structure}. Each underlying metric is drawn from prior work (cited at its definition below). The three-family taxonomy itself is our framing. For each (model, dataset) we form a single frozen embedding matrix $X \in \mathbb{R}^{n \times d}$, the row-embedding matrix for that table (for target-table SSL encoders this is the same matrix the encoder was trained on; for frozen transfer and prior-based encoders it is the inference-time row matrix), and evaluate all eight diagnostics from its singular value decomposition
\begin{equation}
X \;=\; U\,\Sigma\,V^{\top},
\end{equation}
and from the eigenspectrum $\lambda_1 \ge \cdots \ge \lambda_d \ge 0$ of the centred covariance
\begin{equation}
C \;=\; \tfrac{1}{n}\,X_c^{\top}\,X_c,
\qquad X_c \;=\; X - \mathbf{1}\,\bar{x}^{\top},
\qquad \bar{x} \;=\; \tfrac{1}{n}\,X^{\top}\mathbf{1}.
\end{equation}
Let $\sigma_1 \ge \sigma_2 \ge \cdots$ denote the singular values of $X$ and $r$ its numerical rank. All eight diagnostics are deterministic functions of $X$.

\textbf{Spectral Spread.} If variance in an embedding matrix concentrates in only a handful of singular directions, most of the ambient dimensions are redundant: the representation effectively lives on a low-dimensional subspace, and the remaining capacity is unavailable to the downstream head. Spectral Spread diagnostics measure how evenly variance is allocated across the spectrum. Higher values mean more independent directions actively carry information, a necessary condition for rich, transferable embeddings.

\textbf{RankMe} \citep{garrido2023rankme}.
\begin{equation}
\mathrm{RankMe}(X) \;=\; \exp\!\Bigl(-\sum_{i=1}^{\min(n,d)} p_i \log p_i\Bigr),
\qquad p_i = \frac{\sigma_i}{\sum_j \sigma_j}.
\end{equation}
The exponentiated entropy of the singular-value distribution. It equals $1$ for a rank-one spectrum and $\min(n,d)$ when all singular values are equal.

\textbf{RankMe$\boldsymbol{^{\star}}$} \citep{garrido2023rankme,tsitsulin2023}.
\begin{equation}
\mathrm{RankMe}^{\star}(X) \;=\; \frac{\mathrm{RankMe}(X)}{\min(n,d)} \;\in\; [0, 1].
\end{equation}
A dimension-normalised variant: the fraction of the available embedding dimensions the representation actually uses.

\textbf{NESum.}
\begin{equation}
\mathrm{NESum}(X) \;=\; \frac{1}{\lambda_1}\sum_{i=1}^{d} \lambda_i.
\end{equation}
Total variance divided by leading variance. It equals $1$ under rank-one collapse and approaches $\min(n,d)$ as the covariance spectrum flattens.

\textbf{Spectral Shape.} Two embeddings can have the same effective rank and still look very different along the spectrum: one may drop off sharply after the top few singular directions, while another decays as a slow power-law that keeps weak but non-trivial signal in many more directions. Spectral Shape diagnostics capture this tail profile, which determines how much low-variance structure survives to support tasks whose discriminative signal is not confined to the dominant directions.

\textbf{Pseudo $\boldsymbol{\kappa}$.}
\begin{equation}
\kappa(X) \;=\; \frac{\sigma_{\max}}{\sigma_{\min\text{ nonzero}}}.
\end{equation}
A coarse condition-number proxy. Large $\kappa$ signals a near-degenerate spectrum in which a few directions dominate.

\textbf{$\boldsymbol{\alpha_{\mathrm{req}}}$} \citep{agrawal2022alphareq}.
\begin{equation}
\alpha_{\mathrm{req}}(X) \;=\; -\hat{\beta}_1,
\qquad
\log \lambda_i \;=\; \beta_0 + \beta_1 \log i + \epsilon_i,\quad i=1,\ldots,r,
\end{equation}
where $(\hat{\beta}_0, \hat{\beta}_1)$ is the ordinary least-squares estimator. That is, $\alpha_{\mathrm{req}}$ is the decay exponent of a power-law fit $\lambda_i \propto i^{-\alpha}$ to the centred-covariance spectrum. Larger $\alpha_{\mathrm{req}}$ means a faster-decaying, more heavy-tailed spectrum.

\emph{Convention note on $\alpha_{\mathrm{req}}$.} We follow the original \citet{agrawal2022alphareq} definition, which fits the slope on covariance eigenvalues $\lambda_i$. \citet{tsitsulin2023} restate the same metric on singular values $\sigma_i$. Because $\lambda_i = \sigma_i^2 / n$, the two fitted slopes differ by exactly a factor of two, so our reported $\alpha_{\mathrm{req}}$ values live on the Agrawal scale and should be divided by $2$ to compare against Tsitsulin-scale results. To prevent power-law fits through float-precision noise on (nearly) collapsed spectra we drop eigenvalues below $\varepsilon \cdot d \cdot \lambda_1$ (numpy's default rank tolerance) before the fit. When fewer than two eigenvalues survive we record $\alpha_{\mathrm{req}} = \mathrm{NaN}$ rather than a spurious finite slope.

\textbf{Spatial Structure.} Two embeddings can share an identical singular-value spectrum yet lay their points out very differently: one spreading them uniformly, another concentrating them on a low-dimensional manifold or in tight clusters. Spatial Structure diagnostics measure these point-cloud-level properties, which are invisible to the spectrum alone and reveal whether a representation has acquired a meaningful geometric organisation (beneficial for retrieval- and clustering-style downstream tasks) or has instead collapsed structure that a purely spectral view would not detect.

\textbf{$\boldsymbol{\hat{d}_{\mathrm{TwoNN}}}$} \citep{facco2017twonn,ansuini2019intrinsic}.
\begin{equation}
\hat{d}_{\mathrm{TwoNN}}(X) \;=\; \hat{\beta}_1,
\qquad
-\log\!\bigl(1 - \tfrac{k}{N}\bigr) \;=\; \beta_0 + \beta_1 \log \mu_{(k)} + \epsilon_k,\quad k=1,\ldots,\lfloor 0.9 N\rfloor,
\end{equation}
where $(\hat{\beta}_0, \hat{\beta}_1)$ is the ordinary least-squares estimator, $\mu_i = d_{2,i}/d_{1,i}$ with $d_{1,i}, d_{2,i}$ the distances from row $i$ to its first and second nearest neighbours, $\mu_{(k)}$ the sorted statistic, and $N$ the sample size. This is the manifold-hypothesis \emph{intrinsic dimension}: the number of independent directions the data actually spans. We average over $20$ random $90\%$ subsets of at most $2000$ rows (seed $42$).

\textbf{Coherence $\boldsymbol{\mu_0}$.}
\begin{equation}
\mu_0(X) \;=\; \max\!\left(\frac{n}{r}\,\max_i \|U_{i,:}\|_2^2,\;\; \frac{d}{r}\,\max_j \|V_{j,:}\|_2^2\right),
\end{equation}
computed on the top-$r$ singular vectors. Coherence is high when a small number of rows or dimensions disproportionately drive the representation, and low when energy is spread evenly across all of them.

\textbf{Self-Cluster} \citep{tsitsulin2023}.
\begin{equation}
\mathrm{SC}(X) \;=\; \frac{d\,\|\tilde{X}\tilde{X}^{\top}\|_F^2 - n(d + n - 1)}{(d - 1)(n - 1)\,n},
\qquad \tilde{X}_{i,:} = \frac{X_{i,:}}{\|X_{i,:}\|_2}.
\end{equation}
Zero in expectation when the row-normalised vectors are i.i.d.\ isotropic, positive under clustering, so it measures how much the embedding departs from a uniform distribution on the sphere. Rows with zero norm (from upstream non-finite sanitisation) are dropped before normalisation.
Table~\ref{tab:taskfree_overview} reports the Spearman rank correlation ($\rho$) between eight diagnostics, grouped into the three families above, and downstream performance on the two row-level task categories. Each cell is a \emph{per-task average}: we compute Spearman $\rho$ within every task (across 13--14 row-capable models per task after task-specific filtering, with \textsc{random} excluded throughout), then average the per-task $\rho$'s. The $p$-values come from Wilcoxon signed-rank tests of the per-task $\rho$ distribution against zero. This matches the aggregation used in the pertask breakdown tables (Tabs.~\ref{tab:corr_record_linkage}--\ref{tab:corr_regression}), so the headline numbers and the per-head detail tell a single consistent story. The \textbf{Row Prediction} column combines classification and regression tasks, using macro-$F_1$ as the performance-oriented score for classification rows and $-\text{nRMSE}$ for regression rows. The per-task-type breakdowns (pure-classification vs.\ pure-regression, per-head) are in Tables~\ref{tab:corr_classification} and~\ref{tab:corr_regression}.
Table~\ref{tab:taskfree_rp_reg} provides the per-task breakdown for Row Prediction (Regression) with MLP and linear heads side by side.
Tables~\ref{tab:corr_record_linkage}--\ref{tab:corr_regression} report per-head detail for all three task types, combining MLP and Linear heads side by side for direct comparison.

\begin{figure*}[t]
\centering
\begin{subfigure}[t]{0.32\textwidth}
  \centering
  \includegraphics[width=\linewidth]{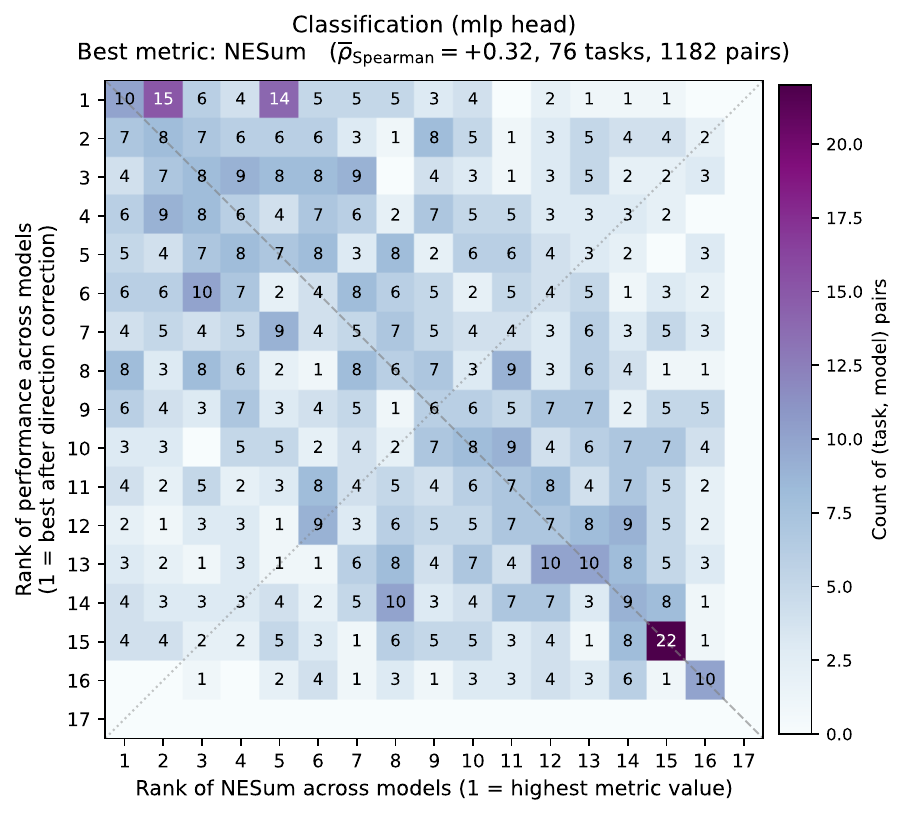}
  \caption{\textbf{Classification} (MLP). Best-correlated prior metric: NESum ($\bar{\rho}=+0.32$, 76 tasks, with one of 77 classification targets dropped for constant performance across models on MLP). Density on the main diagonal.}
  \label{fig:rankrank_classification}
\end{subfigure}
\hfil
\begin{subfigure}[t]{0.32\textwidth}
  \centering
  \includegraphics[width=\linewidth]{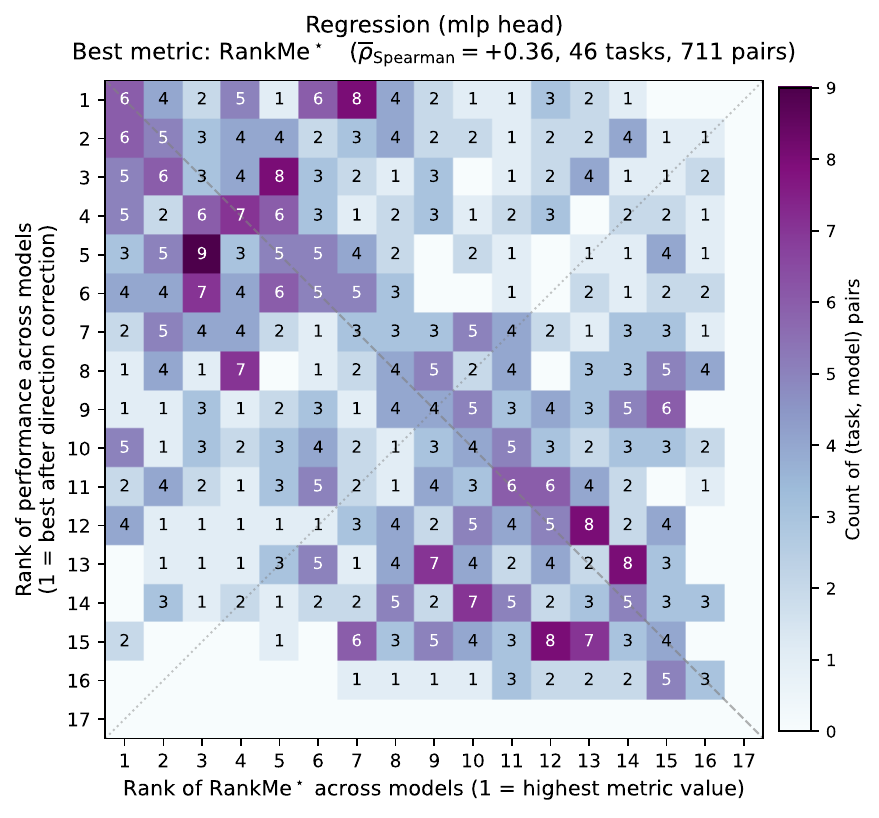}
  \caption{\textbf{Regression} (MLP). Best-correlated prior metric: RankMe* ($\bar{\rho}=+0.36$, 46 tasks). Density on the main diagonal.}
  \label{fig:rankrank_regression}
\end{subfigure}
\hfil
\begin{subfigure}[t]{0.32\textwidth}
  \centering
  \includegraphics[width=\linewidth]{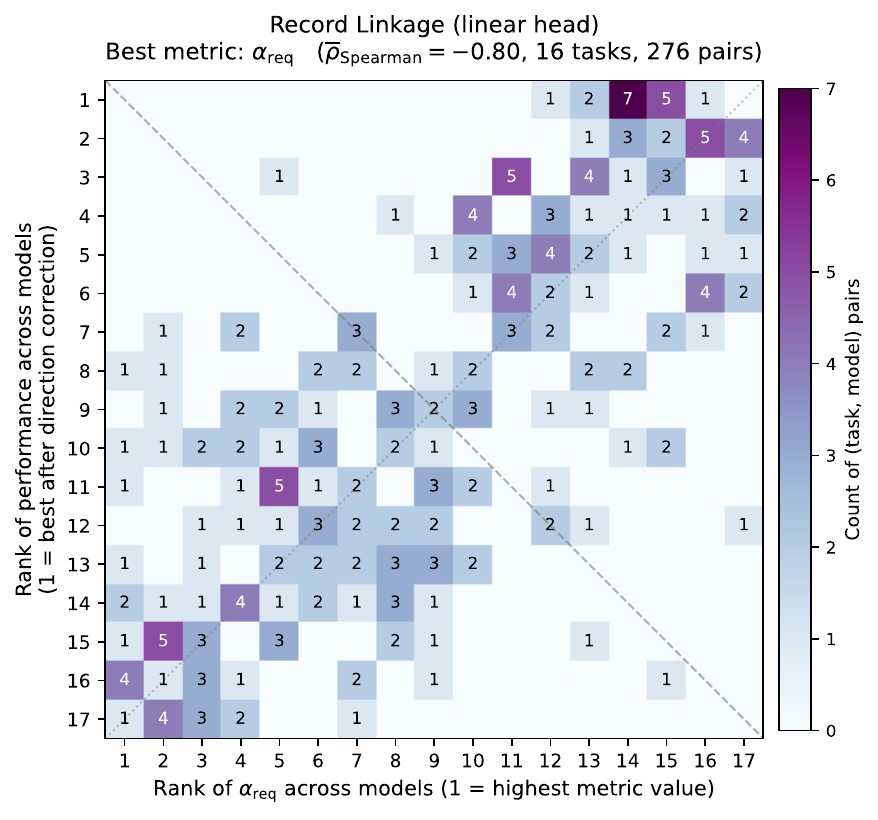}
  \caption{\textbf{Record Linkage} (Linear). Best-correlated prior metric: $\alpha_{\mathrm{req}}$ ($\bar{\rho}=-0.80$, 16 tasks). Density on the anti-diagonal.}
  \label{fig:rankrank_record_linkage}
\end{subfigure}
\caption{\textbf{Rank--rank density heatmaps for row tasks.} For every task, models are ranked by the best-correlated intrinsic-geometry diagnostic (x-axis, $1=$ highest value) and by direction-corrected downstream performance (y-axis, $1=$ best). Cell numbers count $(\text{task},\text{model})$ pairs per rank bin. Diagonal concentration indicates positive rank agreement. Anti-diagonal concentration indicates negative agreement.}
\label{fig:rankrank_all}
\end{figure*}

\input{tables/taskfree_overview}

\textbf{Spectral Spread is the most predictive family.}
RankMe ($\bar\rho = +0.485$, $p < 0.001$), RankMe* ($+0.471$), and NESum ($+0.460$) are the strongest observed positive correlates: embeddings with more uniform singular-value distributions (i.e., higher effective rank) consistently rank higher on both tasks.
The $\alpha_{\mathrm{req}}$ metric in Spectral Shape is the strongest observed negative correlate overall ($\bar\rho = -0.372$), but its signal is concentrated on Record Linkage ($\rho = -0.746$, $p < 0.001$). On Row Prediction the per-task mean is essentially zero ($\rho = +0.003$, $p = 0.94$) because the classification and regression signs partially cancel (see the per-head breakdown in Tab.~\ref{tab:taskfree_rp_reg} and the companion classification table). Heavy-tailed spectral decay therefore predicts entity-matching performance but is an ambivalent signal for feature-based prediction.
Within Spatial Structure, $\hat{d}_{\mathrm{TwoNN}}$ is informative for Record Linkage ($\rho = +0.398$, $p < 0.001$) but shows no significant signal for Row Prediction ($\rho = -0.042$, $p = 0.095$), indicating task-specific utility.
Correlations are uniformly stronger for Record Linkage than Row Prediction, implying that intrinsic quality metrics are better proxies for entity-matching tasks than for feature-based prediction.

\input{tables/taskfree_rp_reg}

\textbf{Row Prediction (Regression) is the most variable.}
Spectral Spread is the strongest positive signal: RankMe and RankMe* reach MLP Pearson $+0.317$/$+0.341$ (Spearman $+0.328$/$+0.356$, ${\sim}80\%$ directional), with the linear head slightly weaker (Pearson $+0.235$/$+0.261$). NESum follows closely (MLP Pearson $+0.248$, Spearman $+0.250$).
Spectral Shape is much weaker: Pseudo $\kappa$ registers MLP Pearson $+0.118$ (Spearman $+0.036$), and $\alpha_{\mathrm{req}}$ is essentially flat on the MLP head (Pearson $+0.012$, Spearman $+0.080$), emerging only on the linear head (Pearson $+0.119$, Spearman $+0.186$).
The dominant negative signal is in Spatial Structure: Self-Cluster reaches MLP Pearson $-0.316$ and Spearman $-0.327$ (${\sim}80\%$/${83\%}$ directional), meaning that tighter intra-model clustering of row embeddings co-varies with \emph{worse} regression. $\hat{d}_{\mathrm{TwoNN}}$ is consistently weakly negative (MLP Pearson $-0.120$, Spearman $-0.105$), and Coherence $\mu_0$ is near-zero.
Relative to classification, regression amplifies both the Spectral Spread positive signal and the Self-Cluster / $\hat{d}_{\mathrm{TwoNN}}$ negative signals. $\alpha_{\mathrm{req}}$ flips from weakly negative (classification, MLP Pearson $-0.105$) to flat or weakly positive (regression).
\emph{Note:} the per-task correlations here (and in Tables~\ref{tab:corr_record_linkage}--\ref{tab:corr_regression}) exclude the \textsc{random} baseline, matching Table~\ref{tab:taskfree_overview}. \textsc{random} is a high-leverage NESum outlier ($\overline{\mathrm{NESum}}\approx 409$ vs.\ ${\approx}4.1$ for real encoders, roughly $10^{2}\times$ on this aggregation basis) and can exert undue leverage on scale-sensitive Pearson fits.

\clearpage
\subsection{Intrinsic-Geometry Diagnostics: Per-Head Breakdowns}
\label{app:taskfree_pertask}

The following three tables combine MLP and Linear heads side by side for each task type, enabling direct comparison of how probe complexity modulates the predictive power of each quality metric.

\input{tables/taskfree_pertask}

\clearpage
\section{DLTE Operator Specification}
\label{app:dlte-operators}

This section specifies the complete DLTE pipeline operators summarized in Sec.~\ref{sec:dlte}. Algorithm~\ref{alg:dlte} is the end-to-end procedure. The following subsections give the CSLS formula, Stage-2 threshold grid, Stage-3 match-profile scalars, and the union-appended-row second-pass in Stage~3. Nothing below is trained end-to-end: Stage-2 calibrates five scalars \emph{per (Stage-1 backbone, column model) pair} on the dev split by grid search over macro-$F_1$. Stage-1 and Stage-3 have no tunable parameters beyond the fixed profile scalars reported here.

\begin{algorithm}[H]
\small
\caption{\textsc{TRL-DLTE} pipeline (per query $q$, lake $\mathcal{L}$).}
\label{alg:dlte}
\begin{algorithmic}[1]
\Require Frozen encoder outputs $e^{\mathrm{tbl}}, e^{\mathrm{col}}, e^{\mathrm{row}}$; retrieval depth $K=100$;
\Statex \hspace{\algorithmicindent}Stage-2 thresholds $\tau_{\mathrm{floor}},\tau_{\mathrm{u}},\tau_{\mathrm{us}},\tau_{\mathrm{jm}},\tau_{\mathrm{ks}}$ calibrated per (Stage-1 backbone, column model) pair (Table~\ref{tab:dlte_s2_grid});
\Statex \hspace{\algorithmicindent}Stage-3 match profiles $\Pi^{\mathrm{u}}$ (union) and $\Pi^{\mathrm{j}}$ (join) from Table~\ref{tab:dlte_s3_profiles}.
\State \textbf{Stage~1 (table retrieval).} L2-normalize $e^{\mathrm{tbl}}(q)$ and query the pre-built FAISS~\cite{faiss} inner-product index over the lake's normalized table embeddings; return top-$K$ candidates $\mathcal{C}_K(q)$.
\State \textbf{Stage~2 (column alignment + relation classification).}
\ForAll{$c \in \mathcal{C}_K(q)$}
  \State Form cost $D_{ij}=1-\cos(e^{\mathrm{col}}_{q,i},e^{\mathrm{col}}_{c,j})$; solve Hungarian assignment on $D$ to get matched cosine similarities $S=(s_1,\ldots,s_L)$.
  \State Let $n_\star=|\{s_k\ge\tau_{\mathrm{floor}}\}|$, $r=n_\star/|C(q)|$, $\mu=\mathrm{mean}\{s_k:s_k\ge\tau_{\mathrm{floor}}\}$, $m=\max S$.
  \If{$r \ge \tau_{\mathrm{u}}$ \textbf{and} $\mu \ge \tau_{\mathrm{us}}$}
    \State $\hat y(q,c) \gets \textsc{union}$.
  \ElsIf{$r \le \tau_{\mathrm{jm}}$ \textbf{and} $m \ge \tau_{\mathrm{ks}}$ \textbf{and} $n_\star\in\{1,2,3\}$}
    \State $\hat y(q,c) \gets \textsc{join}$;\ \ key pair $\gets \arg\max_k s_k$.
  \Else
    \State $\hat y(q,c) \gets \textsc{none}$.
  \EndIf
\EndFor
\State \textbf{Stage~3 (row matching + merge).}
\State Pick $c^{\mathrm{u}} \gets \arg\max_{\hat y(q,c)=\textsc{union}} \kappa_{\mathrm{u}}(q,c)$ and $c^{\mathrm{j}} \gets \arg\max_{\hat y(q,c)=\textsc{join}} \kappa_{\mathrm{j}}(q,c)$ over $\mathcal{C}_K(q)$, if any; ties broken by Stage-1 rank (see Sec.~\ref{app:dlte_op_stage3} for $\kappa$).
\State Initialize $q' \gets q$.
\If{$c^{\mathrm{u}}$ exists}
  \State \emph{Dedup union.} Reciprocal-match $R(c^{\mathrm{u}})$ against $R(q)$ via CSLS + profile $\Pi^{\mathrm{u}}$; append unmatched rows of $c^{\mathrm{u}}$ to $q'$ using the Stage-2 column alignment.
\EndIf
\If{$c^{\mathrm{j}}$ exists}
  \State \emph{Join.} Iteratively reciprocal-match seed rows $R(q)$ against $R(c^{\mathrm{j}})$ via CSLS + $\Pi^{\mathrm{j}}$ for up to $I^{\mathrm{j}}=10$ rounds; copy non-key join columns into $q'$.
  \State \emph{Second pass.} Reciprocal-match union-appended rows against as-yet-unmatched $R(c^{\mathrm{j}})$ (Sec.~\ref{app:dlte_op_second_pass}).
\EndIf
\State \Return enriched query $q'$.
\end{algorithmic}
\end{algorithm}

\subsection{Stage-1 Retrieval: Scoring and Pool}
\label{app:dlte_op_stage1}

Given table embeddings $e^{\mathrm{tbl}}(t)\in\mathbb{R}^{d}$, we L2-normalize each vector and build a FAISS \texttt{IndexFlatIP} over the 47{,}772-table lake. Inner-product search on unit vectors returns cosine-ranked candidates without approximation error. For the Stage-1 retrieval pool specifically, column-capable encoders are pooled to a table embedding via column mean (variant \texttt{column\_mean}). The two native table encoders use their native variants (\textsc{TAPEX}: \texttt{table\_embedding}, \textsc{TUTA}: \texttt{cls\_embedding}). We retrieve the top $K=100$ candidates and pass all of them to Stage~2.

\subsection{Stage-2 Alignment and Classification: Thresholds and Grid}
\label{app:dlte_op_stage2}

Hungarian assignment is solved with \texttt{scipy.optimize.linear\_sum\_assignment} on cost $D_{ij}=1-\cos\!\bigl(e^{\mathrm{col}}_{q,i},e^{\mathrm{col}}_{c,j}\bigr)$ with $L=\min(|C(q)|,|C(c)|)$ matched pairs. Column embeddings are L2-normalized before cosine. \textsc{none} is the default label and the per-pair statistics used for classification are the matched similarities $S$ and derived scalars $n_\star, r, \mu, m$ defined in Algorithm~\ref{alg:dlte}. When $n_\star=0$ (no matched pair clears $\tau_{\mathrm{floor}}$) we set $\mu \gets 0$ by convention.

Five thresholds are grid-searched \emph{per (Stage-1 backbone, column model) pair} on the \emph{dev} split to maximize three-way macro-$F_1$ over $\{\textsc{union},\textsc{join},\textsc{none}\}$. The objective is macro-$F_1$ because $\textsc{none}$ dominates (${\sim}98\%$ of pairs) and accuracy optimization collapses to the majority class. Calibration is per-pair because different Stage-1 backbones yield different candidate distributions entering Stage-2, so a per-pair operating point isolates Stage-2 classification quality conditional on the retrieval geometry rather than conflating the two. The grid (Table~\ref{tab:dlte_s2_grid}) visits $5\times6\times6\times5\times5=4500$ combinations per pair across $10\times8=80$ (Stage-1, Stage-2) pairs; total calibration cost is small because per-pair alignments are computed once and only the threshold-dependent per-pair statistics are revisited across the grid. The resulting threshold vector is held fixed across all $14$ Stage-3 row models paired with that (Stage-1, Stage-2) pair, and is stored alongside the predictions in the released code. Both calibrated thresholds and downstream metrics are deterministic given the embeddings.

\begin{table}[h]
\centering
\small
\setlength{\tabcolsep}{10pt}
\caption{Stage-2 threshold search space. Reported results always use the dev-selected threshold vector of the corresponding (Stage-1, Stage-2) pair. No shared default vector is used in evaluation.}
\label{tab:dlte_s2_grid}
\begin{tabular}{lll}
\toprule
\textbf{Symbol} & \textbf{Interpretation} & \textbf{Grid range (step)} \\
\midrule
$\tau_{\mathrm{floor}}$ & Min.\ per-pair similarity to count as matched & $[0.70, 0.90]\ (0.05)$ \\
$\tau_{\mathrm{u}}$     & Min.\ match ratio $r$ for \textsc{union}      & $[0.50, 1.00]\ (0.10)$ \\
$\tau_{\mathrm{us}}$    & Min.\ mean matched similarity $\mu$ for \textsc{union} & $[0.70, 0.95]\ (0.05)$ \\
$\tau_{\mathrm{jm}}$    & Max.\ match ratio $r$ for \textsc{join}       & $[0.20, 0.60]\ (0.10)$ \\
$\tau_{\mathrm{ks}}$    & Min.\ key-column similarity $m$ for \textsc{join} & $[0.75, 0.95]\ (0.05)$ \\
\bottomrule
\end{tabular}
\end{table}

\subsection{Stage-3 Row Matching: CSLS and Profiles}
\label{app:dlte_op_stage3}

\paragraph{Candidate-selection scores.}
Stage~3 consumes, for each query $q$, the highest-confidence \textsc{union} and \textsc{join} candidate from the $K$ Stage-2 outputs. We use a class-conditional score $\kappa$ matched to the decision rule: for a candidate classified as \textsc{union} we set $\kappa_{\mathrm{u}}(q,c)=\min(r,\mu)$ (the binding scalar of the union rule), and for \textsc{join} we set $\kappa_{\mathrm{j}}(q,c)=m$ (the key-column similarity). Ties are broken by Stage-1 retrieval rank. All Stage-3 operations reuse the precomputed frozen row embeddings of the source tables; after union append or join merge, the enriched table $q'$ is never re-encoded.

\paragraph{CSLS similarity.}
Let $M\in\mathbb{R}^{|A|\times|B|}$ be the raw cosine matrix between two row sets $A,B$. For $k_{\mathrm{csls}}=5$, let $r_i$ be the mean of the top-$k_{\mathrm{csls}}$ entries of row $i$ of $M$ and $c_j$ the mean of the top-$k_{\mathrm{csls}}$ entries of column $j$. The CSLS-normalized similarity is
\begin{equation}
  s_{\mathrm{CSLS}}(i,j) \;=\; 2\,M_{ij} \;-\; r_i \;-\; c_j.
  \label{eq:csls}
\end{equation}
CSLS~\cite{csls} discounts hub-like rows/columns whose neighborhoods are dense and makes mutual top-1 pairs more robustly reciprocal.

\paragraph{Reciprocal matching with local confidence filters.}
Given $s_{\mathrm{CSLS}}$, a pair $(i,j)$ is \emph{mutual top-1} if $j=\arg\max_{j'}s_{\mathrm{CSLS}}(i,j')$ and $i=\arg\max_{i'}s_{\mathrm{CSLS}}(i',j)$. For each row $i$, let $s_{i(1)}\ge s_{i(2)}$ denote its top-two CSLS scores and let $\mu_i,\sigma_i$ be the mean and standard deviation of the $i$-th row of $s_{\mathrm{CSLS}}$. We define the standardized signals
\begin{equation}
  z^{\text{best}}_i = \frac{s_{i(1)}-\mu_i}{\sigma_i^{+}}, \qquad
  z^{\text{margin}}_i = \frac{s_{i(1)}-s_{i(2)}}{\sigma_i^{+}}, \qquad
  \sigma_i^{+} \equiv
  \begin{cases} \sigma_i & \text{if } \sigma_i \ge \epsilon\\ 1 & \text{otherwise}\end{cases}
  \label{eq:zfilter}
\end{equation}
with $\epsilon=10^{-12}$ (the second case only fires on degenerate rows where every CSLS score is numerically identical), and the analogous candidate-side quantities $z^{\text{best}}_j, z^{\text{margin}}_j$ computed over the column axis of $s_{\mathrm{CSLS}}$. A mutual top-1 pair $(i,j)$ is \emph{accepted} iff $\min(z^{\text{best}}_i, z^{\text{best}}_j) \ge z^{\text{best}}_{\min}$ and $\min(z^{\text{margin}}_i, z^{\text{margin}}_j) \ge z^{\text{margin}}_{\min}$. Iterative matching removes accepted pairs and re-computes $s_{\mathrm{CSLS}}$ on the remaining rows/columns for up to a profile-specific number of rounds.

\paragraph{Profiles $\Pi^{\mathrm{u}}$ and $\Pi^{\mathrm{j}}$.}
The two enrichment paths have asymmetric costs: false union appends (duplicate rows) are costlier than missed joins (unmatched new columns on some rows). We therefore use two fixed profile vectors (Table~\ref{tab:dlte_s3_profiles}) rather than a single shared scalar. Profiles are \emph{not} tuned per model. They are held fixed across all Stage-3 row encoders, so the reported Stage-3 variation reflects row-embedding geometry rather than operator calibration.

\begin{table}[h]
\centering
\small
\setlength{\tabcolsep}{8pt}
\caption{Stage-3 match profiles. Both profiles use CSLS with $k_{\mathrm{csls}}=5$. Profiles are fixed across all row models.}
\label{tab:dlte_s3_profiles}
\begin{tabular}{lcc}
\toprule
\textbf{Scalar} & $\Pi^{\mathrm{u}}$ (union, precision-first) & $\Pi^{\mathrm{j}}$ (join, recall-first) \\
\midrule
$I_{\max}$ (iterations) & $3$ & $10$ \\
$z^{\text{best}}_{\min}$ & $1.00$ & $0.75$ \\
$z^{\text{margin}}_{\min}$ & $0.25$ & $0.10$ \\
$s_{\min}$ (absolute floor) & disabled & disabled \\
\bottomrule
\end{tabular}
\end{table}

\subsection{Stage-3 Second-Pass Join on Union-Appended Rows}
\label{app:dlte_op_second_pass}

The join phase first matches the seed rows of $q$ against $R(c^{\mathrm{j}})$. This leaves union-appended rows (new rows introduced by the union path) without join-side coverage. A second reciprocal pass then matches those appended rows against the \emph{remaining} rows of $c^{\mathrm{j}}$ (those not yet consumed by the first pass) under the same profile $\Pi^{\mathrm{j}}$. Cells are filled for newly matched pairs using the Stage-2 key-pair alignment, and the second-pass match count is logged separately. This mechanism is what produces the \emph{hard-region} recall (Table~\ref{tab:oracle_ra}) of the enriched-table quadrant where new rows meet new columns; without it, hard-region recall collapses to zero for all row models.

\subsection{What is and is not tuned}
\label{app:dlte_op_honesty}

Stages~1 and~3 are fully fixed across models and splits: all scalars in Table~\ref{tab:dlte_s3_profiles} and the FAISS retrieval depth $K$ are constants, independent of the encoder under evaluation. Stage~2 calibrates a five-scalar operating point per (Stage-1 backbone, column model) pair ($80$ calibration runs in total) on the dev split, using Stage-2 three-way macro-$F_1$ over $\{\textsc{union},\textsc{join},\textsc{none}\}$ as the sole objective. Neither test labels nor end-to-end Cell-$F_1$ / $\mathrm{UJ\text{-}H}$, and no Stage-3 row-model choice, enter this calibration. The resulting dev-selected threshold vector of each pair is held fixed across all $14$ Stage-3 row models paired with that (Stage-1, Stage-2) pair throughout the full pipeline evaluation. Headline pipeline selection (Sec.~\ref{sec:exp-dlte}) is a separate model-selection step that uses dev $\mathrm{UJ\text{-}H}$ as the sole criterion and does not enter this Stage-2 calibration.

\section{DLTE Detailed Rankings}
\label{app:dlte-rankings}

\begin{figure*}[!h]
\centering
\begin{subfigure}[t]{0.49\textwidth}
  \centering
  \includegraphics[width=\linewidth]{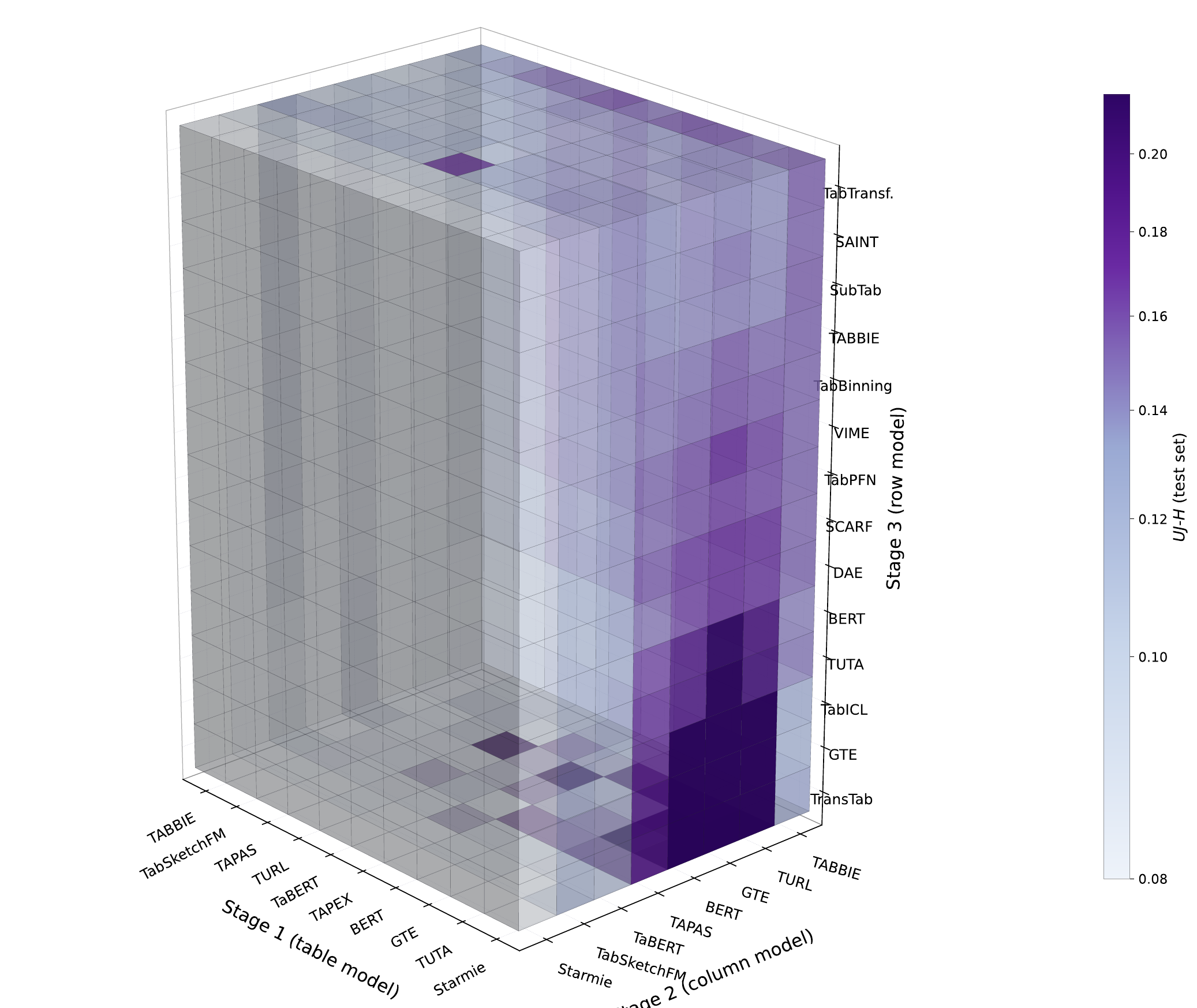}
  \caption{Full cube: all $10 \times 8 \times 14 = 1120$ pipelines. Low-$\mathrm{UJ\text{-}H}$ voxels fade into light blue, high-$\mathrm{UJ\text{-}H}$ voxels pop in deep purple.}
  \label{fig:dlte_voxel_full}
\end{subfigure}
\hfill
\begin{subfigure}[t]{0.49\textwidth}
  \centering
  \includegraphics[width=\linewidth]{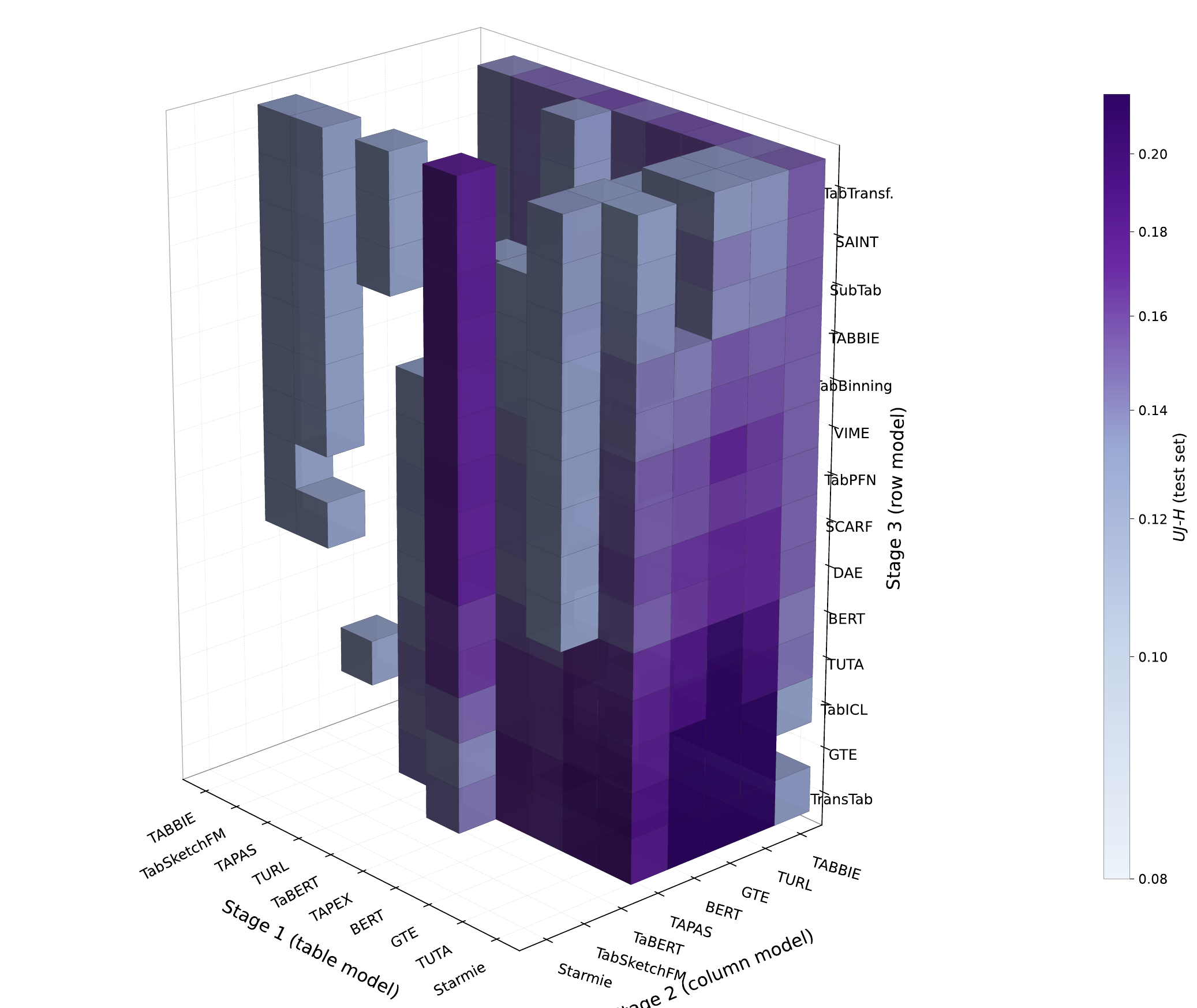}
  \caption{Top $40\%$ pipelines only ($\mathrm{UJ\text{-}H} \geq$ 60th percentile), highlighting the strong-performing regions of the cube.}
  \label{fig:dlte_voxel_top40}
\end{subfigure}
\caption{\textbf{Voxel visualisation of the DLTE Stage-3 pipeline space over $\mathrm{UJ\text{-}H}$.} Axes: Stage~1 (table model, $10$) $\times$ Stage~2 (column model, $8$) $\times$ Stage~3 (row model, $14$). Colour encodes $\mathrm{UJ\text{-}H}$ (light blue $\to$ deep purple). Axes are reordered by marginal-mean $\mathrm{UJ\text{-}H}$ so the best-performing corner is contiguous. See the full per-pipeline breakdown in Tables~\ref{tab:ablation_dlte3} (Cell~$F_1$) and~\ref{tab:ablation_dlte3_ujh} ($\mathrm{UJ\text{-}H}$) (Sec.~\ref{app:ablation_dlte3}).}
\label{fig:dlte_voxel}
\end{figure*}

\paragraph{Dev/test rank stability.}
The dev-selection protocol of Sec.~\ref{sec:exp-dlte} relies on dev-test pipeline rank similarity. Across all 1{,}120 canonical pipelines (5-round mean $\mathrm{UJ\text{-}H}$ per pipeline), Spearman $\rho(\text{dev},\text{test}) = 0.96$ ($p \ll 10^{-100}$, $n=1{,}120$; Kendall $\tau = 0.84$), and the top-50 by dev $\mathrm{UJ\text{-}H}$ and the top-50 by test $\mathrm{UJ\text{-}H}$ share 42 of 50 pipelines. This justifies treating dev-selected pipelines as descriptive of the broader test landscape.

\clearpage
\subsection{Cell $F_1$ as a Complementary Diagnostic}
\label{app:cellf1}

Cell~$F_1$ is the multiset $F_1$ score over recovered cells, pooling the removed-row and removed-column blocks for each query. With $C_p(q)$ and $C_g(q)$ the multisets of cells in the pipeline's predicted enrichment and the ground-truth blocks for query $q$,
\[
  \mathrm{Cell}\,F_1(q) \;=\; \frac{2\,|C_p(q) \cap C_g(q)|}{|C_p(q)| + |C_g(q)|}\,,
\]
averaged over queries. It measures how well a pipeline reconstructs parent-table cells in raw cell terms, regardless of how that recovery is distributed between the union and join paths. This makes Cell~$F_1$ a complement to $\mathrm{UJ\text{-}H}$, the primary end-to-end score for joint recovery of the union and join targets, and we use it here to diagnose stage behavior and high-volume recovery modes in the same 1{,}120-pipeline space.

\paragraph{Per-stage observations.}
At Stage~1 (table retrieval), the Cell~$F_1$ marginal identifies \textsc{Starmie} as the strongest retriever (0.601), followed by \textsc{TUTA} (0.593) and \textsc{GTE} (0.591). This agrees with the top of the $\mathrm{UJ\text{-}H}$ marginal ranking and shows that retrieval quality is a shared driver under both pooled-cell and joint-recovery views (Table~\ref{tab:dlte_marginal}). At Stage~2 (column alignment plus union/join/none decisions), \textsc{TaBERT} leads on Cell~$F_1$ (0.628), ahead of \textsc{GTE} (0.601) and \textsc{TAPAS} (0.600). The Stage~2 Cell~$F_1$ span is 0.084, the largest of the three stages, indicating that pooled cell recovery is especially sensitive to the column-side model. At Stage~3 (row matching and merge), the Cell~$F_1$ marginals are compressed (span 0.026), with \textsc{TabTransformer} (0.591), \textsc{SubTab} (0.591), and \textsc{SAINT} (0.590) leading. Oracle-RA (Table~\ref{tab:oracle_ra}) clarifies the mechanism: these row models obtain their Cell~$F_1$ primarily from near-complete union-side recovery when retrieval and alignment are supplied, separating a union-preservation behavior from the identity-resolution behavior surfaced by Robust Linkage and Oracle-RA $\mathrm{UJ\text{-}H}$.

\paragraph{Pipeline-level signature.}
The highest-Cell~$F_1$ pipelines share a consistent composition: all top-20 use \textsc{TaBERT} at Stage~2, and Stage~3 concentrates on \textsc{TabTransformer}, \textsc{SubTab}, \textsc{SAINT}, and \textsc{TABBIE} (Table~\ref{tab:dlte_top20_cf1}). The best Cell~$F_1$ pipeline is \textsc{Starmie}/\textsc{TaBERT}/\textsc{TabTransformer} at 0.679. The signature is therefore strong table retrieval, \textsc{TaBERT} column-side decisions, and row models that preserve union-side cells. This is an auxiliary lens for workloads that prioritize total recovered cell yield, or for diagnosing which stage limits pooled cell recovery. Overall, Cell~$F_1$ adds a practical diagnostic layer: it confirms the Stage~1 retrieval signal, identifies pooled-cell yield as most sensitive to Stage~2, and exposes a Stage~3 union-preservation mode. The full marginal, pipeline, Oracle-RA, and source-split tables report both metrics so the joint-recovery and pooled-cell views can be read directly (Tables~\ref{tab:dlte_marginal}, \ref{tab:dlte_stage1_full}--\ref{tab:dlte_stage3_full}, \ref{tab:dlte_top20_cf1}, \ref{tab:ablation_dlte3}, \ref{tab:oracle_ra}; Appendix~\ref{app:dlte-source-split}).

\clearpage
\subsection{Pipeline Component Sensitivity}
\label{app:ablation_dlte_pipeline}

\input{tables/ablation_dlte_pipeline}

Table~\ref{tab:ablation_dlte} reports end-to-end Cell~$F_1$ as a function of Stage~1 (retrieval) and Stage~2 (column alignment) model choices, with Stage~3 fixed to the best Cell~$F_1$ row model in this ablation, \textsc{TabTransformer}.
Among Stage~1 models, Starmie leads (row average 0.609), followed by TaBERT (0.600) and the native table encoder TUTA (0.599). TAPEX trails at 0.559 and TABBIE at 0.565, indicating that weak retrieval creates a hard ceiling for downstream performance. The 10-model Stage~1 pool covers all table-capable encoders: the 8 column-capable models (whose column embeddings are pooled to a table embedding) plus the two native table encoders TAPEX and TUTA.
Among Stage~2 models, TaBERT dominates column alignment (column average 0.650), far ahead of the next-best TAPAS (0.610). Starmie at Stage~2 collapses to a flat 0.544 regardless of retrieval model. The wide Stage~2 spread (0.106) confirms that column alignment choice has the largest downstream effect. These results isolate column alignment under a fixed Stage~3 model (\textsc{TabTransf.}). They are conditional, not the unconditional Stage-2 marginal used in Sec.~\ref{sec:exp-dlte} (which leads with \textsc{TABBIE} on test and \textsc{BERT} on dev). The discrepancy is expected under non-additive composition: a column model's apparent strength depends on the upstream retriever and downstream row matcher with which it is paired.

\subsection{Per-Stage Marginal Analysis}

Table~\ref{tab:dlte_marginal} decomposes end-to-end performance into per-stage marginal contributions. For a fixed table model, scores are averaged over all 112 compatible pipelines, for a fixed column model over all 140, and for a fixed row model over all 80.

\input{tables/dlte_marginal_table.tex}

Stage~2 (column model) exhibits the widest $\mathrm{UJ\text{-}H}$ span (0.060) and the widest Cell~$F_1$ span (0.084), confirming it has the largest average downstream effect under the current pipeline. Stage~1 (table model) shows a notable disconnect between retrieval recall and end-to-end contribution: \textsc{Starmie} ranks 1st on marginal Cell~$F_1$ despite only 3rd-best recall@100, while \textsc{GTE} achieves the highest recall (0.801) but ranks 3rd on Cell~$F_1$, behind \textsc{TUTA} (2nd at 0.593). \textsc{TAPEX} further illustrates the disconnect in the opposite direction: it has the lowest marginal Cell~$F_1$ (0.558) and near-lowest recall (0.247 R@100, above only \textsc{TABBIE}'s 0.108), consistent with poor retrieval limiting downstream quality. Stage~3 (row model) has the narrowest span (0.026 Cell~$F_1$) and an even smaller $\mathrm{UJ\text{-}H}$ span (0.013), confirming that upstream errors largely mask row-model differences in the full pipeline. Marginal rankings are main-effect summaries rather than globally optimal compositions. For a pipeline $p=(t,c,r)$ with end-to-end score $y(p)$, the per-stage marginals are $m_T(t)=\mathbb{E}_{c,r}[y(t,c,r)]$, $m_C(c)=\mathbb{E}_{t,r}[y(t,c,r)]$, and $m_R(r)=\mathbb{E}_{t,c}[y(t,c,r)]$, and the additive main-effect score $m_T(t)+m_C(c)+m_R(r)-2\bar{y}$ is maximized by the per-stage rank-1 assembly. On test, this assembly is \textsc{Starmie}/\textsc{TABBIE}/\textsc{TransTab} and scores 0.134 $\mathrm{UJ\text{-}H}$, while the test rank-1 pipeline \textsc{Starmie}/\textsc{GTE}/\textsc{GTE} scores 0.253 and the dev-selected headline \textsc{TUTA}/\textsc{GTE}/\textsc{GTE} scores 0.229. On development, the marginal-leader assembly changes to \textsc{Starmie}/\textsc{BERT}/\textsc{TransTab} and is competitive on test (0.231), confirming that marginal main effects carry signal. The change in the leader assembly across splits, \textsc{TABBIE}'s absence from the top-50, and the gap between the marginal-leader assembly and the end-to-end optima together show that the top of the DLTE space is shaped by residual non-additive stage interactions. We refer to this residual structure as compositional fit.

\begin{figure}[t]
\centering
\includegraphics[width=\textwidth]{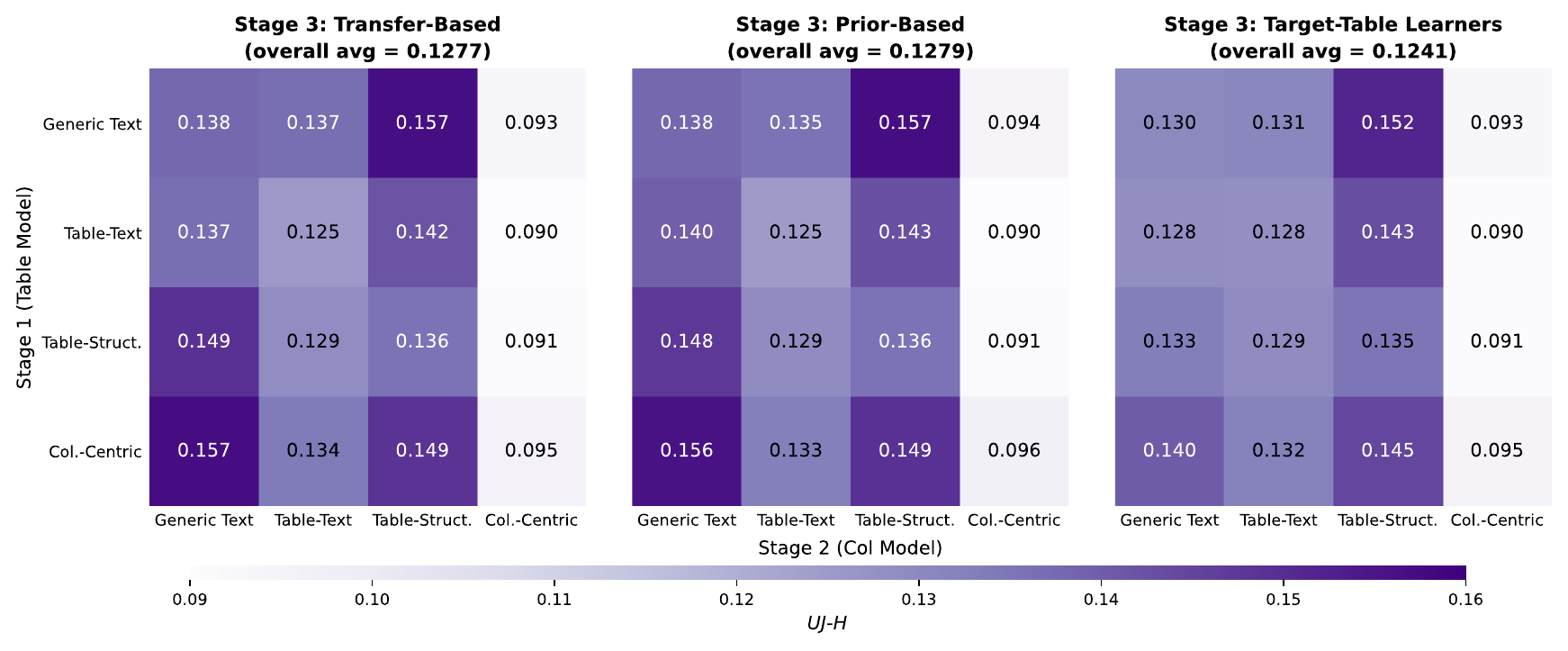}
\caption{DLTE category-level $\mathrm{UJ\text{-}H}$ heatmap (5-round average, test set). Panels correspond to Stage~3 (row model) families. Rows = Stage~1 (table model), columns = Stage~2 (column model). The column-driven gradient confirms Stage~2's dominant effect. Near-identical panels show Stage~3 differences are largely masked end to end.}
\label{fig:dlte_heatmap}
\end{figure}

\subsection{Oracle-RA Row-Model Diagnostic}
\label{app:oracle_ra}

Table~\ref{tab:oracle_ra} reports Oracle-RA results on the test split. Oracle-RA bypasses Stage~1 (retrieval) and Stage~2 (alignment) with ground-truth data, isolating Stage~3 row matching quality. The $\mathrm{UJ\text{-}H}$ spread across row models is 0.546, compared with a marginal end-to-end span of just 0.013 (Table~\ref{tab:dlte_marginal}), confirming that upstream errors mask most row-model differences in the full pipeline.

\begin{table}[h]
\centering
\small
\setlength{\tabcolsep}{4pt}
\caption{Oracle-RA row-model diagnostic (test set). Stages~1--2 use ground truth. Only Stage~3 row matching varies.}
\label{tab:oracle_ra}
\begin{tabular}{clccccc}
\toprule
\textbf{\#} & \textbf{Row Model} & \textbf{$\mathrm{UJ\text{-}H}$$\uparrow$} & \textbf{Cell~$F_1$$\uparrow$} & \textbf{Union$\uparrow$} & \textbf{Join$\uparrow$} & \textbf{Hard$\uparrow$} \\
\midrule
1  & GTE            & \textbf{0.683} & 0.802 & 0.696 & \textbf{0.763} & 0.550 \\
2  & TransTab       & 0.658 & 0.802 & 0.698 & 0.721 & 0.607 \\
3  & TabICL         & 0.606 & 0.789 & 0.609 & 0.664 & \textbf{0.737} \\
4  & TUTA           & 0.487 & 0.743 & 0.829 & 0.403 & 0.418 \\
5  & BERT           & 0.454 & 0.735 & 0.817 & 0.386 & 0.376 \\
6  & SCARF          & 0.340 & 0.709 & 0.864 & 0.246 & 0.318 \\
7  & DAE            & 0.333 & 0.713 & 0.893 & 0.234 & 0.305 \\
8  & VIME           & 0.318 & 0.710 & 0.900 & 0.223 & 0.289 \\
9  & TabPFN         & 0.293 & 0.706 & 0.925 & 0.196 & 0.256 \\
10 & TabBinning     & 0.259 & 0.697 & 0.902 & 0.181 & 0.232 \\
11 & TABBIE         & 0.231 & 0.695 & 0.935 & 0.154 & 0.197 \\
12 & SAINT          & 0.168 & 0.684 & 0.940 & 0.116 & 0.144 \\
13 & SubTab         & 0.164 & 0.685 & 0.957 & 0.109 & 0.133 \\
14 & TabTransformer & 0.137 & 0.680 & 0.961 & 0.092 & 0.111 \\
\bottomrule
\end{tabular}
\end{table}

\paragraph{Key observations.}
Row models divide into two groups: \emph{identity-resolving} models (GTE, TransTab, TabICL) that achieve balanced union/join recovery, and \emph{union-dedup specialists} (TabTransformer, SAINT, SubTab) with near-perfect union recall but near-zero join recall. This distinction is reflected in the Cell~$F_1$ / $\mathrm{UJ\text{-}H}$ pipeline-level pattern: Cell~$F_1$ captures pooled cell-recovery yield (well-served by high union recall), while $\mathrm{UJ\text{-}H}$ captures balanced recovery of both removed blocks and therefore requires both recalls to be high (Appendix~\ref{app:cellf1}). TabICL attains the best hard-region recall (0.737), showing that the second-pass join mechanism can recover cells in the new-rows$\times$new-columns quadrant.

\paragraph{Per-noise-tier breakdown.}
Table~\ref{tab:oracle_ra_per_tier_uj_h} decomposes Oracle-RA $\mathrm{UJ\text{-}H}$ across the four \textsc{TRL-DLTE} noise tiers (cumulative clean $\to$ schema $\to$ cell $\to$ hard; see Sec.~\ref{sec:dlte}). The key observation is that the cross-row-model $\mathrm{UJ\text{-}H}$ span is large and largely tier-invariant (0.562/0.563/0.506/0.553 for clean/schema/cell/hard), and the top/bottom row-model families are stable at every corruption level: the identity-resolving row models (\textsc{GTE}, \textsc{TransTab}, \textsc{TabICL}) occupy the top three positions in every tier, and the union-dedup specialists (\textsc{TabTransformer}, \textsc{SAINT}, \textsc{SubTab}) occupy the bottom three in every tier. The Stage~3 separability exposed by Oracle-RA is therefore not a noise-sensitivity artifact: it persists across all levels of upstream corruption. The fact that union-dedup specialists rank strongly on end-to-end Cell~$F_1$ (Table~\ref{tab:dlte_stage3_full}) but rank last under Oracle-RA on both $\mathrm{UJ\text{-}H}$ and join recall indicates that their end-to-end Cell~$F_1$ advantage reflects union-side recovery behavior under upstream error, not strong identity-resolution Stage~3 behavior on balanced enrichment.

\input{tables/oracle_ra_per_tier/oracle_ra_per_tier_uj_h_test.tex}

\paragraph{Cross-validation against RBench robust linkage.}
The identity-resolving/union-dedup split identified by Oracle-RA is not an artifact of the DLTE pipeline: the same taxonomy is visible in atomic cross-table record linkage. Over all 14 row models, the Oracle-RA $\mathrm{UJ\text{-}H}$ ranking and the RBench Robust Linkage NR ranking (aggregated over DM-D and WDC; Table~\ref{tab:r_main_results}) are strongly correlated: $|\rho_{\mathrm{Spearman}}| = 0.80$ ($p = 6.3{\times}10^{-4}$) and $|\tau_{\mathrm{Kendall}}| = 0.63$ ($p = 1.2{\times}10^{-3}$). The top two row models agree across both views (\textsc{GTE} and \textsc{TransTab}: Robust Linkage NR $=0.048$ and $0.096$, and Oracle-RA $\mathrm{UJ\text{-}H}$ also leads with these two), with \textsc{TabICL} third on Oracle-RA but fifth on Robust Linkage NR ($=0.394$), trailing \textsc{GTE} and \textsc{TransTab} as well as \textsc{TUTA} ($=0.154$) and \textsc{BERT} ($=0.163$). The union-dedup specialists highlighted by Oracle-RA anchor the other end: \textsc{SubTab} and \textsc{TabTransformer} occupy the bottom two positions of the Robust Linkage NR column (NR $= 0.962$, $0.942$), with their near-zero join recall in Oracle-RA mirrored by near-zero WDC $F_1$ (e.g., \textsc{TabTransformer} WDC $F_1 = 0.020$). \textsc{SAINT} sits at NR $= 0.606$, mid-pack on Robust Linkage despite its union-heavy profile. Rank orderings within each family differ for the same reason: DM-C (clean linkage) does not fully separate identity-resolution behavior, which is why \textsc{BERT} leads there but drops to 5th in Oracle-RA. Even so, the two tests agree at both ends of the ranking. This is consistent with a shared identity-resolution capability of frozen row embeddings, surfaced consistently by both entity matching (RBench) and compositional enrichment (DLTE).

\subsection{Full Per-Stage Model Rankings}

Tables~\ref{tab:dlte_stage1_full}--\ref{tab:dlte_stage3_full} report the complete marginal rankings for each DLTE stage (test split, 5-round mean $\pm$ std) over the full $10\times8\times14 = 1120$ table $\times$ column $\times$ row search space. The 10-model Stage~1 pool covers the 8 column-capable encoders (whose column embeddings are pooled to a table embedding) plus the two native table encoders \textsc{TAPEX} and \textsc{TUTA}. Stage~2 remains the 8 column-capable encoders since TAPEX/TUTA do not expose compatible column embeddings. For a fixed table model, scores are averaged over all $8\times14=112$ compatible pipelines, for a fixed column model over all $10\times14=140$ pipelines, and for a fixed row model over all $10\times8=80$ pipelines.

\input{tables/dlte_marginals/dlte_stage1_marginals.tex}

\input{tables/dlte_marginals/dlte_stage2_marginals.tex}

\input{tables/dlte_marginals/dlte_stage3_marginals.tex}

\paragraph{Retrieval recall does not dictate downstream contribution.}
Table~\ref{tab:dlte_stage1_full} exposes a non-monotone relationship between raw Stage~1 retrieval quality and downstream end-to-end contribution. \textsc{GTE} attains the highest recall-at-100 ($0.801$) among the ten table models but does not lead downstream. \textsc{Starmie} has lower recall-at-100 ($0.740$) yet attains the best downstream Cell~$F_1$ ($0.601$) and $\mathrm{UJ\text{-}H}$ ($0.144$). The ordering is neither retrieval-dominated nor inverse: \textsc{TABBIE} is worst on recall-at-100 ($0.108$) while \textsc{TAPEX} is lowest on Cell~$F_1$ ($0.558$), and \textsc{TaBERT} and \textsc{BERT} sit in the middle of both orderings. This non-monotone pattern shows that Stage~1's downstream contribution is not determined by recall volume alone. It also depends on the structure of the retrieved candidate set and on whether those candidates are usable by the downstream column-alignment and row-matching models. This is a Stage-1 instance of compositional fit beyond per-stage marginal rank.

\clearpage
\subsection{Stage~3 Row Matching: Full Pipeline Rankings}
\label{app:ablation_dlte3}

\input{tables/ablation_dlte3_merged}

Table~\ref{tab:ablation_dlte3} shows Cell~$F_1$ for all (Stage~1, Stage~2) configurations across all ten Stage~1 models, crossed with all 14 Stage~3 row models.
The best pipelines use TaBERT at Stage~2 (top rows of each group), confirming Stage~2 dominance regardless of Stage~1 choice.
Starmie leads as Stage~1 (best avg 0.650), while \textsc{TAPEX} has the lowest average Cell~$F_1$ as Stage~1 (0.558), narrowly below \textsc{TABBIE} (0.560), with both lagging the rest of the pool by a clear margin.
Within any given (Stage~1, Stage~2) pair, the Stage~3 spread is narrow: for the best configuration (Starmie~$\to$~TaBERT), Cell~$F_1$ ranges from 0.589--0.679, a span of only $\sim$0.090.
The global average Cell~$F_1$ by Stage~3 column is nearly flat (0.566--0.591), with SAINT, SubTab, and TabTransf.\ consistently at the top and TabICL at the bottom, confirming that Stage~3 row model choice has only marginal end-to-end impact and that, within this Cell~$F_1$ ablation, the dominant lever is the Stage~2 column alignment model.

\clearpage
\subsection{Top-20 Pipeline Combinations}

Tables~\ref{tab:dlte_top20_cf1} and~\ref{tab:dlte_top20_ujh} list the top-20 pipeline combinations by Cell~$F_1$ and $\mathrm{UJ\text{-}H}$, respectively. The two lists rank pipelines along different axes, consistent with their definitions (Appendix~\ref{app:cellf1}): Cell~$F_1$ captures pooled cell-recovery yield, while $\mathrm{UJ\text{-}H}$ captures balanced recovery of both removed blocks.

\begin{table*}[!h]
\centering
\scriptsize
\setlength{\tabcolsep}{3pt}
\caption{Top-20 DLTE combinations by Cell~$F_1$ (5-round average, test set). All entries use TaBERT as Stage~2 column model. Their $\mathrm{UJ\text{-}H}$ ranks (\#513--\#671) reflect the different axis Cell~$F_1$ captures relative to $\mathrm{UJ\text{-}H}$ (Appendix~\ref{app:cellf1}).}
\label{tab:dlte_top20_cf1}
\begin{tabular}{c lll lll cc c}
\toprule
\textbf{\#} & \textbf{Stage~1 (Tbl)} & \textbf{Fam.} & \textbf{Stage~2 (Col)} & \textbf{Fam.} & \textbf{Stage~3 (Row)} & \textbf{Fam.} & \textbf{Cell~$F_1$} & \textbf{$\mathrm{UJ\text{-}H}$} & \textbf{$\mathrm{UJ\text{-}H}$ \#} \\
\midrule
1  & Starmie & Col-Cen & TaBERT & Tbl-Txt & TabTransf. & Tgt-Tbl & \textbf{0.679} & 0.128 & \#532 \\
2  & Starmie & Col-Cen & TaBERT & Tbl-Txt & SubTab & Tgt-Tbl & 0.677 & 0.128 & \#523 \\
3  & GTE & Gen.\ Txt & TaBERT & Tbl-Txt & TabTransf. & Tgt-Tbl & 0.677 & 0.125 & \#589 \\
4  & TAPAS & Tbl-Txt & TaBERT & Tbl-Txt & TabTransf. & Tgt-Tbl & 0.676 & 0.127 & \#544 \\
5  & GTE & Gen.\ Txt & TaBERT & Tbl-Txt & SubTab & Tgt-Tbl & 0.676 & 0.125 & \#597 \\
6  & TaBERT & Tbl-Txt & TaBERT & Tbl-Txt & TabTransf. & Tgt-Tbl & 0.676 & 0.125 & \#608 \\
7  & TAPAS & Tbl-Txt & TaBERT & Tbl-Txt & SubTab & Tgt-Tbl & 0.675 & 0.127 & \#549 \\
8  & Starmie & Col-Cen & TaBERT & Tbl-Txt & SAINT & Tgt-Tbl & 0.674 & 0.126 & \#573 \\
9  & TaBERT & Tbl-Txt & TaBERT & Tbl-Txt & SubTab & Tgt-Tbl & 0.674 & 0.125 & \#616 \\
10 & BERT & Gen.\ Txt & TaBERT & Tbl-Txt & TabTransf. & Tgt-Tbl & 0.673 & 0.126 & \#575 \\
11 & GTE & Gen.\ Txt & TaBERT & Tbl-Txt & SAINT & Tgt-Tbl & 0.672 & 0.123 & \#656 \\
12 & BERT & Gen.\ Txt & TaBERT & Tbl-Txt & SubTab & Tgt-Tbl & 0.672 & 0.126 & \#577 \\
13 & TAPAS & Tbl-Txt & TaBERT & Tbl-Txt & SAINT & Tgt-Tbl & 0.672 & 0.126 & \#581 \\
14 & Starmie & Col-Cen & TaBERT & Tbl-Txt & TABBIE & Transfer & 0.671 & 0.127 & \#541 \\
15 & TaBERT & Tbl-Txt & TaBERT & Tbl-Txt & SAINT & Tgt-Tbl & 0.671 & 0.123 & \#671 \\
16 & Starmie & Col-Cen & TaBERT & Tbl-Txt & TabPFN & Prior & 0.670 & 0.129 & \#513 \\
17 & GTE & Gen.\ Txt & TaBERT & Tbl-Txt & TABBIE & Transfer & 0.669 & 0.124 & \#651 \\
18 & TAPAS & Tbl-Txt & TaBERT & Tbl-Txt & TABBIE & Transfer & 0.669 & 0.125 & \#596 \\
19 & TaBERT & Tbl-Txt & TaBERT & Tbl-Txt & TABBIE & Transfer & 0.669 & 0.123 & \#659 \\
20 & BERT & Gen.\ Txt & TaBERT & Tbl-Txt & SAINT & Tgt-Tbl & 0.669 & 0.124 & \#620 \\
\bottomrule
\end{tabular}
\end{table*}

\begin{table*}[!h]
\centering
\scriptsize
\setlength{\tabcolsep}{3pt}
\caption{Top-20 DLTE combinations by $\mathrm{UJ\text{-}H}$ (5-round average, test set). Stage~2 is dominated by GTE, BERT, and TURL. Stage~1 is led by Starmie (11 entries), with \textsc{TUTA} appearing 7 times (from the native table-encoder pool), \textsc{TURL} once (\#17), and \textsc{TAPEX} once (\#19). Zero overlap with the Cell~$F_1$ top-20 list (Table~\ref{tab:dlte_top20_cf1}).}
\label{tab:dlte_top20_ujh}
\begin{tabular}{c lll lll cc c}
\toprule
\textbf{\#} & \textbf{Stage~1 (Tbl)} & \textbf{Fam.} & \textbf{Stage~2 (Col)} & \textbf{Fam.} & \textbf{Stage~3 (Row)} & \textbf{Fam.} & \textbf{Cell~$F_1$} & \textbf{$\mathrm{UJ\text{-}H}$} & \textbf{CF1 \#} \\
\midrule
1  & Starmie & Col-Cen & GTE & Gen.\ Txt & GTE & Transfer & 0.632 & \textbf{0.253} & \#114 \\
2  & Starmie & Col-Cen & GTE & Gen.\ Txt & TransTab & Tgt-Tbl & 0.634 & 0.251 & \#108 \\
3  & Starmie & Col-Cen & GTE & Gen.\ Txt & TabICL & Prior & 0.614 & 0.236 & \#184 \\
4  & Starmie & Col-Cen & BERT & Gen.\ Txt & TransTab & Tgt-Tbl & 0.610 & 0.231 & \#226 \\
5  & TUTA & Tbl-Struct & GTE & Gen.\ Txt & GTE & Transfer & 0.621 & 0.229 & \#149 \\
6  & Starmie & Col-Cen & BERT & Gen.\ Txt & GTE & Transfer & 0.606 & 0.228 & \#313 \\
7  & Starmie & Col-Cen & BERT & Gen.\ Txt & TabICL & Prior & 0.595 & 0.228 & \#504 \\
8  & Starmie & Col-Cen & TURL & Tbl-Struct & GTE & Transfer & 0.612 & 0.225 & \#201 \\
9  & Starmie & Col-Cen & TURL & Tbl-Struct & TransTab & Tgt-Tbl & 0.613 & 0.225 & \#186 \\
10 & Starmie & Col-Cen & TURL & Tbl-Struct & TabICL & Prior & 0.601 & 0.220 & \#429 \\
11 & TUTA & Tbl-Struct & GTE & Gen.\ Txt & TransTab & Tgt-Tbl & 0.621 & 0.218 & \#153 \\
12 & TUTA & Tbl-Struct & BERT & Gen.\ Txt & GTE & Transfer & 0.603 & 0.216 & \#380 \\
13 & Starmie & Col-Cen & GTE & Gen.\ Txt & TUTA & Transfer & 0.639 & 0.215 & \#85 \\
14 & Starmie & Col-Cen & GTE & Gen.\ Txt & BERT & Transfer & 0.637 & 0.211 & \#97 \\
15 & TUTA & Tbl-Struct & BERT & Gen.\ Txt & TransTab & Tgt-Tbl & 0.600 & 0.210 & \#437 \\
16 & TUTA & Tbl-Struct & GTE & Gen.\ Txt & TabICL & Prior & 0.603 & 0.210 & \#386 \\
17 & TURL & Tbl-Struct & GTE & Gen.\ Txt & GTE & Transfer & 0.607 & 0.205 & \#284 \\
18 & TUTA & Tbl-Struct & GTE & Gen.\ Txt & TUTA & Transfer & 0.628 & 0.204 & \#120 \\
19 & TAPEX & Tbl-Txt & GTE & Gen.\ Txt & GTE & Transfer & 0.599 & 0.204 & \#461 \\
20 & TUTA & Tbl-Struct & BERT & Gen.\ Txt & TabICL & Prior & 0.589 & 0.204 & \#545 \\
\bottomrule
\end{tabular}
\end{table*}

\clearpage
\subsection{Source Split: TabFact vs.\ WTQ}
\label{app:dlte-source-split}

To test whether the \textsc{TRL-DLTE} findings are tied to a particular parent source, we partition the 345 test parents into TabFact (246) and WTQ (99) and recompute every pipeline-level metric separately on each partition. All 1,120 canonical pipelines ($10$~Stage-1~$\times$~$8$~Stage-2~$\times$~$14$~Stage-3 models) are evaluated under the same 5 rounds, so the source-split numbers are directly comparable to the main-text aggregates.

\input{tables/dlte_source_split/source_split_headline.tex}
\input{tables/dlte_source_split/source_split_marginals.tex}

\paragraph{Headline pipelines hold in both sources.}
Table~\ref{tab:dlte_source_split_headline} shows that the hybrid-vs-monolith gap persists separately on TabFact and WTQ: the dev-selected best hybrid (\textsc{TUTA}/\textsc{GTE}/\textsc{GTE}) beats the dev-selected best monolith (\textsc{BERT}/\textsc{BERT}/\textsc{BERT}) by $+0.086$ $\mathrm{UJ\text{-}H}$ on TabFact and by $+0.099$ on WTQ. The gap is if anything \emph{wider} on WTQ, so the hybrid advantage is not a TabFact-specific artifact. The unconstrained test maximum (\textsc{Starmie}/\textsc{GTE}/\textsc{GTE}, dev rank 8) reaches 0.242 / 0.281 on TabFact / WTQ with the same wider-on-WTQ pattern. The unconstrained test-set max monolith shifts to \textsc{GTE}/\textsc{GTE}/\textsc{GTE} (0.140 / 0.157, pooled 0.145), still well below every hybrid pipeline reported here.

\paragraph{Per-stage marginal top models are largely source-invariant.}
Table~\ref{tab:dlte_source_split_marginals} shows the top-5 models at each stage under each source. Stage~1 is \textsc{Starmie}-led in both sources, with \textsc{TUTA} second. Stage~2 has a \textsc{TABBIE}$\leftrightarrow$\textsc{TURL} swap at rank 1, and four of the top-5 column models (\textsc{TABBIE}, \textsc{TURL}, \textsc{GTE}, \textsc{BERT}) are common to both sources. Stage~3 is \textsc{TransTab}-led in both, with \textsc{GTE}, \textsc{TabICL}, and \textsc{TUTA} occupying ranks~2--4 in both sources (in varying order). No stage changes leader family under the source split.

\paragraph{Strong rank agreement across sources.}
Across all 1,120 canonical pipelines, Spearman $\rho(\text{TabFact}, \text{WTQ}) = 0.871$ ($p \ll 10^{-100}$), confirming that pipeline rankings are largely source-agnostic. For Oracle-RA, which isolates Stage~3 identity resolution by replacing Stages~1--2 with ground truth (Table~\ref{tab:oracle_ra}), the row-model ranking agrees at Spearman $\rho = 0.987$ ($p = 7.4 \times 10^{-11}$, $n = 14$): \textsc{GTE}, \textsc{TransTab}, \textsc{TabICL} occupy the top three positions in both sources, and \textsc{TabTransformer}, \textsc{SubTab}, \textsc{SAINT} the bottom three in both. Table~\ref{tab:oracle_ra_per_source_uj_h} reports the full per-source Oracle-RA $\mathrm{UJ\text{-}H}$ for every row model. Combined with the per-noise-tier breakdown in Table~\ref{tab:oracle_ra_per_tier_uj_h}, the Stage-3 row-model ranking is stable on both the noise-tier axis (cross-tier spans 0.506--0.563) and the parent-source axis (cross-source spans 0.533--0.551), so the identity-resolving/union-dedup split exposed at Stage~3 does not depend on either the particular noise regime or the particular parent source.

\input{tables/dlte_source_split/oracle_ra_per_source.tex}

\subsection{Robustness of the \textsc{TRL-DLTE} Composition Result}
\label{app:dlte_robustness}

This appendix reports selection analyses for the composition result of Sec.~\ref{sec:exp-dlte}, per-side union and join metrics, checks of the per-stage rankings on independently constructed data, and a lake rebuilt from other parent tables.

\noindent\textbf{Selection analyses.}
Three analyses test whether the dev-selected composition result of Sec.~\ref{sec:exp-dlte} depends on selection (test split, 5-round means). Without selection, 38.3\% of the 1{,}117 hybrid pipelines score above the best monolith (\textsc{BERT}/\textsc{BERT}/\textsc{BERT}) on Cell~$F_1$, against 25\% expected if the three monoliths were interchangeable random draws. With selection matched, the dev-selected best of three random hybrids beats the dev-selected monolith on test Cell~$F_1$ in 76\% of draws, against 50\% if hybrids and monoliths were exchangeable. With selection fixed in advance by rule, the assembly of each stage's development-marginal leader, \textsc{Starmie}/\textsc{BERT}/\textsc{TransTab}, reaches 0.231 test $\mathrm{UJ\text{-}H}$ (rank 4 of 1{,}120), against 0.253 for the test optimum and 0.145 for the best monolith on test (\textsc{GTE}/\textsc{GTE}/\textsc{GTE}) (Appendix~\ref{app:dlte-rankings}).

\noindent\textbf{Per-side union and join metrics.}
Table~\ref{tab:dlte_per_side} reports precision and $F_1$ separately for the union target (removed rows in the seed columns) and the join target (removed columns for the seed rows), computed per query and averaged over queries as for $\mathrm{UJ\text{-}H}$ (test split, 5-round means). The best pipeline is a hybrid under each per-side metric: \textsc{Starmie}/\textsc{TaBERT}/\textsc{TabTransformer} for union $F_1$ (0.916) and \textsc{TAPEX}/\textsc{GTE}/\textsc{GTE} for join $F_1$ (0.203). \textsc{TaBERT} columns and the union-dedup row models lead union $F_1$, and \textsc{GTE} columns and the identity-resolving row models lead join $F_1$, the split that Appendix~\ref{app:oracle_ra} describes.

\input{tables/dlte_per_side}

\noindent\textbf{Stage rankings on independently constructed data.}
Each DLTE stage has a counterpart whose gold was not produced by fragmenting a parent table. Stage~1: on \textsc{Pylon}~\cite{pylon}, 1{,}746 GitHub tables with 124 queries in human-curated union cliques (Appendix~\ref{app:pylon_union}), the R@60 ranking of the eight column models under the standard union-search harness (Table~\ref{tab:pylon_union}) agrees with their Stage-1 marginal $\mathrm{UJ\text{-}H}$ (Table~\ref{tab:dlte_stage1_full}) at Spearman $\rho=0.98$ (exact permutation $p=0.0004$). Stage-1 recall of the union and join targets also agrees with the benchmark's search tasks, at $\rho=0.81$ ($p=0.022$) against Union Search on its five corpora and $\rho=0.88$ ($p=0.007$) against Join Search over the four OpenData lakes (eight column models, exact permutation tests). Stage~2: Valentine builds most of its 550 schema-matching pairs by partitioning source tables, and its nine real-world pairs (seven Magellan, two WikiData) preserve the model ordering on all 550 pairs (Spearman $\rho=0.93$). Stage~3: the Oracle-RA ranking agrees with Robust Linkage NR, whose match labels come from the source benchmarks rather than from fragmentation ($|\rho|=0.80$, Appendix~\ref{app:oracle_ra}). These checks concern single-stage rankings. No organic gold exists for the composed pipeline.

\noindent\textbf{A lake rebuilt from other parents.}
We reran the same generation pipeline on parent tables from GitTables~\citep{gittables} instead of TabFact and WTQ, with CKAN distractors, and evaluated all 1{,}120 pipelines once. The per-stage conclusions reproduce: the five Stage-3 models with the highest marginal $\mathrm{UJ\text{-}H}$ are the same five as on the original lake, four of the top five Stage-2 models recur, and \textsc{Starmie} again leads Stage~1. The assembly of each stage's development Cell~$F_1$ leader ranks 7th of 1{,}120 on test Cell~$F_1$, while the identity of the single best pipeline changes between lakes, so a pipeline should be re-selected on a small development split of each new lake.

\clearpage
\section{Embedding Ablation: Retrieval vs.\ Structural Grounding}
\label{app:openai_ablation}

We evaluate three OpenAI embedding variants~\cite{openai_embeddings,openai_te3}: \textsc{TE3-Small} (text-embedding-3-small, 768-d), \textsc{TE3-Large} (text-embedding-3-large, 768-d), and \textsc{Ada-002} (text-embedding-ada-002, 1536-d). Alongside this proprietary, access-restricted trio (parameter counts undisclosed), we evaluate one open embedder at 8B-class scale, \textsc{Qwen3-Embedding-8B}~\cite{qwen3embedding}, well beyond the main leaderboard's $\sim$0.5M--500M eligibility band. We also evaluate three newer open encoders. \textsc{Qwen3-Embedding-0.6B}~\cite{qwen3embedding}, an MTEB-strong embedder at roughly five times \textsc{BERT}'s size, sits just above the band. \textsc{ModernBERT}~\cite{modernbert} at base size, both raw and embedding-trained (\textsc{GTE-ModernBERT}), sits inside it, beside \textsc{BERT} and \textsc{GTE}, the pool's generic-text representatives. The OpenAI trio and \textsc{Qwen3-Embedding-0.6B} cover all 13 \textsc{TRL-CTbench} tasks. \textsc{Qwen3-Embedding-8B}, 13 times the size of the 0.6B model, covers the four tasks that a tabular specialist leads in Table~\ref{tab:main_results} (Schema Matching, Union Search, Table Subset, and Table QA) and Table Retrieval. The two \textsc{ModernBERT} variants cover the retrieval subset only: the five union-search datasets and Valentine schema matching. At row level, the trio and \textsc{Qwen3-Embedding-8B} cover Row Prediction and Record Linkage, and \textsc{Qwen3-Embedding-0.6B} covers the 50-table row-prediction suite. TE3-Small and TE3-Large are requested at 768-d via OpenAI's \texttt{dimensions} API parameter for dimensional parity with the 768-d open-source encoders in our pool, and the two Qwen3 embedders are used without an instruction prefix at 768-d, within the supported output-dimension range of their Matryoshka-trained representations. Ada-002 does not expose this option and retains its native 1536-d. The two \textsc{ModernBERT} variants keep their native 768-d, with mean pooling for \textsc{ModernBERT} and the CLS embedding for \textsc{GTE-ModernBERT}. The open models are the Hugging Face checkpoints \texttt{Qwen/Qwen3-Embedding-8B}, \texttt{Qwen/Qwen3-Embedding-0.6B}, \texttt{answerdotai/ModernBERT-base}, and \texttt{Alibaba-NLP/gte-modernbert-base}. Table~\ref{tab:openai_ablation} reports the results on the 13 column- and table-level tasks alongside the open-source models from the main tables, Table~\ref{tab:openai_ablation_union} gives Union Search per dataset, and Tables~\ref{tab:openai_ablation_row} and~\ref{tab:qwen3_row_suite} give the row-level results. As in Table~\ref{tab:main_results}, table-level cells use the aggregation (and, for Table Retrieval, the query encoder) selected on development data, and Union Search averages the five \benchmark\ union-search datasets (SANTOS, TUS, TUS-hard, UGEN-v1, UGEN-v2).

\input{tables/openai_ablation}
\input{tables/openai_ablation_union}

\paragraph{Column/table level.}
All three proprietary variants lead Table Retrieval, occupying ranks 1--3 (\textsc{Ada-002}: 0.529 MRR, vs.\ \textsc{GTE}: 0.476). This is consistent with their training objective: large-scale semantic retrieval over diverse text, which transfers directly to table-to-query matching. \textsc{Qwen3-Embedding-8B} reaches 0.487 MRR on Table Retrieval: statistically indistinguishable from the nearest proprietary variant (\textsc{TE3-Small}: 0.490), clearly below \textsc{Ada-002} (0.529), and above the strongest open-source pool model (\textsc{GTE}: 0.476, +0.011). On Table QA, however, the same three variants rank 4--9 out of 13, at or behind \textsc{TURL} (0.277), \textsc{TABBIE} (0.276), \textsc{TaBERT} (0.267), and \textsc{Starmie} (0.266), models whose pretraining encodes table structure. \textsc{Ada-002} ties \textsc{Starmie} at 0.266. \textsc{Qwen3-Embedding-8B} and \textsc{Qwen3-Embedding-0.6B} (both 0.264) fall in the same band, between \textsc{TE3-Large} (0.265) and \textsc{TE3-Small} (0.260). The contrast within a single task family shows that retrieval quality and structural understanding are distinct capabilities.

\paragraph{All 13 tasks.}
Across the 13 tasks of Table~\ref{tab:openai_ablation}, an embedder beyond the pool has the best value in nine columns. In seven of them the pool leader in Table~\ref{tab:main_results} is \textsc{BERT} or \textsc{GTE} (Column Type, Column Relation, Join Search, Join Classification, Union Classification, Union Regression, and Table Retrieval), so a stronger text encoder replaces a text encoder. Of the four tasks that a tabular specialist leads in Table~\ref{tab:main_results}, the specialist keeps two. \textsc{Starmie} keeps Schema Matching at 0.764, against at most 0.651 for every embedder, and \textsc{TURL} keeps Table QA at 0.277, ahead of every embedder for all five seeds (paired $t$-tests over the five seeds, $p \le 0.027$). On Table Subset, \textsc{Ada-002} reaches 0.574 against \textsc{TAPAS}'s 0.567, a difference that is not significant ($p = 0.15$, with \textsc{Ada-002} ahead for four of the five seeds), and \textsc{TAPAS} stays ahead of the other four embedders for all five seeds ($p \le 0.012$). On the Union Search mean, \textsc{Qwen3-Embedding-8B} (0.670), \textsc{Qwen3-Embedding-0.6B} (0.666), and \textsc{TE3-Large} (0.664) sit up to 0.008 above \textsc{Starmie} (0.662). They gain on SANTOS, TUS, and the two UGEN datasets and fall behind on the adversarial TUS-hard split, where \textsc{Starmie}'s 0.523 stays above \textsc{BERT}, \textsc{GTE}, and every embedder beyond the pool (at most 0.307, Table~\ref{tab:openai_ablation_union}). \textsc{Qwen3-Embedding-0.6B} stays below the specialist on Schema Matching (0.630), Table Subset (0.547), and Table QA (0.264), and on the nine tasks that \textsc{BERT} or \textsc{GTE} leads in Table~\ref{tab:main_results} its cells lie within 0.04 of the better of the two. \textsc{ModernBERT} and \textsc{GTE-ModernBERT} were run on the retrieval subset only. Without embedding training, \textsc{ModernBERT} trails \textsc{BERT} on both of its tasks (Union Search 0.515 against 0.564, Schema Matching 0.391 against 0.423). \textsc{GTE-ModernBERT} matches \textsc{GTE} on TUS-hard (0.293 each), reaches 0.661 on the Union Search mean against \textsc{Starmie}'s 0.662, and stays far below \textsc{Starmie} on Schema Matching (0.474 against 0.764).

\input{tables/openai_ablation_row}
\input{tables/qwen3_row_suite}

\paragraph{Row level.}
The pattern extends to rows (Table~\ref{tab:openai_ablation_row}). On Record Linkage, an entity-matching task driven by text similarity, the proprietary models again occupy the top three ranks (\textsc{TE3-Large}: 0.426 $F_1$, vs.\ \textsc{GTE}: 0.403). \textsc{Qwen3-Embedding-8B} reaches 0.405, level with \textsc{GTE} (0.403) and below the full proprietary trio. On Row Prediction, however, \textsc{TE3-Small} reaches 0.801 AUROC (rank 2) but trails \textsc{TabICL} (0.816), a meta-pretrained model whose in-context conditioning adapts to each task. \textsc{Qwen3-Embedding-8B} reaches 0.794 AUROC, below both \textsc{TE3-Small} and \textsc{TE3-Large} (0.797) but ahead of \textsc{Ada-002} (0.789). On the 50-table row-prediction suite (Table~\ref{tab:qwen3_row_suite}), \textsc{Qwen3-Embedding-0.6B} lands within the generic-text class: its classification AUROC (0.789) is level with \textsc{BERT} (0.791) and \textsc{TabPFN} (0.793), its regression $R^2$ (0.410) lies between \textsc{BERT} (0.378) and \textsc{TabPFN} (0.486), and \textsc{TabICL} ranks first on both (mean rank 1.77 and 1.39, against 3.17 and 3.26). Across both granularities, proprietary embeddings lead \emph{matching} tasks (retrieval, linkage) but do not displace specialized models on \emph{understanding} tasks (QA, prediction), reinforcing that no single model family, open or proprietary, is universally dominant.

\paragraph{Scope and model selection.}
The general benchmark scope rationale is discussed in Appendix~\ref{app:extended-related}. The scale ablation above, spanning both proprietary and open embedders, should therefore be read as a controlled scaling study within the embedding-only regime, not as a proxy for evaluating $7$B--$70$B generative systems.

At 13 times the size of the 0.6B model, \textsc{Qwen3-Embedding-8B} widens the Schema Matching gap (0.533, against 0.630 for the 0.6B model and 0.764 for \textsc{Starmie}), leaves the adversarial TUS-hard split nearly unmoved (0.295, against 0.282 and 0.523), and scores level with the 0.6B model on Table Subset (0.550 against 0.547, both below \textsc{TAPAS}'s 0.567). Under the same wrapper, it still reaches 0.98--1.00 MAP on SANTOS and TUS.

\clearpage
\section{Robustness}
\label{app:robustness}
This section reports three auxiliary stability diagnostics introduced by
Observatory~\citep{observatory}, namely \textit{Sample Fidelity},
\textit{Perturbation Robustness}, and \textit{Row/Column Order Insignificance},
applied directly to the models in our pool. The diagnostic definitions,
evaluation datasets, and headline metrics all follow the Observatory originals.
These diagnostics are not part of our contribution and are not counted toward
the 16 benchmark tasks. All numerical results below are recomputed from our
finalized embeddings.

All headline statistics use a unified table-first aggregation protocol.
For \textit{Sample Fidelity} and \textit{Row/Column Order Insignificance},
we compute two canonical metrics,
\texttt{table\_cosine\_similarity} and \texttt{table\_mcv}, and then report
the dataset-level mean and standard deviation over valid tables.
For \textit{Sample Fidelity}, each table cell below follows the format
\emph{cosine mean $\pm$ std / MCV mean $\pm$ std}.
For \textit{Row/Column Order Insignificance}, cosine similarity and MCV are
reported in separate tables.
For \textit{Perturbation Robustness}, following the original Observatory
protocol, we report changed-only cosine similarity only.
Cosine similarity is the primary metric for cross-task comparison.

\subsection{Unified Aggregation Protocol}

Let $T^{(0)}$ be the original table and
$\{T^{(1)}, \ldots, T^{(K)}\}$ be its transformed variants.
For a table item $i$ (a column, a row, or the whole table depending on the
task) with embedding $z_i^{(k)}$ under variant $k$, we compute
\begin{align}
s_i &= \frac{1}{|V_i|} \sum_{k \in V_i}
\cos\!\left(z_i^{(0)}, z_i^{(k)}\right), \\
c_i &= \mathrm{MCV}\!\left(\{z_i^{(0)}\} \cup
\{z_i^{(k)} \mid k \in V_i\}\right),
\end{align}
where $V_i$ is the set of valid transformed variants for item $i$.
The table-level metrics are then
\begin{align}
\mathrm{table\_cosine} &= \frac{1}{m} \sum_{i=1}^{m} s_i, \\
\mathrm{table\_mcv} &= \frac{1}{m} \sum_{i=1}^{m} c_i,
\end{align}
with $m$ being the number of valid items in the table.
Finally, the appendix tables report the mean and standard deviation of these
table-level metrics over the evaluation set. MCV is computed with the
Observatory multivariate coefficient of variation implementation
$\sqrt{\mu^\top \Sigma \mu / (\mu^\top \mu)^2}$.

For \textit{Perturbation Robustness}, we use a strict changed-only protocol:
only columns that were actually modified by a perturbation are included in the
headline cosine metric; tables without changed columns are excluded from the
dataset-level aggregate.

\Needspace{20\baselineskip}
\subsection{Task and Dataset Summary}

\begin{table}[H]
\centering
\small
\caption{Overview of the three appendix diagnostics.}
\label{tab:appendix_task_overview}
\begin{tabularx}{\linewidth}{lYYY}
\toprule
Diagnostic & Core question & Data and scale & Embeddings and model coverage \\
\midrule
Sample Fidelity &
Whether a table representation remains stable after observing only a subset of
rows. &
\texttt{wiki\_tables}; 4,964 base tables; sampling ratios
$0.25$, $0.50$, and $0.75$. &
Column embeddings; the 8 column-capable models from our pool
(shared across all three robustness diagnostics):
\texttt{bert}, \texttt{gte}, \texttt{starmie}, \texttt{tabbie},
\texttt{tabert}, \texttt{tabsketchfm}, \texttt{tapas}, \texttt{turl}. \\
\midrule
Perturbation Robustness &
Whether column representations remain stable under schema and content
perturbations. &
Database tables; 80 base tables; three perturbation families:
\texttt{DB\_\allowbreak{}schema\_\allowbreak{}synonym}, \texttt{DB\_\allowbreak{}schema\_\allowbreak{}abbreviation}, and
\texttt{DB\_\allowbreak{}DBcontent\_\allowbreak{}equivalence}. &
Column embeddings; the same 8 column-capable models as above. \\
\midrule
Row/Column Order Insignificance &
Whether row, column, and table representations are invariant to row-order and
column-order permutations. &
\texttt{wiki\_tables}; 4,964 base tables; each table has either 11 or 6
shuffle variants depending on how many unique permutations are available. &
Column, row, and table embeddings; the 8 column-capable models above, 14
row-capable models, and 10 table-capable models from our pool. \\
\bottomrule
\end{tabularx}
\end{table}

\subsection{Sample Fidelity}

The Sample Fidelity diagnostic measures whether the semantic representation of a
table remains stable when only a subset of rows is observed.
Following the Observatory Sample Fidelity protocol~\citep{observatory}, we construct
subsampled variants at ratios $0.25$, $0.50$, and $0.75$, and compare each
sampled table against the original table through column embeddings.

For each column, we compute the average cosine similarity between the original
column embedding and all sampled variants, and the MCV over the set consisting
of the original column embedding plus all sampled versions.
The final table score is the mean over all columns in the same table.

\begin{table}[H]
\centering
\small
\caption{Dataset parameters for Sample Fidelity. The $0.25$ setting has fewer
unique subsamples for small tables. The $0.50$ and $0.75$ settings use 11
versions per base table throughout.}
\label{tab:appendix_sample_params}
\begin{tabular}{lrrrr}
\toprule
Sampling ratio & Base tables & 11 variants & 8 variants & 7/6 variants \\
\midrule
$0.25$ & 4,964 & 3,816 & 335 & 475 / 338 \\
$0.50$ & 4,964 & 4,964 & 0 & 0 / 0 \\
$0.75$ & 4,964 & 4,964 & 0 & 0 / 0 \\
\bottomrule
\end{tabular}
\end{table}

\begin{table}[t]
\centering
\small
\setlength{\tabcolsep}{4pt}
\caption{Full Sample Fidelity results. Each cell reports
\emph{cosine mean $\pm$ std} on the first line and
\emph{MCV mean $\pm$ std} on the second line.}
\label{tab:appendix_sample_results}
\begin{adjustbox}{max width=\linewidth}
\begin{tabular}{lccc}
\toprule
Model & $0.25$ & $0.50$ & $0.75$ \\
\midrule
\texttt{bert} &
\score{0.9138}{0.0284}{0.0211}{0.0083} &
\score{0.9619}{0.0175}{0.0144}{0.0073} &
\score{0.9828}{0.0108}{0.0110}{0.0076} \\
\texttt{gte} &
\score{0.8290}{0.0564}{0.0316}{0.0134} &
\score{0.9180}{0.0342}{0.0212}{0.0116} &
\score{0.9620}{0.0199}{0.0175}{0.0124} \\
\texttt{starmie} &
\score{0.9198}{0.1161}{0.0306}{0.0407} &
\score{0.9846}{0.0291}{0.0123}{0.0175} &
\score{0.9923}{0.0194}{0.0084}{0.0146} \\
\texttt{tabbie} &
\score{0.9567}{0.0363}{0.0099}{0.0087} &
\score{0.9873}{0.0136}{0.0032}{0.0032} &
\score{0.9959}{0.0052}{0.0014}{0.0017} \\
\texttt{tabert} &
\score{0.9933}{0.0029}{0.0025}{0.0012} &
\score{0.9965}{0.0015}{0.0017}{0.0007} &
\score{0.9982}{0.0009}{0.0013}{0.0006} \\
\texttt{tabsketchfm} &
\score{0.4692}{0.0984}{0.9489}{0.1555} &
\score{0.6596}{0.0916}{0.7524}{0.1485} &
\score{0.8084}{0.0720}{0.5623}{0.1356} \\
\texttt{tapas} &
\score{0.7961}{0.0668}{0.3975}{0.1023} &
\score{0.9146}{0.0407}{0.2619}{0.0666} &
\score{0.9602}{0.0260}{0.1827}{0.0533} \\
\texttt{turl} &
\score{0.8679}{0.0736}{0.4658}{0.1899} &
\score{0.9506}{0.0313}{0.2669}{0.0988} &
\score{0.9782}{0.0197}{0.1735}{0.0726} \\
\bottomrule
\end{tabular}
\end{adjustbox}
\end{table}

\paragraph{Analysis.}
Three patterns are especially clear.
First, all eight models become more stable as the retained row fraction grows
from $0.25$ to $0.75$, which means that sample fidelity is strongly tied to
how much of the original row distribution remains visible.
Second, \texttt{tabert} is the most stable model at every sampling ratio,
indicating that its column representations are highly insensitive to the
removal of rows.
Third, \texttt{tabbie} is the most balanced non-\texttt{tabert} model,
combining high cosine similarity with low variance across all three settings,
whereas \texttt{tabsketchfm} is much more sample-sensitive and degrades
substantially when only $25\%$ of rows are retained.

\subsection{Perturbation Robustness}

The Perturbation Robustness diagnostic evaluates whether column representations remain
stable after semantically valid modifications to the table schema or content.
Following the Observatory Perturbation Robustness protocol~\citep{observatory}, we
use the same three perturbation families:
\texttt{DB\_schema\_synonym} (schema names replaced by synonyms),
\texttt{DB\_schema\_abbreviation} (schema names replaced by abbreviations), and
\texttt{DB\_DBcontent\_equivalence} (schema or content rewritten in a
semantically equivalent form).

The analysis uses a strict changed-only headline protocol.
For each original table and each column that is modified by a perturbation, we
compute the average cosine similarity between the original column embedding and
all changed variants.
The table-level score is the mean over all changed columns in that table.

\begin{table}[t]
\centering
\small
\caption{Dataset parameters for Perturbation Robustness.
``Valid tables'' are tables with at least one changed column under the given
perturbation type.}
\label{tab:appendix_perturbation_params}
\begin{tabular}{lrrrrrr}
\toprule
Perturbation & Original & Valid & Skipped & Changed & Unchanged & Changed \\
type & tables & tables & tables & pairs & pairs & columns \\
\midrule
Content equivalence & 80 & 29 & 51 & 238 & 1,320 & 101 \\
Schema abbreviation & 80 & 76 & 4 & 691 & 1,456 & 232 \\
Schema synonym & 80 & 69 & 11 & 453 & 1,708 & 202 \\
\bottomrule
\end{tabular}
\end{table}

\begin{table}[t]
\centering
\small
\setlength{\tabcolsep}{4pt}
\caption{Full Perturbation Robustness results.
Each cell reports changed-only cosine mean with small-font standard deviation.}
\label{tab:appendix_perturbation_results}
\begin{adjustbox}{max width=\linewidth}
\begin{tabular}{lccc}
\toprule
Model & Content equivalence & Schema abbreviation & Schema synonym \\
\midrule
\texttt{bert} &
$0.8420$\,\std{0.0358} &
$0.9453$\,\std{0.0294} &
$0.9524$\,\std{0.0277} \\
\texttt{gte} &
$0.5777$\,\std{0.0961} &
$0.7577$\,\std{0.0914} &
$0.8283$\,\std{0.0644} \\
\texttt{starmie} &
$0.9261$\,\std{0.0715} &
$0.9998$\,\std{0.0008} &
$0.9998$\,\std{0.0009} \\
\texttt{tabbie} &
$0.9783$\,\std{0.0224} &
$0.9982$\,\std{0.0028} &
$0.9987$\,\std{0.0017} \\
\texttt{tabert} &
$0.9451$\,\std{0.0181} &
$0.9507$\,\std{0.0160} &
$0.9635$\,\std{0.0134} \\
\texttt{tabsketchfm} &
$0.2939$\,\std{0.1748} &
$0.6471$\,\std{0.2820} &
$0.5459$\,\std{0.2646} \\
\texttt{tapas} &
$0.6605$\,\std{0.0866} &
$0.9545$\,\std{0.0342} &
$0.9584$\,\std{0.0309} \\
\texttt{turl} &
$0.7551$\,\std{0.0965} &
$0.9857$\,\std{0.0425} &
$0.9905$\,\std{0.0344} \\
\bottomrule
\end{tabular}
\end{adjustbox}
\end{table}

\paragraph{Analysis.}
The hardest perturbation is content equivalence, followed by the two
schema-only perturbations.
This shows that semantic re-expression of column content is much more difficult
than renaming a schema attribute with a synonym or abbreviation.
\texttt{tabbie} is the strongest overall model and remains the most robust on
the hardest content-equivalence setting.
\texttt{starmie} is nearly invariant to schema-level changes, but drops more
than \texttt{tabbie} under content-level perturbations, suggesting that its
robustness is particularly strong at the schema level.
By contrast, \texttt{gte} and especially \texttt{tabsketchfm} are clearly more
sensitive to semantic perturbations.
Note that the content equivalence evaluation uses only 29 valid tables
(Table~\ref{tab:appendix_perturbation_params}), so these comparisons should
be treated as indicative rather than definitive.

\subsection{Row/Column Order Insignificance}

The Row/Column Order Insignificance diagnostic evaluates whether learned
representations are invariant to row-order and column-order permutations.
Following the Observatory Row/Column Order Insignificance protocol~\citep{observatory},
each base table is paired with multiple shuffled variants from
\texttt{wiki\_tables}.
In our data, 4,451 base tables have 11 variants
(the original table plus 10 shuffles), while 513 tables have 6 variants.

We study six evaluation conditions:
\texttt{column/column},
\texttt{column/row},
\texttt{row/column},
\texttt{row/row},
\texttt{table/column}, and
\texttt{table/row}.
The first term denotes the embedding granularity and the second term denotes
the applied shuffle type.
Thus, \texttt{column/column} measures the invariance of column embeddings to
column permutation, \texttt{column/row} tests whether column semantics change
under row shuffling, \texttt{row/column} and \texttt{row/row} evaluate the
stability of row embeddings under column-order and row-order changes,
respectively, and \texttt{table/column} and \texttt{table/row} test whole-table
invariance to column and row permutations.
When needed, shuffled items are realigned to their original indices before
comparison; models that already canonicalize the relevant order do not require
this extra step.

\begin{table}[t]
\centering
\scriptsize
\setlength{\tabcolsep}{3pt}
\caption{Cosine similarity results for the Row/Column Order Insignificance diagnostic.
The six metric columns are grouped by embedding granularity and shuffle type.
\textsc{na} means that the corresponding embedding granularity is not available
for that model.}
\label{tab:appendix_rowcolumn_cosine}
\begin{adjustbox}{max width=\linewidth}
\begin{tabular}{lcccccc}
\toprule
& \multicolumn{2}{c}{Column embeddings} &
\multicolumn{2}{c}{Row embeddings} &
\multicolumn{2}{c}{Table embeddings} \\
\cmidrule(lr){2-3}\cmidrule(lr){4-5}\cmidrule(l){6-7}
Model & column shuffle & row shuffle & column shuffle & row shuffle & column shuffle & row shuffle \\
\midrule
\texttt{bert} & $1.0000$\,\std{0.0000} & $0.9809$\,\std{0.0147} & $0.9754$\,\std{0.0102} & $1.0000$\,\std{0.0000} & $0.9731$\,\std{0.0144} & $0.9824$\,\std{0.0168} \\
\texttt{dae} & \NA & \NA & $0.5015$\,\std{0.0750} & $0.9770$\,\std{0.0082} & \NA & \NA \\
\texttt{gte} & $1.0000$\,\std{0.0000} & $0.9704$\,\std{0.0190} & $0.9556$\,\std{0.0204} & $1.0000$\,\std{0.0000} & $0.9590$\,\std{0.0174} & $0.9648$\,\std{0.0199} \\
\texttt{saint} & \NA & \NA & $0.7533$\,\std{0.1345} & $0.8857$\,\std{0.0823} & \NA & \NA \\
\texttt{scarf} & \NA & \NA & $0.5371$\,\std{0.0700} & $0.9741$\,\std{0.0092} & \NA & \NA \\
\texttt{starmie} & $0.9988$\,\std{0.0027} & $0.9708$\,\std{0.0346} & \NA & \NA & $0.9988$\,\std{0.0027} & $0.9580$\,\std{0.0518} \\
\texttt{subtab} & \NA & \NA & $0.8233$\,\std{0.0762} & $0.9245$\,\std{0.0282} & \NA & \NA \\
\texttt{tabbie} & $0.9966$\,\std{0.0039} & $0.9986$\,\std{0.0021} & $0.9941$\,\std{0.0029} & $1.0000$\,\std{0.0000} & $0.9975$\,\std{0.0028} & $0.9988$\,\std{0.0021} \\
\texttt{tabert} & $0.9706$\,\std{0.0056} & $0.9947$\,\std{0.0021} & \NA & \NA & $0.9958$\,\std{0.0012} & $0.9984$\,\std{0.0009} \\
\texttt{tabicl} & \NA & \NA & $0.6155$\,\std{0.0854} & $1.0000$\,\std{0.0000} & \NA & \NA \\
\texttt{tabpfn} & \NA & \NA & $0.9998$\,\std{0.0001} & $1.0000$\,\std{0.0000} & \NA & \NA \\
\texttt{tabsketchfm} & $0.9598$\,\std{0.0255} & $1.0000$\,\std{0.0000} & \NA & \NA & $0.9902$\,\std{0.0063} & $1.0000$\,\std{0.0000} \\
\texttt{tabtransformer} & \NA & \NA & $0.4226$\,\std{0.0808} & $0.4370$\,\std{0.0761} & \NA & \NA \\
\texttt{tabular\_binning} & \NA & \NA & $0.5587$\,\std{0.0592} & $0.9725$\,\std{0.0084} & \NA & \NA \\
\texttt{tapas} & $0.9277$\,\std{0.0242} & $0.9821$\,\std{0.0108} & \NA & \NA & $0.9461$\,\std{0.0680} & $0.9705$\,\std{0.0624} \\
\texttt{tapex} & \NA & \NA & \NA & \NA & $0.9944$\,\std{0.0045} & $0.9988$\,\std{0.0029} \\
\texttt{transtab} & \NA & \NA & $0.8712$\,\std{0.0621} & $0.9627$\,\std{0.0349} & \NA & \NA \\
\texttt{turl} & $0.9974$\,\std{0.0125} & $0.9970$\,\std{0.0128} & \NA & \NA & $0.9974$\,\std{0.0128} & $0.9972$\,\std{0.0130} \\
\texttt{tuta} & \NA & \NA & $0.9279$\,\std{0.0262} & $1.0000$\,\std{0.0000} & $0.9278$\,\std{0.0547} & $0.9387$\,\std{0.0549} \\
\texttt{vime} & \NA & \NA & $0.5365$\,\std{0.0757} & $0.9783$\,\std{0.0078} & \NA & \NA \\
\bottomrule
\end{tabular}
\end{adjustbox}
\end{table}

\begin{table}[t]
\centering
\scriptsize
\setlength{\tabcolsep}{3pt}
\caption{MCV results for the Row/Column Order Insignificance task.
The six metric columns are grouped by embedding granularity and shuffle type.
\textsc{na} means that the corresponding embedding granularity is not available
for that model.}
\label{tab:appendix_rowcolumn_mcv}
\begin{adjustbox}{max width=\linewidth}
\begin{tabular}{lcccccc}
\toprule
& \multicolumn{2}{c}{Column embeddings} &
\multicolumn{2}{c}{Row embeddings} &
\multicolumn{2}{c}{Table embeddings} \\
\cmidrule(lr){2-3}\cmidrule(lr){4-5}\cmidrule(l){6-7}
Model & column shuffle & row shuffle & column shuffle & row shuffle & column shuffle & row shuffle \\
\midrule
\texttt{bert} & $0.0000$\,\std{0.0000} & $0.0097$\,\std{0.0054} & $0.0127$\,\std{0.0054} & $0.0000$\,\std{0.0000} & $0.0129$\,\std{0.0063} & $0.0102$\,\std{0.0057} \\
\texttt{dae} & \NA & \NA & $0.1699$\,\std{0.0371} & $0.0412$\,\std{0.0135} & \NA & \NA \\
\texttt{gte} & $0.0000$\,\std{0.0000} & $0.0052$\,\std{0.0031} & $0.0074$\,\std{0.0035} & $0.0000$\,\std{0.0000} & $0.0064$\,\std{0.0033} & $0.0052$\,\std{0.0031} \\
\texttt{saint} & \NA & \NA & $0.0545$\,\std{0.0357} & $0.0289$\,\std{0.0247} & \NA & \NA \\
\texttt{scarf} & \NA & \NA & $0.1551$\,\std{0.0317} & $0.0324$\,\std{0.0086} & \NA & \NA \\
\texttt{starmie} & $0.0028$\,\std{0.0039} & $0.0222$\,\std{0.0158} & \NA & \NA & $0.0028$\,\std{0.0039} & $0.0250$\,\std{0.0236} \\
\texttt{subtab} & \NA & \NA & $0.3267$\,\std{0.1011} & $0.2524$\,\std{0.0550} & \NA & \NA \\
\texttt{tabbie} & $0.0010$\,\std{0.0012} & $0.0004$\,\std{0.0005} & $0.0037$\,\std{0.0013} & $0.0000$\,\std{0.0000} & $0.0014$\,\std{0.0013} & $0.0009$\,\std{0.0009} \\
\texttt{tabert} & $0.0038$\,\std{0.0009} & $0.0019$\,\std{0.0008} & \NA & \NA & $0.0010$\,\std{0.0004} & $0.0006$\,\std{0.0003} \\
\texttt{tabicl} & \NA & \NA & $0.0439$\,\std{0.0167} & $0.0000$\,\std{0.0000} & \NA & \NA \\
\texttt{tabpfn} & \NA & \NA & $0.0001$\,\std{0.0000} & $0.0000$\,\std{0.0000} & \NA & \NA \\
\texttt{tabsketchfm} & $0.0066$\,\std{0.0034} & $0.0000$\,\std{0.0000} & \NA & \NA & $0.0013$\,\std{0.0008} & $0.0000$\,\std{0.0000} \\
\texttt{tabtransformer} & \NA & \NA & $0.2524$\,\std{0.0582} & $0.2470$\,\std{0.0540} & \NA & \NA \\
\texttt{tabular\_binning} & \NA & \NA & $0.1514$\,\std{0.0326} & $0.0394$\,\std{0.0111} & \NA & \NA \\
\texttt{tapas} & $0.0276$\,\std{0.0088} & $0.0142$\,\std{0.0048} & \NA & \NA & $0.0288$\,\std{0.0245} & $0.0197$\,\std{0.0230} \\
\texttt{tapex} & \NA & \NA & \NA & \NA & $0.0191$\,\std{0.0096} & $0.0046$\,\std{0.0036} \\
\texttt{transtab} & \NA & \NA & $0.0279$\,\std{0.0204} & $0.0136$\,\std{0.0100} & \NA & \NA \\
\texttt{turl} & $0.0027$\,\std{0.0091} & $0.0029$\,\std{0.0089} & \NA & \NA & $0.0044$\,\std{0.0148} & $0.0046$\,\std{0.0143} \\
\texttt{tuta} & \NA & \NA & $0.0587$\,\std{0.0190} & $0.0000$\,\std{0.0000} & $0.0514$\,\std{0.0348} & $0.0430$\,\std{0.0307} \\
\texttt{vime} & \NA & \NA & $0.1638$\,\std{0.0378} & $0.0430$\,\std{0.0144} & \NA & \NA \\
\bottomrule
\end{tabular}
\end{adjustbox}
\end{table}

\paragraph{Analysis.}
The strongest qualitative pattern is that many models are almost perfectly
stable on \texttt{row/row}, whereas the more difficult condition is often
\texttt{row/column}, which requires row semantics to survive a change in column
order.
Among shared column/table models, \texttt{tabbie} is the most balanced model
across all available conditions, while \texttt{turl} and \texttt{tabert} also
show very strong order robustness.
\texttt{tabsketchfm} is especially notable because it is extremely stable under
order perturbations even though it is much weaker on the other two tasks.
For row representations, \texttt{tabpfn} is almost ideal, \texttt{tabbie},
\texttt{bert}, and \texttt{gte} form a strong second tier, and
\texttt{tabtransformer} is the least robust model in this setting,
consistent with its use of learned column positional embeddings that make
row representations inherently sensitive to column order.

\subsection{Cross-Task Comparison of Shared Models}

To avoid confounding caused by different model coverage across tasks, the
cross-task comparison uses only the eight models that appear in all three
evaluations:
\texttt{bert}, \texttt{gte}, \texttt{starmie}, \texttt{tabbie},
\texttt{tabert}, \texttt{tabsketchfm}, \texttt{tapas}, and \texttt{turl}.
We rank models within each task by their average cosine performance and then
compute the average rank across tasks (an ordinal summary that reflects
relative positioning rather than the magnitude of score differences).

\begin{table}[t]
\centering
\small
\caption{Cross-task ranking over the eight column-capable models from our pool that appear in all three robustness evaluations. Lower average rank
means better overall robustness.}
\label{tab:appendix_cross_task_rank}
\begin{tabular}{lrrrr}
\toprule
Model & Sample rank & Perturbation rank & Order rank & Average rank \\
\midrule
\texttt{tabbie} & 2 & 1 & 1 & 1.33 \\
\texttt{tabert} & 1 & 3 & 3 & 2.33 \\
\texttt{starmie} & 3 & 2 & 6 & 3.67 \\
\texttt{turl} & 5 & 5 & 2 & 4.00 \\
\texttt{bert} & 4 & 4 & 5 & 4.33 \\
\texttt{tabsketchfm} & 8 & 8 & 4 & 6.67 \\
\texttt{gte} & 6 & 7 & 7 & 6.67 \\
\texttt{tapas} & 7 & 6 & 8 & 7.00 \\
\bottomrule
\end{tabular}
\end{table}

\paragraph{Analysis.}
\texttt{tabbie} is the most balanced model across tasks.
It is not always the single best model on every individual condition, but it
remains near the top under sampling, semantic perturbation, and structural
reordering, which makes it the strongest all-round choice in our study.
\texttt{tabert} is the best model for sample fidelity, while
\texttt{starmie} is especially strong for schema-level perturbations.
\texttt{turl} is consistently robust without being the single best model in any
one dimension.
Finally, \texttt{tabsketchfm} is the clearest example of a task-specialized
behavior: it is highly order-invariant, but much weaker under row subsampling
and semantic perturbation.

\subsection{Implementation Notes and Caveats}

All numbers in this section are recomputed from finalized embeddings using a unified
table-first headline protocol rather than copied from older logs.
This recomputation includes three practical decisions that are important for
interpretation:
\begin{itemize}
    \item \textbf{Table-first aggregation.} All three tasks first aggregate to
    per-table scores and only then compute dataset-level mean and standard
    deviation.
    \item \textbf{Changed-only perturbation scoring.} For
    \textit{Perturbation Robustness}, only changed columns contribute to the
    headline cosine metric. We do not report MCV for this task, matching the
    original Observatory perturbation protocol.
    \item \textbf{MCV comparability.} MCV is computed from different numbers
    of embedding variants in the sample-fidelity and order-insignificance
    tasks (up to 11 for order shuffles), and the covariance estimate is rank-deficient in all
    cases ($K{+}1$ samples in $d$-dimensional space, $K{+}1 \ll d$).
    MCV magnitudes should therefore be compared only within the same task.
\end{itemize}

These caveats do not change the main qualitative conclusions, but they matter
for correct interpretation of the absolute numbers and for reproducible
comparison across properties.

\section{Computational Efficiency}
\label{app:efficiency}

\input{tables/effbench_column.tex}
\input{tables/effbench_table.tex}

We report per-model embedding generation cost across three workloads (column, table, row) on a controlled efficiency test suite. All measurements use a single NVIDIA L40S GPU with 32\,GB RAM per job.

\paragraph{Efficiency test suite.}
The suite comprises: (i)~\textbf{Eff-Real}: 8~anchor tables selected from the 50 \textsc{TRL-Rbench} OpenML tables via metadata-space clustering (covering 1\,000--71\,518 rows and 18--1\,775 columns with diverse type mixes and missingness rates); (ii)~\textbf{Eff-Scale}: 47~semi-synthetic tables generated by varying one factor at a time from a baseline (row track: $N \in \{500, \ldots, 100\text{k}\}$, $D \in \{8, \ldots, 256\}$, categorical share, cardinality, missingness; column track: $C \in \{4, \ldots, 128\}$, context rows, cell token length, type mix); and (iii)~\textbf{Bridge}: 3~tables valid for both row and column workloads. Each model is timed using its unmodified production embedding script via a thin wall-clock wrapper, ensuring that the measured cost exactly matches the actual benchmark pipeline.

\paragraph{Column and table embedding cost.}
Table~\ref{tab:eff_column} reports column-level results. Generic text encoders are fastest (\textsc{BERT} 5.3\,s, \textsc{GTE} 5.7\,s median), while \textsc{TaBERT} (13.2\,s) and \textsc{TabSketchFM} (14.1\,s) are slowest, a 2.7$\times$ spread. For table embeddings (Table~\ref{tab:eff_table}), \textsc{TAPEX} takes 6.1\,s.

\paragraph{Row embedding cost.}
Table~\ref{tab:eff_row} reports row-level results. The 101$\times$ spread between the fastest (\textsc{TabICL}, 8.7\,s) and slowest (\textsc{TransTab}, 875\,s) models is driven primarily by the frozen-vs.-trained regime distinction: all target-table self-supervised models include per-table training, which dominates their wall-clock cost. Among frozen and prior-based models, \textsc{TabICL} and \textsc{TabPFN} (20.7\,s) are fastest despite requiring a fit step, because their meta-learned priors avoid gradient-based training.

\input{tables/effbench_row.tex}

\begin{figure}[t]
\centering
\begin{subfigure}[t]{0.48\linewidth}
\includegraphics[width=\linewidth]{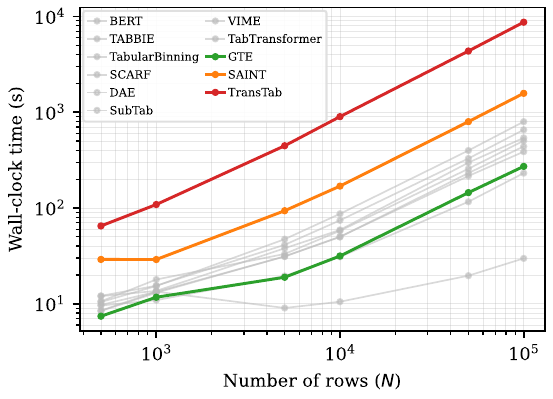}
\caption{Row scaling (varying $N$)}
\end{subfigure}
\hfill
\begin{subfigure}[t]{0.48\linewidth}
\includegraphics[width=\linewidth]{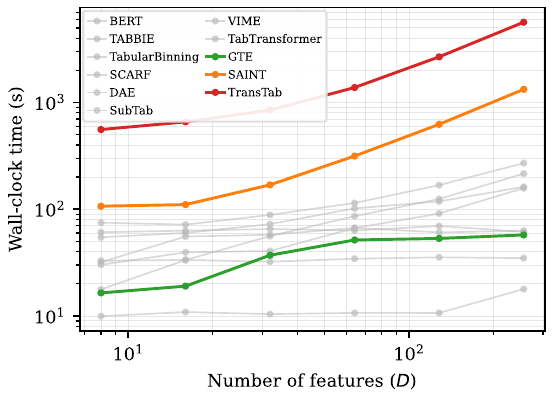}
\caption{Feature scaling (varying $D$)}
\end{subfigure}
\caption{\textbf{Embedding generation cost vs.\ table size} (log-log scale). Training-based models scale super-linearly with rows.}
\label{fig:eff_scaling}
\end{figure}

\paragraph{Scaling behavior.}
Figure~\ref{fig:eff_scaling} shows how row embedding cost scales with table size. Training-based models (\textsc{TransTab}, \textsc{SAINT}) scale super-linearly with row count, while frozen inference models scale approximately linearly. Feature count scaling is more uniform across the models in the sweep. Column-embedding cost (Table~\ref{tab:eff_column}) is dominated by per-column tokenization overhead: \textsc{TaBERT} and \textsc{TabSketchFM} are roughly $2.5$--$2.7\times$ slower per job than \textsc{BERT}/\textsc{GTE}. The column track of \textsc{Eff-Scale} ($C \in \{4, \ldots, 128\}$) confirms that this gap widens approximately linearly with column count (full curves reproducible with the code release).

\paragraph{Support envelope.}
Not all models can handle all scales under the 1-hour budget. \textsc{TURL} runs out of memory on tables with $>$208 columns or $>$256 features. \textsc{TabPFN} cost rises sharply with table size on \textsc{Eff-Real} anchors (18\,s for 1\,000 rows $\times$ 18 features, 263\,s for 71\,518 $\times$ 43, 1\,121\,s for 50\,000 $\times$ 208, and a timeout for 3\,751 $\times$ 1\,775). \textsc{SAINT} and \textsc{TabTransformer} fail on the widest anchor table (1\,775 columns). These limits are important for practitioners selecting models for large-scale deployment.

\paragraph{Methodology.}
\textsc{TabICL} and \textsc{TabPFN} are timed only on \textsc{Eff-Real} anchor tables. Their context-fit step is most representative of production usage when run on real labeled splits, while the synthetic \textsc{Eff-Scale} suite is unlabeled. This labeled-split timing reflects deployment cost. The row-embedding matrix consumed by every \textsc{TRL-Rbench} task is extracted from a target-agnostic forward pass conditioned only on the unlabeled $X$ rows, so the same matrix is reused across all curated targets within a table. Each measurement times the unmodified production script end-to-end (including model loading, preprocessing, and any per-table training) via a subprocess wrapper that records wall-clock time and polls \texttt{nvidia-smi} for peak GPU VRAM at 0.5\,s intervals. Results are recorded as individual JSON files with full provenance (hostname, GPU type, SLURM job ID, return code, output verification). The code that runs and analyzes the suite is included in the code release (\url{https://github.com/LOGO-CUHKSZ/TRL-Bench}).

\section{Reproducibility Details}
\label{app:reproducibility}

\paragraph{Data access.}
All source datasets are publicly available. OpenML tables are accessed via the OpenML API using the dataset IDs listed in Appendix~\ref{app:dataset-inventory}. DeepMatcher datasets are from the original DeepMatcher release. WDC Products data is from the WDC Product Data Corpus (the LSPM v2 release). CTBench datasets (SATO, SOTAB, SANTOS, Valentine, etc.) are from their respective original releases. DLTE parent tables are derived from TabFact and WTQ.

\paragraph{Splits.}
Row prediction uses fixed $80/10/10$ train/valid/test splits generated by row index under a fixed per-table random seed (seed $0$ for all but a small number of tables, with per-table seed overrides used to preserve usable class structure for classification targets), frozen once and reused across all experiments. Record linkage retains the original DeepMatcher (3:1:1 train/valid/test) and WDC splits; both benchmarks define exactly two tables per dataset, so table-disjoint evaluation does not apply.

For the four CTBench pairwise tasks marked $\dagger$ (join classification, column overlap, union classification, union regression), we generate table-disjoint splits as follows. All pairs from the source pair-level random splits are pooled, and the set of unique tables is partitioned randomly (seed~42) into disjoint train/valid/test sets at a 70/15/15 ratio. Each pair is then assigned to the split containing \emph{both} of its tables, and cross-partition pairs are discarded. For spider\_join, tables are grouped by database prefix before partitioning, so all tables from the same database land in the same split. Retention rates range from 53\% to 100\% of original pairs depending on the dataset. The remaining supervised table-pair task (table subset) already has table-disjoint splits in the source data.

DLTE uses a parent-table-level 827/207/345 train/dev/test split fixed at benchmark construction (before any pipeline evaluation) by two successive calls to \texttt{train\_test\_split}: first 75/25 train+dev vs.\ test, then 80/20 train vs.\ dev within the train+dev portion, yielding an effective 60/15/25 split. The split is random, not stratified by source (TabFact vs.\ WTQ). Given the 989/390 source composition, each split is expected to contain roughly 72\% TabFact and 28\% WTQ parents. Split manifests are included in the released code. All 1,120 canonical pipelines completed across all 5 rounds.

\paragraph{Hyperparameters.}
Supervised probe training follows the unified protocol of Sec.~\ref{sec:benchmark-protocol}: for each supervised probe task, we train both a linear head and a one-hidden-layer MLP with hidden size 256 using Adam (learning rate $10^{-3}$) for up to 100 epochs with early stopping on the validation set, and the headline score is the arithmetic average of the two heads.

\paragraph{Seeds.}
All supervised probe results, including record linkage, are averaged over 5 random seeds: 42, 52, 62, 72, 82.

\paragraph{Model wrappers and launch configurations.}
Exact model-specific wrapper parameters, truncation rules, and embedding-generation launch configurations are provided in the released codebase. Appendix~\ref{app:input-pipelines} states the benchmark-level policy. The codebase contains the executable per-model settings used to produce the released embeddings. For the two prior-based models, whose released weights evolve over time, we report both the package version and the checkpoint behind the reported results. \textsc{TabPFN} ran with \texttt{tabpfn} 6.4.1 (and \texttt{tabpfn-extensions} 0.2.2) and the package's default classifier checkpoint \texttt{tabpfn-v2.5-classifier-v2.5\_default.ckpt} (TabPFN-2.5~\cite{tabpfn25}). \textsc{TabICL} ran with \texttt{tabicl} 0.1.4 and checkpoint \texttt{tabicl-classifier-v1.1-0506.ckpt}. Table~\ref{tab:full-models} lists the checkpoint of every model. The upstream \textsc{TabSketchFM} release contains no checkpoint, so its weights come from our run of the upstream pretraining code, initialized from \texttt{bert-base-uncased}, and we release them with the benchmark. \textsc{Starmie} and the target-table SSL models are trained on each dataset and have no shared checkpoint.

\paragraph{Third-party model packages.}
TabPFN and TabICL are used as provided by their respective authors. Their internal weights are not modified during evaluation. The benchmark consumes only the exported intermediate row representations used by the shared probe protocol of Sec.~\ref{sec:benchmark-protocol}, never their task predictions as final benchmark outputs.

\section{Statistical Reporting}
\label{app:statistical}

For column/table tasks, we report means over 5 random seeds for all supervised probe results. Standard deviations are reported per cell in the main results tables (Tables~\ref{tab:main_results}, \ref{tab:r_main_results}) and in the aggregation ablation tables (Tables~\ref{tab:aggregation_ablation}, \ref{tab:aggregation_ablation_linear}). Training-free tasks (column clustering, schema matching) are deterministic given fixed embeddings. For row prediction, the reported metrics are macro-averaged across all 123 targets. The sole exception is \textsc{TabTransformer}, which covers 63 targets due to its categorical-feature requirement. For DLTE, end-to-end results are 5-round averages. We do not report confidence intervals for the normalized-rank aggregates, as these are summary statistics over heterogeneous per-task metrics. Per-task raw scores with standard deviations are the appropriate unit of statistical comparison.

\paragraph{Significance tests.}
To assess whether model differences are statistically meaningful, we apply Friedman omnibus tests and Holm-corrected pairwise Wilcoxon signed-rank tests, using per-task mean scores (averaged over 5~seeds) as the unit of analysis. Table~\ref{tab:significance} summarizes the results.

\begin{table}[h]
\centering
\small
\setlength{\tabcolsep}{5pt}
\caption{Friedman omnibus tests and Holm-corrected pairwise Wilcoxon signed-rank results. Each row uses per-task means (averaged over 5 seeds) as the unit of analysis. ``Sig.\ pairs'' reports the number of pairwise comparisons with $p_{\mathrm{adj}} < 0.05$ after Holm correction, out of $\binom{k}{2}$ total pairs for $k$ models. \textsc{TabTransformer} is excluded from the RBench rows because its partial target coverage (63 of 123) breaks the matched-sample requirement of Friedman/Wilcoxon tests, leaving 13 of the 14 non-baseline row models.}
\label{tab:significance}
\begin{tabular}{llccrc}
\toprule
\textbf{Scope} & \textbf{Suite} & \textbf{Tasks} & \textbf{Models} & \makecell{\textbf{Friedman}\\$\chi^2$\;($p$)} & \makecell{\textbf{Sig.\ pairs}\\(Holm)} \\
\midrule
All tasks         & CTBench & 13 &  8 & 28.4\;($1.9{\times}10^{-4}$)       & 4\,/\,28 \\
\quad Schema      & CTBench &  3 &  8 & 18.0\;(0.012)           & --- \\
\quad Join        & CTBench &  3 &  8 & 9.7\;(0.208)           & --- \\
\quad Union       & CTBench &  5 &  8 & 11.2\;(0.130)           & --- \\
\addlinespace[2pt]
Classification    & RBench  & 77 & 13 & 209.4\;($3.7{\times}10^{-38}$) & 26\,/\,78 \\
Regression        & RBench  & 46 & 13 & 154.6\;($6.5{\times}10^{-27}$) & 38\,/\,78 \\
\bottomrule
\end{tabular}
\end{table}

\begin{figure}[h]
\centering
\includegraphics[width=0.9\linewidth]{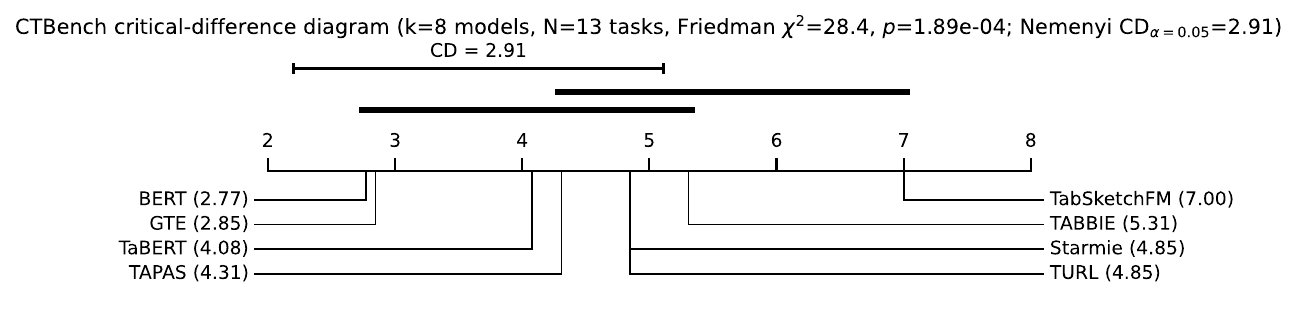}
\caption{CTBench critical-difference diagram (Dem\v{s}ar style). The horizontal axis shows mean rank across 13 CTBench tasks for the 8 fully-supported models. Horizontal bars group models whose mean-rank differences are below the Nemenyi critical difference (CD$=2.91$), so models sharing a bar are not significantly different at $\alpha=0.05$. Only \textsc{TabSketchFM}'s pairs against \textsc{BERT}, \textsc{GTE}, and \textsc{TaBERT} clear the CD threshold. Most of the top-eight CTBench models live in a single overlapping clique.}
\label{fig:ctbench_cd}
\end{figure}

On \textsc{TRL-CTbench} (13 tasks, 8 fully supported models), the Friedman test rejects the null hypothesis that all models perform equally ($\chi^2 = 28.4$, $p < 0.001$). However, only 4 of 28 pairwise Wilcoxon signed-rank comparisons are significant after Holm correction (all $p_{\mathrm{adj}} < 0.05$), and all four involve \textsc{TabSketchFM} being significantly weaker than the top models. The remaining models (\textsc{BERT}, \textsc{GTE}, \textsc{TaBERT}, \textsc{TAPAS}, \textsc{TURL}, \textsc{TABBIE}, and \textsc{Starmie}) are not significantly different from each other across the full task set. Per-family Friedman tests are significant for Schema ($p = 0.012$) but not for Join ($p = 0.208$) or Union ($p = 0.130$). Figure~\ref{fig:ctbench_cd} renders the corresponding Dem\v{s}ar critical-difference (CD) diagram~\cite{demsar2006} using post-hoc Nemenyi at $\alpha=0.05$: \textsc{BERT} holds the best mean rank (2.77), and under the CD threshold only \textsc{TabSketchFM}'s pairs against \textsc{BERT}, \textsc{GTE}, and \textsc{TaBERT} clear significance. Most CTBench models form a single overlapping clique, visually confirming the near-tie pattern.

The \textsc{TRL-CTbench} model$\times$task grid is ragged: not every model is applicable to every task, so a model's normalized-rank average can cover a different task subset than another model's. As a robustness check on the family orderings of Table~\ref{tab:main_results}, we re-aggregate each family with a Bradley-Terry paired-comparison model~\citep{hunter2004} that uses only observed head-to-head cells, with no imputation and no dropped models. Table~\ref{tab:bt_family} reports each model's Bradley-Terry rank and 95\% bootstrap rank interval within each of the four families. The re-aggregation agrees closely with the normalized-rank orderings (Spearman 0.994/0.927/1.000/0.939 for Schema/Join/Union/Grounding). Every family keeps its rank-1 model, and every pair of models that the two aggregations order differently has overlapping rank intervals. The intervals also show how certain each lead is. \textsc{BERT}'s Schema lead holds in every bootstrap resample (interval $[1,1]$, $P(\text{rank}\,1)=1.00$). The leads of \textsc{GTE} on Join ($[1,3]$, $P(\text{rank}\,1)=0.50$) and \textsc{TaBERT} on Grounding ($[1,3]$, $0.51$) are less certain. \textsc{TAPEX}'s Union lead rests on its three table-level Union tasks and has the widest interval of the four leads ($[1,8]$, $0.43$). We report this per-family view only and do not combine the four families into one leaderboard.

\input{tables/bt_family.tex}

On \textsc{TRL-Rbench}, the larger number of targets provides substantially more statistical power. Classification (77~targets, 13~models; \textsc{TabTransformer} excluded due to partial target coverage) yields $\chi^2 = 209.4$, $p = 3.7 \times 10^{-38}$, with 26/78 pairwise comparisons significant. Regression (46~targets, 13~models) yields $\chi^2 = 154.6$, $p = 6.5 \times 10^{-27}$, with 38/78 pairs significant. \textsc{TabICL}'s advantage over all other models is confirmed on both classification and regression (all $p_{\mathrm{adj}} < 0.001$). Among the remaining transfer-based and target-table learners, most pairwise differences are non-significant ($p_{\mathrm{adj}} = 1.0$), supporting the paper's conclusion that training regime and task family matter more than individual model choice within a regime.

\paragraph{\textsc{TRL-DLTE} per-stage significance.}
For DLTE we test whether each stage's model identity has a significant end-to-end effect, blocking by the remaining two stages and using the aggregate of 5-round mean scores per pipeline as the unit of analysis. All three stages have balanced complete-block designs: Stage~1 has 112 blocks (8 column $\times$ 14 row other-stage combinations) with 10 target models, Stage~2 has 140 blocks (10 tables $\times$ 14 rows) with 8 target models, and Stage~3 has 80 blocks (10 tables $\times$ 8 columns) with 14 target models. Table~\ref{tab:dlte_significance} reports Friedman omnibus statistics with Kendall's concordance $W$ as an effect-size summary, together with Holm-corrected pairwise Wilcoxon signed-rank tests for all three stages on the primary metric $\mathrm{UJ\text{-}H}$ and the complementary Cell~$F_1$ diagnostic. All six stage-metric combinations reject the null at $p \le 1.1{\times}10^{-13}$ (Stage~3 $\mathrm{UJ\text{-}H}$), with the other five well below $10^{-16}$, but the interesting picture is the effect-size pattern rather than the p-values. Stage~2 (column alignment) carries the single largest effect on both metrics (Kendall's $W = 0.79$ on Cell~$F_1$, $0.66$ on $\mathrm{UJ\text{-}H}$, with $26$/$28$ and $24$/$28$ Holm-significant pairs), consistent with the finding that column alignment is the most sensitive lever in the pipeline. Stage~1 (retrieval) has a moderate effect on both metrics ($W = 0.57$ on Cell~$F_1$ and $W = 0.33$ on $\mathrm{UJ\text{-}H}$, with $38$/$45$ and $38$/$45$ Holm-significant pairs). Stage~3 exhibits a striking metric-dependent asymmetry. On Cell~$F_1$ the effect is strong ($W = 0.61$, $\chi^2 = 638.7$) with $87$ of $91$ pairwise comparisons Holm-significant, reflecting the sharp separation between union-dedup specialists (\textsc{TabTransformer}, \textsc{SubTab}, \textsc{SAINT}) and the other row models on union-side raw cell recovery. On $\mathrm{UJ\text{-}H}$, by contrast, the effect collapses to $W = 0.09$ (weak) and only $31$ of $91$ pairwise comparisons survive Holm correction, even though the omnibus remains significant. This is not evidence of absent Stage~3 signal: the Oracle-RA diagnostic (Table~\ref{tab:oracle_ra}) shows a latent cross-row-model $\mathrm{UJ\text{-}H}$ spread of $0.546$, far above the end-to-end marginal span of $\sim 0.013$. The metric-dependent asymmetry instead reflects the DLTE composition bottleneck quantified in Sec.~\ref{sec:exp-dlte}: Cell~$F_1$ is sensitive to Stage~3 union-side behavior, which passes through the pipeline largely intact, while $\mathrm{UJ\text{-}H}$ additionally requires join-side recovery, which is masked by upstream retrieval/alignment error and therefore compresses the visible Stage~3 effect within the end-to-end pipeline. These tests quantify average stage effects, not the additivity of best compositions. Large average effects do not by themselves determine which model combination is best, and top DLTE quality depends on non-additive compositional fit.

\input{tables/dlte_significance/dlte_significance_combined.tex}

\paragraph{Table-clustered re-test of the row-prediction significance tests.}
As a further robustness check, we re-ran the \textsc{TRL-Rbench} row-prediction significance tests of Table~\ref{tab:significance} with the table, rather than the individual target, as the clustering unit, addressing the statistical dependence among multiple targets drawn from the same table. The omnibus conclusion is unchanged: the Friedman test remains decisive on both classification and regression ($p < 10^{-18}$), and \textsc{TabICL}'s lead over the strongest model of every other adaptation regime (Sec.~\ref{sec:exp-row}) still holds at $p < 0.001$.

\section{Dataset Counting Protocol}
\label{app:counting}

The total dataset count depends on the grouping granularity. At the finest level, treating each distinct dataset source as one entry, the benchmark contains 87 datasets (20 CTBench + 50 OpenML row-prediction + 16 record-linkage + 1 DLTE enrichment lake). Alternative aggregations include: 84 (grouping the 4 OpenData regional splits as one source), 77 (additionally grouping the 4 WDC Products size variants as one entry and pairing each DeepMatcher dirty variant with its clean counterpart), and 28 (additionally grouping all 50 OpenML row-prediction tables as one curated collection). Throughout the main paper, we use the dataset-source level as the default unless otherwise noted.

\section{Broader Impact}
\label{app:broader-impact}

A benchmark for tabular representations can improve scientific comparability and reduce evaluation fragmentation. It can also help practitioners identify when specialized table-aware models are necessary and when simpler frozen encoders suffice.

At the same time, stronger table representations can be used in settings involving sensitive records, schema inference, or large-scale data linkage. Benchmark releases should therefore carefully respect data licenses, privacy constraints, and documentation requirements. Because the released assets are drawn from many public sources under heterogeneous terms, we document per-source provenance, license, and redistribution status in an asset manifest, and where a source license does not permit redistribution, we release reconstruction scripts and manifests rather than mirrored raw files (Appendix~\ref{app:licenses}). Anyone building on the rehosted corpora inherits these same license terms and should consult the manifest before further redistribution. In particular, record linkage and data lake retrieval settings can raise concerns around surveillance, re-identification, and inappropriate dataset fusion. \textsc{TRL-DLTE} specifically evaluates the composition of retrieval, column-alignment, and row-matching capabilities that link and merge records across otherwise-unrelated tables, and techniques that transfer well on this benchmark could equally lower the cost of unauthorized cross-source entity resolution outside of it. A responsible release should document dataset provenance and usage restrictions.

\section{Licenses and Asset Documentation}
\label{app:licenses}

\begin{table}[t]
\centering
\footnotesize
\setlength{\tabcolsep}{3pt}
\renewcommand{\arraystretch}{1.05}
\caption{Licenses and provenance for assets used in \benchmark. License entries reflect the upstream code/data licenses observed at the official source as of preparation. Where the source repository or release page does not state an explicit dataset license, we mark the cell accordingly. ``Code'' and ``data'' licenses are listed separately when they differ. For benchmarks redistributed through LakeBench, both the LakeBench redistribution license and the upstream license are noted.}
\label{tab:licenses}
\begin{tabularx}{\textwidth}{>{\raggedright\arraybackslash}p{0.18\textwidth} l Y Y}
\toprule
\textbf{Asset} & \textbf{Citation} & \textbf{License} & \textbf{Notes} \\
\midrule
OpenML tables & \cite{openml} & Per-dataset (commonly CC0 / CC-BY) & Via OpenML API; licenses are set per dataset by the uploader \\
SATO & \cite{sato} & Apache 2.0 & Public release; tables from the WebTables corpus within VizNet \\
SOTAB & \cite{sotab} & Not explicitly stated on dataset page (CEUR paper itself: CC-BY 4.0) & Public release on Web Data Commons; benchmark data must not appear in training corpora (per WDC notice) \\
WikiCT (relation) & \cite{turl} & Apache 2.0 (TURL code); CC-BY-SA inherited from upstream Wikipedia content & Relation-extraction split from TURL; tables from the WikiTables / TabEL corpus \\
Wiki Containment / Wiki Union & \cite{lakebench_srinivas} & CC-BY 4.0 (LakeBench Zenodo record); CC-BY-SA 4.0 (label files and Wikipedia content) & Derived from Wikipedia tables; splits from LakeBench (Zenodo archives \texttt{wiki-containment} and \texttt{wiki-union}) \\
SANTOS & \cite{santos} & BSD-3-Clause (code); CC-BY 4.0 (data on Zenodo) & Public release \\
UGEN & \cite{ugen} & MIT & GitHub: northeastern-datalab/gen \\
TUS / TUS-hard & \cite{tus_original} & Unspecified (source repo has no LICENSE file) & TUS from Nargesian et al.; TUS-hard is a repo-derived low-overlap subset (Hungarian-matched containment $<$ 0.70) \\
Pylon & \cite{pylon} & Unspecified (upstream repo has no LICENSE file) & Downloaded from the official source (Appendix~\ref{app:pylon_union}) \\
Valentine & \cite{valentine} & Apache 2.0 & Public release; benchmark assets in a separate (also Apache 2.0) repo \\
Spider Join & \cite{spider,lakebench_srinivas} & CC-BY-SA 4.0 (Spider upstream; ShareAlike propagates to Spider-derived artifacts); CC-BY 4.0 (LakeBench-original metadata) & Databases from Spider; join-classification benchmark from LakeBench \\
OpenData (main / CAN / USA / UK / SG) & \cite{lakebench_deng} & Unspecified (BIT-DataLab/LakeBench source repo has no LICENSE; data hosted on Google Drive) & Join- and union-search benchmarks from the Deng et al.\ LakeBench; per-portal open-data licenses for source tables (Canadian / UK / US / Singapore open-data portals) \\
ECB Union / CKAN Subset & \cite{lakebench_srinivas} & CC-BY 4.0 (LakeBench Zenodo); per-portal open licenses for source tables & Benchmarks from LakeBench; source data from open govt.\ portals \\
DeepMatcher & \cite{deepmatcher} & BSD-3-Clause (code); benchmarks released without explicit dataset license & Public release \\
WDC Products & \cite{wdcproducts,wdcproducts_lspm} & Not explicitly stated on Web Data Commons; publicly released & Training/gold-standard release at WWW '19 Companion (ECNLP); LSPM v2 extension from WIMS 2020 \\
WTQ & \cite{wikitablequestions} & CC-BY-SA 4.0 & Table QA + DLTE parent tables \\
NQ-Tables & \cite{nqtables} & Apache 2.0 (TAPAS release artifact, NQ source repo); CC-BY-SA 3.0 (Wikipedia text upstream, pre-2023 NQ era) & Public release \\
TabFact & \cite{tabfact} & CC-BY 4.0 (data); MIT (code) & DLTE parent tables \\
GitTables & \cite{gittables} & CC-BY 4.0 (GitTables site); CC0 1.0 (Zenodo record of GitTables 1M); per-table licenses inherited from the source GitHub repositories & Parent tables of the rebuilt \textsc{TRL-DLTE} lake (Appendix~\ref{app:dlte_robustness}); the corpus keeps only tables whose source repository license allows distribution and records each table's license in its metadata \\
CKAN distractors & \cite{lakebench_srinivas} & CC-BY 4.0 (LakeBench Zenodo); per-portal open licenses for source tables & \benchmark-derived selection from the LakeBench CKAN Subset pool; not a separately published LakeBench split \\
LakeBench (Srinivas et al.) & \cite{lakebench_srinivas} & CC BY-NC-ND 4.0 (code at \texttt{IBM/tabsketchfm}); CC-BY 4.0 / CC-BY-SA 4.0 (data on Zenodo) & Public release; \benchmark\ derives only from the CC-BY 4.0 Zenodo data, not the NC-ND code \\
TabArena & \cite{tabarena} & Apache 2.0 & Public release; per-table licenses inherited from upstream sources (UCI, Kaggle, OpenML, etc.) \\
\bottomrule
\end{tabularx}
\end{table}

All source datasets are used under their original licenses. \benchmark\ creates derived benchmark assets from public source data, including curated target selections, rewritten row-pair tasks, new table-disjoint split manifests, and \textsc{TRL-DLTE} table fragments. For each source, we document the transformation applied (label repairs, split regeneration, row-pair rewrites, fragmentation) and respect the redistribution terms of the original license: where source licenses permit redistribution we release the derived files directly, and where they do not we release reconstruction scripts and manifests rather than mirrored raw files. For platform-hosted collections (e.g., OpenML, CKAN / open government portals), licenses can vary by dataset. The release therefore includes a per-asset manifest with the original source, URL, license, and redistribution status. The benchmark code is released under an open-source license.

%% file: tables/row_curation_per_table.tex
\begin{scriptsize}
\setlength{\tabcolsep}{2.5pt}
\begin{longtable}{@{}r >{\raggedright\arraybackslash}p{2.35cm} l r r >{\raggedright\arraybackslash}p{3.75cm} >{\raggedright\arraybackslash}p{2.75cm}@{}}
\caption{\textbf{Per-table curation of the 50 \textsc{TRL-Rbench} row-prediction tables.} Released targets with task type (C = classification, R = regression) and the repair-log actions of Table~\ref{tab:repair_log}. Domains abbreviate Figure~\ref{fig:row_prediction_pie}(a). Inputs = columns left after removing all targets. $^{\ast}$\,Near-trivial target kept as an exception.} \\
\label{tab:row_per_table} \\
\toprule
\textbf{ID} & \textbf{Name} & \textbf{Domain} & \textbf{Rows} & \textbf{Inputs} & \textbf{Released targets} & \textbf{Repair log} \\
\midrule
\endfirsthead
\toprule
\textbf{ID} & \textbf{Name} & \textbf{Domain} & \textbf{Rows} & \textbf{Inputs} & \textbf{Released targets} & \textbf{Repair log} \\
\midrule
\endhead
\midrule
\multicolumn{7}{r}{\emph{(continued on next page)}} \\
\endfoot
\bottomrule
\endlastfoot
3 & kr-vs-kp & Games/other & 3{,}196 & 34 & \texttt{class} (C), \texttt{katri} (C) & Removed \texttt{hdchk} \\
38 & sick & Healthcare & 3{,}772 & 25 & \texttt{Class} (C), \texttt{TSH} (R), \texttt{referral\_\allowbreak{}source} (C) &  \\
458 & analcatdata\_\allowbreak{}authorship & Games/other & 841 & 69 & \texttt{Author} (C)$^{\ast}$, \texttt{the} (R) & Kept near-trivial target$^{\ast}$ \\
1063 & kc2 & Software & 522 & 20 & \texttt{problems} (C), \texttt{loc} (R) & Re-split (seed 2); \texttt{v(g)}, \texttt{branchCount} removed (still unusable); \texttt{loc} added \\
1486 & nomao & Software & 34{,}465 & 117 & \texttt{Class} (C), \texttt{V89} (R) &  \\
4534 & PhishingWebsites & Software & 11{,}055 & 28 & \texttt{Result} (C), \texttt{Submitting\_\allowbreak{}to\_\allowbreak{}email} (C), \texttt{SSLfinal\_\allowbreak{}State} (C) &  \\
6332 & cylinder-bands & Engineering & 540 & 33 & \texttt{band\_\allowbreak{}type} (C), \texttt{press\_\allowbreak{}speed} (R), \texttt{paper\_\allowbreak{}type} (C) &  \\
40668 & connect-4 & Games/other & 67{,}557 & 41 & \texttt{class} (C), \texttt{d1} (C)$^{\ast}$ & Kept near-trivial target$^{\ast}$ \\
40966 & MiceProtein & Healthcare & 1{,}080 & 75 & \texttt{PSD95\_\allowbreak{}N} (R), \texttt{IL1B\_\allowbreak{}N} (R), \texttt{Tau\_\allowbreak{}N} (R) & Removed \texttt{class} \\
40978 & Internet-Advertisements & Business & 3{,}279 & 1{,}557 & \texttt{class} (C), \texttt{local} (C) &  \\
44958 & auction\_\allowbreak{}verification & Finance & 2{,}043 & 5 & \texttt{verification.time} (R), \texttt{property.winner} (C), \texttt{property.price} (C) &  \\
44967 & student\_\allowbreak{}performance\_\allowbreak{}por & Education & 649 & 28 & \texttt{G3} (R), \texttt{Walc} (C), \texttt{internet} (C) &  \\
44975 & wave\_\allowbreak{}energy & Sciences & 72{,}000 & 47 & \texttt{energy\_\allowbreak{}total} (R), \texttt{energy1} (R) &  \\
44984 & cps88wages & Finance & 28{,}155 & 5 & \texttt{wage} (R), \texttt{smsa} (C) &  \\
44992 & fps\_\allowbreak{}benchmark & Games/other & 24{,}624 & 38 & \texttt{FPS} (R), \texttt{GpuMemoryType} (C)$^{\ast}$ & Removed \texttt{GpuName}, \texttt{CpuName}; kept near-trivial target$^{\ast}$ \\
46906 & anneal & Engineering & 898 & 29 & \texttt{classes} (C), \texttt{width} (R) & Removed \texttt{packing} \\
46907 & Another-Dataset-on-used-Fiat-500 & Engineering & 1{,}538 & 6 & \texttt{price} (R), \texttt{previous\_\allowbreak{}owners} (C) &  \\
46908 & APSFailure & Engineering & 76{,}000 & 167 & \texttt{AirPressureSystemFailure} (C), \texttt{cs\_\allowbreak{}000} (R) & Removed \texttt{as\_000}; original target removed, then restored \\
46910 & bank-marketing & Finance & 45{,}211 & 11 & \texttt{SubscribeTermDeposit} (C), \texttt{education} (C), \texttt{housing} (C) &  \\
46911 & Bank\_\allowbreak{}Customer\_\allowbreak{}Churn & Finance & 10{,}000 & 9 & \texttt{churn} (C), \texttt{country} (C) &  \\
46912 & Bioresponse & Healthcare & 3{,}751 & 1{,}775 & \texttt{MoleculeElicitsResponse} (C), \texttt{D52} (R) &  \\
46915 & churn & Business & 5{,}000 & 17 & \texttt{CustomerChurned} (C), \texttt{total\_\allowbreak{}day\_\allowbreak{}calls} (R), \texttt{international\_\allowbreak{}plan} (C) &  \\
46916 & coil2000\_\allowbreak{}insurance\_\allowbreak{}policies & Finance & 9{,}822 & 83 & \texttt{MobileHomePolicy} (C), \texttt{numberOfHouses} (R), \texttt{customerSubtype} (C) &  \\
46918 & credit-g & Finance & 1{,}000 & 18 & \texttt{good\_\allowbreak{}or\_\allowbreak{}bad\_\allowbreak{}customer} (C), \texttt{credit\_\allowbreak{}amount} (R), \texttt{credit\_\allowbreak{}purpose} (C) &  \\
46919 & credit\_\allowbreak{}card\_\allowbreak{}clients\_\allowbreak{}default & Finance & 30{,}000 & 21 & \texttt{Default\allowbreak{}On\allowbreak{}Payment\allowbreak{}Next\allowbreak{}Month} (C), \texttt{BILL\_\allowbreak{}AMT6} (R), \texttt{PAY\_\allowbreak{}6} (C) &  \\
46920 & customer\_\allowbreak{}satisfaction\_\allowbreak{}in\_\allowbreak{}airline & Business & 129{,}880 & 19 & \texttt{satisfaction} (C), \texttt{FlightDistance} (R), \texttt{Class} (C) &  \\
46922 & Diabetes130US & Healthcare & 71{,}518 & 43 & \texttt{EarlyReadmission} (C), \texttt{num\_\allowbreak{}lab\_\allowbreak{}procedures} (R) &  \\
46923 & diamonds & Finance & 53{,}940 & 8 & \texttt{price} (R), \texttt{cut} (C) &  \\
46927 & Fitness\_\allowbreak{}Club & Business & 1{,}500 & 5 & \texttt{attended} (C), \texttt{weight} (R) &  \\
46929 & GiveMeSomeCredit & Finance & 150{,}000 & 9 & \texttt{Financial\allowbreak{}Distress\allowbreak{}Next\allowbreak{}Two\allowbreak{}Years} (C), \texttt{Number\allowbreak{}Of\allowbreak{}Open\allowbreak{}Credit\allowbreak{}Lines\allowbreak{}And\allowbreak{}Loans} (R) &  \\
46930 & hazelnut-spread-contaminant-detection & Engineering & 2{,}400 & 29 & \texttt{Contaminated} (C), \texttt{s32} (R) &  \\
46932 & heloc & Finance & 10{,}459 & 22 & \texttt{RiskPerformance} (C), \texttt{AverageMInFile} (R) &  \\
46933 & hiva\_\allowbreak{}agnostic & Healthcare & 3{,}845 & 1{,}616 & \texttt{CompoundActivity} (C), \texttt{molecule\_\allowbreak{}structure\_\allowbreak{}property\_\allowbreak{}1564} (C)$^{\ast}$ & Kept near-trivial target$^{\ast}$ \\
46934 & houses & Finance & 20{,}640 & 6 & \texttt{LnMedianHouseValue} (R), \texttt{HousingMedianAge} (R), \texttt{TotalBedrooms} (R) &  \\
46935 & HR\_\allowbreak{}Analytics\_\allowbreak{}Job\_\allowbreak{}Change\_\allowbreak{}of\_\allowbreak{}Data\_\allowbreak{}Scientists & Business & 19{,}158 & 10 & \texttt{LookingForJobChange} (C), \texttt{training\_\allowbreak{}hours} (R), \texttt{education\_\allowbreak{}level} (C) &  \\
46937 & in\_\allowbreak{}vehicle\_\allowbreak{}coupon\_\allowbreak{}recommendation & Business & 12{,}684 & 23 & \texttt{AcceptCoupon} (C), \texttt{CoffeeHouse} (C) &  \\
46939 & kddcup09\_\allowbreak{}appetency & Business & 50{,}000 & 208 & \texttt{Var28} (R), \texttt{Var7} (C) & Removed \texttt{appetency}; re-split (seed 51) \\
46940 & Marketing\_\allowbreak{}Campaign & Business & 2{,}240 & 23 & \texttt{Response} (C), \texttt{Year\_\allowbreak{}Birth} (R), \texttt{Education} (C) &  \\
46950 & polish\_\allowbreak{}companies\_\allowbreak{}bankruptcy & Finance & 5{,}910 & 62 & \texttt{company\_\allowbreak{}bankrupt} (C), \texttt{log\_\allowbreak{}total\_\allowbreak{}assets} (R), \texttt{operating\_\allowbreak{}profit\_\allowbreak{}to\_\allowbreak{}financial\_\allowbreak{}expenses} (R) &  \\
46952 & qsar-biodeg & Sciences & 1{,}054 & 39 & \texttt{Biodegradable} (C), \texttt{Weighted\_\allowbreak{}Hyper\allowbreak{}Wiener\_\allowbreak{}Index\_\allowbreak{}Burden\_\allowbreak{}Matrix} (R), \texttt{Num\_\allowbreak{}Esters} (C) & Re-split (seed 129) \\
46955 & SDSS17 & Sciences & 78{,}053 & 10 & \texttt{ObjectType} (C), \texttt{redshift} (R) &  \\
46956 & seismic-bumps & Engineering & 2{,}584 & 13 & \texttt{HighEnergySeismicBump} (C), \texttt{gpuls} (R), \texttt{ghazard} (C) &  \\
46958 & splice & Sciences & 3{,}190 & 59 & \texttt{SiteType} (C), \texttt{position\_\allowbreak{}6} (C) & Re-split (seed 166) \\
46960 & students\_\allowbreak{}dropout\_\allowbreak{}and\_\allowbreak{}academic\_\allowbreak{}success & Education & 4{,}424 & 34 & \texttt{AcademicOutcome} (C), \texttt{Previous\_\allowbreak{}qualification\_\allowbreak{}grade} (R), \texttt{Scholarship\_\allowbreak{}holder} (C) &  \\
46961 & superconductivity & Sciences & 21{,}263 & 80 & \texttt{critical\_\allowbreak{}temp} (R), \texttt{number\_\allowbreak{}of\_\allowbreak{}elements} (C) &  \\
46963 & website\_\allowbreak{}phishing & Software & 1{,}353 & 7 & \texttt{WebsiteType} (C), \texttt{web\_\allowbreak{}traffic} (C), \texttt{SSLfinal\_\allowbreak{}State} (C) &  \\
46964 & wine\_\allowbreak{}quality & Sciences & 6{,}497 & 10 & \texttt{median\_\allowbreak{}wine\_\allowbreak{}quality} (R), \texttt{wine\_\allowbreak{}color} (C)$^{\ast}$, \texttt{pH} (R) & Kept near-trivial target$^{\ast}$ \\
46969 & NATICUSdroid & Software & 7{,}491 & 84 & \texttt{Malware} (C), \texttt{android.permission.CAMERA} (C), \texttt{android.permission.ACCESS\_\allowbreak{}FINE\_\allowbreak{}LOCATION} (C) &  \\
46979 & jm1 & Software & 10{,}885 & 20 & \texttt{defects} (C), \texttt{iv(g)} (R) &  \\
46980 & MIC & Healthcare & 1{,}699 & 109 & \texttt{LET\_\allowbreak{}IS} (C), \texttt{ROE} (R), \texttt{DLIT\_\allowbreak{}AG} (C) &  \\
\end{longtable}
\end{scriptsize}

%% file: tables/baselines_coverage.tex
\begin{table}[t]
\centering
\small
\setlength{\tabcolsep}{3pt}
\caption{Per-task applicability of task-local baselines in \benchmark. \cmark\ = baseline is evaluated on the task. Blank = not applicable. Coverage rationale is in Appendix~\ref{app:baselines}.}
\label{tab:baselines_coverage}
\begin{tabular}{l ccccccccc}
\toprule
 & \textbf{Rnd.} & \textbf{TF-IDF} & \textbf{Chance} & \textbf{Dum.} & \makecell{\textbf{Inv.}\\\textbf{Idx.}} & \makecell{\textbf{Hung.}\\\textbf{Set}} & \textbf{Jacc.} & \textbf{Dist.} & \makecell{\textbf{Cos.}\\\textbf{Thr.}} \\
\textbf{Task} & & & & & & & & & \\
\midrule
Column Clustering & \cmark & \cmark & \cmark &  &  &  &  &  &  \\
Column Type Prediction & \cmark & \cmark & \cmark & \cmark &  &  &  &  &  \\
Column Relation Prediction & \cmark &  & \cmark & \cmark &  &  &  &  &  \\
Join Search (cosine) & \cmark &  & \cmark &  & \cmark &  &  &  &  \\
Join Search (learned) & \cmark &  & \cmark &  &  &  &  &  &  \\
Column Overlap & \cmark &  & \cmark & \cmark &  &  &  &  &  \\
Union Search & \cmark &  & \cmark &  &  & \cmark &  &  &  \\
Schema Matching & \cmark &  & \cmark &  &  &  & \cmark & \cmark &  \\
Table QA & \cmark &  & \cmark &  &  &  &  &  &  \\
Join Classification & \cmark &  & \cmark & \cmark &  &  &  &  &  \\
Table Subset & \cmark &  & \cmark & \cmark &  &  &  &  &  \\
Union Classification & \cmark &  & \cmark & \cmark &  &  &  &  &  \\
Union Regression & \cmark &  & \cmark & \cmark &  &  &  &  &  \\
Table Retrieval & \cmark &  & \cmark &  &  &  &  &  &  \\
Record Linkage & \cmark &  & \cmark & \cmark &  &  &  &  & \cmark \\
Row Prediction & \cmark &  & \cmark & \cmark &  &  &  &  &  \\
\bottomrule
\end{tabular}
\end{table}

%% file: tables/baselines_nonneural.tex
\begin{table}[t]
\centering
\small
\setlength{\tabcolsep}{5pt}
\caption{Non-neural matching baselines, per-dataset 5-seed means. Values are deterministic given fixed inputs. Reported $\pm$ indicates seed-to-seed variation in the upstream data pipeline. Metrics: col-MAP for Inv.-Index Containment, MAP@10 for Hungarian Set Match, Recall@GT for Valentine matchers.}
\label{tab:baselines_nonneural}
\begin{tabular}{ll c r}
\toprule
\textbf{Baseline} & \textbf{Dataset} & \textbf{Value} & \textbf{$n_\text{seeds}$} \\
\midrule
Inv.-Index Cont. & \texttt{opendata} & $0.148$\,\std{0.000} & 5 \\
 & \texttt{opendata\_CAN} & $0.182$\,\std{0.000} & 5 \\
 & \texttt{opendata\_USA} & $0.138$\,\std{0.000} & 5 \\
 & \texttt{opendata\_UK\_SG} & $0.152$\,\std{0.000} & 5 \\
\midrule
Hungarian Set Match & \texttt{santos} & $0.975$\,\std{0.000} & 5 \\
 & \texttt{tus} & $1.000$\,\std{0.000} & 5 \\
 & \texttt{tus\_hard} & $0.008$\,\std{0.000} & 5 \\
 & \texttt{ugen\_v1} & $0.647$\,\std{0.000} & 5 \\
 & \texttt{ugen\_v2} & $0.239$\,\std{0.000} & 5 \\
\midrule
Jaccard (Valentine) & \texttt{valentine} & $0.473$\,\std{0.000} & 5 \\
\midrule
Distribution (Valentine) & \texttt{valentine} & $0.394$\,\std{0.002} & 5 \\
\bottomrule
\end{tabular}
\end{table}

%% file: tables/baselines_analytical.tex
\begin{table}[t]
\centering
\small
\setlength{\tabcolsep}{4pt}
\caption{Random, Dummy, and TF-IDF baselines per probe task (5-seed mean $\pm$ std). Values are computed under the paper's main convention (Tables~\ref{tab:main_results} and \ref{tab:r_main_results}): avg(MLP, linear) probe where both heads exist, best table-aggregation per task, strict (table-disjoint) splits for the $\dagger$-marked pairwise tasks (\textsc{ColOverlap}, \textsc{JoinCls}, \textsc{UnionCls}, \textsc{UnionReg}), \textsc{TblRet} via the model\_only retrieval pipeline, and binary match-class $F_1$ for record linkage. \textsc{na} = baseline not applicable to the task per Table~\ref{tab:baselines_coverage}.}
\label{tab:baselines_analytical}
\begin{tabular}{l l c c c}
\toprule
\textbf{Task (metric)} & & \textbf{Random} & \textbf{Dummy} & \textbf{TF-IDF} \\
\midrule
ColType ($F_1$) & & $0.132$\,\std{0.003} & $0.178$\,\std{0.000} & $0.813$\,\std{0.007} \\
ColClust (NMI) & & $0.032$\,\std{0.001} & \textsc{na} & $0.400$\,\std{0.001} \\
ColRel ($F_1$) & & $0.015$\,\std{0.001} & $0.000$\,\std{0.000} & \textsc{na} \\
ColOverlap (nRMSE) & & $1.014$\,\std{0.003} & $1.000$\,\std{0.000} & \textsc{na} \\
JoinCls ($F_1$) & & $0.516$\,\std{0.014} & $0.414$\,\std{0.000} & \textsc{na} \\
UnionCls ($F_1$) & & $0.500$\,\std{0.004} & $0.331$\,\std{0.000} & \textsc{na} \\
UnionReg (nRMSE) & & $1.138$\,\std{0.024} & $1.001$\,\std{0.000} & \textsc{na} \\
TblSubset ($F_1$) & & $0.458$\,\std{0.027} & $0.369$\,\std{0.000} & \textsc{na} \\
TblQA (Acc) & & $0.204$\,\std{0.004} & \textsc{na} & \textsc{na} \\
TblRet (MRR) & & $0.131$\,\std{0.120} & \textsc{na} & \textsc{na} \\
\midrule
RowPred (AUROC) & & $0.506$\,\std{0.001} & $0.500$\,\std{0.000} & \textsc{na} \\
RowPred (Macro-$F_1$) & & $0.348$\,\std{0.001} & $0.304$\,\std{0.000} & \textsc{na} \\
RowPred Reg. (SGM$\downarrow$) & & $1.103$\,\std{0.000} & $1.004$\,\std{0.000} & \textsc{na} \\
\midrule
RecLink DM-C ($F_1$) & & $0.179$\,\std{0.007} & $0.000$\,\std{0.000} & \textsc{na} \\
RecLink DM-D ($F_1$) & & $0.223$\,\std{0.003} & $0.000$\,\std{0.000} & \textsc{na} \\
RecLink WDC ($F_1$) & & $0.128$\,\std{0.003} & $0.000$\,\std{0.000} & \textsc{na} \\
\bottomrule
\end{tabular}
\end{table}

%% file: tables/baselines_cosine.tex
\begin{table}[t]
\centering
\small
\setlength{\tabcolsep}{5pt}
\caption{Cosine-threshold baseline on record linkage (binary match-class $F_1$, 5-seed mean $\pm$ std). An unsupervised baseline that thresholds cosine similarity between frozen row embeddings. The same numbers appear in the ``Cos'' column of Table~\ref{tab:ablation_rl_head}.}
\label{tab:baselines_cosine}
\begin{tabular}{l cccc}
\toprule
\textbf{Model} & \textbf{DM-C} & \textbf{DM-D} & \textbf{WDC} & \textbf{All} \\
\midrule
BERT & $0.390$\,\std{0.000} & $0.315$\,\std{0.000} & $0.390$\,\std{0.000} & $0.371$\,\std{0.000} \\
GTE & $0.698$\,\std{0.000} & $0.728$\,\std{0.000} & $0.511$\,\std{0.000} & $0.659$\,\std{0.000} \\
TABBIE & $0.309$\,\std{0.000} & $0.296$\,\std{0.000} & $0.389$\,\std{0.000} & $0.326$\,\std{0.000} \\
TUTA & $0.363$\,\std{0.000} & $0.397$\,\std{0.000} & $0.354$\,\std{0.000} & $0.369$\,\std{0.000} \\
TabICL & $0.377$\,\std{0.000} & $0.328$\,\std{0.000} & $0.341$\,\std{0.000} & $0.356$\,\std{0.000} \\
TabPFN & $0.260$\,\std{0.007} & $0.285$\,\std{0.000} & $0.387$\,\std{0.000} & $0.298$\,\std{0.004} \\
VIME & $0.242$\,\std{0.000} & $0.295$\,\std{0.000} & $0.423$\,\std{0.000} & $0.301$\,\std{0.000} \\
SCARF & $0.350$\,\std{0.001} & $0.352$\,\std{0.002} & $0.357$\,\std{0.000} & $0.352$\,\std{0.001} \\
DAE & $0.265$\,\std{0.000} & $0.297$\,\std{0.000} & $0.429$\,\std{0.000} & $0.314$\,\std{0.000} \\
TabBinning & $0.387$\,\std{0.002} & $0.421$\,\std{0.002} & $0.383$\,\std{0.000} & $0.395$\,\std{0.002} \\
SAINT & $0.251$\,\std{0.002} & $0.286$\,\std{0.008} & $0.429$\,\std{0.000} & $0.304$\,\std{0.003} \\
SubTab & $0.275$\,\std{0.000} & $0.331$\,\std{0.000} & $0.428$\,\std{0.000} & $0.327$\,\std{0.000} \\
TabTransformer & $0.257$\,\std{0.000} & $0.301$\,\std{0.000} & $0.426$\,\std{0.000} & $0.310$\,\std{0.000} \\
TransTab & $0.410$\,\std{0.001} & $0.311$\,\std{0.000} & $0.567$\,\std{0.004} & $0.425$\,\std{0.001} \\
\bottomrule
\end{tabular}
\end{table}

%% file: tables/label_efficiency.tex
\begin{table}[t]
\centering
\small
\caption{Label-efficiency reference on \textsc{TRL-Rbench} classification targets (unranked). Win rate is the fraction of targets on which the AUROC of the best of four frozen-embedding linear probes (\textsc{BERT}, \textsc{GTE}, \textsc{TabICL}, \textsc{TabPFN}) exceeds that of a tree model trained on raw features with the same $n$ training rows (5 resamples per budget, rows shared by all methods, fixed test split). The trees are untuned (200 trees, learning rate 0.1, maximum depth 6 for XGBoost and 31 leaves for LightGBM). The bottom row gives the number of targets contributing at each budget.}
\label{tab:label_efficiency}
\begin{tabular}{lccccccc}
\toprule
 & \multicolumn{7}{c}{Label budget $n$} \\
\cmidrule(lr){2-8}
\textbf{Statistic} & 32 & 64 & 128 & 256 & 512 & 1{,}024 & 2{,}048 \\
\midrule
Win rate vs.\ LightGBM & 0.69 & 0.68 & 0.57 & 0.57 & 0.54 & 0.49 & 0.48 \\
Win rate vs.\ XGBoost & 0.62 & 0.58 & 0.55 & 0.53 & 0.44 & 0.44 & 0.42 \\
\midrule
Targets contributing (of 77) & 77 & 77 & 77 & 77 & 72 & 63 & 52 \\
\bottomrule
\end{tabular}
\end{table}

%% file: tables/table_size_buckets.tex
\begin{small}
\setlength{\tabcolsep}{4pt}
\begin{longtable}{l l r r r r}
\caption{
\textbf{Per-dataset cell-footprint statistics across TRL-Bench suites.}
Each row is one of the 87 dataset-source entries in the benchmark, grouped by
benchmark category (Schema, Joinability, Unionability, Grounding, Row Prediction,
Record Linkage, DLTE).
Cell footprint $F_{\mathrm{cell}}(T) = n_{\mathrm{row}}(T)\,n_{\mathrm{col}}(T)$,
i.e.\ row count times column count, is computed once per counted loadable table
input under its dataset role. Labeled pairs and train/valid/test split indices
are not expanded into additional table inputs.
Mean, median, and Std.\ are computed over the counted inputs for that dataset.
Std.\ uses the population convention (\texttt{ddof}=0) when at least two inputs
are present.
Row Prediction entries contain one counted feature table each, so Mean equals
Median. Their Std.\ cells are shown with a dash because there is no within-dataset
dispersion to summarize.
} \\
\label{tab:table-size-buckets} \\
\toprule
Dataset & Category & \# table inputs & Mean (cells) & Median (cells) & Std.\ (cells) \\
\midrule
\endfirsthead
\toprule
Dataset & Category & \# table inputs & Mean (cells) & Median (cells) & Std.\ (cells) \\
\midrule
\endhead
\midrule
\multicolumn{6}{r}{\emph{(continued on next page)}} \\
\endfoot
\bottomrule
\endlastfoot
\multicolumn{6}{l}{\emph{Schema} \hfill (col)} \\
  SOTAB & Schema & 72,629 & 1,887 & 260 & 9,922 \\
  WikiCT (rel.) & Schema & 53,567 & 42 & 22 & 105 \\
  sato & Schema & 78,733 & 29 & 10 & 120 \\
\midrule
\multicolumn{6}{l}{\emph{Joinability} \hfill (col/tbl)} \\
  OpenData (main) & Joinability & 16,823 & 1,025,435 & 258,792 & 3,585,105 \\
  OpenData CAN & Joinability & 4,960 & 1,245,433 & 359,628 & 3,858,684 \\
  OpenData UK/SG & Joinability & 3,090 & 272,228 & 54,610 & 2,082,714 \\
  OpenData USA & Joinability & 5,165 & 1,087,988 & 263,340 & 4,209,460 \\
  Spider Join & Joinability & 15,996 & 104,286 & 1,000 & 852,833 \\
  Wiki Containment & Joinability & 39,084 & 134 & 110 & 89 \\
\midrule
\multicolumn{6}{l}{\emph{Unionability} \hfill (col/tbl)} \\
  CKAN Subset & Unionability & 36,846 & 33,039 & 6,696 & 87,250 \\
  ECB Union & Unionability & 4,226 & 11,457 & 1,134 & 49,893 \\
  SANTOS & Unionability & 600 & 96,975 & 17,116 & 244,244 \\
  TUS & Unionability & 1,651 & 41,815 & 37,746 & 30,518 \\
  TUS-hard & Unionability & 2,769 & 44,109 & 41,850 & 29,579 \\
  UGEN v1 & Unionability & 1,050 & 77 & 70 & 44 \\
  UGEN v2 & Unionability & 1,050 & 307 & 140 & 437 \\
  Valentine & Unionability & 1,098 & 315,128 & 210,000 & 248,934 \\
  Wiki Union & Unionability & 40,752 & 133 & 110 & 89 \\
\midrule
\multicolumn{6}{l}{\emph{Grounding} \hfill (col/tbl)} \\
  NQ-Tables & Grounding & 169,885 & 47 & 18 & 155 \\
  WTQ & Grounding & 2,108 & 173 & 90 & 292 \\
\midrule
\multicolumn{6}{l}{\emph{Row Prediction} \hfill (row)} \\
  kc2 & Row Prediction & 1 & 10,440 & 10,440 & -- \\
  nomao & Row Prediction & 1 & 4,032,405 & 4,032,405 & -- \\
  kr-vs-kp & Row Prediction & 1 & 108,664 & 108,664 & -- \\
  sick & Row Prediction & 1 & 94,300 & 94,300 & -- \\
  connect-4 & Row Prediction & 1 & 2,769,837 & 2,769,837 & -- \\
  MiceProtein & Row Prediction & 1 & 81,000 & 81,000 & -- \\
  Internet-Ads & Row Prediction & 1 & 5,105,403 & 5,105,403 & -- \\
  auction-verification & Row Prediction & 1 & 10,215 & 10,215 & -- \\
  student-perf-por & Row Prediction & 1 & 18,172 & 18,172 & -- \\
  wave-energy & Row Prediction & 1 & 3,384,000 & 3,384,000 & -- \\
  cps88wages & Row Prediction & 1 & 140,775 & 140,775 & -- \\
  fps-benchmark & Row Prediction & 1 & 935,712 & 935,712 & -- \\
  PhishingWebsites & Row Prediction & 1 & 309,540 & 309,540 & -- \\
  analcatdata-authorship & Row Prediction & 1 & 58,029 & 58,029 & -- \\
  anneal & Row Prediction & 1 & 26,042 & 26,042 & -- \\
  Fiat-500 & Row Prediction & 1 & 9,228 & 9,228 & -- \\
  APSFailure & Row Prediction & 1 & 12,692,000 & 12,692,000 & -- \\
  bank-marketing & Row Prediction & 1 & 497,321 & 497,321 & -- \\
  Bank-Churn & Row Prediction & 1 & 90,000 & 90,000 & -- \\
  Bioresponse & Row Prediction & 1 & 6,658,025 & 6,658,025 & -- \\
  churn & Row Prediction & 1 & 85,000 & 85,000 & -- \\
  coil2000 & Row Prediction & 1 & 815,226 & 815,226 & -- \\
  credit-g & Row Prediction & 1 & 18,000 & 18,000 & -- \\
  credit-card-default & Row Prediction & 1 & 630,000 & 630,000 & -- \\
  airline-satisfaction & Row Prediction & 1 & 2,467,720 & 2,467,720 & -- \\
  Diabetes130US & Row Prediction & 1 & 3,075,274 & 3,075,274 & -- \\
  diamonds & Row Prediction & 1 & 431,520 & 431,520 & -- \\
  Fitness-Club & Row Prediction & 1 & 7,500 & 7,500 & -- \\
  GiveMeSomeCredit & Row Prediction & 1 & 1,350,000 & 1,350,000 & -- \\
  hazelnut-spread & Row Prediction & 1 & 69,600 & 69,600 & -- \\
  heloc & Row Prediction & 1 & 230,098 & 230,098 & -- \\
  hiva-agnostic & Row Prediction & 1 & 6,213,520 & 6,213,520 & -- \\
  houses & Row Prediction & 1 & 123,840 & 123,840 & -- \\
  HR-Analytics & Row Prediction & 1 & 191,580 & 191,580 & -- \\
  in-vehicle-coupon & Row Prediction & 1 & 291,732 & 291,732 & -- \\
  kddcup09-appetency & Row Prediction & 1 & 10,400,000 & 10,400,000 & -- \\
  Marketing-Campaign & Row Prediction & 1 & 51,520 & 51,520 & -- \\
  polish-bankruptcy & Row Prediction & 1 & 366,420 & 366,420 & -- \\
  qsar-biodeg & Row Prediction & 1 & 41,106 & 41,106 & -- \\
  SDSS17 & Row Prediction & 1 & 780,530 & 780,530 & -- \\
  seismic-bumps & Row Prediction & 1 & 33,592 & 33,592 & -- \\
  splice & Row Prediction & 1 & 188,210 & 188,210 & -- \\
  students-dropout & Row Prediction & 1 & 150,416 & 150,416 & -- \\
  superconductivity & Row Prediction & 1 & 1,701,040 & 1,701,040 & -- \\
  website-phishing & Row Prediction & 1 & 9,471 & 9,471 & -- \\
  wine-quality & Row Prediction & 1 & 64,970 & 64,970 & -- \\
  NATICUSdroid & Row Prediction & 1 & 629,244 & 629,244 & -- \\
  jm1 & Row Prediction & 1 & 217,700 & 217,700 & -- \\
  MIC & Row Prediction & 1 & 185,191 & 185,191 & -- \\
  cylinder-bands & Row Prediction & 1 & 17,820 & 17,820 & -- \\
\midrule
\multicolumn{6}{l}{\emph{Record Linkage} \hfill (row)} \\
  DM-abt\_buy & Record Linkage & 2 & 3,260 & 3,260 & 16 \\
  DM-amazon\_google & Record Linkage & 2 & 6,884 & 6,884 & 2,794 \\
  DM-beer & Record Linkage & 2 & 14,690 & 14,690 & 2,690 \\
  DM-dblp\_acm & Record Linkage & 2 & 9,820 & 9,820 & 644 \\
  DM-dblp\_acm\_dirty & Record Linkage & 2 & 9,820 & 9,820 & 644 \\
  DM-dblp\_scholar & Record Linkage & 2 & 133,758 & 133,758 & 123,294 \\
  DM-dblp\_scholar\_dirty & Record Linkage & 2 & 133,758 & 133,758 & 123,294 \\
  DM-fodors\_zagats & Record Linkage & 2 & 2,160 & 2,160 & 505 \\
  DM-itunes\_amazon & Record Linkage & 2 & 251,320 & 251,320 & 196,064 \\
  DM-itunes\_amazon\_dirty & Record Linkage & 2 & 251,320 & 251,320 & 196,064 \\
  DM-walmart\_amazon & Record Linkage & 2 & 61,570 & 61,570 & 48,800 \\
  DM-walmart\_amazon\_dirty & Record Linkage & 2 & 61,570 & 61,570 & 48,800 \\
  WDC-large & Record Linkage & 2 & 97,136 & 97,136 & 32 \\
  WDC-medium & Record Linkage & 2 & 79,688 & 79,688 & 245 \\
  WDC-small & Record Linkage & 2 & 56,948 & 56,948 & 3,440 \\
  WDC-xlarge & Record Linkage & 2 & 100,380 & 100,380 & 798 \\
\midrule
\multicolumn{6}{l}{\emph{DLTE} \hfill (dlte)} \\
  DLTE-Lake & DLTE & 47,772 & 23,967 & 2,532 & 63,592 \\
\end{longtable}
\end{small}

%% file: tables/ctbench_lexical/ctbench_lexical_proxies.tex
\begin{table}[h]
\centering
\small
\setlength{\tabcolsep}{4pt}
\caption{Observational proxies for surface-text signal in \textsc{TRL-CTbench} tasks, ordered by descending generic-text advantage. \emph{Best baseline} is the strongest task-local non-neural baseline from Table~\ref{tab:main_results} (TF-IDF, Jaccard, Hungarian/Valentine, value-overlap, etc.). \emph{Best neural} is the strongest CTBench encoder on that task. \emph{Gap} is the metric-direction-corrected difference (best-neural $-$ best-baseline, with positive meaning neural is better). \emph{Generic-text advantage} is the mean of \textsc{BERT}, \textsc{GTE} minus the mean of tabular specialists \textsc{TaBERT}, \textsc{TAPAS}, \textsc{Starmie}, \textsc{TURL} on the same task (direction-corrected). Tasks near the top of the list are those most consistent with the claim that generic text encoders remain competitive on CTBench through surface-text signal. Tasks near the bottom are those where tabular pretraining provides a substantial objective-specific gain.}
\label{tab:ctbench_lexical_proxies}
\begin{tabular}{lccccc}
\toprule
\textbf{Task} & \textbf{Dir.} & \textbf{Best baseline} & \textbf{Best neural (model)} & \textbf{Gap} & \textbf{Generic-text adv.} \\
\midrule
Table Retrieval & $\uparrow$ & $0.131$ & $0.476$ (GTE) & $+0.345$ & $+0.200$ \\
Join Search & $\uparrow$ & $0.155$ & $0.469$ (GTE) & $+0.314$ & $+0.116$ \\
Col Type & $\uparrow$ & $0.813$ & $0.926$ (BERT) & $+0.113$ & $+0.088$ \\
Col Rel & $\uparrow$ & $0.015$ & $0.826$ (BERT) & $+0.811$ & $+0.072$ \\
Col Clust & $\uparrow$ & $0.400$ & $0.516$ (BERT) & $+0.116$ & $+0.048$ \\
Union Reg. & $\downarrow$ & $1.138$ & $0.595$ (BERT) & $+0.543$ & $+0.039$ \\
Col Overlap & $\downarrow$ & $1.014$ & $0.786$ (BERT) & $+0.228$ & $+0.032$ \\
Union Class & $\uparrow$ & $0.500$ & $0.857$ (BERT) & $+0.357$ & $+0.032$ \\
Table Subset & $\uparrow$ & $0.458$ & $0.567$ (TAPAS) & $+0.109$ & $+0.006$ \\
Join Class & $\uparrow$ & $0.516$ & $0.535$ (GTE) & $+0.019$ & $+0.003$ \\
Union Search & $\uparrow$ & $0.574$ & $0.662$ (Starmie) & $+0.088$ & $-0.010$ \\
Table QA & $\uparrow$ & $0.204$ & $0.277$ (TURL) & $+0.073$ & $-0.016$ \\
Schema Match & $\uparrow$ & $0.473$ & $0.764$ (Starmie) & $+0.291$ & $-0.067$ \\
\bottomrule
\end{tabular}
\end{table}

%% file: tables/ablation_head_col.tex
\begin{table*}[h!]
\centering
\caption{%
  \textbf{Ablation: Probe head complexity} for column/table-level tasks.
  MLP = two-layer MLP head. Linear = logistic regression / ridge.
  Dummy = majority-class or mean prediction.
  Best embedding per model, 5-seed average. Table-level tasks use each model's best aggregation on the test split, since this table compares probe heads rather than models.
  $\Delta_{\text{M-L}}$ = MLP $-$ Linear (positive = MLP is better).
  $\Delta_{\text{L-D}}$ = Linear $-$ Dummy (capacity of the \emph{frozen} embedding itself).
}
\label{tab:ablation_head}
\resizebox{\textwidth}{!}{%
\begin{tabular}{l ccc>{\columncolor{gray!15}}c>{\columncolor{gray!15}}c | ccc>{\columncolor{gray!15}}c>{\columncolor{gray!15}}c | ccc>{\columncolor{gray!15}}c>{\columncolor{gray!15}}c | ccc>{\columncolor{gray!15}}c>{\columncolor{gray!15}}c | ccc>{\columncolor{gray!15}}c>{\columncolor{gray!15}}c}
\toprule
&
\multicolumn{5}{c|}{\makecell{\textbf{Col}\\[-2pt]\textbf{Type}$^{\text{c}}$\\\met{$F_1\uparrow$}}} &
\multicolumn{5}{c|}{\makecell{\textbf{Col}\\[-2pt]\textbf{Rel}$^{\text{c}}$\\\met{$F_1\uparrow$}}} &
\multicolumn{5}{c|}{\makecell{\textbf{Join}\\[-2pt]\textbf{Class.}$^{\text{t}\dagger}$\\\met{$F_1\uparrow$}}} &
\multicolumn{5}{c|}{\makecell{\textbf{Union}\\[-2pt]\textbf{Class.}$^{\text{t}\dagger}$\\\met{$F_1\uparrow$}}} &
\multicolumn{5}{c}{\makecell{\textbf{Tbl}\\[-2pt]\textbf{Subset}$^{\text{t}}$\\\met{$F_1\uparrow$}}} \\

\cmidrule(lr){2-6}\cmidrule(lr){7-11}\cmidrule(lr){12-16}\cmidrule(lr){17-21}\cmidrule(lr){22-26}

\textbf{Model}
  & \met{MLP} & \met{Lin} & \met{Dum} & \met{$\Delta_{\text{M-L}}$} & \met{$\Delta_{\text{L-D}}$}
  & \met{MLP} & \met{Lin} & \met{Dum} & \met{$\Delta_{\text{M-L}}$} & \met{$\Delta_{\text{L-D}}$}
  & \met{MLP} & \met{Lin} & \met{Dum} & \met{$\Delta_{\text{M-L}}$} & \met{$\Delta_{\text{L-D}}$}
  & \met{MLP} & \met{Lin} & \met{Dum} & \met{$\Delta_{\text{M-L}}$} & \met{$\Delta_{\text{L-D}}$}
  & \met{MLP} & \met{Lin} & \met{Dum} & \met{$\Delta_{\text{M-L}}$} & \met{$\Delta_{\text{L-D}}$} \\
\midrule

BERT     & 0.928 & 0.924 & 0.178 & \met{+0.00} & \met{+0.75} & 0.834 & 0.819 & 0.000 & \met{+0.01} & \met{+0.82} & 0.529 & 0.578 & 0.414 & \met{$-$0.05} & \met{+0.16} & 0.962 & 0.752 & 0.331 & \met{+0.21} & \met{+0.42} & 0.665 & 0.424 & 0.369 & \met{+0.24} & \met{+0.06} \\
GTE      & 0.922 & 0.923 & 0.178 & \met{$-$0.00} & \met{+0.74} & 0.826 & 0.797 & 0.000 & \met{+0.03} & \met{+0.80} & 0.519 & 0.551 & 0.414 & \met{$-$0.03} & \met{+0.14} & 0.950 & 0.736 & 0.331 & \met{+0.21} & \met{+0.41} & 0.662 & 0.425 & 0.369 & \met{+0.24} & \met{+0.06} \\
TaBERT   & 0.874 & 0.874 & 0.178 & \met{+0.00} & \met{+0.70} & 0.771 & 0.749 & 0.000 & \met{+0.02} & \met{+0.75} & 0.468 & 0.527 & 0.414 & \met{$-$0.06} & \met{+0.11} & 0.818 & 0.702 & 0.331 & \met{+0.12} & \met{+0.37} & 0.677 & 0.403 & 0.369 & \met{+0.27} & \met{+0.03} \\
TAPAS    & 0.869 & 0.867 & 0.178 & \met{+0.00} & \met{+0.69} & 0.782 & 0.755 & 0.000 & \met{+0.03} & \met{+0.76} & 0.555 & 0.532 & 0.414 & \met{+0.02} & \met{+0.12} & 0.931 & 0.741 & 0.331 & \met{+0.19} & \met{+0.41} & 0.695 & 0.439 & 0.369 & \met{+0.26} & \met{+0.07} \\
TAPEX    & --- & --- & --- & --- & --- & --- & --- & --- & --- & --- & 0.530 & 0.497 & 0.414 & \met{+0.03} & \met{+0.08} & 0.951 & 0.757 & 0.331 & \met{+0.19} & \met{+0.43} & 0.718 & 0.396 & 0.369 & \met{+0.32} & \met{+0.03} \\
TABBIE   & 0.881 & 0.904 & 0.178 & \met{$-$0.02} & \met{+0.73} & 0.777 & 0.793 & 0.000 & \met{$-$0.02} & \met{+0.79} & 0.523 & 0.560 & 0.414 & \met{$-$0.04} & \met{+0.15} & 0.931 & 0.736 & 0.331 & \met{+0.20} & \met{+0.40} & 0.697 & 0.396 & 0.369 & \met{+0.30} & \met{+0.03} \\
TURL     & 0.836 & 0.792 & 0.178 & \met{+0.04} & \met{+0.61} & 0.777 & 0.740 & 0.000 & \met{+0.04} & \met{+0.74} & 0.510 & 0.554 & 0.414 & \met{$-$0.04} & \met{+0.14} & 0.942 & 0.686 & 0.331 & \met{+0.26} & \met{+0.36} & 0.635 & 0.379 & 0.369 & \met{+0.26} & \met{+0.01} \\
TUTA     & --- & --- & --- & --- & --- & --- & --- & --- & --- & --- & 0.465 & 0.470 & 0.414 & \met{$-$0.01} & \met{+0.06} & 0.906 & 0.713 & 0.331 & \met{+0.19} & \met{+0.38} & 0.472 & 0.421 & 0.369 & \met{+0.05} & \met{+0.05} \\
Starmie  & 0.767 & 0.811 & --- & \met{$-$0.04} & --- & 0.695 & 0.701 & --- & \met{$-$0.01} & --- & 0.512 & 0.509 & --- & \met{+0.00} & --- & 0.949 & 0.757 & --- & \met{+0.19} & --- & 0.672 & 0.407 & --- & \met{+0.26} & --- \\
TabSketchFM & 0.583 & 0.550 & 0.178 & \met{+0.03} & \met{+0.37} & 0.390 & 0.356 & 0.000 & \met{+0.03} & \met{+0.36} & 0.482 & 0.550 & 0.414 & \met{$-$0.07} & \met{+0.14} & 0.785 & 0.689 & 0.331 & \met{+0.10} & \met{+0.36} & 0.663 & 0.443 & 0.369 & \met{+0.22} & \met{+0.07} \\
\midrule
Avg.     & 0.832 & 0.830 & 0.178 & \met{+0.00} & \met{+0.65} & 0.731 & 0.714 & 0.000 & \met{+0.02} & \met{+0.72} & 0.509 & 0.533 & 0.414 & \met{$-$0.02} & \met{+0.12} & 0.913 & 0.727 & 0.331 & \met{+0.19} & \met{+0.39} & 0.656 & 0.413 & 0.369 & \met{+0.24} & \met{+0.05} \\

\bottomrule
\end{tabular}%
}
\end{table*}

%% file: tables/dev_split_sizes.tex
\begin{table}[t]
\centering
\caption{Development-split size and construction rule for each \textsc{TRL-CTbench} table-level task's aggregation selection (Table~\ref{tab:main_results}).}
\label{tab:dev_split_sizes}
\resizebox{\textwidth}{!}{%
\begin{tabular}{llrl}
\toprule
\textbf{Task} & \textbf{Dataset} & \textbf{Dev size} & \textbf{Construction} \\
\midrule
Join Classification & spider\_join & 216 & table-disjoint 70/15/15 split (strict variant); valid partition \\
Union Classification & wiki\_union & 8,068 & table-disjoint 70/15/15 split (strict variant); valid partition \\
Union Regression & ecb\_union & 395 & table-disjoint 70/15/15 split (strict variant); valid partition \\
Table Subset & ckan\_subset & 3,010 & source-provided table-disjoint split; valid partition \\
Table Retrieval & nq\_tables & 966 of 9,594 & question-disjoint carve from training questions (seed 42) \\
\bottomrule
\end{tabular}%
}
\end{table}

%% file: tables/aggregation_ablation.tex
\begin{table*}[t]
\centering
\caption{%
  Effect of table-level embedding aggregation (MLP probe).
  For each model we evaluate every supported aggregation strategy:
  \textsc{cls} (\texttt{[CLS]} token from the linearized table),
  \textsc{col-mean} (mean of per-column embeddings), and
  \textsc{tok-mean} (mean of all non-padding token hidden states).
  Metrics match Table~\ref{tab:main_results}: $F_1$ = macro $F_1$, nRMSE = $\sqrt{1-R^2}$.
  Table~\ref{tab:main_results} reports avg(MLP, linear) for the aggregation selected on development data. Tables~\ref{tab:aggregation_ablation}--\ref{tab:aggregation_ablation_linear} break down the per-probe results. Table Retrieval cells use the query encoder that is better on the test split; Table~\ref{tab:main_results} selects it on development data (Appendix~\ref{app:ablation_qe}), which changes only \textsc{TUTA}'s cell (0.235 in Table~\ref{tab:main_results}).
  Values are mean $\pm$ std over 5 random seeds.
  \textbf{Bold} / \underline{underlined} values highlight best/second-best aggregation for each model and metric (shown only when $\geq$2 variants exist).
  $\dagger$\,Table-disjoint split.
  Dashes indicate the model does not produce that embedding variant.
}
\label{tab:aggregation_ablation}
\resizebox{\textwidth}{!}{%
\begin{tabular}{lll ccccc}
\toprule
\textbf{Family} & \textbf{Model} & \textbf{Agg.} & \makecell{JoinCls${}^{\dagger}$} & \makecell{UnionCls${}^{\dagger}$} & \makecell{UnionReg${}^{\dagger}$} & \makecell{TblSubset} & \makecell{TblRet} \\
 & &  & \met{$F_1\,\uparrow$} & \met{$F_1\,\uparrow$} & \met{nRMSE$\,\downarrow$} & \met{$F_1\,\uparrow$} & \met{MRR$\,\uparrow$} \\
\midrule
\multirow{6}{*}{Generic Text} & \multirow{3}{*}{BERT} & \textsc{cls} & $\underline{0.503}_{\pm0.033}$ & $0.956_{\pm0.003}$ & $0.509_{\pm0.009}$ & $0.615_{\pm0.010}$ & $0.321_{\pm0.013}$ \\
 &  & \textsc{col-mean} & $0.493_{\pm0.040}$ & $\underline{0.961}_{\pm0.002}$ & $\underline{0.454}_{\pm0.009}$ & $\textbf{0.665}_{\pm0.009}$ & $\underline{0.357}_{\pm0.009}$ \\
 &  & \textsc{tok-mean} & $\textbf{0.529}_{\pm0.019}$ & $\textbf{0.962}_{\pm0.004}$ & $\textbf{0.449}_{\pm0.018}$ & $\underline{0.653}_{\pm0.008}$ & $\textbf{0.367}_{\pm0.008}$ \\
\addlinespace[2pt]
 & \multirow{3}{*}{GTE} & \textsc{cls} & $\underline{0.491}_{\pm0.029}$ & $\underline{0.946}_{\pm0.005}$ & $0.640_{\pm0.011}$ & $\underline{0.567}_{\pm0.014}$ & $\textbf{0.476}_{\pm0.003}$ \\
 &  & \textsc{col-mean} & $\textbf{0.519}_{\pm0.037}$ & $0.944_{\pm0.002}$ & $\textbf{0.475}_{\pm0.019}$ & $\textbf{0.662}_{\pm0.004}$ & $0.450_{\pm0.013}$ \\
 &  & \textsc{tok-mean} & $0.453_{\pm0.035}$ & $\textbf{0.950}_{\pm0.004}$ & $\underline{0.578}_{\pm0.018}$ & $0.543_{\pm0.044}$ & $\underline{0.473}_{\pm0.008}$ \\
\cmidrule(lr){2-8}
\multirow{6}{*}{Table-Text} & \multirow{1}{*}{TaBERT} & \textsc{col-mean} & $0.468_{\pm0.068}$ & $0.818_{\pm0.007}$ & $0.536_{\pm0.017}$ & $0.677_{\pm0.007}$ & $0.372_{\pm0.013}$ \\
\addlinespace[2pt]
 & \multirow{3}{*}{TAPAS} & \textsc{cls} & $0.460_{\pm0.035}$ & $0.856_{\pm0.006}$ & $0.576_{\pm0.018}$ & $0.689_{\pm0.007}$ & $0.265_{\pm0.009}$ \\
 &  & \textsc{col-mean} & $\textbf{0.555}_{\pm0.022}$ & $\underline{0.929}_{\pm0.007}$ & $\underline{0.494}_{\pm0.008}$ & $\textbf{0.695}_{\pm0.009}$ & $\underline{0.285}_{\pm0.013}$ \\
 &  & \textsc{tok-mean} & $\underline{0.503}_{\pm0.031}$ & $\textbf{0.931}_{\pm0.006}$ & $\textbf{0.488}_{\pm0.010}$ & $\underline{0.693}_{\pm0.011}$ & $\textbf{0.295}_{\pm0.006}$ \\
\addlinespace[2pt]
 & \multirow{2}{*}{TAPEX} & \textsc{cls} & $\textbf{0.530}_{\pm0.046}$ & $\underline{0.942}_{\pm0.002}$ & $\textbf{0.468}_{\pm0.012}$ & $\textbf{0.718}_{\pm0.009}$ & --- \\
 &  & \textsc{tok-mean} & $\underline{0.526}_{\pm0.043}$ & $\textbf{0.951}_{\pm0.004}$ & $\underline{0.471}_{\pm0.010}$ & $\underline{0.704}_{\pm0.007}$ & --- \\
\cmidrule(lr){2-8}
\multirow{4}{*}{Table-Struct.} & \multirow{2}{*}{TABBIE} & \textsc{cls} & $\textbf{0.523}_{\pm0.043}$ & $\textbf{0.931}_{\pm0.003}$ & $\textbf{0.542}_{\pm0.002}$ & $\textbf{0.697}_{\pm0.007}$ & $\textbf{0.170}_{\pm0.004}$ \\
 &  & \textsc{col-mean} & $\underline{0.474}_{\pm0.034}$ & $\underline{0.919}_{\pm0.002}$ & $\underline{0.665}_{\pm0.011}$ & $\underline{0.681}_{\pm0.009}$ & $\underline{0.102}_{\pm0.002}$ \\
\addlinespace[2pt]
 & \multirow{1}{*}{TURL} & \textsc{col-mean} & $0.510_{\pm0.028}$ & $0.942_{\pm0.002}$ & $0.499_{\pm0.018}$ & $0.635_{\pm0.007}$ & $0.199_{\pm0.010}$ \\
\addlinespace[2pt]
 & \multirow{1}{*}{TUTA} & \textsc{cls} & $0.465_{\pm0.019}$ & $0.906_{\pm0.006}$ & $0.511_{\pm0.013}$ & $0.472_{\pm0.016}$ & $0.260_{\pm0.013}$ \\
\cmidrule(lr){2-8}
\multirow{4}{*}{Col.-Centric} & \multirow{1}{*}{Starmie} & \textsc{col-mean} & $0.512_{\pm0.017}$ & $0.949_{\pm0.004}$ & $0.536_{\pm0.006}$ & $0.672_{\pm0.010}$ & $0.018_{\pm0.002}$ \\
\addlinespace[2pt]
 & \multirow{3}{*}{TabSketchFM} & \textsc{cls} & $\textbf{0.482}_{\pm0.032}$ & $0.779_{\pm0.004}$ & $0.549_{\pm0.024}$ & $\underline{0.659}_{\pm0.015}$ & $\textbf{0.218}_{\pm0.011}$ \\
 &  & \textsc{col-mean} & $0.471_{\pm0.026}$ & $\textbf{0.785}_{\pm0.002}$ & $\textbf{0.512}_{\pm0.021}$ & $0.647_{\pm0.004}$ & $\underline{0.197}_{\pm0.014}$ \\
 &  & \textsc{tok-mean} & $\underline{0.476}_{\pm0.021}$ & $\underline{0.783}_{\pm0.004}$ & $\underline{0.513}_{\pm0.013}$ & $\textbf{0.663}_{\pm0.009}$ & $0.193_{\pm0.035}$ \\
\midrule
\addlinespace[2pt]
\multicolumn{2}{l}{\textit{Avg.\ across models}} & \textsc{cls} & $0.494$ & $0.902$ & $0.542$ & $0.631$ & $\underline{0.285}$ \\
\multicolumn{2}{l}{} & \textsc{col-mean} & $\textbf{0.500}$ & $\underline{0.906}$ & $\underline{0.521}$ & $\textbf{0.667}$ & $0.248$ \\
\multicolumn{2}{l}{} & \textsc{tok-mean} & $\underline{0.497}$ & $\textbf{0.915}$ & $\textbf{0.500}$ & $\underline{0.651}$ & $\textbf{0.332}$ \\
\bottomrule
\end{tabular}%
}
\end{table*}

%% file: tables/aggregation_ablation_linear.tex
\begin{table*}[t]
\centering
\caption{%
  Effect of table-level embedding aggregation (LINEAR probe).
  For each model we evaluate every supported aggregation strategy:
  \textsc{cls} (\texttt{[CLS]} token from the linearized table),
  \textsc{col-mean} (mean of per-column embeddings), and
  \textsc{tok-mean} (mean of all non-padding token hidden states).
  Metrics match Table~\ref{tab:main_results}: $F_1$ = macro $F_1$, nRMSE = $\sqrt{1-R^2}$.
  Table~\ref{tab:main_results} reports the aggregation selected on development data.
  Values are mean $\pm$ std over 4--5 random seeds.
  \textbf{Bold} / \underline{underlined} values highlight best/second-best aggregation for each model and metric (shown only when $\geq$2 variants exist).
  $\dagger$\,Table-disjoint split.
  Dashes indicate the model does not produce that embedding variant.
}
\label{tab:aggregation_ablation_linear}
\resizebox{\textwidth}{!}{%
\begin{tabular}{lll cccc}
\toprule
\textbf{Family} & \textbf{Model} & \textbf{Agg.} & \makecell{JoinCls${}^{\dagger}$} & \makecell{UnionCls${}^{\dagger}$} & \makecell{UnionReg${}^{\dagger}$} & \makecell{TblSubset} \\
 & &  & \met{$F_1\,\uparrow$} & \met{$F_1\,\uparrow$} & \met{nRMSE$\,\downarrow$} & \met{$F_1\,\uparrow$} \\
\midrule
\multirow{6}{*}{Generic Text} & \multirow{3}{*}{BERT} & \textsc{cls} & $\underline{0.526}_{\pm0.000}$ & $0.745_{\pm0.000}$ & $0.770_{\pm0.000}$ & $\textbf{0.436}_{\pm0.000}$ \\
 &  & \textsc{col-mean} & $0.500_{\pm0.000}$ & $\underline{0.745}_{\pm0.000}$ & $\underline{0.736}_{\pm0.000}$ & $\underline{0.424}_{\pm0.000}$ \\
 &  & \textsc{tok-mean} & $\textbf{0.578}_{\pm0.000}$ & $\textbf{0.752}_{\pm0.000}$ & $\textbf{0.735}_{\pm0.000}$ & $0.408_{\pm0.000}$ \\
\addlinespace[2pt]
 & \multirow{3}{*}{GTE} & \textsc{cls} & $\underline{0.503}_{\pm0.000}$ & $0.726_{\pm0.000}$ & $0.830_{\pm0.000}$ & $\underline{0.421}_{\pm0.000}$ \\
 &  & \textsc{col-mean} & $\textbf{0.551}_{\pm0.000}$ & $\underline{0.735}_{\pm0.000}$ & $\textbf{0.726}_{\pm0.000}$ & $\textbf{0.425}_{\pm0.000}$ \\
 &  & \textsc{tok-mean} & $0.478_{\pm0.000}$ & $\textbf{0.736}_{\pm0.000}$ & $\underline{0.754}_{\pm0.000}$ & $0.416_{\pm0.000}$ \\
\cmidrule(lr){2-7}
\multirow{6}{*}{Table-Text} & \multirow{1}{*}{TaBERT} & \textsc{col-mean} & $0.527_{\pm0.000}$ & $0.702_{\pm0.000}$ & $0.694_{\pm0.000}$ & $0.403_{\pm0.001}$ \\
\addlinespace[2pt]
 & \multirow{3}{*}{TAPAS} & \textsc{cls} & $\underline{0.531}_{\pm0.000}$ & $0.707_{\pm0.000}$ & $0.760_{\pm0.000}$ & $0.432_{\pm0.000}$ \\
 &  & \textsc{col-mean} & $\textbf{0.532}_{\pm0.000}$ & $\textbf{0.746}_{\pm0.000}$ & $\underline{0.731}_{\pm0.000}$ & $\textbf{0.439}_{\pm0.001}$ \\
 &  & \textsc{tok-mean} & $0.516_{\pm0.000}$ & $\underline{0.741}_{\pm0.000}$ & $\textbf{0.726}_{\pm0.000}$ & $\underline{0.436}_{\pm0.001}$ \\
\addlinespace[2pt]
 & \multirow{2}{*}{TAPEX} & \textsc{cls} & $\underline{0.497}_{\pm0.000}$ & $\underline{0.747}_{\pm0.000}$ & $\textbf{0.750}_{\pm0.000}$ & $\underline{0.396}_{\pm0.000}$ \\
 &  & \textsc{tok-mean} & $\textbf{0.550}_{\pm0.000}$ & $\textbf{0.757}_{\pm0.000}$ & $\underline{0.772}_{\pm0.000}$ & $\textbf{0.411}_{\pm0.000}$ \\
\cmidrule(lr){2-7}
\multirow{4}{*}{Table-Struct.} & \multirow{2}{*}{TABBIE} & \textsc{cls} & $\underline{0.560}_{\pm0.000}$ & $\underline{0.736}_{\pm0.000}$ & $\underline{0.784}_{\pm0.000}$ & $\textbf{0.396}_{\pm0.000}$ \\
 &  & \textsc{col-mean} & $\textbf{0.581}_{\pm0.000}$ & $\textbf{0.737}_{\pm0.000}$ & $\textbf{0.757}_{\pm0.000}$ & $\underline{0.387}_{\pm0.000}$ \\
\addlinespace[2pt]
 & \multirow{1}{*}{TURL} & \textsc{col-mean} & $0.554_{\pm0.000}$ & $0.686_{\pm0.000}$ & $0.814_{\pm0.000}$ & $0.379_{\pm0.000}$ \\
\addlinespace[2pt]
 & \multirow{1}{*}{TUTA} & \textsc{cls} & $0.470_{\pm0.000}$ & $0.713_{\pm0.000}$ & $0.792_{\pm0.000}$ & $0.421_{\pm0.001}$ \\
\cmidrule(lr){2-7}
\multirow{4}{*}{Col.-Centric} & \multirow{1}{*}{Starmie} & \textsc{col-mean} & $0.509_{\pm0.031}$ & $0.757_{\pm0.000}$ & $0.789_{\pm0.000}$ & $0.407_{\pm0.000}$ \\
\addlinespace[2pt]
 & \multirow{3}{*}{TabSketchFM} & \textsc{cls} & $\textbf{0.550}_{\pm0.007}$ & $\textbf{0.695}_{\pm0.000}$ & $0.864_{\pm0.001}$ & $0.432_{\pm0.002}$ \\
 &  & \textsc{col-mean} & $0.513_{\pm0.013}$ & $\underline{0.689}_{\pm0.000}$ & $\textbf{0.825}_{\pm0.001}$ & $\textbf{0.445}_{\pm0.003}$ \\
 &  & \textsc{tok-mean} & $\underline{0.533}_{\pm0.015}$ & $0.684_{\pm0.000}$ & $\underline{0.830}_{\pm0.001}$ & $\underline{0.443}_{\pm0.001}$ \\
\midrule
\addlinespace[2pt]
\multicolumn{2}{l}{\textit{Avg.\ across models}} & \textsc{cls} & $0.520$ & $0.724$ & $0.793$ & $\underline{0.419}$ \\
\multicolumn{2}{l}{} & \textsc{col-mean} & $\textbf{0.533}$ & $\underline{0.725}$ & $\textbf{0.759}$ & $0.414$ \\
\multicolumn{2}{l}{} & \textsc{tok-mean} & $\underline{0.531}$ & $\textbf{0.734}$ & $\underline{0.763}$ & $\textbf{0.423}$ \\
\bottomrule
\end{tabular}%
}
\end{table*}

%% file: tables/join_search_learned_table.tex
\begin{table*}[t]
\centering
\caption{%
  Join search MAP: direct cosine similarity vs.\ learned linear projection, averaged across 5 embedding rounds.
  The learned projection trains a shared \texttt{Linear}$(d,d)$ head with multi-positive InfoNCE loss
  on a fixed 20\%/80\% query-role-disjoint split and evaluates on the held-out 80\%.
  Results are column-level MAP, macro-averaged over queries.
  \colorbox{rankfirst}{\strut\textbf{Bold orange}} / \colorbox{ranksecond}{\strut\uline{Underlined blue}} / \colorbox{rankthird}{\strut Light purple} highlights indicate best/second-best/third-best per column (non-baseline).
}
\label{tab:join_search_learned}
\resizebox{\textwidth}{!}{%
\begin{tabular}{ll cc cc cc cc}
\toprule
& &
  \multicolumn{2}{c}{\textbf{All}} &
  \multicolumn{2}{c}{\textbf{CAN}} &
  \multicolumn{2}{c}{\textbf{USA}} &
  \multicolumn{2}{c}{\textbf{UK+SG}} \\
\cmidrule(lr){3-4} \cmidrule(lr){5-6} \cmidrule(lr){7-8} \cmidrule(lr){9-10}
\textbf{Family} & \textbf{Model} & Cos. & Proj. & Cos. & Proj. & Cos. & Proj. & Cos. & Proj. \\
\midrule
 \multirow{2}{*}{Generic Text} & BERT & \cells\uline{0.387} & \cellt0.431 & \cells\uline{0.300} & \cellt0.328 & \cells\uline{0.447} & \cellt0.508 & \cells\uline{0.431} & \cells\uline{0.470} \\
  & GTE & \cellf\textbf{0.411} & \cellf\textbf{0.461} & \cellf\textbf{0.331} & \cellf\textbf{0.370} & \cellf\textbf{0.460} & \cellf\textbf{0.535} & \cellf\textbf{0.478} & \cellf\textbf{0.508} \\
\midrule
\multicolumn{10}{l}{\textit{Tabular-Pretrained}} \\
 \multirow{2}{*}{\quad Table-Text} & TaBERT & \cellt0.326 & \cells\uline{0.458} & \cellt0.295 & \cells\uline{0.346} & \cellt0.377 & \cells\uline{0.525} & 0.262 & \cellt0.296 \\
  & TAPAS & 0.246 & 0.332 & 0.230 & 0.295 & 0.285 & 0.383 & 0.208 & 0.272 \\
\cmidrule(lr){2-10}
 \multirow{2}{*}{\quad Table-Struct.} & TURL & 0.271 & 0.292 & 0.248 & 0.267 & 0.299 & 0.341 & \cellt0.269 & 0.295 \\
  & TABBIE & 0.189 & 0.218 & 0.174 & 0.176 & 0.207 & 0.231 & 0.206 & 0.206 \\
\cmidrule(lr){2-10}
 \multirow{2}{*}{\quad Col.-Centric} & Starmie & 0.248 & 0.298 & 0.230 & 0.295 & 0.290 & 0.379 & 0.206 & 0.293 \\
  & TabSketchFM & 0.231 & 0.277 & 0.195 & 0.227 & 0.276 & 0.339 & 0.188 & 0.218 \\
\bottomrule
\end{tabular}%
}
\end{table*}

%% file: tables/tus_hard_comparison.tex
\begin{table}[t]
\centering
\small
\caption{%
  Union search MAP on TUS vs.\ TUS-hard (5-seed average).
  TUS-hard filters out positive pairs with high directed containment
  ($\geq 0.70$), creating a low-overlap variant that better separates lexical overlap from broader union signal in our frozen-transfer setting.
  Ranks are among neural models only.
  \colorbox{rankfirst}{\strut\textbf{Bold orange}} / \colorbox{ranksecond}{\strut\uline{Underlined blue}} / \colorbox{rankthird}{\strut Light purple} highlights indicate best/second-best/third-best on TUS-hard.
  Spearman $\rho = -0.67$ between the two rankings.
}
\label{tab:tus_hard}
\begin{tabular}{llcccccc}
\toprule
\textbf{Type} & \textbf{Model}
  & \makecell{MAP\\TUS$\,\uparrow$} & \makecell{Rank\\TUS}
  & \makecell{MAP\\TUS-hard$\,\uparrow$} & \makecell{Rank\\Hard}
  & \makecell{Drop\\(\%)} \\
\midrule
\multirow{2}{*}{\textit{Baseline}}
 & Random     & $0.209$ & --- & $0.079$ & --- & $62.3$ \\
 & Val.\ Ovlp & $1.000$ & --- & $0.008$ & --- & $99.2$ \\
\midrule
\multirow{2}{*}{Generic Text}
 & BERT  & $0.959$ & 1 & $0.307$ & 6 & $68.0$ \\
 & GTE   & $0.954$ & 2 & $0.293$ & 8 & $69.2$ \\
\addlinespace[2pt]
\multirow{2}{*}{Table-Text}
 & TaBERT & $0.926$ & 5 & \cells\uline{$0.436$} & 2 & $52.9$ \\
 & TAPAS  & $0.891$ & 6 & \cellt$0.376$ & 3 & $57.8$ \\
\addlinespace[2pt]
\multirow{2}{*}{Table-Struct.}
 & TABBIE & $0.700$ & 8 & $0.317$ & 5 & $54.8$ \\
 & TURL   & $0.953$ & 3 & $0.304$ & 7 & $68.1$ \\
\addlinespace[2pt]
\multirow{2}{*}{Col.-Centric}
 & Starmie & $0.844$ & 7 & \cellf$\textbf{0.523}$ & 1 & $38.1$ \\
 & TabSketchFM  & $0.941$ & 4 & \cellt$0.376$ & 4 & $60.1$ \\
\bottomrule
\end{tabular}
\end{table}

%% file: tables/pylon_union.tex
\begin{table}[t]
\centering
\small
\caption{%
  \textsc{Pylon}~\cite{pylon} union search under the benchmark's standard multi-column protocol (per-column Munkres bipartite matching on column embeddings; hnsw/$K$/threshold$=0.1$, matching the \textsc{TUS}/\textsc{TUS-hard} operating point).
  MAP@10 and R@60 are each the mean of two runs, identical to 4 decimal places.
  R@60 is the smallest swept $k$ at or above Pylon's largest topic clique (48); ranks and \colorbox{rankfirst}{\strut\textbf{bold orange}}/\colorbox{ranksecond}{\strut\uline{underlined blue}}/\colorbox{rankthird}{\strut light purple} highlights are computed on R@60 only. MAP@10 uses the $K{=}10$ cutoff of SANTOS and UGEN and is not separately ranked here.
  $^\dagger$\textsc{TabSketchFM}'s embeddings cover 1742/1746 tables (4 unlabeled distractor tables excluded by a preprocessing skip-and-log guard; 0 queries or gold labels affected); every other model covers all 1746/1746 tables.
}
\label{tab:pylon_union}
\begin{tabular}{llccc}
\toprule
\textbf{Type} & \textbf{Model} & \makecell{MAP@10\\[-2pt]$\uparrow$} & \makecell{R@60\\[-2pt]$\uparrow$} & \makecell{Rank\\R@60} \\
\midrule
\multirow{2}{*}{Generic Text}
 & BERT & $0.8523$ & $0.3619$ & 4 \\
 & GTE & $0.8724$ & \cells\uline{$0.4371$} & 2 \\
\addlinespace[2pt]
\multirow{2}{*}{Table-Text}
 & TaBERT & $0.8606$ & \cellt$0.4139$ & 3 \\
 & TAPAS & $0.7297$ & $0.2102$ & 6 \\
\addlinespace[2pt]
\multirow{2}{*}{Table-Struct.}
 & TABBIE & $0.5283$ & $0.1201$ & 8 \\
 & TURL & $0.7456$ & $0.2969$ & 5 \\
\addlinespace[2pt]
\multirow{2}{*}{Col.-Centric}
 & Starmie & $0.9074$ & \cellf$\textbf{0.6253}$ & 1 \\
 & TabSketchFM$^\dagger$ & $0.6483$ & $0.1954$ & 7 \\
\bottomrule
\end{tabular}
\end{table}

%% file: tables/hybrid_ablation.tex
\begin{table}[t]
\centering
\small
\caption{%
  Table retrieval MRR: model-only vs.\ hybrid mode.
  In model-only mode, the projection head operates on the model's own table embedding.
  In hybrid mode, the model's table embedding is concatenated with the query encoder's
  (mpnet or sentence-t5) table embedding before projection.
  For each model, we report the best aggregation $\times$ sentence-encoder combination on the test split, since this table compares retrieval modes rather than models.
  Values are mean $\pm$ std over 5 seeds.
}
\label{tab:hybrid_ablation}
\begin{tabular}{llccc}
\toprule
\textbf{Family} & \textbf{Model} & \makecell{Model-Only\\MRR$\,\uparrow$} & \makecell{Hybrid\\MRR$\,\uparrow$} & $\Delta$ \\
\midrule
\multirow{2}{*}{Generic Text}
 & BERT & $0.367_{\pm0.008}$ & $0.553_{\pm0.007}$ & $+0.186$ \\
 & GTE & $0.476_{\pm0.003}$ & $0.533_{\pm0.008}$ & $+0.057$ \\
\addlinespace[2pt]
\multirow{3}{*}{Table-Text}
 & TaBERT & $0.372_{\pm0.013}$ & $0.555_{\pm0.006}$ & $+0.183$ \\
 & TAPAS & $0.295_{\pm0.006}$ & $0.526_{\pm0.008}$ & $+0.231$ \\
 & TAPEX & $0.332_{\pm0.005}$ & $0.536_{\pm0.028}$ & $+0.204$ \\
\addlinespace[2pt]
\multirow{3}{*}{Table-Struct.}
 & TABBIE & $0.170_{\pm0.004}$ & $0.516_{\pm0.007}$ & $+0.347$ \\
 & TURL & $0.199_{\pm0.010}$ & $0.521_{\pm0.031}$ & $+0.322$ \\
 & TUTA & $0.260_{\pm0.013}$ & $0.509_{\pm0.012}$ & $+0.249$ \\
\addlinespace[2pt]
\multirow{2}{*}{Col.-Centric}
 & Starmie & $0.018_{\pm0.002}$ & $0.527_{\pm0.012}$ & $+0.509$ \\
 & TabSketchFM & $0.218_{\pm0.011}$ & $0.522_{\pm0.008}$ & $+0.304$ \\
\bottomrule
\end{tabular}
\end{table}

%% file: tables/ablation_qe.tex
\begin{table}[h!]
\centering
\footnotesize
\setlength{\tabcolsep}{3pt}
\caption{%
  \textbf{Ablation: Query encoder} for grounding tasks.
  Table retrieval uses MRR ($\uparrow$, model\_only).
  Semantic parsing uses accuracy ($\uparrow$).
  Each table model is paired with two query encoders.
  $\Delta$ = MPNet $-$ ST5.
  5-seed average. Table retrieval uses best embedding per model.
}
\label{tab:ablation_qe}
\resizebox{0.6\columnwidth}{!}{%
\begin{tabular}{l cc>{\columncolor{gray!15}}c | cc>{\columncolor{gray!15}}c}
\toprule
&
\multicolumn{3}{c|}{\makecell{\textbf{Tbl}\\[-2pt]\textbf{Ret.}$^{\text{t}}$\\\met{MRR$\uparrow$}}} &
\multicolumn{3}{c}{\makecell{\textbf{Tbl}\\[-2pt]\textbf{QA}$^{\text{c}}$\\\met{Acc$\uparrow$}}} \\

\cmidrule(lr){2-4}\cmidrule(lr){5-7}

\textbf{Model}
  & \met{MPNet} & \met{ST5} & $\Delta$
  & \met{MPNet} & \met{ST5} & $\Delta$ \\
\midrule

BERT       & 0.368 & 0.352 & \met{+0.02} & 0.238 & 0.271 & \met{$-$0.03} \\[-3pt]
           & \std{0.007} & \std{0.007} &  & \std{0.005} & \std{0.004} &  \\
GTE        & 0.478 & 0.440 & \met{+0.04} & 0.233 & 0.256 & \met{$-$0.02} \\[-3pt]
           & \std{0.004} & \std{0.005} &  & \std{0.004} & \std{0.005} &  \\
TaBERT     & 0.372 & 0.324 & \met{+0.05} & 0.252 & 0.281 & \met{$-$0.03} \\[-3pt]
           & \std{0.013} & \std{0.007} &  & \std{0.007} & \std{0.005} &  \\
TAPAS      & 0.296 & 0.276 & \met{+0.02} & 0.240 & 0.269 & \met{$-$0.03} \\[-3pt]
           & \std{0.006} & \std{0.007} &  & \std{0.005} & \std{0.008} &  \\
TAPEX      & 0.332 & 0.309 & \met{+0.02} & --- & --- & --- \\[-3pt]
           & \std{0.005} & \std{0.006} &  &  &  &  \\
TABBIE     & 0.170 & 0.145 & \met{+0.02} & 0.261 & 0.290 & \met{$-$0.03} \\[-3pt]
           & \std{0.004} & \std{0.002} &  & \std{0.004} & \std{0.008} &  \\
TURL       & 0.197 & 0.199 & \met{$-$0.00} & 0.274 & 0.281 & \met{$-$0.01} \\[-3pt]
           & \std{0.014} & \std{0.010} &  & \std{0.006} & \std{0.004} &  \\
TUTA       & 0.260 & 0.235 & \met{+0.03} & --- & --- & --- \\[-3pt]
           & \std{0.013} & \std{0.013} &  &  &  &  \\
Starmie    & 0.018 & 0.012 & \met{+0.01} & 0.248 & 0.283 & \met{$-$0.03} \\[-3pt]
           & \std{0.002} & \std{0.002} &  & \std{0.008} & \std{0.008} &  \\
TabSketchFM     & 0.221 & 0.191 & \met{+0.03} & 0.216 & 0.253 & \met{$-$0.04} \\[-3pt]
           & \std{0.013} & \std{0.006} &  & \std{0.006} & \std{0.006} &  \\
\midrule
Avg.       & 0.271 & 0.248 & \met{+0.02} & 0.245 & 0.273 & \met{$-$0.03} \\

\bottomrule
\end{tabular}%
}
\end{table}

%% file: tables/strict_ablation.tex
\begin{table*}[t]
\centering
\small
\caption{%
  Pair-level random vs.\ table-disjoint split comparison on the four tasks that
  support both protocols.  Each cell reports the 5-seed average (avg.\ of MLP and
  linear probes). Table-level tasks use each model's best aggregation on the test split, since this table compares split protocols rather than models.
  $\Delta$ = table-disjoint $-$ pair-random. Negative $\Delta$ for $F_1$ and positive $\Delta$
  for nRMSE both indicate that table-disjoint evaluation is harder.
  Values are mean $\pm$ std over 5 seeds.
  Dashes indicate the model does not produce that embedding variant.
  Avg.\ is computed over all models with available data per task.
}
\label{tab:strict_ablation}
\setlength{\tabcolsep}{4pt}
\resizebox{\textwidth}{!}{%
\begin{tabular}{ll rrr rrr rrr rrr}
\toprule
& &
\multicolumn{3}{c}{\makecell{JoinCls\\[-2pt]\met{$F_1\uparrow$}}} &
\multicolumn{3}{c}{\makecell{ColOverlap\\[-2pt]\met{nRMSE$\downarrow$}}} &
\multicolumn{3}{c}{\makecell{UnionCls\\[-2pt]\met{$F_1\uparrow$}}} &
\multicolumn{3}{c}{\makecell{UnionReg\\[-2pt]\met{nRMSE$\downarrow$}}} \\
\cmidrule(lr){3-5}\cmidrule(lr){6-8}\cmidrule(lr){9-11}\cmidrule(lr){12-14}
\textbf{Type} & \textbf{Model}
  & P-R & T-D & $\Delta$
  & P-R & T-D & $\Delta$
  & P-R & T-D & $\Delta$
  & P-R & T-D & $\Delta$ \\
\midrule
Baseline & Random & $0.704$ & $0.516$ & $-0.189$ & $0.972$ & $1.012$ & $+0.040$ & $0.666$ & $0.500$ & $-0.166$ & $0.868$ & $1.138$ & $+0.270$ \\
\midrule
\multirow{4}{*}{Generic Text}
 & BERT    & $0.768$ & $0.553$ & $-0.215$ & $0.758$ & $0.786$ & $+0.028$ & $0.877$ & $0.857$ & $-0.020$ & $0.485$ & $0.592$ & $+0.107$ \\[-3pt]
 &         & \std{0.013} & \std{0.010} &  & \std{0.001} & \std{0.001} &  & \std{0.001} & \std{0.002} &  & \std{0.004} & \std{0.009} &  \\
 & GTE     & $0.760$ & $0.535$ & $-0.225$ & $0.775$ & $0.817$ & $+0.042$ & $0.868$ & $0.843$ & $-0.025$ & $0.504$ & $0.600$ & $+0.096$ \\[-3pt]
 &         & \std{0.027} & \std{0.019} &  & \std{0.000} & \std{0.002} &  & \std{0.001} & \std{0.002} &  & \std{0.005} & \std{0.009} &  \\
\addlinespace[2pt]
\multirow{6}{*}{Table-Text}
 & TaBERT  & $0.759$ & $0.498$ & $-0.261$ & $0.812$ & $0.855$ & $+0.044$ & $0.819$ & $0.760$ & $-0.059$ & $0.520$ & $0.615$ & $+0.095$ \\[-3pt]
 &         & \std{0.021} & \std{0.034} &  & \std{0.000} & \std{0.002} &  & \std{0.001} & \std{0.004} &  & \std{0.003} & \std{0.009} &  \\
 & TAPAS   & $0.740$ & $0.544$ & $-0.196$ & $0.782$ & $0.823$ & $+0.041$ & $0.867$ & $0.837$ & $-0.030$ & $0.502$ & $0.607$ & $+0.105$ \\[-3pt]
 &         & \std{0.010} & \std{0.011} &  & \std{0.000} & \std{0.002} &  & \std{0.001} & \std{0.004} &  & \std{0.002} & \std{0.005} &  \\
 & TAPEX   & $0.784$ & $0.538$ & $-0.246$ & {---} & {---} & {---} & $0.877$ & $0.854$ & $-0.023$ & $0.494$ & $0.609$ & $+0.115$ \\[-3pt]
 &         & \std{0.010} & \std{0.021} &  &  &  &  & \std{0.000} & \std{0.002} &  & \std{0.002} & \std{0.006} &  \\
\addlinespace[2pt]
\multirow{6}{*}{Table-Struct.}
 & TABBIE  & $0.763$ & $0.542$ & $-0.221$ & $0.832$ & $0.862$ & $+0.030$ & $0.861$ & $0.833$ & $-0.028$ & $0.540$ & $0.663$ & $+0.123$ \\[-3pt]
 &         & \std{0.009} & \std{0.022} &  & \std{0.001} & \std{0.002} &  & \std{0.001} & \std{0.002} &  & \std{0.003} & \std{0.001} &  \\
 & TURL    & $0.626$ & $0.532$ & $-0.094$ & $0.778$ & $0.809$ & $+0.031$ & $0.842$ & $0.814$ & $-0.028$ & $0.549$ & $0.657$ & $+0.108$ \\[-3pt]
 &         & \std{0.014} & \std{0.014} &  & \std{0.001} & \std{0.001} &  & \std{0.001} & \std{0.001} &  & \std{0.003} & \std{0.009} &  \\
 & TUTA    & $0.692$ & $0.468$ & $-0.224$ & {---} & {---} & {---} & $0.850$ & $0.810$ & $-0.041$ & $0.531$ & $0.652$ & $+0.121$ \\[-3pt]
 &         & \std{0.028} & \std{0.010} &  &  &  &  & \std{0.001} & \std{0.003} &  & \std{0.005} & \std{0.007} &  \\
\addlinespace[2pt]
\multirow{4}{*}{Col.-Centric}
 & Starmie & $0.715$ & $0.510$ & $-0.205$ & $0.808$ & $0.847$ & $+0.039$ & $0.876$ & $0.853$ & $-0.023$ & $0.560$ & $0.662$ & $+0.102$ \\[-3pt]
 &         & \std{0.005} & \std{0.019} &  & \std{0.000} & \std{0.001} &  & \std{0.000} & \std{0.002} &  & \std{0.003} & \std{0.003} &  \\
 & TabSketchFM  & $0.750$ & $0.516$ & $-0.234$ & $0.887$ & $0.946$ & $+0.059$ & $0.794$ & $0.737$ & $-0.057$ & $0.545$ & $0.668$ & $+0.123$ \\[-3pt]
 &         & \std{0.015} & \std{0.015} &  & \std{0.001} & \std{0.001} &  & \std{0.001} & \std{0.002} &  & \std{0.002} & \std{0.010} &  \\
\midrule
\multicolumn{2}{l}{\textit{Avg.}}
 & $0.736$ & $0.523$ & $-0.212$ & $0.804$ & $0.843$ & $+0.039$ & $0.853$ & $0.820$ & $-0.033$ & $0.523$ & $0.633$ & $+0.110$ \\
\bottomrule
\end{tabular}%
}
\end{table*}

%% file: tables/row_regime/row_regime_table.tex
\begin{table}[h]
\centering
\small
\setlength{\tabcolsep}{5pt}
\caption{Regime-wise normalized-rank summary for \textsc{TRL-RBench}. For each of the four sub-tasks we report mean $\pm$ std normalized rank within each adaptation regime (lower is better), together with a Kruskal--Wallis $H$-test for the null that all regimes have equal NR distributions and the fraction of total sub-task variance explained by between-regime variance. Values are computed from the per-model NR aggregates in Table~\ref{tab:r_main_results}. The ``between / total'' column quantifies how much of the cross-model variation is captured by regime identity alone.}
\label{tab:row_regime_variance}
\begin{tabular}{lccccc}
\toprule
\textbf{Sub-task} & \textbf{Transfer} & \textbf{Target-Table} & \textbf{Prior-Based} & \makecell{\textbf{Kruskal--Wallis}\\$H$\;($p$)} & \textbf{Between\,/\,Total} \\
\midrule
Classification & $0.504_{\pm0.073}$ & $0.481_{\pm0.134}$ & $0.328_{\pm0.164}$ & $2.0$\;($0.373$) & $0.17$ \\
Regression & $0.637_{\pm0.055}$ & $0.473_{\pm0.129}$ & $0.319_{\pm0.180}$ & $4.8$\;($0.090$) & $0.41$ \\
Clean Linkage & $0.188_{\pm0.060}$ & $0.654_{\pm0.193}$ & $0.509_{\pm0.086}$ & $8.6$\;($0.014$) & $0.64$ \\
Robust Linkage & $0.192_{\pm0.130}$ & $0.647_{\pm0.249}$ & $0.528_{\pm0.135}$ & $5.5$\;($0.064$) & $0.48$ \\
\bottomrule
\end{tabular}
\end{table}

%% file: tables/ablation_dim.tex
\begin{table}[h!]
\centering
\caption{%
  \textbf{Ablation: Embedding dimension for record linkage.}
  Binary $F_1$ (match class, see Appendix~\ref{app:metrics}), 5-seed average, linear probe head.
  Results are broken down by dataset group:
  \textit{DM-C} = 8 clean DeepMatcher pairs,
  \textit{DM-D} = 4 dirty DeepMatcher pairs,
  \textit{WDC} = 4 WDC-Products pairs,
  \textit{All} = unweighted mean over all 16 pairs.
  \textbf{Bold} = best dimension per model, \uline{underline} = second best.
}
\label{tab:ablation_dim}
\resizebox{\textwidth}{!}{%
\begin{tabular}{l ccccc | ccccc | ccccc | ccccc}
\toprule
&
\multicolumn{5}{c|}{\textbf{DM-C} \met{$F_1$$\uparrow$}} &
\multicolumn{5}{c|}{\textbf{DM-D} \met{$F_1$$\uparrow$}} &
\multicolumn{5}{c|}{\textbf{WDC} \met{$F_1$$\uparrow$}} &
\multicolumn{5}{c}{\textbf{All (16 pairs)} \met{$F_1$$\uparrow$}} \\

\cmidrule(lr){2-6}\cmidrule(lr){7-11}\cmidrule(lr){12-16}\cmidrule(lr){17-21}

\textbf{Model}
  & \met{64} & \met{128} & \met{256} & \met{512} & \met{768}
  & \met{64} & \met{128} & \met{256} & \met{512} & \met{768}
  & \met{64} & \met{128} & \met{256} & \met{512} & \met{768}
  & \met{64} & \met{128} & \met{256} & \met{512} & \met{768} \\
\midrule

VIME         & 0.095 & 0.101 & 0.112 & 0.150 & 0.191 & 0.043 & 0.110 & 0.130 & 0.162 & 0.225 & 0.007 & 0.009 & 0.018 & 0.047 & 0.075 & 0.060 & 0.081 & 0.093 & 0.127 & 0.171 \\[-3pt]
             & \std{0.001} & \std{0.026} & \std{0.000} & \std{0.002} & \std{0.000} & \std{0.000} & \std{0.008} & \std{0.000} & \std{0.001} & \std{0.000} & \std{0.000} & \std{0.001} & \std{0.000} & \std{0.002} & \std{0.000} & \std{0.000} & \std{0.011} & \std{0.000} & \std{0.002} & \std{0.000} \\
SCARF        & 0.097 & 0.150 & 0.158 & 0.159 & 0.175 & 0.041 & 0.073 & 0.130 & 0.120 & 0.143 & 0.009 & 0.008 & 0.021 & 0.040 & 0.056 & 0.061 & 0.095 & 0.117 & 0.120 & 0.138 \\[-3pt]
             & \std{0.001} & \std{0.006} & \std{0.000} & \std{0.006} & \std{0.000} & \std{0.000} & \std{0.000} & \std{0.000} & \std{0.010} & \std{0.000} & \std{0.000} & \std{0.000} & \std{0.000} & \std{0.001} & \std{0.001} & \std{0.000} & \std{0.003} & \std{0.000} & \std{0.000} & \std{0.000} \\
DAE          & 0.078 & 0.126 & 0.119 & 0.160 & 0.183 & 0.072 & 0.092 & 0.127 & 0.157 & 0.200 & 0.007 & 0.008 & 0.019 & 0.049 & 0.075 & 0.059 & 0.088 & 0.096 & 0.131 & 0.160 \\[-3pt]
             & \std{0.000} & \std{0.018} & \std{0.000} & \std{0.002} & \std{0.000} & \std{0.000} & \std{0.009} & \std{0.000} & \std{0.004} & \std{0.000} & \std{0.000} & \std{0.000} & \std{0.000} & \std{0.001} & \std{0.000} & \std{0.000} & \std{0.011} & \std{0.000} & \std{0.000} & \std{0.000} \\
TabBinning   & 0.087 & 0.124 & 0.117 & 0.161 & 0.149 & 0.099 & 0.087 & 0.125 & 0.152 & 0.161 & 0.008 & 0.012 & 0.025 & 0.046 & 0.064 & 0.070 & 0.087 & 0.096 & 0.130 & 0.131 \\[-3pt]
             & \std{0.000} & \std{0.000} & \std{0.000} & \std{0.001} & \std{0.000} & \std{0.000} & \std{0.001} & \std{0.000} & \std{0.010} & \std{0.000} & \std{0.000} & \std{0.000} & \std{0.000} & \std{0.000} & \std{0.000} & \std{0.000} & \std{0.000} & \std{0.000} & \std{0.003} & \std{0.000} \\
SAINT        & 0.058 & 0.078 & 0.144 & 0.187 & 0.204 & 0.058 & 0.061 & 0.093 & 0.194 & 0.168 & 0.000 & 0.000 & 0.011 & 0.034 & 0.064 & 0.044 & 0.054 & 0.098 & 0.150 & 0.160 \\[-3pt]
             & \std{0.000} & \std{0.000} & \std{0.000} & \std{0.003} & \std{0.000} & \std{0.000} & \std{0.000} & \std{0.000} & \std{0.013} & \std{0.000} & \std{0.000} & \std{0.000} & \std{0.000} & \std{0.003} & \std{0.000} & \std{0.000} & \std{0.000} & \std{0.000} & \std{0.001} & \std{0.000} \\
SubTab       & 0.000 & 0.033 & 0.014 & 0.035 & 0.043 & 0.000 & 0.010 & 0.017 & 0.017 & 0.043 & 0.000 & 0.000 & 0.000 & 0.000 & 0.001 & 0.000 & 0.019 & 0.011 & 0.022 & 0.032 \\[-3pt]
             & \std{0.000} & \std{0.004} & \std{0.000} & \std{0.003} & \std{0.000} & \std{0.000} & \std{0.022} & \std{0.000} & \std{0.000} & \std{0.000} & \std{0.000} & \std{0.000} & \std{0.000} & \std{0.000} & \std{0.000} & \std{0.000} & \std{0.003} & \std{0.000} & \std{0.001} & \std{0.000} \\
TabTransf.   & 0.064 & 0.075 & 0.071 & 0.103 & 0.100 & 0.000 & 0.010 & 0.042 & 0.062 & 0.056 & 0.002 & 0.002 & 0.008 & 0.000 & 0.010 & 0.032 & 0.041 & 0.048 & 0.067 & 0.067 \\[-3pt]
             & \std{0.000} & \std{0.000} & \std{0.000} & \std{0.012} & \std{0.000} & \std{0.000} & \std{0.000} & \std{0.000} & \std{0.006} & \std{0.000} & \std{0.000} & \std{0.001} & \std{0.000} & \std{0.000} & \std{0.000} & \std{0.000} & \std{0.000} & \std{0.000} & \std{0.004} & \std{0.000} \\
TransTab     & 0.179 & 0.205 & 0.271 & 0.266 & 0.286 & 0.198 & 0.239 & 0.310 & 0.393 & 0.369 & 0.000 & 0.000 & 0.016 & 0.052 & 0.075 & 0.139 & 0.162 & 0.217 & 0.245 & 0.254 \\[-3pt]
             & \std{0.003} & \std{0.011} & \std{0.001} & \std{0.002} & \std{0.013} & \std{0.003} & \std{0.003} & \std{0.008} & \std{0.007} & \std{0.001} & \std{0.000} & \std{0.000} & \std{0.001} & \std{0.002} & \std{0.001} & \std{0.002} & \std{0.006} & \std{0.002} & \std{0.003} & \std{0.006} \\
\midrule
Avg.         & 0.082 & 0.112 & 0.126 & 0.153 & 0.166 & 0.064 & 0.085 & 0.122 & 0.157 & 0.171 & 0.004 & 0.005 & 0.015 & 0.034 & 0.052 & 0.058 & 0.078 & 0.097 & 0.124 & 0.139 \\

\bottomrule
\end{tabular}%
}
\end{table}

%% file: tables/pca_linkage.tex
\begin{table}[h!]
\centering
\caption{%
  \textbf{Common-dimension ablation for record linkage.}
  Binary $F_1$ (match class; Appendix~\ref{app:metrics}) with the MLP probe head, mean$\pm$SD over 5 seeds
  (42/52/62/72/82), each model's score averaged over all 16 record-linkage pairs
  (12 DeepMatcher + 4 WDC-Products). \textbf{Native} is each model's own pretrained embedding width
  (192--768d, Table~\ref{tab:full-models}); \textbf{PCA-192} projects the native
  embedding onto its top-192 principal components (fit on the training split,
  seed-invariant except where the embedding generation itself is seed-sensitive); \textbf{RP-192} is a criterion-free Gaussian
  random-projection control at the same 192-d budget (fresh draw per seed).
  Native values come from this analysis's own run and can differ slightly from the MLP column of Table~\ref{tab:ablation_rl_head}.
  $^{\dagger}$\textsc{TabPFN}'s native embedding is already 192-d, so it is the
  fixed reference point and has no projected arm. The bottom two rows report
  Spearman rank correlation between the native ranking and each projected arm's
  ranking (mean$\pm$SD over 5 seeds) on the 13 projected models, reproduced from
  the source analysis (not recomputed for the per-model rows above); DM-12 = the
  12 DeepMatcher pairs, All-16 = all 16 pairs including WDC-Products.
}
\label{tab:pca_linkage}
\resizebox{\textwidth}{!}{%
\begin{tabular}{l ccc}
\toprule
\textbf{Model} & \textbf{Native}\,\met{$F_1{\uparrow}$} & \textbf{PCA-192}\,\met{$F_1{\uparrow}$} & \textbf{RP-192}\,\met{$F_1{\uparrow}$} \\
\midrule
BERT & $0.474$\,\std{.006} & $0.419$\,\std{.007} & $0.462$\,\std{.006} \\
GTE & $0.510$\,\std{.004} & $0.514$\,\std{.006} & $0.518$\,\std{.009} \\
TABBIE & $0.343$\,\std{.008} & $0.334$\,\std{.004} & $0.338$\,\std{.010} \\
TUTA & $0.439$\,\std{.011} & $0.394$\,\std{.006} & $0.416$\,\std{.008} \\
\midrule
TabICL & $0.399$\,\std{.010} & $0.350$\,\std{.003} & $0.382$\,\std{.003} \\
TabPFN & $0.305$\,\std{.007} & \textemdash$^{\dagger}$ & \textemdash$^{\dagger}$ \\
\midrule
VIME & $0.326$\,\std{.013} & $0.291$\,\std{.008} & $0.315$\,\std{.016} \\
SCARF & $0.321$\,\std{.010} & $0.303$\,\std{.010} & $0.309$\,\std{.006} \\
DAE & $0.318$\,\std{.006} & $0.288$\,\std{.008} & $0.306$\,\std{.009} \\
TabBinning & $0.304$\,\std{.006} & $0.288$\,\std{.008} & $0.294$\,\std{.005} \\
SAINT & $0.195$\,\std{.008} & $0.167$\,\std{.007} & $0.165$\,\std{.008} \\
SubTab & $0.134$\,\std{.013} & $0.250$\,\std{.011} & $0.123$\,\std{.006} \\
TabTransformer & $0.079$\,\std{.014} & $0.194$\,\std{.005} & $0.055$\,\std{.014} \\
TransTab & $0.425$\,\std{.007} & $0.432$\,\std{.005} & $0.420$\,\std{.006} \\
\midrule
Spearman $\rho$(native, arm) --- DM-12 pool & \textemdash & $0.940$\,\std{.019} & $0.947$\,\std{.016} \\
Spearman $\rho$(native, arm) --- All-16 pool & \textemdash & $0.941$\,\std{.022} & $0.985$\,\std{.010} \\
\bottomrule
\end{tabular}%
}
\end{table}

%% file: tables/ablation_rl_head.tex
\begin{table}[h!]
\centering
\caption{%
  \textbf{Ablation: Probe head for record linkage.}
  Four evaluation protocols on frozen row embeddings (base 768-dim):
  \textit{Cosine} = cosine-similarity thresholding (unsupervised).
  \textit{Linear} = logistic regression probe.
  \textit{MLP} = one-hidden-layer MLP probe (hidden size 256).
  \textit{Dummy} = majority-class baseline.
  Binary $F_1$ (match class, see Appendix~\ref{app:metrics}), 5-seed average, grouped by dataset family.
  $\Delta_{\text{L-C}}$ = Linear $-$ Cosine (gain from supervised probing).
}
\label{tab:ablation_rl_head}
\resizebox{\textwidth}{!}{%
\begin{tabular}{l cccc>{\columncolor{gray!15}}c | cccc>{\columncolor{gray!15}}c | cccc>{\columncolor{gray!15}}c | cccc>{\columncolor{gray!15}}c}
\toprule
&
\multicolumn{5}{c|}{\textbf{DM-C (8 pairs)} \met{$F_1$$\uparrow$}} &
\multicolumn{5}{c|}{\textbf{DM-D (4 pairs)} \met{$F_1$$\uparrow$}} &
\multicolumn{5}{c|}{\textbf{WDC (4 pairs)} \met{$F_1$$\uparrow$}} &
\multicolumn{5}{c}{\textbf{All (16 pairs)} \met{$F_1$$\uparrow$}} \\

\cmidrule(lr){2-6}\cmidrule(lr){7-11}\cmidrule(lr){12-16}\cmidrule(lr){17-21}

\textbf{Model}
  & \met{Cos} & \met{Lin} & \met{MLP} & \met{Dum} & \met{$\Delta_{\text{L-C}}$}
  & \met{Cos} & \met{Lin} & \met{MLP} & \met{Dum} & \met{$\Delta_{\text{L-C}}$}
  & \met{Cos} & \met{Lin} & \met{MLP} & \met{Dum} & \met{$\Delta_{\text{L-C}}$}
  & \met{Cos} & \met{Lin} & \met{MLP} & \met{Dum} & \met{$\Delta_{\text{L-C}}$} \\
\midrule

BERT         & 0.390 & 0.349 & 0.487 & 0.000 & \met{$-$0.04} & 0.315 & 0.388 & 0.541 & 0.000 & \met{+0.07} & 0.390 & 0.093 & 0.379 & 0.000 & \met{$-$0.30} & 0.371 & 0.295 & 0.473 & 0.000 & \met{$-$0.08} \\
GTE          & 0.698 & 0.334 & 0.451 & 0.000 & \met{$-$0.36} & 0.728 & 0.442 & 0.589 & 0.000 & \met{$-$0.29} & 0.511 & 0.072 & 0.550 & 0.000 & \met{$-$0.44} & 0.659 & 0.295 & 0.510 & 0.000 & \met{$-$0.36} \\
TUTA         & 0.363 & 0.317 & 0.437 & 0.000 & \met{$-$0.05} & 0.397 & 0.370 & 0.532 & 0.000 & \met{$-$0.03} & 0.354 & 0.090 & 0.365 & 0.000 & \met{$-$0.26} & 0.369 & 0.273 & 0.443 & 0.000 & \met{$-$0.10} \\
TABBIE       & 0.309 & 0.309 & 0.421 & 0.000 & \met{+0.00} & 0.296 & 0.302 & 0.359 & 0.000 & \met{+0.01} & 0.389 & 0.092 & 0.188 & 0.000 & \met{$-$0.30} & 0.326 & 0.253 & 0.347 & 0.000 & \met{$-$0.07} \\
TabICL       & 0.377 & 0.187 & 0.444 & 0.000 & \met{$-$0.19} & 0.328 & 0.193 & 0.442 & 0.000 & \met{$-$0.14} & 0.341 & 0.048 & 0.245 & 0.000 & \met{$-$0.29} & 0.356 & 0.154 & 0.394 & 0.000 & \met{$-$0.20} \\
TabPFN       & 0.260 & 0.165 & 0.343 & 0.000 & \met{$-$0.10} & 0.285 & 0.153 & 0.348 & 0.000 & \met{$-$0.13} & 0.387 & 0.023 & 0.152 & 0.000 & \met{$-$0.36} & 0.298 & 0.126 & 0.296 & 0.000 & \met{$-$0.17} \\
\midrule
TransTab     & 0.410 & 0.258 & 0.419 & 0.000 & \met{$-$0.15} & 0.311 & 0.363 & 0.482 & 0.000 & \met{+0.05} & 0.567 & 0.041 & 0.760 & 0.000 & \met{$-$0.53} & 0.425 & 0.230 & 0.520 & 0.000 & \met{$-$0.19} \\
VIME         & 0.242 & 0.150 & 0.364 & 0.000 & \met{$-$0.09} & 0.295 & 0.148 & 0.370 & 0.000 & \met{$-$0.15} & 0.423 & 0.050 & 0.149 & 0.000 & \met{$-$0.37} & 0.301 & 0.125 & 0.311 & 0.000 & \met{$-$0.18} \\
SCARF        & 0.350 & 0.148 & 0.384 & 0.000 & \met{$-$0.20} & 0.352 & 0.107 & 0.408 & 0.000 & \met{$-$0.24} & 0.357 & 0.037 & 0.102 & 0.000 & \met{$-$0.32} & 0.352 & 0.110 & 0.320 & 0.000 & \met{$-$0.24} \\
DAE          & 0.265 & 0.152 & 0.330 & 0.000 & \met{$-$0.11} & 0.297 & 0.145 & 0.360 & 0.000 & \met{$-$0.15} & 0.429 & 0.051 & 0.145 & 0.000 & \met{$-$0.38} & 0.314 & 0.125 & 0.291 & 0.000 & \met{$-$0.19} \\
TabBinning   & 0.387 & 0.141 & 0.371 & 0.000 & \met{$-$0.25} & 0.421 & 0.141 & 0.417 & 0.000 & \met{$-$0.28} & 0.383 & 0.041 & 0.095 & 0.000 & \met{$-$0.34} & 0.395 & 0.116 & 0.313 & 0.000 & \met{$-$0.28} \\
SAINT        & 0.251 & 0.169 & 0.165 & 0.000 & \met{$-$0.08} & 0.286 & 0.163 & 0.189 & 0.000 & \met{$-$0.12} & 0.429 & 0.032 & 0.235 & 0.000 & \met{$-$0.40} & 0.304 & 0.133 & 0.188 & 0.000 & \met{$-$0.17} \\
SubTab       & 0.275 & 0.034 & 0.154 & 0.000 & \met{$-$0.24} & 0.331 & 0.014 & 0.229 & 0.000 & \met{$-$0.32} & 0.428 & 0.000 & 0.017 & 0.000 & \met{$-$0.43} & 0.327 & 0.020 & 0.138 & 0.000 & \met{$-$0.31} \\
TabTransf.   & 0.257 & 0.097 & 0.068 & 0.000 & \met{$-$0.16} & 0.301 & 0.066 & 0.111 & 0.000 & \met{$-$0.24} & 0.426 & 0.022 & 0.018 & 0.000 & \met{$-$0.40} & 0.310 & 0.071 & 0.066 & 0.000 & \met{$-$0.24} \\
\midrule
Avg.         & 0.345 & 0.201 & 0.346 & 0.000 & \met{$-$0.14} & 0.353 & 0.214 & 0.384 & 0.000 & \met{$-$0.14} & 0.415 & 0.049 & 0.243 & 0.000 & \met{$-$0.37} & 0.365 & 0.166 & 0.330 & 0.000 & \met{$-$0.20} \\

\bottomrule
\end{tabular}%
}
\end{table}

%% file: tables/linkage_split_audit/linkage_split_audit.tex
\begin{table*}[t]
\centering
\scriptsize
\setlength{\tabcolsep}{4pt}
\caption{Source-split row overlap audit for the 16 record-linkage datasets in \textsc{TRL-Rbench}. Pair overlap is the fraction of test pairs that also appear in the train pair list. Row overlap reports the fraction of distinct test-side tableA / tableB rows that already appear in the train+valid pair lists. The original DeepMatcher~\cite{deepmatcher} and WDC LSPM~\cite{wdcproducts_lspm} splits are pair-disjoint by construction (last column), but most sources keep individual rows across splits because each row participates in many candidate pairs. Beer, iTunes-Amazon, iTunes-Amazon-D, WDC-medium, and WDC-small are the only sources where fewer than half of the test-side rows are seen during training on at least one side.}
\label{tab:linkage_split_audit}
\begin{tabular}{l r r r r r r}
\toprule
 & & & & \multicolumn{2}{c}{\textbf{Test rows in train+valid}} & \\
\cmidrule(lr){5-6}
\textbf{Dataset} & \textbf{train} & \textbf{valid} & \textbf{test} & \textbf{tableA (\%)} & \textbf{tableB (\%)} & \textbf{Pair overlap (\%)} \\
\midrule
\multicolumn{7}{l}{\emph{DeepMatcher Clean (DM-C)}} \\
Abt-Buy & 5,743 & 1,916 & 1,916 & 94.2 & 95.0 & 0.00 \\
Amazon-Google & 6,874 & 2,293 & 2,293 & 90.9 & 85.5 & 0.00 \\
Beer & 268 & 91 & 91 & 40.8 & 51.8 & 0.00 \\
DBLP-ACM & 7,417 & 2,473 & 2,473 & 87.0 & 87.7 & 0.00 \\
DBLP-Scholar & 17,223 & 5,742 & 5,742 & 95.8 & 69.5 & 0.00 \\
Fodors-Zagats & 567 & 190 & 189 & 81.0 & 86.8 & 0.00 \\
iTunes-Amazon & 321 & 109 & 109 & 25.0 & 15.1 & 0.00 \\
Walmart-Amazon & 6,144 & 2,049 & 2,049 & 87.9 & 54.8 & 0.00 \\
\midrule
\multicolumn{7}{l}{\emph{DeepMatcher Dirty (DM-D)}} \\
DBLP-ACM-D & 7,417 & 2,473 & 2,473 & 87.0 & 87.7 & 0.00 \\
DBLP-Scholar-D & 17,223 & 5,742 & 5,742 & 95.8 & 69.5 & 0.00 \\
iTunes-Amazon-D & 321 & 109 & 109 & 25.0 & 15.1 & 0.00 \\
Walmart-Amazon-D & 6,144 & 2,049 & 2,049 & 87.9 & 54.8 & 0.00 \\
\midrule
\multicolumn{7}{l}{\emph{WDC Products LSPM v2}} \\
WDC-small & 7,230 & 1,808 & 4,398 & 22.0 & 21.5 & 0.00 \\
WDC-medium & 20,453 & 5,114 & 4,398 & 53.4 & 49.1 & 0.00 \\
WDC-large & 82,714 & 20,683 & 4,398 & 74.2 & 68.9 & 0.00 \\
WDC-xlarge & 171,714 & 42,947 & 4,398 & 75.0 & 70.8 & 0.02 \\
\bottomrule
\end{tabular}
\end{table*}

%% file: tables/linkage_split_audit/strict_absolute.tex
\begin{table}[ht]
\centering
\small
\setlength{\tabcolsep}{4pt}
\caption{Strict-test row-disjoint ablation at seed~42: per-source pair counts and the per-model linkage $F_1$ (avg of MLP and linear probes) of the two top-ranked Robust Linkage row models (\textsc{GTE} and \textsc{TransTab}), compared with the legacy 5-seed avg.\ on the same source. Strict-test rows are absent from train+valid by construction (Sec.~\ref{app:linkage_split_audit}). ``Strict $F_1$'' is single-seed (seed~42).}
\label{tab:strict_absolute}
\begin{adjustbox}{max width=\textwidth}
\begin{tabular}{l rr c | ccc | ccc}
\toprule
& & & & \multicolumn{3}{c|}{\textsc{GTE}} & \multicolumn{3}{c}{\textsc{TransTab}} \\
\cmidrule(lr){5-7} \cmidrule(lr){8-10}
\textbf{Source} & \textbf{Strict pairs} & \textbf{Retained} & \textbf{\% pos} & \textbf{Strict $F_1$} & \textbf{Legacy $F_1$} & $\Delta$ & \textbf{Strict $F_1$} & \textbf{Legacy $F_1$} & $\Delta$ \\
\midrule
Amazon-Google & 50 & 2.2\% & 74.0\% & 0.700 & 0.310 & +0.389 & 0.636 & 0.281 & +0.354 \\
Beer & 30 & 33.0\% & 33.3\% & 0.502 & 0.411 & +0.092 & 0.297 & 0.285 & +0.011 \\
iTunes-Amazon & 70 & 64.2\% & 30.0\% & 0.565 & 0.665 & -0.100 & 0.493 & 0.574 & -0.080 \\
iTunes-Amazon-D & 70 & 64.2\% & 30.0\% & 0.640 & 0.677 & -0.037 & 0.648 & 0.649 & -0.001 \\
Walmart-Amazon & 105 & 5.1\% & 45.7\% & 0.389 & 0.199 & +0.190 & 0.401 & 0.169 & +0.232 \\
Walmart-Amazon-D & 105 & 5.1\% & 45.7\% & 0.415 & 0.204 & +0.210 & 0.258 & 0.171 & +0.087 \\
WDC-small & 2718 & 61.8\% & 26.9\% & 0.289 & 0.313 & -0.025 & 0.264 & 0.402 & -0.139 \\
WDC-medium & 1195 & 27.2\% & 23.8\% & 0.306 & 0.307 & -0.001 & 0.228 & 0.391 & -0.163 \\
WDC-large & 587 & 13.3\% & 15.8\% & 0.303 & 0.314 & -0.011 & 0.262 & 0.406 & -0.144 \\
WDC-xlarge & 577 & 13.1\% & 15.3\% & 0.309 & 0.311 & -0.002 & 0.190 & 0.401 & -0.212 \\
\bottomrule
\end{tabular}
\end{adjustbox}
\end{table}

%% file: tables/taskfree_overview.tex
\begin{table}[h!]
\centering
\footnotesize
\setlength{\tabcolsep}{3pt}
\caption{Spearman rank correlation ($\rho$) between embedding quality
metrics and downstream task performance, reported as
\emph{per-task averages} so the numbers align with the pertask
breakdown tables (Tabs.~\ref{tab:corr_record_linkage}--\ref{tab:corr_regression}).
\textbf{Record Linkage}: Spearman $\rho$ computed per (dataset, head)
across 16 models, then averaged over
$n=32$ (dataset, head) tasks (16 datasets $\times$ \{MLP, linear\}).
\textbf{Row Prediction}: Spearman $\rho$ per (dataset, label, head)
across models, averaged over $n=241$ task-head cells:
classification (76 MLP + 73 linear) and
regression (46 MLP + 46 linear)
from 77 classification and 46 regression targets crossed with
\{MLP, linear\} probe heads,
less 5 classification rows dropped for near-constant performance
(undefined $\rho$).
\textbf{Overall}: unweighted mean of Record Linkage and Row Prediction $\rho$.
Reported $p$ is a two-sided Wilcoxon signed-rank test of the per-task
$\rho$ distribution against zero. Overall $p$ is the smaller of the two
(anti-conservative). Bold $\rho$ indicates $p < 0.05$.
The \textsc{random} baseline encoder is excluded throughout.}
\label{tab:taskfree_overview}
\vspace{4pt}
\begin{tabular}{ll cc cc cc}
\toprule
& & \multicolumn{2}{c}{\textbf{Record Linkage}} & \multicolumn{2}{c}{\textbf{Row Prediction}} & \multicolumn{2}{c}{\textbf{Overall}} \\
\cmidrule(lr){3-4} \cmidrule(lr){5-6} \cmidrule(lr){7-8}
\textbf{Family} & \textbf{Metric} & $\rho$ & $p$ & $\rho$ & $p$ & $\bar\rho$ & $p$ \\
\midrule
  \multirow{3}{*}{\textbf{Spec.\ Spread}} & RankMe & \textbf{+0.714} & {$<$.001} & \textbf{+0.256} & {$<$.001} & \textbf{+0.485} & {$<$.001} \\
   & RankMe* & \textbf{+0.684} & {$<$.001} & \textbf{+0.258} & {$<$.001} & \textbf{+0.471} & {$<$.001} \\
   & NESum & \textbf{+0.657} & {$<$.001} & \textbf{+0.262} & {$<$.001} & \textbf{+0.460} & {$<$.001} \\
\midrule
  \multirow{2}{*}{\textbf{Spec.\ Shape}} & Pseudo $\kappa$ & \textbf{+0.116} & 0.036 & \textbf{+0.032} & 0.039 & \textbf{+0.074} & 0.036 \\
   & $\alpha_{\mathrm{req}}$ & \textbf{-0.746} & {$<$.001} & +0.003 & 0.939 & \textbf{-0.372} & {$<$.001} \\
\midrule
  \multirow{3}{*}{\textbf{Spatial Struct.}} & $\hat{d}_{\mathrm{TwoNN}}$ & \textbf{+0.398} & {$<$.001} & -0.042 & 0.095 & \textbf{+0.178} & {$<$.001} \\
   & Coherence $\mu_0$ & \textbf{-0.549} & {$<$.001} & -0.030 & 0.235 & \textbf{-0.289} & {$<$.001} \\
   & Self-Cluster & \textbf{-0.182} & {$<$.001} & \textbf{-0.237} & {$<$.001} & \textbf{-0.210} & {$<$.001} \\
\bottomrule
\end{tabular}
\end{table}

%% file: tables/taskfree_rp_reg.tex
\begin{table*}[!ht]
\centering
\small
\setlength{\tabcolsep}{4pt}
\caption{Row Prediction, Regression: per-task correlation between embedding metrics and $-\mathrm{nRMSE}$, computed across models within each (dataset, label), then aggregated.}
\label{tab:taskfree_rp_reg}
\vspace{4pt}
\resizebox{\textwidth}{!}{%
\begin{tabular}{ll | rr rr rr | rr rr rr}
\toprule
& & \multicolumn{6}{c|}{\textbf{MLP Head} (46 tasks)}
& \multicolumn{6}{c}{\textbf{Linear Head} (46 tasks)} \\
\cmidrule(lr){3-8} \cmidrule(lr){9-14}
& & \multicolumn{2}{c}{Pearson} & \multicolumn{2}{c}{Spearman} & \multicolumn{2}{c|}{Dir.\ Rate} & \multicolumn{2}{c}{Pearson} & \multicolumn{2}{c}{Spearman} & \multicolumn{2}{c}{Dir.\ Rate} \\
\cmidrule(lr){3-4} \cmidrule(lr){5-6} \cmidrule(lr){7-8} \cmidrule(lr){9-10} \cmidrule(lr){11-12} \cmidrule(lr){13-14}
\textbf{Family} & \textbf{Metric} & Mean & Med. & Mean & Med. & \scriptsize{P} & \scriptsize{S} & Mean & Med. & Mean & Med. & \scriptsize{P} & \scriptsize{S} \\
\midrule
  \multirow{3}{*}{\textbf{Spec.\ Spread}} & RankMe & +0.317 & +0.410 & +0.328 & +0.480 & 80.4\% & 78.3\% & +0.235 & +0.395 & +0.257 & +0.418 & 78.3\% & 76.1\% \\
   & RankMe* & +0.341 & +0.462 & +0.356 & +0.502 & 80.4\% & 78.3\% & +0.261 & +0.415 & +0.287 & +0.472 & 78.3\% & 73.9\% \\
   & NESum & +0.248 & +0.346 & +0.250 & +0.296 & 78.3\% & 76.1\% & +0.225 & +0.303 & +0.231 & +0.315 & 76.1\% & 80.4\% \\
\midrule
  \multirow{2}{*}{\textbf{Spec.\ Shape}} & Pseudo $\kappa$ & +0.118 & -0.021 & +0.036 & +0.118 & 50.0\% & 56.5\% & +0.093 & -0.038 & -0.006 & +0.054 & 56.5\% & 54.3\% \\
   & $\alpha_{\mathrm{req}}$ & +0.012 & +0.014 & +0.080 & +0.144 & 52.2\% & 63.0\% & +0.119 & +0.141 & +0.186 & +0.229 & 63.0\% & 76.1\% \\
\midrule
  \multirow{3}{*}{\textbf{Spatial Struct.}} & $\hat{d}_{\mathrm{TwoNN}}$ & -0.120 & -0.288 & -0.105 & -0.230 & 65.2\% & 63.0\% & -0.165 & -0.212 & -0.174 & -0.266 & 65.2\% & 69.6\% \\
   & Coherence $\mu_0$ & -0.042 & -0.044 & +0.034 & +0.079 & 52.2\% & 58.7\% & +0.011 & +0.031 & +0.097 & +0.137 & 56.5\% & 67.4\% \\
   & Self-Cluster & -0.316 & -0.374 & -0.327 & -0.414 & 80.4\% & 82.6\% & -0.281 & -0.408 & -0.296 & -0.368 & 73.9\% & 80.4\% \\
\bottomrule
\end{tabular}%
}
\end{table*}

%% file: tables/taskfree_pertask.tex
\begin{table}[ht]
\centering
\caption{Per-task correlation between embedding metrics and \textbf{record linkage} performance ($F_1$, higher is better). We report MLP, Linear, and Cosine-Threshold heads.}
\label{tab:corr_record_linkage}
\footnotesize
\setlength{\tabcolsep}{3pt}
\begin{adjustbox}{max width=\textwidth}
\begin{tabular}{lccccccccc}
\toprule
 & \multicolumn{3}{c}{MLP head} & \multicolumn{3}{c}{Linear head} & \multicolumn{3}{c}{Cosine Thr. head} \\
\cmidrule(lr){2-4} \cmidrule(lr){5-7} \cmidrule(lr){8-10}
Metric & Spearman & Pearson & SC & Spearman & Pearson & SC & Spearman & Pearson & SC \\
\midrule
NESum & \makecell{$0.67$\\{\scriptsize\textcolor{gray}{$[0.53,\,0.76]$}}} & \makecell{$0.58$\\{\scriptsize\textcolor{gray}{$[0.44,\,0.69]$}}} & 0.94 & \makecell{$0.64$\\{\scriptsize\textcolor{gray}{$[0.58,\,0.71]$}}} & \makecell{$0.59$\\{\scriptsize\textcolor{gray}{$[0.52,\,0.66]$}}} & 1.00 & \makecell{$0.54$\\{\scriptsize\textcolor{gray}{$[0.45,\,0.63]$}}} & \makecell{$0.73$\\{\scriptsize\textcolor{gray}{$[0.62,\,0.83]$}}} & 1.00 \\
RankMe & \makecell{$0.69$\\{\scriptsize\textcolor{gray}{$[0.54,\,0.80]$}}} & \makecell{$0.59$\\{\scriptsize\textcolor{gray}{$[0.42,\,0.71]$}}} & 0.94 & \makecell{$0.74$\\{\scriptsize\textcolor{gray}{$[0.67,\,0.79]$}}} & \makecell{$0.70$\\{\scriptsize\textcolor{gray}{$[0.64,\,0.74]$}}} & 1.00 & \makecell{$0.56$\\{\scriptsize\textcolor{gray}{$[0.48,\,0.64]$}}} & \makecell{$0.83$\\{\scriptsize\textcolor{gray}{$[0.78,\,0.88]$}}} & 1.00 \\
RankMe$^{\star}$ & \makecell{$0.68$\\{\scriptsize\textcolor{gray}{$[0.53,\,0.79]$}}} & \makecell{$0.60$\\{\scriptsize\textcolor{gray}{$[0.44,\,0.71]$}}} & 0.94 & \makecell{$0.69$\\{\scriptsize\textcolor{gray}{$[0.63,\,0.74]$}}} & \makecell{$0.63$\\{\scriptsize\textcolor{gray}{$[0.57,\,0.70]$}}} & 1.00 & \makecell{$0.52$\\{\scriptsize\textcolor{gray}{$[0.43,\,0.61]$}}} & \makecell{$0.77$\\{\scriptsize\textcolor{gray}{$[0.69,\,0.84]$}}} & 1.00 \\
$\alpha_{\mathrm{req}}$ & \makecell{$-0.69$\\{\scriptsize\textcolor{gray}{$[-0.80,\,-0.54]$}}} & \makecell{$-0.69$\\{\scriptsize\textcolor{gray}{$[-0.77,\,-0.59]$}}} & 0.94 & \makecell{$-0.80$\\{\scriptsize\textcolor{gray}{$[-0.85,\,-0.73]$}}} & \makecell{$-0.59$\\{\scriptsize\textcolor{gray}{$[-0.71,\,-0.47]$}}} & 1.00 & \makecell{$-0.52$\\{\scriptsize\textcolor{gray}{$[-0.60,\,-0.45]$}}} & \makecell{$-0.44$\\{\scriptsize\textcolor{gray}{$[-0.51,\,-0.37]$}}} & 1.00 \\
Pseudo $\kappa$ & \makecell{$0.11$\\{\scriptsize\textcolor{gray}{$[-0.01,\,0.25]$}}} & \makecell{$0.16$\\{\scriptsize\textcolor{gray}{$[-0.04,\,0.36]$}}} & 0.62 & \makecell{$0.12$\\{\scriptsize\textcolor{gray}{$[-0.02,\,0.24]$}}} & \makecell{$0.05$\\{\scriptsize\textcolor{gray}{$[-0.11,\,0.20]$}}} & 0.75 & \makecell{$-0.02$\\{\scriptsize\textcolor{gray}{$[-0.13,\,0.10]$}}} & \makecell{$-0.11$\\{\scriptsize\textcolor{gray}{$[-0.22,\,0.00]$}}} & 0.56 \\
$\mu_0$-coherence & \makecell{$-0.57$\\{\scriptsize\textcolor{gray}{$[-0.67,\,-0.45]$}}} & \makecell{$-0.44$\\{\scriptsize\textcolor{gray}{$[-0.53,\,-0.35]$}}} & 0.94 & \makecell{$-0.53$\\{\scriptsize\textcolor{gray}{$[-0.61,\,-0.46]$}}} & \makecell{$-0.40$\\{\scriptsize\textcolor{gray}{$[-0.48,\,-0.32]$}}} & 1.00 & \makecell{$-0.54$\\{\scriptsize\textcolor{gray}{$[-0.61,\,-0.47]$}}} & \makecell{$-0.38$\\{\scriptsize\textcolor{gray}{$[-0.44,\,-0.32]$}}} & 1.00 \\
Self-cluster & \makecell{$-0.25$\\{\scriptsize\textcolor{gray}{$[-0.33,\,-0.17]$}}} & \makecell{$-0.22$\\{\scriptsize\textcolor{gray}{$[-0.29,\,-0.15]$}}} & 0.94 & \makecell{$-0.11$\\{\scriptsize\textcolor{gray}{$[-0.19,\,-0.03]$}}} & \makecell{$-0.05$\\{\scriptsize\textcolor{gray}{$[-0.12,\,0.04]$}}} & 0.81 & \makecell{$-0.18$\\{\scriptsize\textcolor{gray}{$[-0.24,\,-0.12]$}}} & \makecell{$-0.18$\\{\scriptsize\textcolor{gray}{$[-0.24,\,-0.12]$}}} & 0.94 \\
TwoNN ID & \makecell{$0.33$\\{\scriptsize\textcolor{gray}{$[0.20,\,0.46]$}}} & \makecell{$0.13$\\{\scriptsize\textcolor{gray}{$[-0.06,\,0.31]$}}} & 0.81 & \makecell{$0.46$\\{\scriptsize\textcolor{gray}{$[0.35,\,0.57]$}}} & \makecell{$0.32$\\{\scriptsize\textcolor{gray}{$[0.18,\,0.47]$}}} & 1.00 & \makecell{$0.07$\\{\scriptsize\textcolor{gray}{$[-0.08,\,0.20]$}}} & \makecell{$0.10$\\{\scriptsize\textcolor{gray}{$[-0.04,\,0.25]$}}} & 0.53 \\
\bottomrule
\end{tabular}
\end{adjustbox}
\end{table}

\bigskip

\begin{table}[ht]
\centering
\caption{Per-task correlation between embedding metrics and \textbf{classification} performance (macro-$F_1$, higher is better). Spearman / Pearson cells show mean with bootstrap 95\% CI in brackets over per-task correlations. SC is the sign-consistency fraction (fraction of tasks whose Spearman is in the dominant-sign direction).}
\label{tab:corr_classification}
\footnotesize
\setlength{\tabcolsep}{3pt}
\begin{tabular}{lcccccc}
\toprule
 & \multicolumn{3}{c}{MLP head} & \multicolumn{3}{c}{Linear head} \\
\cmidrule(lr){2-4} \cmidrule(lr){5-7}
Metric & Spearman & Pearson & SC & Spearman & Pearson & SC \\
\midrule
NESum & \makecell{$0.32$\\{\scriptsize\textcolor{gray}{$[0.25,\,0.38]$}}} & \makecell{$0.29$\\{\scriptsize\textcolor{gray}{$[0.23,\,0.35]$}}} & 0.86 & \makecell{$0.23$\\{\scriptsize\textcolor{gray}{$[0.17,\,0.29]$}}} & \makecell{$0.27$\\{\scriptsize\textcolor{gray}{$[0.22,\,0.32]$}}} & 0.79 \\
RankMe & \makecell{$0.28$\\{\scriptsize\textcolor{gray}{$[0.20,\,0.35]$}}} & \makecell{$0.26$\\{\scriptsize\textcolor{gray}{$[0.20,\,0.32]$}}} & 0.79 & \makecell{$0.19$\\{\scriptsize\textcolor{gray}{$[0.11,\,0.26]$}}} & \makecell{$0.23$\\{\scriptsize\textcolor{gray}{$[0.17,\,0.29]$}}} & 0.74 \\
RankMe$^{\star}$ & \makecell{$0.27$\\{\scriptsize\textcolor{gray}{$[0.19,\,0.34]$}}} & \makecell{$0.26$\\{\scriptsize\textcolor{gray}{$[0.20,\,0.32]$}}} & 0.76 & \makecell{$0.17$\\{\scriptsize\textcolor{gray}{$[0.09,\,0.25]$}}} & \makecell{$0.22$\\{\scriptsize\textcolor{gray}{$[0.15,\,0.28]$}}} & 0.66 \\
$\alpha_{\mathrm{req}}$ & \makecell{$-0.08$\\{\scriptsize\textcolor{gray}{$[-0.15,\,-0.02]$}}} & \makecell{$-0.11$\\{\scriptsize\textcolor{gray}{$[-0.19,\,-0.02]$}}} & 0.61 & \makecell{$-0.07$\\{\scriptsize\textcolor{gray}{$[-0.14,\,0.00]$}}} & \makecell{$-0.09$\\{\scriptsize\textcolor{gray}{$[-0.17,\,-0.01]$}}} & 0.60 \\
Pseudo $\kappa$ & \makecell{$0.06$\\{\scriptsize\textcolor{gray}{$[-0.01,\,0.13]$}}} & \makecell{$0.14$\\{\scriptsize\textcolor{gray}{$[0.07,\,0.22]$}}} & 0.60 & \makecell{$0.02$\\{\scriptsize\textcolor{gray}{$[-0.04,\,0.09]$}}} & \makecell{$0.16$\\{\scriptsize\textcolor{gray}{$[0.08,\,0.24]$}}} & 0.55 \\
$\mu_0$-coherence & \makecell{$-0.11$\\{\scriptsize\textcolor{gray}{$[-0.18,\,-0.04]$}}} & \makecell{$-0.21$\\{\scriptsize\textcolor{gray}{$[-0.29,\,-0.13]$}}} & 0.64 & \makecell{$-0.07$\\{\scriptsize\textcolor{gray}{$[-0.14,\,0.00]$}}} & \makecell{$-0.19$\\{\scriptsize\textcolor{gray}{$[-0.28,\,-0.10]$}}} & 0.61 \\
Self-cluster & \makecell{$-0.23$\\{\scriptsize\textcolor{gray}{$[-0.31,\,-0.16]$}}} & \makecell{$-0.22$\\{\scriptsize\textcolor{gray}{$[-0.28,\,-0.15]$}}} & 0.76 & \makecell{$-0.15$\\{\scriptsize\textcolor{gray}{$[-0.23,\,-0.07]$}}} & \makecell{$-0.16$\\{\scriptsize\textcolor{gray}{$[-0.23,\,-0.10]$}}} & 0.67 \\
TwoNN ID & \makecell{$0.05$\\{\scriptsize\textcolor{gray}{$[-0.02,\,0.12]$}}} & \makecell{$-0.01$\\{\scriptsize\textcolor{gray}{$[-0.10,\,0.09]$}}} & 0.53 & \makecell{$-0.02$\\{\scriptsize\textcolor{gray}{$[-0.10,\,0.07]$}}} & \makecell{$-0.05$\\{\scriptsize\textcolor{gray}{$[-0.14,\,0.04]$}}} & 0.58 \\
\bottomrule
\end{tabular}
\end{table}

\bigskip

\begin{table}[ht]
\centering
\caption{Per-task correlation between embedding metrics and \textbf{regression} performance ($\text{nRMSE} = \sqrt{1-R^2}$, lower is better, with per-task $\text{SGM}(\text{nRMSE})=\text{nRMSE}$). We correlate against $-\text{nRMSE}$ so positive values mean ``higher metric $\to$ better performance.''}
\label{tab:corr_regression}
\footnotesize
\setlength{\tabcolsep}{3pt}
\begin{tabular}{lcccccc}
\toprule
 & \multicolumn{3}{c}{MLP head} & \multicolumn{3}{c}{Linear head} \\
\cmidrule(lr){2-4} \cmidrule(lr){5-7}
Metric & Spearman & Pearson & SC & Spearman & Pearson & SC \\
\midrule
NESum & \makecell{$0.25$\\{\scriptsize\textcolor{gray}{$[0.14,\,0.36]$}}} & \makecell{$0.25$\\{\scriptsize\textcolor{gray}{$[0.14,\,0.35]$}}} & 0.76 & \makecell{$0.23$\\{\scriptsize\textcolor{gray}{$[0.12,\,0.33]$}}} & \makecell{$0.23$\\{\scriptsize\textcolor{gray}{$[0.12,\,0.33]$}}} & 0.80 \\
RankMe & \makecell{$0.33$\\{\scriptsize\textcolor{gray}{$[0.22,\,0.43]$}}} & \makecell{$0.32$\\{\scriptsize\textcolor{gray}{$[0.22,\,0.41]$}}} & 0.78 & \makecell{$0.26$\\{\scriptsize\textcolor{gray}{$[0.13,\,0.38]$}}} & \makecell{$0.23$\\{\scriptsize\textcolor{gray}{$[0.11,\,0.35]$}}} & 0.76 \\
RankMe$^{\star}$ & \makecell{$0.36$\\{\scriptsize\textcolor{gray}{$[0.24,\,0.46]$}}} & \makecell{$0.34$\\{\scriptsize\textcolor{gray}{$[0.24,\,0.43]$}}} & 0.78 & \makecell{$0.29$\\{\scriptsize\textcolor{gray}{$[0.16,\,0.41]$}}} & \makecell{$0.26$\\{\scriptsize\textcolor{gray}{$[0.14,\,0.37]$}}} & 0.74 \\
$\alpha_{\mathrm{req}}$ & \makecell{$0.08$\\{\scriptsize\textcolor{gray}{$[-0.01,\,0.17]$}}} & \makecell{$0.01$\\{\scriptsize\textcolor{gray}{$[-0.09,\,0.11]$}}} & 0.63 & \makecell{$0.19$\\{\scriptsize\textcolor{gray}{$[0.09,\,0.28]$}}} & \makecell{$0.12$\\{\scriptsize\textcolor{gray}{$[0.02,\,0.22]$}}} & 0.76 \\
Pseudo $\kappa$ & \makecell{$0.04$\\{\scriptsize\textcolor{gray}{$[-0.06,\,0.13]$}}} & \makecell{$0.12$\\{\scriptsize\textcolor{gray}{$[0.02,\,0.22]$}}} & 0.57 & \makecell{$-0.01$\\{\scriptsize\textcolor{gray}{$[-0.09,\,0.08]$}}} & \makecell{$0.09$\\{\scriptsize\textcolor{gray}{$[-0.00,\,0.19]$}}} & 0.54 \\
$\mu_0$-coherence & \makecell{$0.03$\\{\scriptsize\textcolor{gray}{$[-0.09,\,0.15]$}}} & \makecell{$-0.04$\\{\scriptsize\textcolor{gray}{$[-0.16,\,0.07]$}}} & 0.59 & \makecell{$0.10$\\{\scriptsize\textcolor{gray}{$[-0.02,\,0.22]$}}} & \makecell{$0.01$\\{\scriptsize\textcolor{gray}{$[-0.11,\,0.12]$}}} & 0.67 \\
Self-cluster & \makecell{$-0.33$\\{\scriptsize\textcolor{gray}{$[-0.42,\,-0.23]$}}} & \makecell{$-0.32$\\{\scriptsize\textcolor{gray}{$[-0.41,\,-0.22]$}}} & 0.83 & \makecell{$-0.30$\\{\scriptsize\textcolor{gray}{$[-0.41,\,-0.18]$}}} & \makecell{$-0.28$\\{\scriptsize\textcolor{gray}{$[-0.40,\,-0.16]$}}} & 0.80 \\
TwoNN ID & \makecell{$-0.11$\\{\scriptsize\textcolor{gray}{$[-0.23,\,0.02]$}}} & \makecell{$-0.12$\\{\scriptsize\textcolor{gray}{$[-0.25,\,0.02]$}}} & 0.63 & \makecell{$-0.17$\\{\scriptsize\textcolor{gray}{$[-0.31,\,-0.04]$}}} & \makecell{$-0.17$\\{\scriptsize\textcolor{gray}{$[-0.31,\,-0.02]$}}} & 0.70 \\
\bottomrule
\end{tabular}
\end{table}

%% file: tables/ablation_dlte_pipeline.tex
\begin{table*}[h!]
\centering
\caption{%
  \textbf{Ablation: DLTE pipeline component sensitivity.}
  End-to-end cell $F_1$ (a) and $\mathrm{UJ\text{-}H}$ (b) ($\uparrow$) on the test set as a function of
  Stage~1 table retrieval model (rows) and
  Stage~2 column alignment model (columns).
  Stage~3 row matching is held fixed to the best-performing row model
  (\textbf{ TabTransf. }, selected as the Stage~3 model with the highest
  mean cell $F_1$ across all (Stage~1, Stage~2) pairs, in both panels).
  Each cell shows the full pipeline performance for that (retrieval, alignment) pair.
  \textbf{Bold} = best retrieval model per column alignment. \uline{underline} = second.
}
\label{tab:ablation_dlte}
\resizebox{\textwidth}{!}{%
\begin{tabular}{l cccccccc | c}
\toprule

& \multicolumn{8}{c|}{\textbf{Stage 2: Column Alignment Model}} & \\

\cmidrule(lr){2-9}

\makecell[l]{\textbf{Stage 1:}\\\textbf{Retrieval}}
  & \met{BERT} & \met{GTE} & \met{TaBERT} & \met{TAPAS} & \met{TURL} & \met{Starmie} & \met{TabSketchFM} & \met{TABBIE}
  & \met{Avg.} \\
\midrule
\multicolumn{10}{l}{\textit{(a) Cell $F_1$}} \\
BERT        & 0.607 & 0.608 & 0.673 & 0.612 & 0.603 & 0.544 & 0.554 & 0.570 & 0.597 \\
GTE         & 0.608 & 0.609 & \uline{0.677} & 0.612 & 0.604 & 0.544 & 0.555 & \textbf{0.579} & 0.598 \\
TaBERT      & 0.608 & 0.611 & 0.676 & 0.626 & \uline{0.614} & 0.544 & 0.554 & 0.567 & 0.600 \\
TAPAS       & 0.608 & 0.609 & 0.676 & 0.609 & 0.605 & 0.544 & 0.553 & 0.565 & 0.596 \\
TURL        & 0.606 & 0.605 & 0.667 & 0.610 & 0.599 & 0.544 & \uline{0.557} & \uline{0.575} & 0.595 \\
Starmie     & \uline{0.609} & \textbf{0.639} & \textbf{0.679} & \textbf{0.645} & \textbf{0.615} & 0.544 & \textbf{0.566} & 0.573 & 0.609 \\
TabSketchFM & \textbf{0.627} & 0.601 & 0.650 & 0.615 & 0.602 & 0.544 & 0.552 & 0.556 & 0.593 \\
TABBIE      & 0.586 & 0.571 & 0.575 & 0.573 & 0.579 & 0.544 & 0.550 & 0.541 & 0.565 \\
TAPEX       & 0.575 & 0.573 & 0.575 & 0.565 & 0.584 & 0.544 & 0.547 & 0.512 & 0.559 \\
TUTA        & 0.602 & \uline{0.628} & 0.650 & \uline{0.629} & \textbf{0.615} & 0.544 & 0.554 & 0.567 & 0.599 \\
\midrule
Avg.        & 0.604 & 0.605 & 0.650 & 0.610 & 0.602 & 0.544 & 0.554 & 0.560 & 0.591 \\
\midrule
\multicolumn{10}{l}{\textit{(b) $\mathrm{UJ\text{-}H}$}} \\
BERT        & 0.119 & 0.119 & 0.126 & 0.124 & 0.128 & 0.084 & 0.099 & \uline{0.158} & 0.120 \\
GTE         & 0.120 & 0.120 & 0.125 & 0.122 & 0.129 & 0.084 & 0.100 & 0.157 & 0.120 \\
TaBERT      & 0.119 & 0.120 & 0.125 & 0.127 & \textbf{0.135} & 0.084 & 0.098 & \textbf{0.159} & 0.121 \\
TAPAS       & 0.118 & 0.116 & 0.127 & 0.120 & 0.126 & 0.084 & 0.099 & 0.152 & 0.118 \\
TURL        & 0.117 & 0.118 & 0.132 & 0.124 & 0.122 & 0.084 & \uline{0.103} & 0.151 & 0.119 \\
Starmie     & \uline{0.122} & \uline{0.133} & 0.128 & \textbf{0.132} & \textbf{0.135} & 0.084 & \textbf{0.105} & 0.153 & 0.124 \\
TabSketchFM & \textbf{0.125} & 0.113 & 0.134 & 0.122 & 0.125 & 0.084 & 0.099 & 0.139 & 0.118 \\
TABBIE      & 0.116 & 0.103 & 0.134 & 0.106 & 0.110 & 0.084 & 0.091 & 0.129 & 0.109 \\
TAPEX       & 0.114 & 0.114 & \textbf{0.176} & 0.102 & 0.126 & 0.084 & 0.094 & 0.148 & 0.120 \\
TUTA        & 0.119 & \textbf{0.135} & \uline{0.135} & \uline{0.131} & \uline{0.134} & 0.084 & 0.094 & 0.148 & 0.122 \\
\midrule
Avg.        & 0.119 & 0.119 & 0.134 & 0.121 & 0.127 & 0.084 & 0.098 & 0.149 & 0.119 \\

\bottomrule
\end{tabular}%
}
\end{table*}

%% file: tables/dlte_marginal_table.tex
\begin{table*}[t]
\centering
\small
\setlength{\tabcolsep}{4pt}
\caption{%
  Per-stage marginal contributions in \textsc{TRL-DLTE} (5-round average, test set).
  For a fixed table model, scores are averaged over all 112 compatible pipelines. For a fixed column model, over all 140. For a fixed row model, over all 80.
  \colorbox{rankfirst}{\strut\textbf{Bold orange}} / \colorbox{ranksecond}{\strut\uline{Underlined blue}} / \colorbox{rankthird}{\strut Light purple} highlights indicate best/second-best/third-best per column within each stage panel.
  Stage~1 additionally reports target recall@100 (mean fraction of the two relevant targets recovered among top-100 candidates) for reference.
  Performance span = best $-$ worst marginal score within the stage.
}
\label{tab:dlte_marginal}
\resizebox{\textwidth}{!}{%
\begin{tabular}{ll lccc | ll lccl | ll lcc}
\toprule
& & \multicolumn{4}{c|}{\textbf{Stage 1: Table Model}} &
& & \multicolumn{4}{c|}{\textbf{Stage 2: Column Model}} &
& & \multicolumn{3}{c}{\textbf{Stage 3: Row Model}} \\
\cmidrule(lr){3-6} \cmidrule(lr){9-12} \cmidrule(lr){15-17}
& & \textbf{Model} & \makecell{Cell\\[-2pt]$F_1\uparrow$} & \makecell{$UJ$-$H$\\[-2pt]$\uparrow$} & \makecell{Tgt.\\[-2pt]R@100$\uparrow$} &
& & \textbf{Model} & \makecell{Cell\\[-2pt]$F_1\uparrow$} & \makecell{$UJ$-$H$\\[-2pt]$\uparrow$} & &
& & \textbf{Model} & \makecell{Cell\\[-2pt]$F_1\uparrow$} & \makecell{$UJ$-$H$\\[-2pt]$\uparrow$} \\
\midrule
& & Starmie        & \cellf\textbf{.601} & \cellf\textbf{.144} & \cellt.740 &
& & TaBERT         & \cellf\textbf{.628} & .128 & &
& & TabTransf.     & \cellf\textbf{.591} & .119 \\
& & TUTA           & \cells\uline{.593} & \cells\uline{.138} & .585 &
& & GTE            & \cells\uline{.601} & \cellt.141 & &
& & SubTab         & \cells\uline{.591} & .121 \\
& & GTE            & \cellt.591 & \cellt.129 & \cellf\textbf{.801} &
& & TAPAS          & \cellt.600 & .132 & &
& & SAINT          & \cellt.590 & .119 \\
& & TaBERT         & .590 & .124 & .720 &
& & TURL           & .597 & \cells\uline{.143} & &
& & TABBIE         & .589 & .122 \\
\cmidrule(lr){15-17}
& & BERT           & .589 & .128 & \cells\uline{.763} &
& & BERT           & .595 & .135 & &
& & BERT           & .581 & .128 \\
& & TAPAS          & .587 & .118 & .615 &
& & TABBIE         & .554 & \cellf\textbf{.143} & &
& & TransTab       & .574 & \cellf\textbf{.132} \\
& & TURL           & .587 & .124 & .597 &
& & TabSketchFM    & .553 & .100 & &
& & GTE            & .574 & \cells\uline{.131} \\
& & TabSketchFM    & .584 & .116 & .413 &
& & Starmie        & .544 & .084 & &
& & TabICL         & .566 & \cellt.130 \\
& & TABBIE         & .560 & .109 & .108 &
& & & & & &
& & & & \\
& & TAPEX          & .558 & .127 & .247 &
& & & & & &
& & & & \\
\cmidrule(lr){3-6} \cmidrule(lr){9-12} \cmidrule(lr){15-17}
& & \textit{Span}  & \textit{.043} & \textit{.036} & \textit{.693} &
& & \textit{Span}  & \textit{.084} & \textit{.060} & &
& & \textit{Span}  & \textit{.026} & \textit{.013} \\
\bottomrule
\end{tabular}%
}
\vspace{2pt}
{\raggedright\footnotesize
\textbf{Note.}
Stage~3 shows the top 4 and bottom 4 (by Cell~$F_1$) of 14 row models, separated by a rule; the full ranking is in Appendix~\ref{app:dlte-rankings}. Stage~2 has the widest average downstream $UJ\text{-}H$ span (0.060) and the widest Cell~$F_1$ span (0.084) under the current pipeline, indicating that column-model choice has the largest mean effect among the three stages. Stage~1's target recall@100 ranking does not match its downstream Cell~$F_1$ ranking (e.g., \textsc{GTE} has the best retrieval score but \textsc{Starmie} yields the best downstream Cell~$F_1$ marginal), showing that retrieval quality alone does not determine end-to-end enrichment.\par}
\end{table*}

%% file: tables/oracle_ra_per_tier/oracle_ra_per_tier_uj_h_test.tex
\begin{table}[h]
\centering
\small
\setlength{\tabcolsep}{4pt}
\caption{Oracle-RA per-noise-tier $UJ\text{-}H$ and Cell~$F_1$ for each row model (test split, 5-round mean). Tiers are the cumulative noise levels used in \textsc{TRL-DLTE} construction (clean~$\to$~schema~$\to$~cell~$\to$~hard). The final row reports the cross-row-model span within each tier for each metric, summarizing how Stage~3 separability varies with upstream noise. Row models are sorted by mean $UJ\text{-}H$ across tiers.}
\label{tab:oracle_ra_per_tier_uj_h}
\begin{tabular}{lcccccccc}
\toprule
 & \multicolumn{2}{c}{\textbf{Clean}} & \multicolumn{2}{c}{\textbf{Schema}} & \multicolumn{2}{c}{\textbf{Cell}} & \multicolumn{2}{c}{\textbf{Hard}} \\
\cmidrule(lr){2-3}\cmidrule(lr){4-5}\cmidrule(lr){6-7}\cmidrule(lr){8-9}
\textbf{Row model} & $UJ\text{-}H$ & Cell~$F_1$ & $UJ\text{-}H$ & Cell~$F_1$ & $UJ\text{-}H$ & Cell~$F_1$ & $UJ\text{-}H$ & Cell~$F_1$ \\
\midrule
\textsc{GTE} & 0.703 & 0.839 & 0.704 & 0.839 & 0.633 & 0.774 & 0.692 & 0.756 \\
\textsc{TransTab} & 0.652 & 0.831 & 0.675 & 0.839 & 0.612 & 0.777 & 0.691 & 0.760 \\
\textsc{TabICL} & 0.613 & 0.820 & 0.612 & 0.820 & 0.546 & 0.757 & 0.653 & 0.757 \\
\textsc{TUTA} & 0.502 & 0.765 & 0.499 & 0.768 & 0.449 & 0.722 & 0.499 & 0.717 \\
\textsc{BERT} & 0.480 & 0.760 & 0.480 & 0.762 & 0.431 & 0.719 & 0.425 & 0.699 \\
\textsc{SCARF} & 0.351 & 0.728 & 0.338 & 0.725 & 0.308 & 0.692 & 0.363 & 0.693 \\
\textsc{DAE} & 0.323 & 0.728 & 0.323 & 0.728 & 0.296 & 0.696 & 0.392 & 0.698 \\
\textsc{VIME} & 0.307 & 0.725 & 0.318 & 0.728 & 0.275 & 0.694 & 0.373 & 0.695 \\
\textsc{TabPFN} & 0.306 & 0.728 & 0.300 & 0.726 & 0.267 & 0.692 & 0.297 & 0.679 \\
\textsc{TabBinning} & 0.256 & 0.714 & 0.256 & 0.713 & 0.240 & 0.685 & 0.283 & 0.678 \\
\textsc{TABBIE} & 0.236 & 0.712 & 0.229 & 0.713 & 0.207 & 0.683 & 0.255 & 0.673 \\
\textsc{SAINT} & 0.171 & 0.702 & 0.173 & 0.703 & 0.150 & 0.673 & 0.177 & 0.658 \\
\textsc{SubTab} & 0.170 & 0.704 & 0.172 & 0.705 & 0.151 & 0.675 & 0.164 & 0.655 \\
\textsc{TabTransformer} & 0.141 & 0.699 & 0.141 & 0.699 & 0.127 & 0.672 & 0.139 & 0.651 \\
\midrule
\textit{Span (max$-$min)} & 0.562 & 0.140 & 0.563 & 0.141 & 0.506 & 0.106 & 0.553 & 0.108 \\
\bottomrule
\end{tabular}
\end{table}

%% file: tables/dlte_marginals/dlte_stage1_marginals.tex
\begin{table}[h]
\centering
\small
\setlength{\tabcolsep}{4pt}
\caption{Stage~1 (table model) marginal rankings, sorted by Cell~$F_1$ (mean $\pm$ std over 5 rounds). R@100 is Stage~1 retrieval recall against any gold candidate.}
\label{tab:dlte_stage1_full}
\begin{tabular}{clccc}
\toprule
\textbf{Rank} & \textbf{Model (Family)} & \textbf{Cell~$F_1$~$\uparrow$} & \textbf{$\mathrm{UJ\text{-}H}$~$\uparrow$} & \textbf{R@100~$\uparrow$} \\
\midrule
1 & Starmie (Col.-Centric) & \textbf{$0.601_{\pm0.001}$} & \textbf{$0.144_{\pm0.003}$} & $0.740_{\pm0.000}$ \\
2 & TUTA (Table-Struct.) & $0.593_{\pm0.001}$ & $0.138_{\pm0.001}$ & $0.585_{\pm0.000}$ \\
3 & GTE (Generic Text) & $0.591_{\pm0.002}$ & $0.129_{\pm0.001}$ & \textbf{$0.801_{\pm0.000}$} \\
4 & TaBERT (Table-Text) & $0.590_{\pm0.001}$ & $0.124_{\pm0.001}$ & $0.720_{\pm0.000}$ \\
5 & BERT (Generic Text) & $0.589_{\pm0.002}$ & $0.128_{\pm0.002}$ & $0.763_{\pm0.000}$ \\
6 & TAPAS (Table-Text) & $0.587_{\pm0.001}$ & $0.118_{\pm0.001}$ & $0.615_{\pm0.000}$ \\
7 & TURL (Table-Struct.) & $0.587_{\pm0.000}$ & $0.124_{\pm0.001}$ & $0.597_{\pm0.007}$ \\
8 & TabSketchFM (Col.-Centric) & $0.584_{\pm0.001}$ & $0.116_{\pm0.002}$ & $0.413_{\pm0.001}$ \\
9 & TABBIE (Table-Struct.) & $0.560_{\pm0.000}$ & $0.109_{\pm0.001}$ & $0.108_{\pm0.000}$ \\
10 & TAPEX (Table-Text) & $0.558_{\pm0.001}$ & $0.127_{\pm0.001}$ & $0.247_{\pm0.000}$ \\
\bottomrule
\end{tabular}
\end{table}

%% file: tables/dlte_marginals/dlte_stage2_marginals.tex
\begin{table}[h]
\centering
\small
\setlength{\tabcolsep}{4pt}
\caption{Stage~2 (column model) marginal rankings, sorted by Cell~$F_1$ (mean $\pm$ std over 5 rounds). }
\label{tab:dlte_stage2_full}
\begin{tabular}{clcc}
\toprule
\textbf{Rank} & \textbf{Model (Family)} & \textbf{Cell~$F_1$~$\uparrow$} & \textbf{$\mathrm{UJ\text{-}H}$~$\uparrow$} \\
\midrule
1 & TaBERT (Table-Text) & \textbf{$0.628_{\pm0.001}$} & $0.128_{\pm0.000}$ \\
2 & GTE (Generic Text) & $0.601_{\pm0.001}$ & $0.141_{\pm0.002}$ \\
3 & TAPAS (Table-Text) & $0.600_{\pm0.001}$ & $0.132_{\pm0.001}$ \\
4 & TURL (Table-Struct.) & $0.597_{\pm0.004}$ & $0.143_{\pm0.004}$ \\
5 & BERT (Generic Text) & $0.595_{\pm0.000}$ & $0.135_{\pm0.001}$ \\
6 & TABBIE (Table-Struct.) & $0.554_{\pm0.001}$ & \textbf{$0.143_{\pm0.000}$} \\
7 & TabSketchFM (Col.-Centric) & $0.553_{\pm0.001}$ & $0.100_{\pm0.001}$ \\
8 & Starmie (Col.-Centric) & $0.544_{\pm0.000}$ & $0.084_{\pm0.000}$ \\
\bottomrule
\end{tabular}
\end{table}

%% file: tables/dlte_marginals/dlte_stage3_marginals.tex
\begin{table}[h]
\centering
\small
\setlength{\tabcolsep}{4pt}
\caption{Stage~3 (row model) marginal rankings, sorted by Cell~$F_1$ (mean $\pm$ std over 5 rounds). }
\label{tab:dlte_stage3_full}
\begin{tabular}{clcc}
\toprule
\textbf{Rank} & \textbf{Model (Family)} & \textbf{Cell~$F_1$~$\uparrow$} & \textbf{$\mathrm{UJ\text{-}H}$~$\uparrow$} \\
\midrule
1 & TabTransformer (Target-Table) & \textbf{$0.591_{\pm0.001}$} & $0.119_{\pm0.001}$ \\
2 & SubTab (Target-Table) & $0.591_{\pm0.001}$ & $0.121_{\pm0.001}$ \\
3 & SAINT (Target-Table) & $0.590_{\pm0.002}$ & $0.119_{\pm0.001}$ \\
4 & TABBIE (Table-Struct.) & $0.589_{\pm0.000}$ & $0.122_{\pm0.000}$ \\
5 & TabPFN (Prior-Based) & $0.589_{\pm0.000}$ & $0.126_{\pm0.000}$ \\
6 & DAE (Target-Table) & $0.587_{\pm0.000}$ & $0.127_{\pm0.001}$ \\
7 & TabBinning (Target-Table) & $0.587_{\pm0.003}$ & $0.123_{\pm0.003}$ \\
8 & VIME (Target-Table) & $0.587_{\pm0.001}$ & $0.125_{\pm0.001}$ \\
9 & SCARF (Target-Table) & $0.585_{\pm0.001}$ & $0.127_{\pm0.001}$ \\
10 & TUTA (Table-Struct.) & $0.584_{\pm0.000}$ & $0.130_{\pm0.001}$ \\
11 & BERT (Generic Text) & $0.581_{\pm0.000}$ & $0.128_{\pm0.001}$ \\
12 & TransTab (Target-Table) & $0.574_{\pm0.001}$ & \textbf{$0.132_{\pm0.001}$} \\
13 & GTE (Generic Text) & $0.574_{\pm0.000}$ & $0.131_{\pm0.001}$ \\
14 & TabICL (Prior-Based) & $0.566_{\pm0.000}$ & $0.130_{\pm0.001}$ \\
\bottomrule
\end{tabular}
\end{table}

%% file: tables/ablation_dlte3_merged.tex
\begin{table*}[!ht]
\centering
\caption{%
  \textbf{Ablation: DLTE Stage~3 (row matching).}
  Rows are (Stage~1, Stage~2) configurations. Columns are Stage~3 row models.
  All ten Stage~1 models are included (8 column-capable + TAPEX and TUTA native table encoders).
  Cell $F_1$ ($\uparrow$) on test set.
  Column groupings by adaptation regime: \textbf{Transfer} = externally pretrained, used frozen.
  \textbf{Prior} = meta-pretrained prior-fitted.
  \textbf{Learner} = target-table feature-corruption SSL.
}
\label{tab:ablation_dlte3}
\resizebox{\textwidth}{!}{%
\begin{tabular}{ll cccccccccccccc | c}
\toprule
&  & \multicolumn{4}{c}{\textbf{Transfer}} & \multicolumn{2}{c}{\textbf{Prior}} & \multicolumn{8}{c}{\textbf{Learner}} & \\
\cmidrule(lr){3-6}\cmidrule(lr){7-8}\cmidrule(lr){9-16}
\textbf{Stage 1} & \textbf{Stage 2}
  & \rotatebox{30}{BERT} & \rotatebox{30}{GTE} & \rotatebox{30}{TABBIE} & \rotatebox{30}{TUTA} & \rotatebox{30}{TabICL} & \rotatebox{30}{TabPFN} & \rotatebox{30}{TransTab} & \rotatebox{30}{VIME} & \rotatebox{30}{SCARF} & \rotatebox{30}{DAE} & \rotatebox{30}{TabBinning} & \rotatebox{30}{SAINT} & \rotatebox{30}{SubTab} & \rotatebox{30}{TabTransf.} & \rotatebox{30}{Avg.} \\
\midrule
Starmie     & TaBERT      & 0.638 & 0.607 & 0.671 & 0.643 & 0.589 & 0.670 & 0.609 & 0.662 & 0.654 & 0.661 & 0.663 & 0.674 & 0.677 & 0.679 & 0.650 \\
            & GTE         & 0.637 & 0.632 & 0.639 & 0.639 & 0.614 & 0.638 & 0.634 & 0.636 & 0.633 & 0.636 & 0.635 & 0.638 & 0.639 & 0.639 & 0.635 \\
            & TAPAS       & 0.624 & 0.608 & 0.641 & 0.631 & 0.592 & 0.641 & 0.608 & 0.637 & 0.631 & 0.636 & 0.636 & 0.643 & 0.645 & 0.645 & 0.630 \\
            & TURL        & 0.616 & 0.612 & 0.614 & 0.616 & 0.601 & 0.616 & 0.613 & 0.613 & 0.612 & 0.615 & 0.612 & 0.612 & 0.615 & 0.615 & 0.613 \\
            & BERT        & 0.609 & 0.606 & 0.610 & 0.613 & 0.595 & 0.610 & 0.610 & 0.606 & 0.607 & 0.608 & 0.606 & 0.609 & 0.609 & 0.609 & 0.608 \\
            & TabSketchFM & 0.565 & 0.562 & 0.567 & 0.566 & 0.561 & 0.567 & 0.561 & 0.564 & 0.565 & 0.564 & 0.564 & 0.565 & 0.566 & 0.566 & 0.564 \\
            & TABBIE      & 0.559 & 0.550 & 0.570 & 0.563 & 0.543 & 0.569 & 0.550 & 0.567 & 0.565 & 0.567 & 0.568 & 0.572 & 0.573 & 0.573 & 0.564 \\
            & Starmie     & 0.544 & 0.544 & 0.544 & 0.544 & 0.544 & 0.544 & 0.544 & 0.544 & 0.544 & 0.544 & 0.544 & 0.544 & 0.544 & 0.544 & 0.544 \\
\midrule
TAPAS       & TaBERT      & 0.637 & 0.610 & 0.669 & 0.644 & 0.593 & 0.668 & 0.613 & 0.662 & 0.654 & 0.661 & 0.662 & 0.672 & 0.675 & 0.676 & 0.650 \\
            & GTE         & 0.597 & 0.589 & 0.609 & 0.600 & 0.576 & 0.607 & 0.587 & 0.605 & 0.602 & 0.606 & 0.604 & 0.608 & 0.609 & 0.609 & 0.601 \\
            & BERT        & 0.595 & 0.584 & 0.607 & 0.598 & 0.573 & 0.606 & 0.585 & 0.605 & 0.601 & 0.604 & 0.604 & 0.608 & 0.608 & 0.608 & 0.599 \\
            & TAPAS       & 0.593 & 0.582 & 0.606 & 0.596 & 0.572 & 0.607 & 0.583 & 0.603 & 0.599 & 0.604 & 0.603 & 0.608 & 0.609 & 0.609 & 0.598 \\
            & TURL        & 0.592 & 0.582 & 0.602 & 0.595 & 0.570 & 0.602 & 0.583 & 0.599 & 0.596 & 0.600 & 0.599 & 0.603 & 0.605 & 0.605 & 0.595 \\
            & TABBIE      & 0.552 & 0.546 & 0.563 & 0.557 & 0.542 & 0.562 & 0.548 & 0.560 & 0.558 & 0.560 & 0.560 & 0.563 & 0.564 & 0.565 & 0.557 \\
            & TabSketchFM & 0.550 & 0.551 & 0.554 & 0.553 & 0.549 & 0.553 & 0.550 & 0.552 & 0.553 & 0.553 & 0.553 & 0.553 & 0.553 & 0.553 & 0.552 \\
            & Starmie     & 0.544 & 0.544 & 0.544 & 0.544 & 0.544 & 0.544 & 0.544 & 0.544 & 0.544 & 0.544 & 0.544 & 0.544 & 0.544 & 0.544 & 0.544 \\
\midrule
GTE         & TaBERT      & 0.636 & 0.607 & 0.669 & 0.643 & 0.589 & 0.668 & 0.611 & 0.662 & 0.653 & 0.660 & 0.662 & 0.672 & 0.676 & 0.677 & 0.649 \\
            & TURL        & 0.604 & 0.607 & 0.603 & 0.607 & 0.593 & 0.605 & 0.608 & 0.604 & 0.603 & 0.604 & 0.603 & 0.603 & 0.604 & 0.604 & 0.604 \\
            & TAPAS       & 0.599 & 0.594 & 0.608 & 0.603 & 0.580 & 0.611 & 0.594 & 0.607 & 0.603 & 0.607 & 0.606 & 0.610 & 0.611 & 0.612 & 0.603 \\
            & GTE         & 0.601 & 0.592 & 0.608 & 0.601 & 0.581 & 0.607 & 0.591 & 0.605 & 0.604 & 0.606 & 0.605 & 0.608 & 0.610 & 0.609 & 0.602 \\
            & BERT        & 0.600 & 0.589 & 0.606 & 0.602 & 0.575 & 0.606 & 0.591 & 0.601 & 0.600 & 0.602 & 0.602 & 0.607 & 0.607 & 0.608 & 0.600 \\
            & TABBIE      & 0.565 & 0.557 & 0.577 & 0.569 & 0.552 & 0.576 & 0.558 & 0.574 & 0.572 & 0.573 & 0.575 & 0.577 & 0.579 & 0.579 & 0.570 \\
            & TabSketchFM & 0.553 & 0.552 & 0.556 & 0.555 & 0.552 & 0.555 & 0.552 & 0.554 & 0.555 & 0.555 & 0.554 & 0.554 & 0.555 & 0.555 & 0.554 \\
            & Starmie     & 0.544 & 0.544 & 0.544 & 0.544 & 0.544 & 0.544 & 0.544 & 0.544 & 0.544 & 0.544 & 0.544 & 0.544 & 0.544 & 0.544 & 0.544 \\
\midrule
TaBERT      & TaBERT      & 0.636 & 0.610 & 0.669 & 0.644 & 0.592 & 0.667 & 0.612 & 0.661 & 0.653 & 0.660 & 0.662 & 0.671 & 0.674 & 0.676 & 0.649 \\
            & TAPAS       & 0.605 & 0.593 & 0.622 & 0.609 & 0.580 & 0.623 & 0.594 & 0.619 & 0.613 & 0.619 & 0.618 & 0.624 & 0.626 & 0.626 & 0.612 \\
            & TURL        & 0.600 & 0.592 & 0.611 & 0.605 & 0.579 & 0.611 & 0.593 & 0.608 & 0.605 & 0.609 & 0.608 & 0.611 & 0.613 & 0.614 & 0.604 \\
            & GTE         & 0.600 & 0.594 & 0.611 & 0.602 & 0.582 & 0.608 & 0.592 & 0.607 & 0.605 & 0.609 & 0.607 & 0.610 & 0.611 & 0.611 & 0.603 \\
            & BERT        & 0.593 & 0.577 & 0.604 & 0.595 & 0.565 & 0.604 & 0.579 & 0.600 & 0.598 & 0.600 & 0.600 & 0.606 & 0.606 & 0.608 & 0.595 \\
            & TABBIE      & 0.554 & 0.546 & 0.565 & 0.557 & 0.541 & 0.564 & 0.547 & 0.562 & 0.560 & 0.562 & 0.563 & 0.566 & 0.567 & 0.567 & 0.559 \\
            & TabSketchFM & 0.550 & 0.550 & 0.554 & 0.553 & 0.549 & 0.553 & 0.550 & 0.552 & 0.553 & 0.553 & 0.553 & 0.553 & 0.554 & 0.554 & 0.552 \\
            & Starmie     & 0.544 & 0.544 & 0.544 & 0.544 & 0.544 & 0.544 & 0.544 & 0.544 & 0.544 & 0.544 & 0.544 & 0.544 & 0.544 & 0.544 & 0.544 \\
\midrule
BERT        & TaBERT      & 0.634 & 0.606 & 0.666 & 0.641 & 0.587 & 0.664 & 0.608 & 0.658 & 0.650 & 0.657 & 0.659 & 0.669 & 0.672 & 0.673 & 0.646 \\
            & TAPAS       & 0.599 & 0.594 & 0.610 & 0.606 & 0.581 & 0.612 & 0.594 & 0.608 & 0.604 & 0.609 & 0.607 & 0.611 & 0.612 & 0.612 & 0.604 \\
            & TURL        & 0.603 & 0.603 & 0.602 & 0.606 & 0.591 & 0.603 & 0.605 & 0.602 & 0.601 & 0.602 & 0.601 & 0.602 & 0.604 & 0.603 & 0.602 \\
            & GTE         & 0.599 & 0.589 & 0.607 & 0.599 & 0.578 & 0.605 & 0.588 & 0.604 & 0.602 & 0.605 & 0.604 & 0.606 & 0.608 & 0.608 & 0.600 \\
            & BERT        & 0.599 & 0.586 & 0.605 & 0.599 & 0.574 & 0.605 & 0.588 & 0.602 & 0.600 & 0.602 & 0.602 & 0.607 & 0.607 & 0.607 & 0.599 \\
            & TABBIE      & 0.557 & 0.548 & 0.568 & 0.561 & 0.544 & 0.567 & 0.551 & 0.565 & 0.563 & 0.565 & 0.565 & 0.569 & 0.570 & 0.570 & 0.562 \\
            & TabSketchFM & 0.554 & 0.555 & 0.556 & 0.556 & 0.553 & 0.555 & 0.554 & 0.554 & 0.555 & 0.555 & 0.554 & 0.555 & 0.555 & 0.554 & 0.555 \\
            & Starmie     & 0.544 & 0.544 & 0.544 & 0.544 & 0.544 & 0.544 & 0.544 & 0.544 & 0.544 & 0.544 & 0.544 & 0.544 & 0.544 & 0.544 & 0.544 \\
\midrule
TURL        & TaBERT      & 0.632 & 0.608 & 0.661 & 0.638 & 0.590 & 0.660 & 0.610 & 0.654 & 0.646 & 0.654 & 0.654 & 0.663 & 0.666 & 0.667 & 0.643 \\
            & GTE         & 0.607 & 0.607 & 0.605 & 0.607 & 0.591 & 0.605 & 0.603 & 0.604 & 0.603 & 0.603 & 0.603 & 0.604 & 0.605 & 0.605 & 0.604 \\
            & TAPAS       & 0.591 & 0.578 & 0.606 & 0.594 & 0.566 & 0.606 & 0.579 & 0.603 & 0.599 & 0.604 & 0.603 & 0.609 & 0.610 & 0.610 & 0.597 \\
            & BERT        & 0.592 & 0.575 & 0.602 & 0.592 & 0.564 & 0.601 & 0.576 & 0.599 & 0.596 & 0.599 & 0.599 & 0.605 & 0.606 & 0.606 & 0.594 \\
            & TURL        & 0.588 & 0.578 & 0.597 & 0.589 & 0.569 & 0.597 & 0.578 & 0.595 & 0.593 & 0.596 & 0.595 & 0.597 & 0.599 & 0.599 & 0.591 \\
            & TABBIE      & 0.561 & 0.553 & 0.573 & 0.565 & 0.548 & 0.572 & 0.555 & 0.570 & 0.568 & 0.570 & 0.570 & 0.573 & 0.575 & 0.575 & 0.566 \\
            & TabSketchFM & 0.555 & 0.553 & 0.558 & 0.556 & 0.550 & 0.557 & 0.552 & 0.556 & 0.556 & 0.556 & 0.556 & 0.557 & 0.558 & 0.557 & 0.555 \\
            & Starmie     & 0.544 & 0.544 & 0.544 & 0.544 & 0.544 & 0.544 & 0.544 & 0.544 & 0.544 & 0.544 & 0.544 & 0.544 & 0.544 & 0.544 & 0.544 \\
\midrule
TabSketchFM & TaBERT      & 0.618 & 0.598 & 0.645 & 0.626 & 0.583 & 0.644 & 0.600 & 0.638 & 0.634 & 0.639 & 0.640 & 0.647 & 0.649 & 0.650 & 0.629 \\
            & BERT        & 0.606 & 0.590 & 0.623 & 0.610 & 0.575 & 0.623 & 0.590 & 0.619 & 0.615 & 0.619 & 0.620 & 0.624 & 0.626 & 0.627 & 0.612 \\
            & TAPAS       & 0.595 & 0.582 & 0.611 & 0.598 & 0.571 & 0.611 & 0.582 & 0.606 & 0.602 & 0.608 & 0.607 & 0.613 & 0.614 & 0.615 & 0.601 \\
            & TURL        & 0.589 & 0.580 & 0.599 & 0.591 & 0.570 & 0.599 & 0.580 & 0.597 & 0.594 & 0.598 & 0.597 & 0.600 & 0.602 & 0.602 & 0.593 \\
            & GTE         & 0.587 & 0.578 & 0.598 & 0.591 & 0.568 & 0.597 & 0.578 & 0.595 & 0.593 & 0.596 & 0.596 & 0.598 & 0.601 & 0.601 & 0.591 \\
            & TabSketchFM & 0.550 & 0.549 & 0.553 & 0.552 & 0.548 & 0.552 & 0.549 & 0.551 & 0.551 & 0.551 & 0.552 & 0.552 & 0.552 & 0.552 & 0.551 \\
            & TABBIE      & 0.543 & 0.540 & 0.554 & 0.549 & 0.537 & 0.554 & 0.542 & 0.552 & 0.550 & 0.552 & 0.553 & 0.555 & 0.556 & 0.556 & 0.549 \\
            & Starmie     & 0.544 & 0.544 & 0.544 & 0.544 & 0.544 & 0.544 & 0.544 & 0.544 & 0.544 & 0.544 & 0.544 & 0.544 & 0.544 & 0.544 & 0.544 \\
\midrule
TUTA        & TaBERT      & 0.619 & 0.598 & 0.645 & 0.624 & 0.584 & 0.643 & 0.600 & 0.639 & 0.633 & 0.638 & 0.639 & 0.646 & 0.649 & 0.650 & 0.629 \\
            & GTE         & 0.625 & 0.621 & 0.627 & 0.628 & 0.603 & 0.627 & 0.621 & 0.622 & 0.624 & 0.625 & 0.624 & 0.627 & 0.628 & 0.628 & 0.623 \\
            & TAPAS       & 0.616 & 0.604 & 0.626 & 0.619 & 0.591 & 0.626 & 0.605 & 0.622 & 0.618 & 0.621 & 0.621 & 0.626 & 0.629 & 0.629 & 0.618 \\
            & TURL        & 0.614 & 0.611 & 0.614 & 0.618 & 0.598 & 0.616 & 0.609 & 0.613 & 0.613 & 0.615 & 0.613 & 0.613 & 0.616 & 0.615 & 0.613 \\
            & BERT        & 0.604 & 0.603 & 0.600 & 0.610 & 0.589 & 0.602 & 0.600 & 0.599 & 0.600 & 0.601 & 0.600 & 0.602 & 0.602 & 0.602 & 0.601 \\
            & TABBIE      & 0.555 & 0.548 & 0.566 & 0.559 & 0.543 & 0.565 & 0.550 & 0.562 & 0.561 & 0.563 & 0.564 & 0.566 & 0.568 & 0.567 & 0.560 \\
            & TabSketchFM & 0.553 & 0.552 & 0.555 & 0.554 & 0.550 & 0.554 & 0.551 & 0.553 & 0.554 & 0.554 & 0.554 & 0.554 & 0.554 & 0.554 & 0.553 \\
            & Starmie     & 0.544 & 0.544 & 0.544 & 0.544 & 0.544 & 0.544 & 0.544 & 0.544 & 0.544 & 0.544 & 0.544 & 0.544 & 0.544 & 0.544 & 0.544 \\
\midrule
TAPEX       & GTE         & 0.589 & 0.599 & 0.575 & 0.589 & 0.587 & 0.577 & 0.595 & 0.574 & 0.578 & 0.578 & 0.574 & 0.574 & 0.574 & 0.573 & 0.581 \\
            & TURL        & 0.581 & 0.574 & 0.583 & 0.582 & 0.568 & 0.584 & 0.577 & 0.583 & 0.582 & 0.583 & 0.583 & 0.583 & 0.584 & 0.584 & 0.581 \\
            & BERT        & 0.571 & 0.563 & 0.573 & 0.573 & 0.555 & 0.573 & 0.565 & 0.571 & 0.570 & 0.571 & 0.571 & 0.574 & 0.574 & 0.575 & 0.570 \\
            & TaBERT      & 0.567 & 0.558 & 0.573 & 0.570 & 0.556 & 0.573 & 0.561 & 0.571 & 0.571 & 0.572 & 0.572 & 0.574 & 0.574 & 0.575 & 0.569 \\
            & TAPAS       & 0.565 & 0.561 & 0.565 & 0.565 & 0.557 & 0.565 & 0.562 & 0.563 & 0.563 & 0.564 & 0.563 & 0.564 & 0.565 & 0.565 & 0.563 \\
            & TabSketchFM & 0.546 & 0.546 & 0.547 & 0.547 & 0.546 & 0.547 & 0.545 & 0.546 & 0.547 & 0.546 & 0.546 & 0.546 & 0.547 & 0.547 & 0.546 \\
            & Starmie     & 0.544 & 0.544 & 0.544 & 0.544 & 0.544 & 0.544 & 0.544 & 0.544 & 0.544 & 0.544 & 0.544 & 0.544 & 0.544 & 0.544 & 0.544 \\
            & TABBIE      & 0.508 & 0.511 & 0.511 & 0.512 & 0.516 & 0.512 & 0.514 & 0.511 & 0.511 & 0.511 & 0.512 & 0.512 & 0.512 & 0.512 & 0.512 \\
\midrule
TABBIE      & BERT        & 0.574 & 0.559 & 0.584 & 0.574 & 0.549 & 0.580 & 0.559 & 0.580 & 0.576 & 0.579 & 0.579 & 0.584 & 0.585 & 0.586 & 0.575 \\
            & TURL        & 0.571 & 0.562 & 0.577 & 0.571 & 0.557 & 0.577 & 0.563 & 0.576 & 0.574 & 0.576 & 0.576 & 0.577 & 0.579 & 0.579 & 0.572 \\
            & TaBERT      & 0.564 & 0.557 & 0.574 & 0.567 & 0.552 & 0.573 & 0.559 & 0.571 & 0.570 & 0.572 & 0.572 & 0.574 & 0.575 & 0.575 & 0.568 \\
            & TAPAS       & 0.565 & 0.561 & 0.573 & 0.566 & 0.556 & 0.572 & 0.562 & 0.570 & 0.569 & 0.571 & 0.571 & 0.573 & 0.573 & 0.573 & 0.568 \\
            & GTE         & 0.566 & 0.560 & 0.570 & 0.567 & 0.558 & 0.569 & 0.562 & 0.568 & 0.567 & 0.568 & 0.568 & 0.570 & 0.571 & 0.571 & 0.567 \\
            & TabSketchFM & 0.551 & 0.546 & 0.550 & 0.551 & 0.543 & 0.549 & 0.543 & 0.548 & 0.548 & 0.549 & 0.549 & 0.550 & 0.550 & 0.550 & 0.548 \\
            & Starmie     & 0.544 & 0.544 & 0.544 & 0.544 & 0.544 & 0.544 & 0.544 & 0.544 & 0.544 & 0.544 & 0.544 & 0.544 & 0.544 & 0.544 & 0.544 \\
            & TABBIE      & 0.533 & 0.532 & 0.540 & 0.537 & 0.533 & 0.540 & 0.534 & 0.539 & 0.538 & 0.538 & 0.539 & 0.540 & 0.541 & 0.541 & 0.538 \\
\midrule
\multicolumn{2}{l}{Global Avg.} & 0.581 & 0.574 & 0.589 & 0.584 & 0.566 & 0.589 & 0.574 & 0.587 & 0.585 & 0.587 & 0.587 & 0.590 & 0.591 & 0.591 & 0.584 \\
\bottomrule
\end{tabular}%
}
\end{table*}

\begin{table*}[!ht]
\centering
\caption{%
  \textbf{Ablation: DLTE Stage~3 (row matching), $\mathrm{UJ\text{-}H}$.}
  Same rows, columns, and order as Table~\ref{tab:ablation_dlte3}, with $\mathrm{UJ\text{-}H}$ ($\uparrow$) on test set in place of Cell $F_1$.
}
\label{tab:ablation_dlte3_ujh}
\resizebox{\textwidth}{!}{%
\begin{tabular}{ll cccccccccccccc | c}
\toprule
&  & \multicolumn{4}{c}{\textbf{Transfer}} & \multicolumn{2}{c}{\textbf{Prior}} & \multicolumn{8}{c}{\textbf{Learner}} & \\
\cmidrule(lr){3-6}\cmidrule(lr){7-8}\cmidrule(lr){9-16}
\textbf{Stage 1} & \textbf{Stage 2}
  & \rotatebox{30}{BERT} & \rotatebox{30}{GTE} & \rotatebox{30}{TABBIE} & \rotatebox{30}{TUTA} & \rotatebox{30}{TabICL} & \rotatebox{30}{TabPFN} & \rotatebox{30}{TransTab} & \rotatebox{30}{VIME} & \rotatebox{30}{SCARF} & \rotatebox{30}{DAE} & \rotatebox{30}{TabBinning} & \rotatebox{30}{SAINT} & \rotatebox{30}{SubTab} & \rotatebox{30}{TabTransf.} & \rotatebox{30}{Avg.} \\
\midrule
Starmie     & TaBERT      & 0.122 & 0.115 & 0.127 & 0.123 & 0.112 & 0.129 & 0.117 & 0.127 & 0.126 & 0.127 & 0.126 & 0.126 & 0.128 & 0.128 & 0.124 \\
            & GTE         & 0.211 & 0.253 & 0.155 & 0.215 & 0.236 & 0.166 & 0.251 & 0.175 & 0.174 & 0.175 & 0.158 & 0.142 & 0.138 & 0.133 & 0.184 \\
            & TAPAS       & 0.169 & 0.188 & 0.145 & 0.176 & 0.180 & 0.154 & 0.182 & 0.153 & 0.159 & 0.153 & 0.144 & 0.133 & 0.137 & 0.132 & 0.158 \\
            & TURL        & 0.195 & 0.225 & 0.152 & 0.198 & 0.220 & 0.164 & 0.225 & 0.165 & 0.174 & 0.174 & 0.158 & 0.137 & 0.140 & 0.135 & 0.176 \\
            & BERT        & 0.189 & 0.228 & 0.141 & 0.191 & 0.228 & 0.157 & 0.231 & 0.158 & 0.168 & 0.165 & 0.147 & 0.128 & 0.130 & 0.122 & 0.170 \\
            & TabSketchFM & 0.114 & 0.122 & 0.110 & 0.114 & 0.126 & 0.112 & 0.125 & 0.112 & 0.114 & 0.114 & 0.109 & 0.106 & 0.107 & 0.105 & 0.114 \\
            & TABBIE      & 0.144 & 0.128 & 0.153 & 0.147 & 0.132 & 0.152 & 0.134 & 0.151 & 0.151 & 0.152 & 0.151 & 0.152 & 0.153 & 0.153 & 0.147 \\
            & Starmie     & 0.084 & 0.084 & 0.084 & 0.084 & 0.084 & 0.084 & 0.084 & 0.084 & 0.084 & 0.084 & 0.084 & 0.084 & 0.084 & 0.084 & 0.084 \\
\midrule
TAPAS       & TaBERT      & 0.119 & 0.109 & 0.125 & 0.119 & 0.105 & 0.125 & 0.110 & 0.125 & 0.123 & 0.124 & 0.124 & 0.126 & 0.127 & 0.127 & 0.121 \\
            & GTE         & 0.125 & 0.132 & 0.124 & 0.127 & 0.130 & 0.124 & 0.130 & 0.128 & 0.123 & 0.128 & 0.123 & 0.119 & 0.118 & 0.116 & 0.125 \\
            & BERT        & 0.118 & 0.117 & 0.121 & 0.119 & 0.120 & 0.122 & 0.117 & 0.123 & 0.122 & 0.125 & 0.122 & 0.118 & 0.119 & 0.118 & 0.120 \\
            & TAPAS       & 0.121 & 0.113 & 0.120 & 0.121 & 0.113 & 0.122 & 0.116 & 0.119 & 0.121 & 0.121 & 0.119 & 0.120 & 0.120 & 0.120 & 0.119 \\
            & TURL        & 0.131 & 0.133 & 0.128 & 0.136 & 0.133 & 0.132 & 0.134 & 0.131 & 0.133 & 0.131 & 0.127 & 0.125 & 0.126 & 0.126 & 0.130 \\
            & TABBIE      & 0.145 & 0.129 & 0.151 & 0.146 & 0.133 & 0.151 & 0.134 & 0.150 & 0.149 & 0.151 & 0.150 & 0.150 & 0.152 & 0.152 & 0.146 \\
            & TabSketchFM & 0.097 & 0.097 & 0.100 & 0.098 & 0.097 & 0.100 & 0.097 & 0.100 & 0.100 & 0.100 & 0.100 & 0.098 & 0.099 & 0.099 & 0.099 \\
            & Starmie     & 0.084 & 0.084 & 0.084 & 0.084 & 0.084 & 0.084 & 0.084 & 0.084 & 0.084 & 0.084 & 0.084 & 0.084 & 0.084 & 0.084 & 0.084 \\
\midrule
GTE         & TaBERT      & 0.117 & 0.111 & 0.124 & 0.119 & 0.108 & 0.124 & 0.113 & 0.124 & 0.122 & 0.124 & 0.123 & 0.123 & 0.125 & 0.125 & 0.120 \\
            & TURL        & 0.168 & 0.195 & 0.138 & 0.170 & 0.190 & 0.149 & 0.198 & 0.153 & 0.156 & 0.154 & 0.144 & 0.132 & 0.132 & 0.129 & 0.158 \\
            & TAPAS       & 0.151 & 0.170 & 0.130 & 0.161 & 0.162 & 0.142 & 0.167 & 0.138 & 0.141 & 0.138 & 0.131 & 0.125 & 0.128 & 0.122 & 0.143 \\
            & GTE         & 0.136 & 0.145 & 0.127 & 0.138 & 0.145 & 0.130 & 0.146 & 0.132 & 0.134 & 0.136 & 0.129 & 0.124 & 0.125 & 0.120 & 0.133 \\
            & BERT        & 0.149 & 0.160 & 0.127 & 0.152 & 0.159 & 0.139 & 0.163 & 0.132 & 0.141 & 0.140 & 0.130 & 0.122 & 0.123 & 0.120 & 0.140 \\
            & TABBIE      & 0.148 & 0.128 & 0.156 & 0.149 & 0.137 & 0.156 & 0.138 & 0.155 & 0.154 & 0.155 & 0.155 & 0.155 & 0.157 & 0.157 & 0.150 \\
            & TabSketchFM & 0.101 & 0.104 & 0.101 & 0.102 & 0.107 & 0.103 & 0.107 & 0.103 & 0.104 & 0.104 & 0.101 & 0.100 & 0.100 & 0.100 & 0.103 \\
            & Starmie     & 0.084 & 0.084 & 0.084 & 0.084 & 0.084 & 0.084 & 0.084 & 0.084 & 0.084 & 0.084 & 0.084 & 0.084 & 0.084 & 0.084 & 0.084 \\
\midrule
TaBERT      & TaBERT      & 0.117 & 0.109 & 0.123 & 0.118 & 0.105 & 0.123 & 0.111 & 0.122 & 0.121 & 0.122 & 0.122 & 0.123 & 0.125 & 0.125 & 0.119 \\
            & TAPAS       & 0.128 & 0.130 & 0.131 & 0.129 & 0.127 & 0.133 & 0.131 & 0.134 & 0.131 & 0.133 & 0.129 & 0.126 & 0.129 & 0.127 & 0.130 \\
            & TURL        & 0.149 & 0.165 & 0.141 & 0.156 & 0.164 & 0.147 & 0.164 & 0.150 & 0.151 & 0.148 & 0.144 & 0.134 & 0.137 & 0.135 & 0.149 \\
            & GTE         & 0.136 & 0.150 & 0.129 & 0.139 & 0.148 & 0.130 & 0.148 & 0.134 & 0.134 & 0.139 & 0.131 & 0.124 & 0.124 & 0.120 & 0.135 \\
            & BERT        & 0.130 & 0.136 & 0.122 & 0.134 & 0.137 & 0.130 & 0.138 & 0.127 & 0.132 & 0.128 & 0.125 & 0.120 & 0.120 & 0.119 & 0.128 \\
            & TABBIE      & 0.151 & 0.134 & 0.158 & 0.152 & 0.139 & 0.158 & 0.141 & 0.158 & 0.156 & 0.158 & 0.157 & 0.157 & 0.159 & 0.159 & 0.153 \\
            & TabSketchFM & 0.096 & 0.095 & 0.099 & 0.097 & 0.096 & 0.099 & 0.095 & 0.099 & 0.099 & 0.099 & 0.099 & 0.098 & 0.098 & 0.098 & 0.098 \\
            & Starmie     & 0.084 & 0.084 & 0.084 & 0.084 & 0.084 & 0.084 & 0.084 & 0.084 & 0.084 & 0.084 & 0.084 & 0.084 & 0.084 & 0.084 & 0.084 \\
\midrule
BERT        & TaBERT      & 0.119 & 0.114 & 0.125 & 0.122 & 0.111 & 0.125 & 0.116 & 0.126 & 0.125 & 0.124 & 0.125 & 0.124 & 0.126 & 0.126 & 0.122 \\
            & TAPAS       & 0.154 & 0.172 & 0.133 & 0.164 & 0.164 & 0.146 & 0.169 & 0.143 & 0.145 & 0.143 & 0.135 & 0.127 & 0.130 & 0.124 & 0.146 \\
            & TURL        & 0.163 & 0.189 & 0.138 & 0.168 & 0.182 & 0.145 & 0.191 & 0.149 & 0.154 & 0.154 & 0.142 & 0.133 & 0.134 & 0.128 & 0.155 \\
            & GTE         & 0.134 & 0.136 & 0.127 & 0.134 & 0.136 & 0.128 & 0.138 & 0.131 & 0.131 & 0.135 & 0.129 & 0.122 & 0.123 & 0.119 & 0.130 \\
            & BERT        & 0.139 & 0.140 & 0.123 & 0.139 & 0.137 & 0.128 & 0.143 & 0.126 & 0.131 & 0.129 & 0.125 & 0.120 & 0.122 & 0.119 & 0.130 \\
            & TABBIE      & 0.151 & 0.133 & 0.158 & 0.152 & 0.141 & 0.158 & 0.141 & 0.157 & 0.156 & 0.158 & 0.156 & 0.157 & 0.158 & 0.158 & 0.152 \\
            & TabSketchFM & 0.102 & 0.104 & 0.102 & 0.102 & 0.104 & 0.102 & 0.104 & 0.102 & 0.103 & 0.102 & 0.101 & 0.100 & 0.100 & 0.099 & 0.102 \\
            & Starmie     & 0.084 & 0.084 & 0.084 & 0.084 & 0.084 & 0.084 & 0.084 & 0.084 & 0.084 & 0.084 & 0.084 & 0.084 & 0.084 & 0.084 & 0.084 \\
\midrule
TURL        & TaBERT      & 0.121 & 0.109 & 0.130 & 0.122 & 0.109 & 0.130 & 0.113 & 0.129 & 0.128 & 0.129 & 0.129 & 0.131 & 0.132 & 0.132 & 0.125 \\
            & GTE         & 0.170 & 0.205 & 0.134 & 0.174 & 0.193 & 0.146 & 0.198 & 0.152 & 0.153 & 0.152 & 0.137 & 0.123 & 0.124 & 0.118 & 0.156 \\
            & TAPAS       & 0.133 & 0.142 & 0.127 & 0.139 & 0.146 & 0.135 & 0.148 & 0.134 & 0.136 & 0.134 & 0.129 & 0.125 & 0.128 & 0.124 & 0.134 \\
            & BERT        & 0.121 & 0.122 & 0.118 & 0.123 & 0.122 & 0.122 & 0.123 & 0.123 & 0.124 & 0.123 & 0.121 & 0.116 & 0.119 & 0.117 & 0.121 \\
            & TURL        & 0.120 & 0.118 & 0.123 & 0.123 & 0.122 & 0.126 & 0.121 & 0.126 & 0.125 & 0.128 & 0.125 & 0.122 & 0.124 & 0.122 & 0.123 \\
            & TABBIE      & 0.141 & 0.126 & 0.151 & 0.144 & 0.132 & 0.151 & 0.135 & 0.150 & 0.148 & 0.150 & 0.149 & 0.149 & 0.151 & 0.151 & 0.145 \\
            & TabSketchFM & 0.103 & 0.101 & 0.104 & 0.103 & 0.101 & 0.104 & 0.102 & 0.104 & 0.104 & 0.104 & 0.103 & 0.102 & 0.103 & 0.103 & 0.103 \\
            & Starmie     & 0.084 & 0.084 & 0.084 & 0.084 & 0.084 & 0.084 & 0.084 & 0.084 & 0.084 & 0.084 & 0.084 & 0.084 & 0.084 & 0.084 & 0.084 \\
\midrule
TabSketchFM & TaBERT      & 0.124 & 0.113 & 0.132 & 0.126 & 0.113 & 0.132 & 0.117 & 0.131 & 0.130 & 0.131 & 0.131 & 0.133 & 0.134 & 0.134 & 0.127 \\
            & BERT        & 0.118 & 0.110 & 0.124 & 0.119 & 0.111 & 0.124 & 0.111 & 0.124 & 0.123 & 0.124 & 0.123 & 0.124 & 0.125 & 0.125 & 0.120 \\
            & TAPAS       & 0.118 & 0.117 & 0.124 & 0.119 & 0.117 & 0.125 & 0.118 & 0.126 & 0.124 & 0.126 & 0.123 & 0.121 & 0.123 & 0.122 & 0.122 \\
            & TURL        & 0.128 & 0.132 & 0.126 & 0.131 & 0.133 & 0.131 & 0.133 & 0.131 & 0.131 & 0.132 & 0.128 & 0.125 & 0.125 & 0.125 & 0.129 \\
            & GTE         & 0.113 & 0.111 & 0.114 & 0.115 & 0.110 & 0.114 & 0.111 & 0.114 & 0.114 & 0.115 & 0.114 & 0.113 & 0.114 & 0.113 & 0.113 \\
            & TabSketchFM & 0.098 & 0.097 & 0.100 & 0.098 & 0.097 & 0.100 & 0.097 & 0.100 & 0.100 & 0.100 & 0.100 & 0.099 & 0.099 & 0.099 & 0.099 \\
            & TABBIE      & 0.131 & 0.118 & 0.138 & 0.133 & 0.121 & 0.138 & 0.124 & 0.137 & 0.136 & 0.138 & 0.137 & 0.138 & 0.139 & 0.139 & 0.133 \\
            & Starmie     & 0.084 & 0.084 & 0.084 & 0.084 & 0.084 & 0.084 & 0.084 & 0.084 & 0.084 & 0.084 & 0.084 & 0.084 & 0.084 & 0.084 & 0.084 \\
\midrule
TUTA        & TaBERT      & 0.124 & 0.112 & 0.133 & 0.124 & 0.113 & 0.133 & 0.117 & 0.133 & 0.131 & 0.132 & 0.133 & 0.134 & 0.135 & 0.135 & 0.128 \\
            & GTE         & 0.197 & 0.229 & 0.149 & 0.204 & 0.210 & 0.164 & 0.218 & 0.163 & 0.175 & 0.169 & 0.153 & 0.142 & 0.142 & 0.135 & 0.175 \\
            & TAPAS       & 0.168 & 0.186 & 0.141 & 0.172 & 0.176 & 0.153 & 0.181 & 0.149 & 0.151 & 0.149 & 0.142 & 0.133 & 0.136 & 0.131 & 0.155 \\
            & TURL        & 0.181 & 0.201 & 0.145 & 0.190 & 0.194 & 0.158 & 0.198 & 0.155 & 0.166 & 0.164 & 0.152 & 0.138 & 0.140 & 0.134 & 0.165 \\
            & BERT        & 0.177 & 0.216 & 0.130 & 0.194 & 0.204 & 0.153 & 0.210 & 0.147 & 0.160 & 0.153 & 0.141 & 0.124 & 0.124 & 0.119 & 0.161 \\
            & TABBIE      & 0.140 & 0.124 & 0.147 & 0.143 & 0.130 & 0.148 & 0.131 & 0.147 & 0.145 & 0.147 & 0.146 & 0.146 & 0.148 & 0.148 & 0.142 \\
            & TabSketchFM & 0.095 & 0.095 & 0.095 & 0.095 & 0.097 & 0.096 & 0.097 & 0.096 & 0.097 & 0.096 & 0.096 & 0.095 & 0.095 & 0.094 & 0.096 \\
            & Starmie     & 0.084 & 0.084 & 0.084 & 0.084 & 0.084 & 0.084 & 0.084 & 0.084 & 0.084 & 0.084 & 0.084 & 0.084 & 0.084 & 0.084 & 0.084 \\
\midrule
TAPEX       & GTE         & 0.164 & 0.204 & 0.127 & 0.170 & 0.201 & 0.141 & 0.201 & 0.138 & 0.150 & 0.149 & 0.134 & 0.117 & 0.121 & 0.114 & 0.152 \\
            & TURL        & 0.134 & 0.140 & 0.131 & 0.138 & 0.144 & 0.134 & 0.144 & 0.135 & 0.137 & 0.137 & 0.133 & 0.126 & 0.129 & 0.126 & 0.135 \\
            & BERT        & 0.134 & 0.157 & 0.122 & 0.142 & 0.161 & 0.129 & 0.159 & 0.130 & 0.133 & 0.133 & 0.124 & 0.114 & 0.118 & 0.114 & 0.133 \\
            & TaBERT      & 0.163 & 0.139 & 0.174 & 0.165 & 0.151 & 0.175 & 0.146 & 0.173 & 0.173 & 0.174 & 0.174 & 0.174 & 0.176 & 0.176 & 0.167 \\
            & TAPAS       & 0.109 & 0.110 & 0.104 & 0.111 & 0.113 & 0.106 & 0.112 & 0.104 & 0.108 & 0.106 & 0.103 & 0.101 & 0.102 & 0.102 & 0.106 \\
            & TabSketchFM & 0.094 & 0.094 & 0.094 & 0.094 & 0.094 & 0.094 & 0.094 & 0.094 & 0.094 & 0.094 & 0.094 & 0.094 & 0.094 & 0.094 & 0.094 \\
            & Starmie     & 0.084 & 0.084 & 0.084 & 0.084 & 0.084 & 0.084 & 0.084 & 0.084 & 0.084 & 0.084 & 0.084 & 0.084 & 0.084 & 0.084 & 0.084 \\
            & TABBIE      & 0.140 & 0.125 & 0.148 & 0.142 & 0.130 & 0.148 & 0.129 & 0.147 & 0.146 & 0.147 & 0.147 & 0.147 & 0.148 & 0.148 & 0.142 \\
\midrule
TABBIE      & BERT        & 0.123 & 0.131 & 0.121 & 0.122 & 0.129 & 0.121 & 0.129 & 0.123 & 0.124 & 0.123 & 0.118 & 0.116 & 0.117 & 0.116 & 0.122 \\
            & TURL        & 0.112 & 0.113 & 0.112 & 0.114 & 0.113 & 0.112 & 0.113 & 0.113 & 0.113 & 0.113 & 0.110 & 0.108 & 0.110 & 0.110 & 0.112 \\
            & TaBERT      & 0.125 & 0.113 & 0.133 & 0.127 & 0.117 & 0.133 & 0.118 & 0.132 & 0.131 & 0.132 & 0.132 & 0.133 & 0.134 & 0.134 & 0.128 \\
            & TAPAS       & 0.101 & 0.096 & 0.106 & 0.102 & 0.098 & 0.106 & 0.098 & 0.105 & 0.105 & 0.106 & 0.105 & 0.106 & 0.106 & 0.106 & 0.103 \\
            & GTE         & 0.102 & 0.101 & 0.104 & 0.104 & 0.103 & 0.104 & 0.103 & 0.104 & 0.105 & 0.105 & 0.103 & 0.103 & 0.103 & 0.103 & 0.103 \\
            & TabSketchFM & 0.095 & 0.096 & 0.093 & 0.093 & 0.097 & 0.093 & 0.095 & 0.094 & 0.094 & 0.093 & 0.093 & 0.092 & 0.092 & 0.091 & 0.094 \\
            & Starmie     & 0.084 & 0.084 & 0.084 & 0.084 & 0.084 & 0.084 & 0.084 & 0.084 & 0.084 & 0.084 & 0.084 & 0.084 & 0.084 & 0.084 & 0.084 \\
            & TABBIE      & 0.123 & 0.113 & 0.128 & 0.124 & 0.114 & 0.128 & 0.118 & 0.127 & 0.127 & 0.127 & 0.127 & 0.127 & 0.129 & 0.129 & 0.124 \\
\midrule
\multicolumn{2}{l}{Global Avg.} & 0.128 & 0.131 & 0.122 & 0.130 & 0.130 & 0.126 & 0.132 & 0.125 & 0.127 & 0.127 & 0.123 & 0.119 & 0.121 & 0.119 & 0.126 \\
\bottomrule
\end{tabular}%
}
\end{table*}

%% file: tables/dlte_source_split/source_split_headline.tex
\begin{table}[h]
\centering
\small
\setlength{\tabcolsep}{4pt}
\caption{Source-split $\mathrm{UJ\text{-}H}$ and Cell~$F_1$ for headline DLTE pipelines (5-round mean over the test split). The hybrid-vs-monolith gap holds separately on TabFact (246 test parents) and WTQ (99 test parents), and is wider on WTQ. Dev-selected pipelines are the rank-1 hybrid and rank-1 monolithic by dev $\mathrm{UJ\text{-}H}$. The test-set unconstrained block is the rank-1 by $\mathrm{UJ\text{-}H}$ across all 1{,}120 pipelines on test, included as a sensitivity reference. Pipeline notation: Stage~1 (Tbl) / Stage~2 (Col) / Stage~3 (Row).}
\label{tab:dlte_source_split_headline}
\begin{tabular}{lcccccc}
\toprule
 & \multicolumn{3}{c}{$\mathrm{UJ\text{-}H}$} & \multicolumn{3}{c}{Cell~$F_1$} \\
\cmidrule(lr){2-4}\cmidrule(lr){5-7}
\textbf{Pipeline} & \textbf{TabFact} & \textbf{WTQ} & \textbf{All} & \textbf{TabFact} & \textbf{WTQ} & \textbf{All} \\
\midrule
\multicolumn{7}{l}{\textit{Dev-selected (top-1 by dev $\mathrm{UJ\text{-}H}$, evaluated on test):}} \\
Best hybrid: \textsc{TUTA}/\textsc{GTE}/\textsc{GTE} & \textbf{0.225} & \textbf{0.238} & \textbf{0.229} & \textbf{0.617} & \textbf{0.633} & \textbf{0.621} \\
Best monolith: \textsc{BERT}/\textsc{BERT}/\textsc{BERT} & 0.139 & 0.140 & 0.139 & 0.596 & 0.607 & 0.599 \\
Hybrid $-$ monolith & $+0.086$ & $+0.099$ & $+0.090$ & $+0.021$ & $+0.026$ & $+0.022$ \\
\midrule
\multicolumn{7}{l}{\textit{Test-set unconstrained reference (max $\mathrm{UJ\text{-}H}$ over 1{,}120 pipelines):}} \\
Best hybrid: \textsc{Starmie}/\textsc{GTE}/\textsc{GTE} & 0.242 & 0.281 & 0.253 & 0.627 & 0.645 & 0.632 \\
Best monolith: \textsc{GTE}/\textsc{GTE}/\textsc{GTE} & 0.140 & 0.157 & 0.145 & 0.589 & 0.598 & 0.592 \\
Hybrid $-$ monolith & $+0.102$ & $+0.125$ & $+0.108$ & $+0.038$ & $+0.047$ & $+0.040$ \\
\bottomrule
\end{tabular}
\end{table}

%% file: tables/dlte_source_split/source_split_marginals.tex
\begin{table}[h]
\centering
\small
\setlength{\tabcolsep}{4pt}
\caption{Per-stage marginal $\mathrm{UJ\text{-}H}$ top-5 under each source, averaged over all 1,120 canonical pipelines. Each entry gives the model's Cell~$F_1$ marginal on its second line. Stage~1 is \textsc{Starmie}-led in both sources. Stage~2 has a \textsc{TABBIE}~$\leftrightarrow$~\textsc{TURL} swap at rank 1 but the same top models recur. Stage~3 is \textsc{TransTab}-led in both.}
\label{tab:dlte_source_split_marginals}
\begin{adjustbox}{max width=\textwidth}
\begin{tabular}{llccccc}
\toprule
\textbf{Stage} & \textbf{Source} & \textbf{Rank 1} & \textbf{Rank 2} & \textbf{Rank 3} & \textbf{Rank 4} & \textbf{Rank 5} \\
\midrule
\multirow{6}{*}{Stage 1 (Tbl)}
  & TabFact & \makecell{\textsc{Starmie} (0.140)\\0.596} & \makecell{\textsc{TUTA} (0.137)\\0.588} & \makecell{\textsc{GTE} (0.125)\\0.587} & \makecell{\textsc{BERT} (0.124)\\0.585} & \makecell{\textsc{TAPEX} (0.122)\\0.554} \\
  & WTQ & \makecell{\textsc{Starmie} (0.155)\\0.614} & \makecell{\textsc{TUTA} (0.140)\\0.604} & \makecell{\textsc{TAPEX} (0.138)\\0.568} & \makecell{\textsc{GTE} (0.138)\\0.601} & \makecell{\textsc{TURL} (0.137)\\0.601} \\
  & All & \makecell{\textsc{Starmie} (0.144)\\0.601} & \makecell{\textsc{TUTA} (0.138)\\0.593} & \makecell{\textsc{GTE} (0.129)\\0.591} & \makecell{\textsc{BERT} (0.128)\\0.589} & \makecell{\textsc{TAPEX} (0.127)\\0.558} \\
\midrule
\multirow{6}{*}{Stage 2 (Col)}
  & TabFact & \makecell{\textsc{TABBIE} (0.140)\\0.543} & \makecell{\textsc{GTE} (0.137)\\0.597} & \makecell{\textsc{TURL} (0.137)\\0.585} & \makecell{\textsc{BERT} (0.132)\\0.591} & \makecell{\textsc{TAPAS} (0.132)\\0.599} \\
  & WTQ & \makecell{\textsc{TURL} (0.159)\\0.626} & \makecell{\textsc{TABBIE} (0.153)\\0.580} & \makecell{\textsc{GTE} (0.150)\\0.611} & \makecell{\textsc{TaBERT} (0.149)\\0.645} & \makecell{\textsc{BERT} (0.140)\\0.606} \\
  & All & \makecell{\textsc{TABBIE} (0.143)\\0.554} & \makecell{\textsc{TURL} (0.143)\\0.597} & \makecell{\textsc{GTE} (0.141)\\0.601} & \makecell{\textsc{BERT} (0.135)\\0.595} & \makecell{\textsc{TAPAS} (0.132)\\0.600} \\
\midrule
\multirow{6}{*}{Stage 3 (Row)}
  & TabFact & \makecell{\textsc{TransTab} (0.128)\\0.572} & \makecell{\textsc{GTE} (0.127)\\0.572} & \makecell{\textsc{TabICL} (0.127)\\0.564} & \makecell{\textsc{TUTA} (0.126)\\0.580} & \makecell{\textsc{BERT} (0.123)\\0.578} \\
  & WTQ & \makecell{\textsc{TransTab} (0.142)\\0.580} & \makecell{\textsc{TUTA} (0.140)\\0.594} & \makecell{\textsc{GTE} (0.140)\\0.578} & \makecell{\textsc{TabICL} (0.139)\\0.570} & \makecell{\textsc{BERT} (0.138)\\0.589} \\
  & All & \makecell{\textsc{TransTab} (0.132)\\0.574} & \makecell{\textsc{GTE} (0.131)\\0.574} & \makecell{\textsc{TabICL} (0.130)\\0.566} & \makecell{\textsc{TUTA} (0.130)\\0.584} & \makecell{\textsc{BERT} (0.128)\\0.581} \\
\bottomrule
\end{tabular}
\end{adjustbox}
\end{table}

%% file: tables/dlte_source_split/oracle_ra_per_source.tex
\begin{table}[h]
\centering
\small
\setlength{\tabcolsep}{4pt}
\caption{Oracle-RA per-source $\mathrm{UJ\text{-}H}$ and Cell~$F_1$ for each row model (test split, 5-round mean). Sources are the two \textsc{TRL-DLTE} parent pools (TabFact: 246 test parents, WTQ: 99 test parents). The ``All'' column is the pooled mean over the 345 test parents $\times$ 4 noise tiers ($1{,}380$ query-tier evaluations). The final row reports the cross-row-model span within each source for each metric, comparable in magnitude to the per-tier spans (0.506--0.563 for $\mathrm{UJ\text{-}H}$, 0.106--0.141 for Cell~$F_1$) in Table~\ref{tab:oracle_ra_per_tier_uj_h}. Row models are sorted by pooled $\mathrm{UJ\text{-}H}$.}
\label{tab:oracle_ra_per_source_uj_h}
\begin{tabular}{lcccccc}
\toprule
 & \multicolumn{3}{c}{$\mathrm{UJ\text{-}H}$} & \multicolumn{3}{c}{Cell~$F_1$} \\
\cmidrule(lr){2-4}\cmidrule(lr){5-7}
\textbf{Row model} & \textbf{TabFact} & \textbf{WTQ} & \textbf{All} & \textbf{TabFact} & \textbf{WTQ} & \textbf{All} \\
\midrule
\textsc{GTE}            & 0.684 & 0.680 & 0.683 & 0.805 & 0.796 & 0.802 \\
\textsc{TransTab}       & 0.677 & 0.609 & 0.658 & 0.808 & 0.786 & 0.802 \\
\textsc{TabICL}         & 0.615 & 0.585 & 0.606 & 0.792 & 0.779 & 0.789 \\
\textsc{TUTA}           & 0.499 & 0.457 & 0.487 & 0.749 & 0.729 & 0.743 \\
\textsc{BERT}           & 0.467 & 0.421 & 0.454 & 0.741 & 0.719 & 0.735 \\
\textsc{SCARF}          & 0.361 & 0.288 & 0.340 & 0.716 & 0.694 & 0.709 \\
\textsc{DAE}            & 0.346 & 0.302 & 0.333 & 0.718 & 0.701 & 0.713 \\
\textsc{VIME}           & 0.336 & 0.275 & 0.318 & 0.716 & 0.696 & 0.710 \\
\textsc{TabPFN}         & 0.304 & 0.265 & 0.293 & 0.712 & 0.692 & 0.706 \\
\textsc{TabBinning}     & 0.252 & 0.274 & 0.259 & 0.700 & 0.692 & 0.697 \\
\textsc{TABBIE}         & 0.237 & 0.219 & 0.231 & 0.700 & 0.684 & 0.695 \\
\textsc{SAINT}          & 0.161 & 0.185 & 0.168 & 0.687 & 0.677 & 0.684 \\
\textsc{SubTab}         & 0.166 & 0.161 & 0.164 & 0.689 & 0.675 & 0.685 \\
\textsc{TabTransformer} & 0.133 & 0.147 & 0.137 & 0.684 & 0.671 & 0.680 \\
\midrule
\textit{Span (max$-$min)} & 0.551 & 0.533 & 0.546 & 0.124 & 0.125 & 0.122 \\
\bottomrule
\end{tabular}
\end{table}

%% file: tables/dlte_per_side.tex
\begin{table}[h]
\centering
\small
\setlength{\tabcolsep}{4pt}
\caption{Per-side union and join precision and $F_1$ (test split, 5-round means over the 1{,}120 pipelines). Precision is the share of the cells a pipeline adds in the union block (new rows in the seed columns) or the join block (new columns for the seed rows) that match the removed cells of that target. Recall is the target recall used in $\mathrm{UJ\text{-}H}$, and $F_1$ is the per-query harmonic mean of the two, averaged over queries. (a) Per-stage marginal means, sorted by union $F_1$, with the best value of each stage in bold. (b) Headline pipelines with their ranks under each per-side $F_1$. A rank range marks a tie.}
\label{tab:dlte_per_side}
\textbf{(a)} Per-stage marginal means\par\smallskip
\begin{tabular}{lcccc}
\toprule
 & \multicolumn{2}{c}{\textbf{Union}} & \multicolumn{2}{c}{\textbf{Join}} \\
\cmidrule(lr){2-3}\cmidrule(lr){4-5}
\textbf{Model} & \textbf{P} & \textbf{$F_1$} & \textbf{P} & \textbf{$F_1$} \\
\midrule
\multicolumn{5}{l}{\textit{Stage~1 (table model, 112 pipelines each)}} \\
Starmie & \textbf{0.487} & \textbf{0.471} & \textbf{0.072} & \textbf{0.041} \\
TaBERT & 0.432 & 0.418 & 0.026 & 0.016 \\
TUTA & 0.409 & 0.401 & 0.056 & 0.033 \\
GTE & 0.407 & 0.396 & 0.041 & 0.023 \\
BERT & 0.404 & 0.393 & 0.038 & 0.022 \\
TAPAS & 0.404 & 0.392 & 0.016 & 0.011 \\
TURL & 0.388 & 0.379 & 0.028 & 0.017 \\
TabSketchFM & 0.383 & 0.371 & 0.009 & 0.006 \\
TABBIE & 0.190 & 0.190 & 0.010 & 0.006 \\
TAPEX & 0.178 & 0.189 & 0.037 & 0.023 \\
\midrule
\multicolumn{5}{l}{\textit{Stage~2 (column model, 140 pipelines each)}} \\
TaBERT & \textbf{0.755} & \textbf{0.706} & 0.004 & 0.003 \\
GTE & 0.466 & 0.458 & \textbf{0.078} & \textbf{0.048} \\
BERT & 0.446 & 0.446 & 0.060 & 0.034 \\
TAPAS & 0.438 & 0.420 & 0.050 & 0.030 \\
TURL & 0.384 & 0.382 & 0.061 & 0.033 \\
TABBIE & 0.354 & 0.361 & 0.003 & 0.002 \\
TabSketchFM & 0.104 & 0.107 & 0.011 & 0.008 \\
Starmie & 0.000 & 0.000 & 0.000 & 0.000 \\
\midrule
\multicolumn{5}{l}{\textit{Stage~3 (row model, 80 pipelines each)}} \\
TabTransformer & 0.364 & \textbf{0.381} & 0.009 & 0.002 \\
SubTab & 0.364 & 0.380 & 0.018 & 0.006 \\
SAINT & 0.364 & 0.377 & 0.008 & 0.004 \\
TABBIE & 0.366 & 0.377 & 0.035 & 0.013 \\
TabPFN & 0.364 & 0.375 & 0.041 & 0.016 \\
VIME & 0.365 & 0.370 & 0.037 & 0.016 \\
TabBinning & 0.365 & 0.370 & 0.027 & 0.012 \\
DAE & 0.366 & 0.370 & 0.040 & 0.018 \\
SCARF & 0.366 & 0.364 & 0.037 & 0.019 \\
TUTA & 0.372 & 0.361 & 0.043 & 0.028 \\
BERT & 0.374 & 0.355 & 0.043 & 0.027 \\
GTE & \textbf{0.381} & 0.334 & \textbf{0.046} & \textbf{0.042} \\
TransTab & 0.375 & 0.327 & 0.043 & 0.040 \\
TabICL & 0.370 & 0.299 & 0.040 & 0.037 \\
\bottomrule
\end{tabular}

\medskip
\textbf{(b)} Headline pipelines\par\smallskip
\setlength{\tabcolsep}{3pt}
\begin{tabular}{llcccccc}
\toprule
 & & \multicolumn{2}{c}{\textbf{Union}} & \multicolumn{2}{c}{\textbf{Join}} & \multicolumn{2}{c}{\textbf{Rank by $F_1$}} \\
\cmidrule(lr){3-4}\cmidrule(lr){5-6}\cmidrule(lr){7-8}
\textbf{Pipeline} & \textbf{Selection} & \textbf{P} & \textbf{$F_1$} & \textbf{P} & \textbf{$F_1$} & \textbf{Union} & \textbf{Join} \\
\midrule
Starmie/TaBERT/TabTransformer & Best union $F_1$ & 0.914 & 0.916 & 0.001 & 0.000 & 1 & 862 \\
TAPEX/GTE/GTE & Best join $F_1$ & 0.230 & 0.223 & 0.221 & 0.203 & 776 & 1 \\
Starmie/GTE/GTE & Best $\mathrm{UJ\text{-}H}$ & 0.737 & 0.649 & 0.211 & 0.193 & 126 & 2 \\
TUTA/GTE/GTE & Dev-selected hybrid & 0.638 & 0.571 & 0.166 & 0.162 & 158 & 12 \\
\midrule
BERT/BERT/BERT & Monolith & 0.494 & 0.491 & 0.058 & 0.039 & 294 & 180 \\
GTE/GTE/GTE & Monolith & 0.518 & 0.463 & 0.072 & 0.064 & 346 & 95 \\
TABBIE/TABBIE/TABBIE & Monolith & 0.255 & 0.276 & 0.000 & 0.000 & 719 & 890--1120 \\
\bottomrule
\end{tabular}
\end{table}

%% file: tables/openai_ablation.tex
\begin{table*}[t]
\centering
\caption{%
  Embedders beyond the pool on the 13 \textsc{TRL-CTbench} tasks, beside the pool models of Table~\ref{tab:main_results}.
  Metrics and aggregation follow Table~\ref{tab:main_results}: table-level cells use the aggregation (and, for Table Retrieval, the query encoder) selected on development data,
  and Union Search averages SANTOS, TUS, TUS-hard, UGEN-v1, and UGEN-v2 (per dataset in Table~\ref{tab:openai_ablation_union}).
  Values are means over 5 seeds, with the std on the second line. Cells without a std are single deterministic evaluations.
  \textbf{Bold} = best, \uline{underline} = second best.
  Dashes mark an unsupported granularity, and n/e marks a task outside the model's evaluated subset.
  The original Table Retrieval heads of \textsc{Qwen3-Emb-8B} were not kept, so its development-data pick uses heads re-trained with the same recipe.
}
\label{tab:openai_ablation}
\setlength{\ulthick}{1pt}%
\resizebox{\textwidth}{!}{%
\begin{tabular}{ll ccc|ccc|ccccc|cc}
\toprule
& &
\multicolumn{3}{c|}{\makecell{\textbf{Schema}}} &
\multicolumn{3}{c|}{\makecell{\textbf{Join}}} &
\multicolumn{5}{c|}{\makecell{\textbf{Union}}} &
\multicolumn{2}{c}{\makecell{\textbf{Grounding}}} \\
\cmidrule(lr){3-5}\cmidrule(lr){6-8}\cmidrule(lr){9-13}\cmidrule(lr){14-15}
\textbf{Type} & \textbf{Model} &
 \makecell{Col\\[-2pt]Type$^{\text{c}}$\\\met{$F_1\uparrow$}} &
 \makecell{Col\\[-2pt]Clust$^{\text{c}}$\\\met{NMI$\uparrow$}} &
 \makecell{Col\\[-2pt]Rel$^{\text{c}}$\\\met{$F_1\uparrow$}} &
 \makecell{Join\\[-2pt]Search$^{\text{c}}$\\\met{MAP$\uparrow$}} &
 \makecell{Col\\[-2pt]Overlap$^{\text{c}\dagger}$\\\met{nRMSE$\downarrow$}} &
 \makecell{Join\\[-2pt]Class.$^{\text{t}\dagger}$\\\met{$F_1\uparrow$}} &
 \makecell{Union\\[-2pt]Search$^{\text{c}}$\\\met{MAP$\uparrow$}} &
 \makecell{Schema\\[-2pt]Match$^{\text{c}}$\\\met{R@GT$\uparrow$}} &
 \makecell{Union\\[-2pt]Class.$^{\text{t}\dagger}$\\\met{$F_1\uparrow$}} &
 \makecell{Union\\[-2pt]Reg.$^{\text{t}\dagger}$\\\met{nRMSE$\downarrow$}} &
 \makecell{Tbl\\[-2pt]Subset$^{\text{t}}$\\\met{$F_1\uparrow$}} &
 \makecell{Tbl\\[-2pt]QA$^{\text{c}}$\\\met{Acc$\uparrow$}} &
 \makecell{Tbl\\[-2pt]Ret.$^{\text{t}}$\\\met{MRR$\uparrow$}} \\
\midrule
\multirow{4}{*}{Generic Text}
 & BERT & $0.926$ & $\textbf{0.516}$ & $0.826$ & $0.434$ & $\textbf{0.786}$ & $0.496$ & $0.564$ & $0.423$ & $0.857$ & $0.595$ & $0.545$ & $0.255$ & $0.367$ \\[-3pt]
 & & \std{.001} & \std{.000} & \std{.002} & \std{.000} & \std{.001} & \std{.020} & \std{.000} & \std{.000} & \std{.002} & \std{.005} & \std{.005} & \std{.004} & \std{.008} \\
 & GTE & $0.922$ & $0.466$ & $0.811$ & $0.469$ & $0.817$ & $0.535$ & $0.608$ & $0.417$ & $0.839$ & $0.600$ & $0.544$ & $0.245$ & $0.476$ \\[-3pt]
 & & \std{.002} & \std{.001} & \std{.002} & \std{.002} & \std{.002} & \std{.019} & \std{.000} & \std{.000} & \std{.001} & \std{.009} & \std{.002} & \std{.002} & \std{.003} \\
\addlinespace[2pt]
\multirow{6}{*}{Table-Text}
 & TaBERT & $0.874$ & \uline{$0.514$} & $0.760$ & $0.406$ & $0.855$ & $0.498$ & $0.619$ & $0.430$ & $0.760$ & $0.615$ & $0.540$ & $0.267$ & $0.372$ \\[-3pt]
 & & \std{.003} & \std{.003} & \std{.002} & \std{.002} & \std{.002} & \std{.034} & \std{.000} & \std{.000} & \std{.004} & \std{.009} & \std{.004} & \std{.005} & \std{.013} \\
 & TAPAS & $0.868$ & $0.448$ & $0.769$ & $0.320$ & $0.823$ & $0.509$ & $0.529$ & $0.415$ & $0.837$ & $0.612$ & \uline{$0.567$} & $0.254$ & $0.295$ \\[-3pt]
 & & \std{.001} & \std{.001} & \std{.003} & \std{.001} & \std{.002} & \std{.015} & \std{.000} & \std{.000} & \std{.004} & \std{.004} & \std{.004} & \std{.003} & \std{.006} \\
 & TAPEX & --- & --- & --- & --- & --- & $0.514$ & --- & --- & $0.854$ & $0.622$ & $0.557$ & --- & $0.332$ \\[-3pt]
 & &  &  &  &  &  & \std{.023} &  &  & \std{.002} & \std{.005} & \std{.004} &  & \std{.005} \\
\addlinespace[2pt]
\multirow{6}{*}{Table-Struct.}
 & TABBIE & $0.892$ & $0.262$ & $0.785$ & $0.208$ & $0.862$ & $0.527$ & $0.410$ & $0.205$ & $0.833$ & $0.663$ & $0.546$ & \uline{$0.276$} & $0.170$ \\[-3pt]
 & & \std{.004} & \std{.007} & \std{.002} & \std{.001} & \std{.002} & \std{.017} & \std{.000} & \std{.000} & \std{.002} & \std{.001} & \std{.004} & \std{.004} & \std{.004} \\
 & TURL & $0.814$ & $0.406$ & $0.758$ & $0.299$ & $0.809$ & $0.532$ & $0.575$ & $0.340$ & $0.814$ & $0.657$ & $0.507$ & $\textbf{0.277}$ & $0.199$ \\[-3pt]
 & & \std{.001} & \std{.004} & \std{.002} & \std{.000} & \std{.001} & \std{.014} & \std{.000} & \std{.001} & \std{.001} & \std{.009} & \std{.004} & \std{.005} & \std{.010} \\
 & TUTA & --- & --- & --- & --- & --- & $0.468$ & --- & --- & $0.810$ & $0.652$ & $0.447$ & --- & $0.235$ \\[-3pt]
 & &  &  &  &  &  & \std{.010} &  &  & \std{.003} & \std{.007} & \std{.008} &  & \std{.013} \\
\addlinespace[2pt]
\multirow{4}{*}{Col.-Centric}
 & Starmie & $0.789$ & $0.404$ & $0.698$ & $0.316$ & $0.847$ & $0.510$ & $0.662$ & $\textbf{0.764}$ & $0.853$ & $0.662$ & $0.539$ & $0.266$ & $0.018$ \\[-3pt]
 & & \std{.004} & \std{.000} & \std{.001} & \std{.001} & \std{.001} & \std{.019} & \std{.000} & \std{.000} & \std{.002} & \std{.003} & \std{.005} & \std{.005} & \std{.002} \\
 & TabSketchFM & $0.566$ & $0.252$ & $0.373$ & $0.265$ & $0.946$ & $0.505$ & $0.531$ & $0.155$ & $0.737$ & $0.671$ & $0.545$ & $0.235$ & $0.218$ \\[-3pt]
 & & \std{.001} & \std{.007} & \std{.003} & \std{.001} & \std{.001} & \std{.011} & \std{.000} & \std{.000} & \std{.002} & \std{.006} & \std{.007} & \std{.005} & \std{.011} \\
\midrule
\multirow{6}{*}{Proprietary}
 & TE3-Small & $0.936$ & $0.497$ & \uline{$0.850$} & $0.473$ & \uline{$0.788$} & \uline{$0.561$} & $0.660$ & \uline{$0.651$} & $0.857$ & \uline{$0.586$} & $0.558$ & $0.260$ & $0.490$ \\[-3pt]
 & & \std{.002} & \std{.000} & \std{.001} & \std{.001} & \std{.002} & \std{.024} & \std{.000} & \std{.000} & \std{.002} & \std{.004} & \std{.003} & \std{.005} & \std{.007} \\
 & TE3-Large & \uline{$0.939$} & $0.508$ & $\textbf{0.857}$ & \uline{$0.476$} & $0.791$ & $\textbf{0.577}$ & $0.664$ & $0.636$ & $\textbf{0.859}$ & $0.592$ & $0.559$ & $0.265$ & \uline{$0.511$} \\[-3pt]
 & & \std{.001} & \std{.001} & \std{.002} & \std{.001} & \std{.001} & \std{.027} & \std{.000} & \std{.000} & \std{.003} & \std{.006} & \std{.002} & \std{.003} & \std{.006} \\
 & Ada-002 & $\textbf{0.940}$ & $0.513$ & $0.846$ & $0.469$ & $0.795$ & $0.529$ & $0.621$ & $0.584$ & $0.844$ & $\textbf{0.583}$ & $\textbf{0.574}$ & $0.266$ & $\textbf{0.529}$ \\[-3pt]
 & & \std{.002} & \std{.000} & \std{.001} & \std{.001} & \std{.001} & \std{.011} & \std{.000} & \std{.000} & \std{.004} & \std{.009} & \std{.006} & \std{.004} & \std{.006} \\
\addlinespace[2pt]
\multirow{8}{*}{Open}
 & Qwen3-Emb-8B & n/e & n/e & n/e & n/e & n/e & n/e & $\textbf{0.670}$ & $0.533$ & n/e & n/e & $0.550$ & $0.264$ & $0.487$ \\[-3pt]
 & &  &  &  &  &  &  &  &  &  &  & \std{.003} & \std{.003} & \std{.008} \\
 & Qwen3-Emb-0.6B & $0.913$ & $0.477$ & $0.826$ & $\textbf{0.476}$ & $0.808$ & $0.518$ & \uline{$0.666$} & $0.630$ & \uline{$0.858$} & $0.631$ & $0.547$ & $0.264$ & $0.446$ \\[-3pt]
 & & \std{.002} &  & \std{.002} & \std{.001} & \std{.001} & \std{.024} &  &  & \std{.001} & \std{.004} & \std{.006} & \std{.004} & \std{.007} \\
 & ModernBERT & n/e & n/e & n/e & n/e & n/e & n/e & $0.515$ & $0.391$ & n/e & n/e & n/e & n/e & n/e \\[-3pt]
 & &  &  &  &  &  &  &  &  &  &  &  &  &  \\
 & GTE-ModernBERT & n/e & n/e & n/e & n/e & n/e & n/e & $0.661$ & $0.474$ & n/e & n/e & n/e & n/e & n/e \\[-3pt]
 & &  &  &  &  &  &  &  &  &  &  &  &  &  \\
\bottomrule
\end{tabular}%
}
\end{table*}

%% file: tables/openai_ablation_union.tex
\begin{table}[t]
\centering
\small
\caption{%
  Union Search MAP per dataset.
  The mean over the five datasets is the Union Search value of Tables~\ref{tab:main_results} and~\ref{tab:openai_ablation}.
  Pool values and the OpenAI trio's SANTOS, TUS, and TUS-hard values are 5-round means. The remaining values are single deterministic evaluations.
  \textbf{Bold} = best, \uline{underline} = second best.
}
\label{tab:openai_ablation_union}
\setlength{\tabcolsep}{4pt}
\begin{tabular}{ll ccccc|c}
\toprule
\textbf{Type} & \textbf{Model} & SANTOS & TUS & TUS-hard & UGEN-v1 & UGEN-v2 & Mean \\
\midrule
\multirow{2}{*}{Generic Text}
 & BERT & $0.908$ & $0.959$ & $0.307$ & $0.426$ & $0.222$ & $0.564$ \\
 & GTE & $0.891$ & $0.954$ & $0.293$ & $0.415$ & $0.485$ & $0.608$ \\
\addlinespace[2pt]
\multirow{2}{*}{Table-Text}
 & TaBERT & $0.867$ & $0.926$ & \uline{$0.436$} & $0.397$ & $0.468$ & $0.619$ \\
 & TAPAS & $0.891$ & $0.891$ & $0.376$ & $0.409$ & $0.077$ & $0.529$ \\
\addlinespace[2pt]
\multirow{2}{*}{Table-Struct.}
 & TABBIE & $0.721$ & $0.700$ & $0.317$ & $0.236$ & $0.076$ & $0.410$ \\
 & TURL & $0.903$ & $0.953$ & $0.304$ & $0.379$ & $0.335$ & $0.575$ \\
\addlinespace[2pt]
\multirow{2}{*}{Col.-Centric}
 & Starmie & $0.965$ & $0.844$ & $\textbf{0.523}$ & $0.437$ & $0.541$ & $0.662$ \\
 & TabSketchFM & $0.935$ & $0.941$ & $0.376$ & $0.347$ & $0.055$ & $0.531$ \\
\midrule
\multirow{3}{*}{Proprietary}
 & TE3-Small & $\textbf{0.999}$ & $0.980$ & $0.299$ & $\textbf{0.496}$ & $0.525$ & $0.660$ \\
 & TE3-Large & \uline{$0.996$} & $0.978$ & $0.292$ & $0.489$ & $0.567$ & $0.664$ \\
 & Ada-002 & $0.903$ & $0.960$ & $0.298$ & $0.437$ & $0.509$ & $0.621$ \\
\addlinespace[2pt]
\multirow{4}{*}{Open}
 & Qwen3-Emb-8B & $0.981$ & $0.999$ & $0.295$ & $0.485$ & $\textbf{0.591}$ & $\textbf{0.670}$ \\
 & Qwen3-Emb-0.6B & $0.991$ & $\textbf{1.000}$ & $0.282$ & \uline{$0.496$} & $0.561$ & \uline{$0.666$} \\
 & ModernBERT & $0.836$ & $0.996$ & $0.306$ & $0.384$ & $0.051$ & $0.515$ \\
 & GTE-ModernBERT & $0.982$ & \uline{$1.000$} & $0.293$ & $0.457$ & \uline{$0.573$} & $0.661$ \\
\bottomrule
\end{tabular}
\end{table}

%% file: tables/openai_ablation_row.tex
\begin{table}[t]
\centering
\small
\caption{%
  Row-level results of the OpenAI trio and \textsc{Qwen3-Embedding-8B} beside six pool models from Table~\ref{tab:r_main_results}.
  Row metrics average MLP and linear probes. RowPred reports AUROC over the 77 classification targets of the row-prediction suite.
  RecLink reports binary $F_1$ (match class), averaged unweighted over all 16 linkage datasets (Table~\ref{tab:r_main_results} reports the same metric per group).
  Values are mean $\pm$ std over 5 seeds.
  \textbf{Bold} = best, \uline{underline} = second best.
}
\label{tab:openai_ablation_row}
\setlength{\tabcolsep}{4pt}
\begin{tabular}{ll cc}
\toprule
\textbf{Type} & \textbf{Model}
  & \makecell{RowPred\\[-2pt]\met{AUROC$\,\uparrow$}}
  & \makecell{RecLink\\[-2pt]\met{$F_1\,\uparrow$}} \\
\midrule
\multirow{2}{*}{Generic Text}
 & BERT & $0.791_{\pm0.000}$ & $0.384_{\pm0.002}$ \\
 & GTE & $0.770_{\pm0.000}$ & $0.403_{\pm0.003}$ \\
\addlinespace[2pt]
\multirow{2}{*}{Table-Struct.}
 & TABBIE & $0.770_{\pm0.001}$ & $0.300_{\pm0.005}$ \\
 & TUTA & $0.720_{\pm0.000}$ & $0.358_{\pm0.005}$ \\
\addlinespace[2pt]
Meta-Pretr.
 & TabICL & $\textbf{0.816}_{\pm0.001}$ & $0.274_{\pm0.005}$ \\
\addlinespace[2pt]
Tgt-Tbl
 & TransTab & $0.778_{\pm0.001}$ & $0.375_{\pm0.017}$ \\
\midrule
\multirow{3}{*}{Proprietary}
 & TE3-Small & $\underline{0.801_{\pm0.000}}$ & $0.412_{\pm0.004}$ \\
 & TE3-Large & $0.797_{\pm0.001}$ & $\textbf{0.426}_{\pm0.004}$ \\
 & Ada-002 & $0.789_{\pm0.001}$ & $\underline{0.423_{\pm0.002}}$ \\
\addlinespace[2pt]
Open (8B)
 & Qwen3-Emb-8B & $0.794_{\pm0.002}$ & $0.405_{\pm0.004}$ \\
\bottomrule
\end{tabular}
\end{table}

%% file: tables/qwen3_row_suite.tex
\begin{table}[t]
\centering
\small
\caption{%
  \textsc{Qwen3-Embedding-0.6B} on the 50-table row-prediction suite, beside \textsc{BERT}, \textsc{GTE}, \textsc{TabPFN}, and \textsc{TabICL}
  under the frozen-probe protocol of Table~\ref{tab:r_main_results} (MLP and linear probes averaged per target, 5 seeds).
  Means and mean ranks (1 = best, ties averaged) are over the 77 classification targets (47 tables) and the 46 regression targets (39 tables).
  \textbf{Bold} = best, \uline{underline} = second best.
}
\label{tab:qwen3_row_suite}
\setlength{\tabcolsep}{5pt}
\begin{tabular}{l cc|cc}
\toprule
& \multicolumn{2}{c|}{\textbf{Classification}} & \multicolumn{2}{c}{\textbf{Regression}} \\
\cmidrule(lr){2-3}\cmidrule(lr){4-5}
\textbf{Model} & AUROC\,\met{$\uparrow$} & Rank\,\met{$\downarrow$} & $R^2$\,\met{$\uparrow$} & Rank\,\met{$\downarrow$} \\
\midrule
BERT & $0.791$ & \uline{$2.94$} & $0.378$ & $3.22$ \\
GTE & $0.770$ & $3.87$ & $0.312$ & $4.35$ \\
TabPFN & \uline{$0.793$} & $3.26$ & \uline{$0.486$} & \uline{$2.78$} \\
TabICL & $\textbf{0.816}$ & $\textbf{1.77}$ & $\textbf{0.569}$ & $\textbf{1.39}$ \\
Qwen3-Emb-0.6B & $0.789$ & $3.17$ & $0.410$ & $3.26$ \\
\bottomrule
\end{tabular}
\end{table}

%% file: tables/effbench_column.tex
\begin{table}[!h]
\centering
\caption{\textbf{Column embedding generation cost} (median wall-clock seconds $\pm$ IQR across the efficiency test suite). All models are frozen inference.}
\label{tab:eff_column}
\small
\begin{tabular}{lrr}
\toprule
Model & Time (s) & Datasets \\
\midrule
\textsc{BERT}        &  5.3 $\pm$  4.9 & 58 \\
\textsc{GTE}         &  5.7 $\pm$  1.3 & 58 \\
\textsc{TABBIE}      &  7.4 $\pm$  3.2 & 58 \\
\textsc{TAPAS}       &  7.8 $\pm$  3.9 & 58 \\
\textsc{TURL}        & 11.2 $\pm$  1.8 & 55 \\
\textsc{TaBERT}      & 13.2 $\pm$  9.7 & 58 \\
\textsc{TabSketchFM}      & 14.1 $\pm$  5.4 & 58 \\
\bottomrule
\end{tabular}
\end{table}

%% file: tables/effbench_table.tex
\begin{table}[!h]
\centering
\caption{\textbf{Table embedding generation cost} (median wall-clock seconds $\pm$ IQR).}
\label{tab:eff_table}
\small
\begin{tabular}{lrr}
\toprule
Model & Time (s) & Datasets \\
\midrule
\textsc{TAPEX} &   6.1 $\pm$   3.3 & 58 \\
\bottomrule
\end{tabular}
\end{table}

%% file: tables/effbench_row.tex
\begin{table}[t]
\centering
\caption{\textbf{Row embedding generation cost} (median wall-clock seconds $\pm$ IQR). ``Train'' models include per-table self-supervised training. ``Infer'' models are frozen or meta-pretrained.}
\label{tab:eff_row}
\small
\begin{tabular}{llrr}
\toprule
Model & Type & Time (s) & Datasets \\
\midrule
\textsc{TabICL}          & Infer &    8.7 $\pm$    8.8 &  8 \\
\textsc{TabBinning}      & Train &   10.2 $\pm$    1.4 & 38 \\
\textsc{TabPFN}          & Infer &   20.7 $\pm$  125.0 &  7 \\
\textsc{SCARF}           & Train &   30.9 $\pm$    6.1 & 38 \\
\textsc{GTE}             & Infer &   31.1 $\pm$   14.5 & 38 \\
\textsc{DAE}             & Train &   44.8 $\pm$   27.1 & 38 \\
\textsc{BERT}            & Infer &   50.1 $\pm$   17.4 & 38 \\
\textsc{TABBIE}          & Infer &   57.1 $\pm$   24.6 & 38 \\
\textsc{SubTab}          & Train &   62.9 $\pm$   20.0 & 38 \\
\textsc{VIME}            & Train &   71.8 $\pm$   32.2 & 38 \\
\textsc{TabTransformer}  & Train &   86.5 $\pm$   38.0 & 36 \\
\textsc{SAINT}           & Train &  167.9 $\pm$   64.3 & 37 \\
\textsc{TransTab}        & Train &  875.1 $\pm$  512.1 & 38 \\
\bottomrule
\end{tabular}
\end{table}

%% file: tables/bt_family.tex
\begin{table}[t]
\centering
\small
\caption{Per-family Bradley-Terry rank aggregation of the \textsc{TRL-CTbench} results. Each cell gives a model's rank within the family and its 95\% bootstrap rank interval. Ranks come from pairwise wins in the (task, round) cells that both models observe, without imputation. Intervals come from 2{,}000 resamples of the family's tasks. \na\ marks a model with no task in the family. Bold marks the family's rank-1 model.}
\label{tab:bt_family}
\begin{tabular}{lcccc}
\toprule
\textbf{Model} & \textbf{Schema} & \textbf{Join} & \textbf{Union} & \textbf{Grounding} \\
\midrule
BERT & \textbf{1\,[1,1]} & 2\,[1,9] & 2\,[1,5] & 4\,[3,5] \\
GTE & 2\,[2,3] & \textbf{1\,[1,3]} & 3\,[2,5] & 3\,[1,7] \\
TaBERT & 3\,[2,6] & 6\,[3,8] & 6\,[1,8] & \textbf{1\,[1,3]} \\
TAPAS & 4\,[3,5] & 4\,[3,5] & 5\,[1,7] & 7\,[5,7] \\
TAPEX & \na & 5\,[4,5] & \textbf{1\,[1,8]} & 2\,[2,4] \\
TABBIE & 5\,[3,7] & 8\,[3,8] & 8\,[6,10] & 6\,[2,9] \\
TURL & 6\,[5,6] & 3\,[1,6] & 7\,[6,9] & 5\,[1,8] \\
TUTA & \na & 10\,[10,10] & 9\,[7,10] & 8\,[6,8] \\
Starmie & 7\,[6,7] & 7\,[5,8] & 4\,[1,7] & 9\,[4,10] \\
TabSketchFM & 8\,[8,8] & 9\,[7,9] & 10\,[7,10] & 10\,[7,10] \\
\bottomrule
\end{tabular}
\end{table}

%% file: tables/dlte_significance/dlte_significance_combined.tex
\begin{table}[h]
\centering
\small
\setlength{\tabcolsep}{5pt}
\caption{DLTE per-stage statistical significance on the primary metric $UJ\text{-}H$ and the complementary Cell~$F_1$ diagnostic. Each row tests whether stage-specific model identity affects end-to-end performance, blocking by the remaining two stages (one block per (other-stage-a, other-stage-b) combination, with the aggregate 5-round mean score per pipeline as the unit of analysis). Kendall's~$W$ is the effect-size summary of Friedman, $W = \chi^2 / (n(k-1))$, bounded in $[0,1]$. $W > 0.5$ is a strong effect, $W < 0.1$ is weak. ``Sig.\ pairs (Holm)'' counts Holm-corrected pairwise Wilcoxon signed-rank tests with $p_{\mathrm{adj}}<0.05$, out of $\binom{k}{2}$ total pairs for $k$ models. All three stages have complete block designs and yield highly significant Friedman omnibus tests on both metrics. The most informative contrast is the large metric-dependent Stage~3 effect-size gap ($W=0.61$ on Cell~$F_1$ vs.\ $W=0.09$ on $UJ\text{-}H$), which reflects how upstream retrieval/alignment errors mask the join-side Stage~3 signal in the balanced $UJ\text{-}H$ metric.}
\label{tab:dlte_significance}
\begin{tabular}{llccccc}
\toprule
\textbf{Target stage} & \textbf{Metric} & \textbf{Blocks} & \textbf{$k$} & \makecell{\textbf{Friedman}\\$\chi^2$\;($p$)} & \makecell{\textbf{Kendall's}\\$W$} & \makecell{\textbf{Sig.\ pairs}\\(Holm, $<0.05$)} \\
\midrule
\multirow{2}{*}{Stage~1 (table model)}
 & Cell~$F_1$        & 112 & 10  & $569.6$\;($7.1{\times}10^{-117}$) & $0.57$ & 38\,/\,45 \\
 & $UJ\text{-}H$  & 112 & 10  & $332.6$\;($3.2{\times}10^{-66}$) & $0.33$ & 38\,/\,45 \\
\addlinespace[2pt]
\multirow{2}{*}{Stage~2 (column model)}
 & Cell~$F_1$        & 140 & 8  & $776.7$\;($2.0{\times}10^{-163}$) & $0.79$ & 26\,/\,28 \\
 & $UJ\text{-}H$  & 140 & 8  & $649.9$\;($4.3{\times}10^{-136}$) & $0.66$ & 24\,/\,28 \\
\addlinespace[2pt]
\multirow{2}{*}{Stage~3 (row model)}
 & Cell~$F_1$        & 80 & 14  & $638.7$\;($4.3{\times}10^{-128}$) & $0.61$ & 87\,/\,91 \\
 & $UJ\text{-}H$  & 80 & 14  & $90.6$\;($1.1{\times}10^{-13}$) & $0.09$ & 31\,/\,91 \\
\bottomrule
\end{tabular}
\end{table}

%% file: checklist.tex
\section*{NeurIPS Paper Checklist}

\begin{enumerate}

\item {\bf Claims}
    \item[] Question: Do the main claims made in the abstract and introduction accurately reflect the paper's contributions and scope?
    \item[] Answer: \answerYes{}
    \item[] Justification: The abstract and introduction (Sec.~1) state the benchmark's scope (20 models, 16 tasks, cross-granularity coverage) and three main empirical findings, all supported by the experimental results in Sec.~\ref{sec:experiments}. Limitations are discussed in Appendix~\ref{sec:limitations}.
    \item[] Guidelines:
    \begin{itemize}
        \item The answer \answerNA{} means that the abstract and introduction do not include the claims made in the paper.
        \item The abstract and/or introduction should clearly state the claims made, including the contributions made in the paper and important assumptions and limitations. A \answerNo{} or \answerNA{} answer to this question will not be perceived well by the reviewers.
        \item The claims made should match theoretical and experimental results, and reflect how much the results can be expected to generalize to other settings.
        \item It is fine to include aspirational goals as motivation as long as it is clear that these goals are not attained by the paper.
    \end{itemize}

\item {\bf Limitations}
    \item[] Question: Does the paper discuss the limitations of the work performed by the authors?
    \item[] Answer: \answerYes{}
    \item[] Justification: Appendix~\ref{sec:limitations} discusses the benchmark's frozen-evaluation scope, model-specific operating regimes and the wrapper-induced variation they leave, the fixed probe-head settings, and the interpretive limits of aggregate summaries, and Sec.~\ref{sec:exp-dlte} scopes the DLTE conclusions to controlled reconstruction.
    \item[] Guidelines:
    \begin{itemize}
        \item The answer \answerNA{} means that the paper has no limitation while the answer \answerNo{} means that the paper has limitations, but those are not discussed in the paper.
        \item The authors are encouraged to create a separate ``Limitations'' section in their paper.
        \item The paper should point out any strong assumptions and how robust the results are to violations of these assumptions (e.g., independence assumptions, noiseless settings, model well-specification, asymptotic approximations only holding locally). The authors should reflect on how these assumptions might be violated in practice and what the implications would be.
        \item The authors should reflect on the scope of the claims made, e.g., if the approach was only tested on a few datasets or with a few runs. In general, empirical results often depend on implicit assumptions, which should be articulated.
        \item The authors should reflect on the factors that influence the performance of the approach. For example, a facial recognition algorithm may perform poorly when image resolution is low or images are taken in low lighting. Or a speech-to-text system might not be used reliably to provide closed captions for online lectures because it fails to handle technical jargon.
        \item The authors should discuss the computational efficiency of the proposed algorithms and how they scale with dataset size.
        \item If applicable, the authors should discuss possible limitations of their approach to address problems of privacy and fairness.
        \item While the authors might fear that complete honesty about limitations might be used by reviewers as grounds for rejection, a worse outcome might be that reviewers discover limitations that aren't acknowledged in the paper. The authors should use their best judgment and recognize that individual actions in favor of transparency play an important role in developing norms that preserve the integrity of the community. Reviewers will be specifically instructed to not penalize honesty concerning limitations.
    \end{itemize}

\item {\bf Theory assumptions and proofs}
    \item[] Question: For each theoretical result, does the paper provide the full set of assumptions and a complete (and correct) proof?
    \item[] Answer: \answerNA{}
    \item[] Justification: The paper does not include theoretical results; it is an empirical benchmark paper.
    \item[] Guidelines:
    \begin{itemize}
        \item The answer \answerNA{} means that the paper does not include theoretical results.
        \item All the theorems, formulas, and proofs in the paper should be numbered and cross-referenced.
        \item All assumptions should be clearly stated or referenced in the statement of any theorems.
        \item The proofs can either appear in the main paper or the supplemental material, but if they appear in the supplemental material, the authors are encouraged to provide a short proof sketch to provide intuition.
        \item Inversely, any informal proof provided in the core of the paper should be complemented by formal proofs provided in appendix or supplemental material.
        \item Theorems and Lemmas that the proof relies upon should be properly referenced.
    \end{itemize}

    \item {\bf Experimental result reproducibility}
    \item[] Question: Does the paper fully disclose all the information needed to reproduce the main experimental results of the paper to the extent that it affects the main claims and/or conclusions of the paper (regardless of whether the code and data are provided or not)?
    \item[] Answer: \answerYes{}
    \item[] Justification: Sec.~\ref{sec:benchmark-protocol} defines the frozen evaluation protocol; Sec.~\ref{sec:exp-setup} specifies model selection and reporting conventions; Appendix~\ref{app:reproducibility} provides data access instructions, split definitions, hyperparameters, seeds, and pointers to the exact model-specific wrapper and launch configurations in the released code repository. All source datasets are publicly available.
    \item[] Guidelines:
    \begin{itemize}
        \item The answer \answerNA{} means that the paper does not include experiments.
        \item If the paper includes experiments, a \answerNo{} answer to this question will not be perceived well by the reviewers: Making the paper reproducible is important, regardless of whether the code and data are provided or not.
        \item If the contribution is a dataset and\slash or model, the authors should describe the steps taken to make their results reproducible or verifiable.
        \item Depending on the contribution, reproducibility can be accomplished in various ways. For example, if the contribution is a novel architecture, describing the architecture fully might suffice, or if the contribution is a specific model and empirical evaluation, it may be necessary to either make it possible for others to replicate the model with the same dataset, or provide access to the model. In general. releasing code and data is often one good way to accomplish this, but reproducibility can also be provided via detailed instructions for how to replicate the results, access to a hosted model (e.g., in the case of a large language model), releasing of a model checkpoint, or other means that are appropriate to the research performed.
        \item While NeurIPS does not require releasing code, the conference does require all submissions to provide some reasonable avenue for reproducibility, which may depend on the nature of the contribution. For example
        \begin{enumerate}
            \item If the contribution is primarily a new algorithm, the paper should make it clear how to reproduce that algorithm.
            \item If the contribution is primarily a new model architecture, the paper should describe the architecture clearly and fully.
            \item If the contribution is a new model (e.g., a large language model), then there should either be a way to access this model for reproducing the results or a way to reproduce the model (e.g., with an open-source dataset or instructions for how to construct the dataset).
            \item We recognize that reproducibility may be tricky in some cases, in which case authors are welcome to describe the particular way they provide for reproducibility. In the case of closed-source models, it may be that access to the model is limited in some way (e.g., to registered users), but it should be possible for other researchers to have some path to reproducing or verifying the results.
        \end{enumerate}
    \end{itemize}

\item {\bf Open access to data and code}
    \item[] Question: Does the paper provide open access to the data and code, with sufficient instructions to faithfully reproduce the main experimental results, as described in supplemental material?
    \item[] Answer: \answerYes{}
    \item[] Justification: The code is released at \url{https://github.com/LOGO-CUHKSZ/TRL-Bench} and the three suites at \url{https://huggingface.co/datasets/logo-lab/trl-ctbench}, \url{https://huggingface.co/datasets/logo-lab/trl-rbench}, and \url{https://huggingface.co/datasets/logo-lab/trl-dlte}, with the per-model wrappers, launch configurations, and per-suite walkthroughs. All source datasets are publicly available, and Appendix~\ref{app:licenses} lists their licenses, including sources without an explicit license.
    \item[] Guidelines:
    \begin{itemize}
        \item The answer \answerNA{} means that paper does not include experiments requiring code.
        \item Please see the NeurIPS code and data submission guidelines (\url{https://neurips.cc/public/guides/CodeSubmissionPolicy}) for more details.
        \item While we encourage the release of code and data, we understand that this might not be possible, so \answerNo{} is an acceptable answer. Papers cannot be rejected simply for not including code, unless this is central to the contribution (e.g., for a new open-source benchmark).
        \item The instructions should contain the exact command and environment needed to run to reproduce the results. See the NeurIPS code and data submission guidelines (\url{https://neurips.cc/public/guides/CodeSubmissionPolicy}) for more details.
        \item The authors should provide instructions on data access and preparation, including how to access the raw data, preprocessed data, intermediate data, and generated data, etc.
        \item The authors should provide scripts to reproduce all experimental results for the new proposed method and baselines. If only a subset of experiments are reproducible, they should state which ones are omitted from the script and why.
        \item At submission time, to preserve anonymity, the authors should release anonymized versions (if applicable).
        \item Providing as much information as possible in supplemental material (appended to the paper) is recommended, but including URLs to data and code is permitted.
    \end{itemize}

\item {\bf Experimental setting/details}
    \item[] Question: Does the paper specify all the training and test details (e.g., data splits, hyperparameters, how they were chosen, type of optimizer) necessary to understand the results?
    \item[] Answer: \answerYes{}
    \item[] Justification: Sec.~\ref{sec:benchmark-protocol} defines the evaluation protocol (frozen encoders, linear/MLP probes, training-free tasks). Sec.~\ref{sec:exp-setup} specifies model selection and reporting. Appendix~\ref{app:reproducibility} provides splits, hyperparameters, optimizer choice, seeds, and the repository location of the exact per-model wrapper configurations used for embedding extraction.
    \item[] Guidelines:
    \begin{itemize}
        \item The answer \answerNA{} means that the paper does not include experiments.
        \item The experimental setting should be presented in the core of the paper to a level of detail that is necessary to appreciate the results and make sense of them.
        \item The full details can be provided either with the code, in appendix, or as supplemental material.
    \end{itemize}

\item {\bf Experiment statistical significance}
    \item[] Question: Does the paper report error bars suitably and correctly defined or other appropriate information about the statistical significance of the experiments?
    \item[] Answer: \answerYes{}
    \item[] Justification: All supervised results in the main tables (Tables~\ref{tab:main_results} and~\ref{tab:r_main_results}) and the aggregation ablation tables (Tables~\ref{tab:aggregation_ablation} and~\ref{tab:aggregation_ablation_linear}) report mean $\pm$ std over 5 seeds. DLTE results are 5-round averages. Appendix~\ref{app:statistical} provides Friedman omnibus tests and Holm-corrected pairwise Wilcoxon signed-rank tests for both CTBench and RBench, and explains why confidence intervals are not reported for normalized-rank aggregates.
    \item[] Guidelines:
    \begin{itemize}
        \item The answer \answerNA{} means that the paper does not include experiments.
        \item The authors should answer \answerYes{} if the results are accompanied by error bars, confidence intervals, or statistical significance tests, at least for the experiments that support the main claims of the paper.
        \item The factors of variability that the error bars are capturing should be clearly stated (for example, train/test split, initialization, random drawing of some parameter, or overall run with given experimental conditions).
        \item The method for calculating the error bars should be explained (closed form formula, call to a library function, bootstrap, etc.)
        \item The assumptions made should be given (e.g., Normally distributed errors).
        \item It should be clear whether the error bar is the standard deviation or the standard error of the mean.
        \item It is OK to report 1-sigma error bars, but one should state it. The authors should preferably report a 2-sigma error bar than state that they have a 96\% CI, if the hypothesis of Normality of errors is not verified.
        \item For asymmetric distributions, the authors should be careful not to show in tables or figures symmetric error bars that would yield results that are out of range (e.g., negative error rates).
        \item If error bars are reported in tables or plots, the authors should explain in the text how they were calculated and reference the corresponding figures or tables in the text.
    \end{itemize}

\item {\bf Experiments compute resources}
    \item[] Question: For each experiment, does the paper provide sufficient information on the computer resources (type of compute workers, memory, time of execution) needed to reproduce the experiments?
    \item[] Answer: \answerYes{}
    \item[] Justification: Appendix~\ref{app:efficiency} reports standardized hardware and runtime measurements on the efficiency suite, and Sec.~\ref{sec:exp-setup} gives an estimate of the total compute of the full benchmark, derived from these measurements and the run times of the benchmark jobs. The estimate covers the reported benchmark runs, not ablations or preliminary experiments.
    \item[] Guidelines:
    \begin{itemize}
        \item The answer \answerNA{} means that the paper does not include experiments.
        \item The paper should indicate the type of compute workers CPU or GPU, internal cluster, or cloud provider, including relevant memory and storage.
        \item The paper should provide the amount of compute required for each of the individual experimental runs as well as estimate the total compute.
        \item The paper should disclose whether the full research project required more compute than the experiments reported in the paper (e.g., preliminary or failed experiments that didn't make it into the paper).
    \end{itemize}

\item {\bf Code of ethics}
    \item[] Question: Does the research conducted in the paper conform, in every respect, with the NeurIPS Code of Ethics \url{https://neurips.cc/public/EthicsGuidelines}?
    \item[] Answer: \answerYes{}
    \item[] Justification: The research uses only publicly available datasets under their respective licenses, involves no human subjects, and discusses broader impacts including privacy and re-identification risks in Appendix~\ref{app:broader-impact}.
    \item[] Guidelines:
    \begin{itemize}
        \item The answer \answerNA{} means that the authors have not reviewed the NeurIPS Code of Ethics.
        \item If the authors answer \answerNo, they should explain the special circumstances that require a deviation from the Code of Ethics.
        \item The authors should make sure to preserve anonymity (e.g., if there is a special consideration due to laws or regulations in their jurisdiction).
    \end{itemize}

\item {\bf Broader impacts}
    \item[] Question: Does the paper discuss both potential positive societal impacts and negative societal impacts of the work performed?
    \item[] Answer: \answerYes{}
    \item[] Justification: Appendix~\ref{app:broader-impact} discusses positive impacts (scientific comparability, practitioner guidance) and negative impacts (surveillance, re-identification, inappropriate dataset fusion risks from record linkage and data-lake retrieval).
    \item[] Guidelines:
    \begin{itemize}
        \item The answer \answerNA{} means that there is no societal impact of the work performed.
        \item If the authors answer \answerNA{} or \answerNo, they should explain why their work has no societal impact or why the paper does not address societal impact.
        \item Examples of negative societal impacts include potential malicious or unintended uses (e.g., disinformation, generating fake profiles, surveillance), fairness considerations (e.g., deployment of technologies that could make decisions that unfairly impact specific groups), privacy considerations, and security considerations.
        \item The conference expects that many papers will be foundational research and not tied to particular applications, let alone deployments. However, if there is a direct path to any negative applications, the authors should point it out. For example, it is legitimate to point out that an improvement in the quality of generative models could be used to generate Deepfakes for disinformation. On the other hand, it is not needed to point out that a generic algorithm for optimizing neural networks could enable people to train models that generate Deepfakes faster.
        \item The authors should consider possible harms that could arise when the technology is being used as intended and functioning correctly, harms that could arise when the technology is being used as intended but gives incorrect results, and harms following from (intentional or unintentional) misuse of the technology.
        \item If there are negative societal impacts, the authors could also discuss possible mitigation strategies (e.g., gated release of models, providing defenses in addition to attacks, mechanisms for monitoring misuse, mechanisms to monitor how a system learns from feedback over time, improving the efficiency and accessibility of ML).
    \end{itemize}

\item {\bf Safeguards}
    \item[] Question: Does the paper describe safeguards that have been put in place for responsible release of data or models that have a high risk for misuse (e.g., pre-trained language models, image generators, or scraped datasets)?
    \item[] Answer: \answerNA{}
    \item[] Justification: The benchmark releases evaluation code and task reformulations over existing public datasets. It does not release new pretrained models or scraped datasets that pose high misuse risk.
    \item[] Guidelines:
    \begin{itemize}
        \item The answer \answerNA{} means that the paper poses no such risks.
        \item Released models that have a high risk for misuse or dual-use should be released with necessary safeguards to allow for controlled use of the model, for example by requiring that users adhere to usage guidelines or restrictions to access the model or implementing safety filters.
        \item Datasets that have been scraped from the Internet could pose safety risks. The authors should describe how they avoided releasing unsafe images.
        \item We recognize that providing effective safeguards is challenging, and many papers do not require this, but we encourage authors to take this into account and make a best faith effort.
    \end{itemize}

\item {\bf Licenses for existing assets}
    \item[] Question: Are the creators or original owners of assets (e.g., code, data, models), used in the paper, properly credited and are the license and terms of use explicitly mentioned and properly respected?
    \item[] Answer: \answerYes{}
    \item[] Justification: All source datasets and models are cited in the bibliography. Appendix~\ref{app:licenses} (Table~\ref{tab:licenses}) lists the license for each asset. All assets are used under their original terms.
    \item[] Guidelines:
    \begin{itemize}
        \item The answer \answerNA{} means that the paper does not use existing assets.
        \item The authors should cite the original paper that produced the code package or dataset.
        \item The authors should state which version of the asset is used and, if possible, include a URL.
        \item The name of the license (e.g., CC-BY 4.0) should be included for each asset.
        \item For scraped data from a particular source (e.g., website), the copyright and terms of service of that source should be provided.
        \item If assets are released, the license, copyright information, and terms of use in the package should be provided. For popular datasets, \url{paperswithcode.com/datasets} has curated licenses for some datasets. Their licensing guide can help determine the license of a dataset.
        \item For existing datasets that are re-packaged, both the original license and the license of the derived asset (if it has changed) should be provided.
        \item If this information is not available online, the authors are encouraged to reach out to the asset's creators.
    \end{itemize}

\item {\bf New assets}
    \item[] Question: Are new assets introduced in the paper well documented and is the documentation provided alongside the assets?
    \item[] Answer: \answerYes{}
    \item[] Justification: Each suite has a Hugging Face dataset card and a Croissant file with core and responsible-AI fields. The benchmark is further documented through the task summaries (Appendix~\ref{app:task-summary}), per-dataset curation (Appendices~\ref{app:dataset-inventory} and~\ref{app:curation-rationale}), the dataset counting protocol (Appendix~\ref{app:counting}), reproducibility details (Appendix~\ref{app:reproducibility}), and licenses (Appendix~\ref{app:licenses}). Code and data-preparation scripts are released with the benchmark.
    \item[] Guidelines:
    \begin{itemize}
        \item The answer \answerNA{} means that the paper does not release new assets.
        \item Researchers should communicate the details of the dataset\slash code\slash model as part of their submissions via structured templates. This includes details about training, license, limitations, etc.
        \item The paper should discuss whether and how consent was obtained from people whose asset is used.
        \item At submission time, remember to anonymize your assets (if applicable). You can either create an anonymized URL or include an anonymized zip file.
    \end{itemize}

\item {\bf Crowdsourcing and research with human subjects}
    \item[] Question: For crowdsourcing experiments and research with human subjects, does the paper include the full text of instructions given to participants and screenshots, if applicable, as well as details about compensation (if any)?
    \item[] Answer: \answerNA{}
    \item[] Justification: The paper does not involve crowdsourcing or research with human subjects.
    \item[] Guidelines:
    \begin{itemize}
        \item The answer \answerNA{} means that the paper does not involve crowdsourcing nor research with human subjects.
        \item Including this information in the supplemental material is fine, but if the main contribution of the paper involves human subjects, then as much detail as possible should be included in the main paper.
        \item According to the NeurIPS Code of Ethics, workers involved in data collection, curation, or other labor should be paid at least the minimum wage in the country of the data collector.
    \end{itemize}

\item {\bf Institutional review board (IRB) approvals or equivalent for research with human subjects}
    \item[] Question: Does the paper describe potential risks incurred by study participants, whether such risks were disclosed to the subjects, and whether Institutional Review Board (IRB) approvals (or an equivalent approval/review based on the requirements of your country or institution) were obtained?
    \item[] Answer: \answerNA{}
    \item[] Justification: The paper does not involve research with human subjects.
    \item[] Guidelines:
    \begin{itemize}
        \item The answer \answerNA{} means that the paper does not involve crowdsourcing nor research with human subjects.
        \item Depending on the country in which research is conducted, IRB approval (or equivalent) may be required for any human subjects research. If you obtained IRB approval, you should clearly state this in the paper.
        \item We recognize that the procedures for this may vary significantly between institutions and locations, and we expect authors to adhere to the NeurIPS Code of Ethics and the guidelines for their institution.
        \item For initial submissions, do not include any information that would break anonymity (if applicable), such as the institution conducting the review.
    \end{itemize}

\item {\bf Declaration of LLM usage}
    \item[] Question: Does the paper describe the usage of LLMs if it is an important, original, or non-standard component of the core methods in this research? Note that if the LLM is used only for writing, editing, or formatting purposes and does \emph{not} impact the core methodology, scientific rigor, or originality of the research, declaration is not required.
    \item[] Answer: \answerYes{}
    \item[] Justification: An LLM-based assistant drafted candidate auxiliary targets for the 50 row-prediction tables. The drafts were editable suggestions, and five of the authors made every final target decision under a written guideline (Appendix~\ref{app:dataset-inventory}). LLMs are not part of the evaluation protocol.
    \item[] Guidelines:
    \begin{itemize}
        \item The answer \answerNA{} means that the core method development in this research does not involve LLMs as any important, original, or non-standard components.
        \item Please refer to our LLM policy in the NeurIPS handbook for what should or should not be described.
    \end{itemize}

\end{enumerate}